\documentclass{article}

   \usepackage[nonatbib,preprint]{neurips_2026}

\usepackage{marvosym}
\usepackage{fontawesome}

\usepackage[utf8]{inputenc} 
\usepackage[T1]{fontenc}    
\usepackage{hyperref}       
\usepackage{url}            
\usepackage{booktabs}       
\usepackage{amsfonts}       
\usepackage{nicefrac}       
\usepackage{microtype}      

\usepackage{xspace}
\usepackage{graphicx}
\usepackage{subfigure}
\usepackage{booktabs}
\usepackage{multirow}
\usepackage{listings,upquote,soul,framed}
\usepackage[table,xcdraw]{xcolor}
\usepackage{caption}
\usepackage{array}
\usepackage{tabularx}
\usepackage{longtable}
\usepackage{adjustbox}
\usepackage{markdown}
\markdownSetup{
  pipeTables = true
}

\usepackage{pifont}

\newcommand{\hi}[1]{\vspace{.25em} \noindent {\bf #1} }

\newcommand{\llms}{\textsc{LLMs}\xspace}

\newcommand{\oursys}{\textbf{\textsf{X+Slides}}\xspace}

\title{\oursys: Benchmarking Audience-Conditioned\\ Slide Generation}

\author{
    Haodong Chen$^{1}$, 
    \textbf{Xuanhe Zhou \Letter $^{1}$}
    Wei Zhou$^{1}$,  
    Xinyue Shao$^{2}$,
    Yanbing Zhu$^{1}$, \\
    \textbf{Bo Wang$^{3}$,}
    \textbf{Jiawei Hong$^{3}$,}
    \textbf{Anya Jia$^{3}$,}
    \textbf{Fan Wu$^{1}$} \\
    $^1$ Shanghai Jiao Tong University 
    $^2$ Harbin Institute of Technology
    $^3$ SenseTime \\
    \texttt{spectershell@sjtu.edu.cn}
}

\begin{document}

\maketitle

\begin{abstract}
Automatically generating slide decks from source documents is an important application of large language models (\llms). Existing benchmarks primarily assess slide completeness and technical depth, while overlooking \textit{the target audience} as a critical real-world factor. For instance, specialists demand rigorous proofs, whereas decision-makers prioritize actionable conclusions. To bridge this gap, we introduce \oursys, a benchmark specifically designed for audience-conditioned slide generation. Built on a diverse corpus spanning 113 topics and seven presentation scenes, \oursys employs a dynamic evaluation framework constructed from 8,133 deduplicated, source-grounded probes. By assigning audience-specific utility weights to the same source-grounded probes, \oursys reports four complementary metrics: Audience Coverage measures how much audience-essential information is conveyed, Domain-wise Coverage shows which information types are covered, Efficiency measures delivered utility per unit of attention cost, and Correctness verifies whether slide claims are supported by the source. Experiments on DeepPresenter, SlideTailor, and NotebookLM show that current systems can recover a substantial but still incomplete part of audience-essential information: at $\tau_A=0.7$, DeepPresenter reaches a best Audience Coverage of 0.714, SlideTailor reaches 0.594, and the NotebookLM ablation reaches 0.853 while showing clear grounding differences. These results indicate that visual quality and broad topic coverage should not be treated as evidence support without source-grounded evaluation. 
\end{abstract}

\section{Introduction}
\label{sec:introduction}


Slide presentations are a common medium for communicating information in research, business, policy, and education. An effective slide deck is a selective information artifact tailored to the target audience, more than a shortened source document. For the same research paper, a specialist expects proofs, assumptions, and ablations; a learner needs the core idea and examples; and a decision maker focuses on conclusions and actions. Thus, slide generation is fundamentally audience-conditioned.

Recent automated slide generation systems have evolved rapidly from early layout-aware methods~\cite{Autopresent,SlideCoder} to agentic frameworks that incorporate planning, iterative editing, and retrieval augmentation~\cite{PPTAgent,DeepPresenter,SlideBot,DECKBench,SlideTailor}. Other recent advances have also explored narrative reconstruction using discourse trees~\cite{ArcDeck} and pedagogical optimization for educational clarity~\cite{AutoSlides}. Retrieval-augmented and preference-guided systems further improve generation quality~\cite{SlideTailor}. In parallel, benchmarks such as PPTEval~\cite{PPTAgent}, PresentBench~\cite{PresentBench}, SlidesGen-Bench~\cite{SlidesGen-Bench}, PPTArena~\cite{PPTArena}, and PPTBench~\cite{PPTBench} have been introduced to evaluate generation quality across dimensions like layout, design, and multimodal editing.

Despite this progress, most benchmarks implicitly assume a "one-size-fits-all" paradigm, treating all source facts as equally valuable for different audiences. This assumption overlooks the gap between source-level completeness and audience-specific utility. A generated deck may include many correct facts but miss the information its audience needs most: excessive mathematical proofs can obscure an executive briefing, while superficial summaries can weaken a specialist talk. Therefore, evaluating audience-conditioned slide generation requires a fundamental transition from simple source-alignment to granular audience alignment.

To address this gap, we introduce \oursys, a benchmark for audience-conditioned slide generation. Existing benchmarks often mix two different questions: whether a fact exists in the source, and whether that fact is useful for a particular audience. \oursys separates these two steps. It first builds an audience-agnostic, evidence-backed probe bank, where each probe is a verifiable information unit with a question, expected answer, and source evidence. The same probes are then assigned audience-specific utility weights, and a generated deck is scored by whether high-utility probes can be answered from visible slide content while remaining supported by the source.

\begin{figure}[!t]
    \centering
    \includegraphics[width=.95\linewidth]{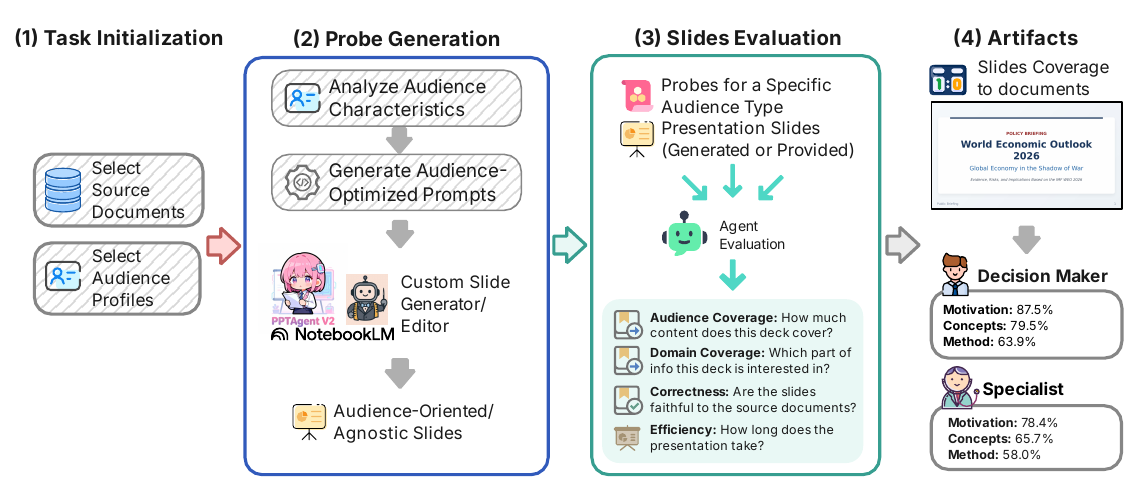}
    \caption{Workflow of \oursys for slides evaluation. Given source documents, audience profiles, usage scenarios, and generated slide decks, \oursys applies specific probe weights, verifies source-supported answerability, and reports audience-conditioned evaluation metrics.}
    \label{fig:benchmark_usage}
\end{figure}

Because the source probes are shared across audiences while their utility weights change with the audience profile, the benchmark can separate source coverage, audience relevance, attention efficiency, and factual support. This decoupled design in \oursys enables the evaluation of four complementary dimensions: \textit{(1) Audience Coverage} quantifies the extent to which audience-essential source information is successfully conveyed in the deck; \textit{(2) Domain-wise Coverage} provides a granular breakdown of Audience Coverage across distinct information types, such as method, limitations, and implementation;
\textit{(3) Efficiency} assesses the density of useful information delivered per unit of audience attention cost;
\textit{(4) Correctness} serves as a critical guardrail, verifying that the claims presented in the deck are factually supported by the source. 
Collectively, these metrics effectively distinguish a faithfully audience-adapted deck from one that is either generically summarized or prone to hallucinations.


To enable this comprehensive evaluation, \oursys uses source documents covering 113 diverse topics, including 50 academic papers and 63 non-academic reports from policy, business, technical, marketing, governance, health, and education domains. It contains 8,133 deduplicated source probes designed to evaluate system performance over three audience profiles within seven presentation scenes. This design fundamentally shifts slide evaluation from document alignment toward audience utility maximization.

In our evaluation of DeepPresenter, SlideTailor, and NotebookLM, current systems recover a substantial part of audience-essential information, but they still show clear limitations in audience-specific information selection and source-grounded correctness. These limitations are difficult to see with audience-agnostic slide metrics, but become visible through \oursys: Audience Coverage shows whether the target audience's important information is selected, Domain-wise Coverage reveals which types of information are over- or under-covered, and Correctness separates useful source-supported content from unsupported claims.

In summary, we make the following contributions.

\noindent(1) We formally conceptualize slide generation as an audience-conditioned task over source-supported information, making explicit that the same source document should be compressed differently when the target reader changes. This is important because it turns audience adaptation from a vague prompting goal into a measurable information-selection problem.

\noindent(2) We develop an audience-agnostic, evidence-backed probe bank that is subsequently weighted to reflect diverse audience utilities.

\noindent(3) We introduce a comprehensive suite of metrics to rigorously evaluate in four dimensions (i.e., Audience Coverage, Domain-wise Coverage, Correctness, and Efficiency).

\noindent(4) We release an effective benchmark using source documents covering 113 topics, 8,133 probes, three distinct audience profiles, and a fully reproducible evaluation pipeline.

\noindent(5) We conduct an in-depth evaluation of DeepPresenter and SlideTailor, with a focused NotebookLM ablation. At $\tau_A=0.7$, DeepPresenter and SlideTailor reach best Audience Coverage scores of 0.714 and 0.594, while the NotebookLM ablation reaches 0.853. More importantly, \oursys shows that audience prompts can change the selected information domains, while Correctness and SafeEfficiency make source-support differences visible.


\section{Related Work}
\label{sec:related}

\hi{Automated Slide Generation.}
Early LLM-based slide generation mainly converts text into slide outlines or layout code. AutoPresent and SlideCoder represent this direction by using code generation or structured layout generation~\cite{Autopresent,SlideCoder}. More recent systems use agent pipelines. PPTAgent and DeepPresenter plan the deck, generate slides, and use visual feedback for refinement~\cite{PPTAgent,DeepPresenter}. SlideBot studies multi-agent slide generation~\cite{SlideBot}. SlideTailor adds preference-conditioned generation for scientific papers~\cite{SlideTailor}. ArcDeck~\cite{ArcDeck} has focused on narrative reconstruction through discourse trees, while Auto-Slides~\cite{AutoSlides} optimizes for pedagogical clarity. These systems improve the ability to produce complete and visually plausible decks. Their objectives, however, usually focus on general slide quality, layout, or user instruction-following rather than on measuring which source information is selected for a specific audience, as \oursys does.

\hi{Slide Evaluation Benchmarks.}
Existing benchmarks cover several important dimensions of presentation quality. PPTEval evaluates content, design, and coherence in PPTAgent~\cite{PPTAgent}. PresentBench evaluates fine-grained checklist completion~\cite{PresentBench}. SlidesGen-Bench proposes computational metrics for generated slides~\cite{SlidesGen-Bench}. PPTBench and PPTArena focus more on multimodal understanding and editing ability~\cite{PPTBench,PPTArena}. Recent work on poster generation has also introduced PaperQuiz~\cite{Paper2Poster}, which uses VLMs to answer content-based questions about a visual artifact. These benchmarks are useful, but they mostly assume that content importance is fixed once the source is given. \oursys instead treats importance as audience-dependent.

\hi{Audience Adaptation.}
Audience adaptation is studied in text simplification, personalized writing, and educational content generation. Most work adapts language difficulty, explanation style, or vocabulary. Recent benchmarks in dialogue and general NLP have introduced persona-aware evaluation~\cite{Personalens} and comprehensive constraint following across hundreds of scenarios~\cite{Cfbench}. Slide generation requires another type of adaptation: deciding which source facts to include at all. DECKBench~\cite{DECKBench} is close to our setting as it uses simulated personas for slide editing. The difference is that it primarily assesses whether a system follows persona-style editing commands, whereas \oursys evaluates whether the final deck covers source information for different audiences.

\hi{LLM-as-Judge Evaluation.}
LLMs and multimodal models are increasingly used as judges for open-ended generation tasks~\cite{JudgeAnything}. We use this paradigm for probe generation, utility weighting, answerability scoring, and correctness checking. To reduce judge-side ambiguity, every probe is tied to source evidence, and a deck receives credit only when the answer can be found in the visible slides and does not contradict the source.

\section{Problem Formulation}
\label{sec:formulation}
\label{sec:problem}


We first define audience-conditioned slide generation as a conditional mapping $f:S \times A \times C \rightarrow T$. Given a source document $S$, an audience profile $A$, and a presentation scene $C$, a generation system produces a slide deck $T=(t_1,\ldots,t_K)$. The source $S$ may be a research paper, report, technical specification, tutorial, or business document. The audience profile $A$ specifies the target role, domain expertise, primary goal, preferred technical depth, and time budget. The presentation scene $C$ describes the delivery setting, such as an academic talk, a policy briefing, an investor pitch, a technical review, or a course lecture. The desired deck should preserve source-supported information that is useful for the target audience under the communication constraints of the scene.
This formulation makes evaluation more demanding than generic slide assessment, because quality depends on the alignment between selected information and audience needs, in addition to fluency, structure, visual coherence, and source grounding. Existing slide-generation benchmarks mainly assess whether the generated deck is fluent, visually coherent, well structured, or grounded in the source. These criteria are necessary, but they do not fully capture the central requirement of audience-conditioned generation: whether the deck selects the \emph{right} information for the intended audience. Most existing benchmarks conflate this requirement with generic source coverage or overall slide quality, without separating source-level information units from audience-specific utility.



To address this gap, we define slide quality at the level of source-supported information slices. An information slice is an atomic, verifiable unit extracted from the source before any audience-specific scoring. In \oursys, each slice is operationalized as an evidence-backed probe
$e_j=(q_j,a_j,Z_j,d_j,m_j,g_j)$, where $q_j$ is a question, $a_j$ is the expected answer, $Z_j$ is the supporting source span, $d_j$ is the depth level, $m_j$ is the evidence modality, and $g_j$ is the information domain.
The depth level $d_j$ describes how detailed the slice is: Level 1 covers context and motivation, Level 2 covers main ideas and main results, Level 3 covers technical details, experiments, and comparisons, and Level 4 covers implementation details, hyperparameters, and edge cases. For example, in a source about cumulative human impact on oceans, the four levels may ask why temporal change is measured, what percentage of the ocean shows increasing impact, how the cumulative-impact score is computed, and why stressor layers are normalized with $\log(X+1)$, respectively. The domain label $g_j$ describes the semantic role of the slice, such as context, method, evidence, limitation, implementation, or implication. These labels are source-side metadata rather than audience labels. Audience conditioning is introduced through the utility weight assigned to the same probe for a specific audience, so score differences are measured over a matched probe bank instead of over different audience-specific question sets. For example, an implementation-detail slice can be essential for a specialist but irrelevant for a decision maker; a motivation slice can remain useful for all audiences but with different importance.

A generated deck communicates a slice well only when the necessary information is visible in the slides and remains supported by the source evidence. Therefore, the generation quality of a deck is not the amount of source text it repeats, but the amount of audience-useful, source-supported slices it successfully conveys under the presentation budget. As shown in Figure~\ref{fig:benchmark_usage}, researchers can use \oursys to evaluate new slide-generation systems over audience-conditioned slide generation, and the complete scoring procedure is formalized in Algorithm~\ref{alg:audience_scoring}.

\begin{table}[!t]
\centering
\refstepcounter{table}\label{alg:audience_scoring}
\small
\begin{tabularx}{\linewidth}{@{}r@{\quad}X@{\quad}p{0.28\linewidth}@{}}
\toprule
\multicolumn{3}{@{}l}{\textbf{Algorithm \thetable} Audience-Conditioned Slide Scoring in \oursys} \\
\midrule
\multicolumn{3}{@{}l}{\textbf{Require:} source $S$; deck $T=(t_1,\ldots,t_K)$; audience $A$; scene $C$;} \\
\multicolumn{3}{@{}l}{\hspace*{4.7em} probe bank $\mathcal{E}$; utility threshold $\tau_A$.} \\
\multicolumn{3}{@{}l}{\textbf{Ensure:} Audience Coverage, Domain-wise Coverage, Efficiency, Correctness, and SafeEfficiency.} \\
1: & $E_A \leftarrow \emptyset$ & $\triangleright$ initialize audience probes \\
2: & \textbf{for all} $e_j=(q_j,a_j,Z_j,d_j,m_j,g_j)\in\mathcal{E}$ \textbf{do} & \\
3: & \quad $w_j \leftarrow \mathrm{UtilityJudge}(e_j,A,C)$ & \\
4: & \quad \textbf{if} $w_j \geq \tau_A$ \textbf{then} $E_A \leftarrow E_A \cup \{(e_j,w_j)\}$ & \\
5: & \textbf{end for} & \\
6: & $R_A \leftarrow 0,\ V_A \leftarrow 0$ & $\triangleright$ aggregate utility \\
7: & \textbf{for all} $(e_j,w_j)\in E_A$ \textbf{do} & \\
8: & \quad $c_j \leftarrow \mathbb{I}[q_j\text{ is answered in }T\text{ and supported by }Z_j]$ & \\
9: & \quad $R_A \leftarrow R_A+w_jc_j,\quad V_A \leftarrow V_A+w_j$ & \\
10: & \textbf{end for} & \\
11: & $K \leftarrow \mathrm{CountSlides}(T)$ & \\
12: & $M \leftarrow 0.25K+\mathrm{Words}(T)/130$ & $\triangleright$ attention cost \\
13: & $\mathrm{AudCov} \leftarrow R_A/V_A$ & $\triangleright$ Audience Coverage \\
14: & \textbf{for all} domain $g$ \textbf{do} & $\triangleright$ domain-wise recall \\
15: & \quad $\mathrm{DomCov}(g) \leftarrow \sum_{e_j:g_j=g} w_jc_j \, / \sum_{e_j:g_j=g} w_j$ & $\triangleright$ Domain-wise Coverage \\
16: & \textbf{end for} & \\
17: & $\mathrm{Eff}_{slide}\leftarrow R_A/K,\quad \mathrm{Eff}_{time}\leftarrow R_A/M$ & \\
18: & $\mathcal{R}\leftarrow \mathrm{ExtractClaims}(T)$ & $\triangleright$ visible slide claims \\
19: & $\mathcal{H}\leftarrow \mathrm{VerifyClaims}(\mathcal{R},S)$ & $\triangleright$ source support labels \\
20: & $\mathrm{Correctness}\leftarrow \mathrm{AggregateSupport}(\mathcal{H})$ & $\triangleright$ claim-level guardrail \\
21: & $\mathrm{SafeEfficiency}\leftarrow \mathrm{Eff}_{time}\times\mathrm{Correctness}$ & \\
22: & \textbf{return} AudCov, DomCov, $\mathrm{Eff}_{slide}$, $\mathrm{Eff}_{time}$, Correctness, SafeEfficiency & \\
\bottomrule
\end{tabularx}
\end{table}

\section{Benchmark Construction}
\label{sec:construction}

\begin{figure}[!t]
    \centering
    \includegraphics[width=\linewidth]{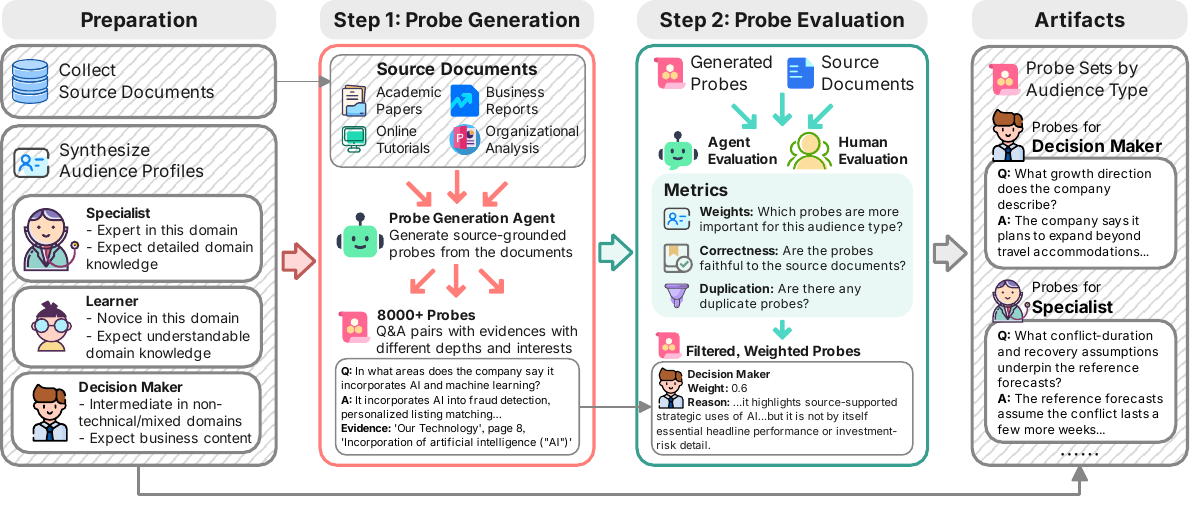}
    \caption{
        Benchmark construction pipeline. \oursys collects and parses diverse sources, builds deduplicated evidence-backed probes, assigns metadata, and attaches audience utility weights.
    }
    \label{fig:pipeline}
\end{figure}

This section describes how we construct the \oursys benchmark. We first present the main design principles, then we describe source document collection, probe generation, audience utility weighting, and deck preparation. The overall dataset construction workflow is shown in Figure~\ref{fig:pipeline}.

\subsection{Design Principles}
\label{subsec:design_principles}

The design of \oursys is based on three measurement principles. These principles are not only implementation choices, but also define what we consider a rigorous and fair evaluation of audience-conditioned slide generation based on real-world observations.

\ding{182} \hi{Separation of Source Facts from Audience Preferences.}
A source document contains many facts, but a given audience needs only a subset of them. Constructing separate probe banks for different audiences would introduce probe difficulty as a confounding factor, making it difficult to attribute score differences to generation quality. Therefore, \oursys first constructs an audience-agnostic probe bank from the source document, and then applies audience-specific utility weights. This ensures that different audiences are evaluated against the same factual baseline, while the scoring changes only through how each audience prioritizes those facts.

\ding{183} \hi{Strict Source Grounding for Credited Units.}
The generated slides may produce statements that appear relevant but are not supported by the source document. Evaluating slides only on audience relevance risks rewarding hallucinations. To avoid this, every probe in \oursys is paired with an expected answer and a precise source evidence span. A slide deck receives credit only when the required answer is visible in the slides and consistent with the source evidence. This ensures that Audience Coverage and Efficiency measure faithful information communication, rather than plausible but unsupported generation.

\ding{184} \hi{Audience Adaptation as Strategic Information Selection.}
Generated decks often include superficial audience-oriented phrases, such as ``tailored for executives'' or ``a beginner's guide''. However, such phrasing does not necessarily mean that the deck selects the right source information. Therefore, \oursys measures audience adaptation as a shift in weighted information recovery. This adaptation has two main axes:
\textit{Content depth} captures the granularity of information and the prerequisite expertise needed to use it. For example, a specialist and a learner may both need method content, but the specialist benefits more from assumptions, ablations, and implementation details, while the learner benefits more from motivation and high-level explanation.
\textit{Information domain focus} captures the semantic role of information. For example, a specialist and a decision maker may both need reliable evidence, but the specialist may value methodological details, while the decision maker may value risks, constraints, and actionable implications. By reporting both Audience Coverage and Domain-wise Coverage, \oursys separates depth adaptation from domain or interest adaptation.

\subsection{Source Document Collection}
\label{subsec:scope}

The benchmark corpus consists of carefully selected source documents covering 113 topics: 50 academic papers and 63 non-academic documents. The academic subset covers computer vision, natural language processing, computer systems, economics, and environmental science. The non-academic subset includes economic and policy reports, corporate investor documents, climate and health reports, technical specifications, marketing materials, governance frameworks, and educational tutorials. This mixture is designed to cover different document types, presentation contexts, and audience utility patterns.


For PDF sources, our parsing pipeline extracts visible text, document titles, section headings, paragraph bodies, tables, and captions. For HTML sources, we extract and clean the main textual content. We use page-level and text-length heuristic checks to control parsing quality. Documents with high text sparsity or substantial table corruption are either parsed with alternative methods or removed from the corpus.

\subsection{Probe Construction and Deck Preparation}
\label{subsec:probe_gen}
\label{subsec:weighting}
\label{subsec:deck_parsing}

In our main experiments, we use three audience profiles: \textit{(1) Specialists}, who require technical depth, explicit assumptions, and detailed limitations; \textit{(2) Learners}, who need motivation, intuitive explanations, and core results; and \textit{(3) Decision Makers}, who prioritize recommendations, risks, and broader business or policy implications. These profiles cover a range of technical depth and information needs, from introductory understanding to expert analysis and strategic decision making.

Probe generation is audience-agnostic. The generation model is instructed to cover all four predefined depth levels, exclude questions answerable only from reference lists, and attach a verifiable source evidence span to every probe. To improve coverage, we run three independent generation passes and merge the results. We then deduplicate the union set: exact duplicates are removed based on evidence spans and answer similarity, while soft duplicates, namely different questions targeting the same source fact, are merged. After deduplication, the probe theme is mapped into one of six coarse information domains: context, method, evidence, limitations, implementation, or implications. This domain label is derived from the source-side probe content and is independent of any generated slide deck.

After deduplication, an LLM-based utility judge assigns a weight to each probe for every target audience. The prompt includes the audience profile and presentation scene, while hiding the broader benchmark condition and source identity. The core prompt templates are shown in Appendix~\ref{subsec:appendix_prompts}, and the full executable prompts are released with the code. The resulting weights are stored with the same source-side probe bank, so different audiences can be compared over matched information units.

Our quantitative experiments benchmark DeepPresenter~\cite{PPTAgent,DeepPresenter} and SlideTailor~\cite{SlideTailor}. We evaluate two generation settings: an \textit{audience-agnostic} setting, where no audience profile is provided, and an \textit{audience-conditioned} setting, where the prompt includes both the presentation scene and the target audience profile. Each generated deck is rendered into high-fidelity slide images, from which we extract visible text. Speaker notes are excluded from scoring. For each deck, we record the number of slides, word count, token count, and estimated presentation duration. Full prompt templates are provided in the appendix and released with our codebase$\footnote{\url{https://github.com/OpenDataBox/X-SlidesBench}}$.

\section{Benchmark Analysis}
\label{sec:benchmark}

This section details the core statistical properties of the \oursys benchmark, illustrating its scale, diversity, and structural composition.

\subsection{Overall Statistics}
\label{subsec:composition}

\oursys uses source documents covering 113 topics, 3 audience profiles, and 7 presentation scenes. It contains 8,133 deduplicated probes. At the main threshold ($\tau_A=0.7$), it retains, on average, 42 essential probes for specialists, 31 for learners, and 24 for decision makers per topic.

\begin{figure}[!t]
    \centering
    \includegraphics[width=\linewidth]{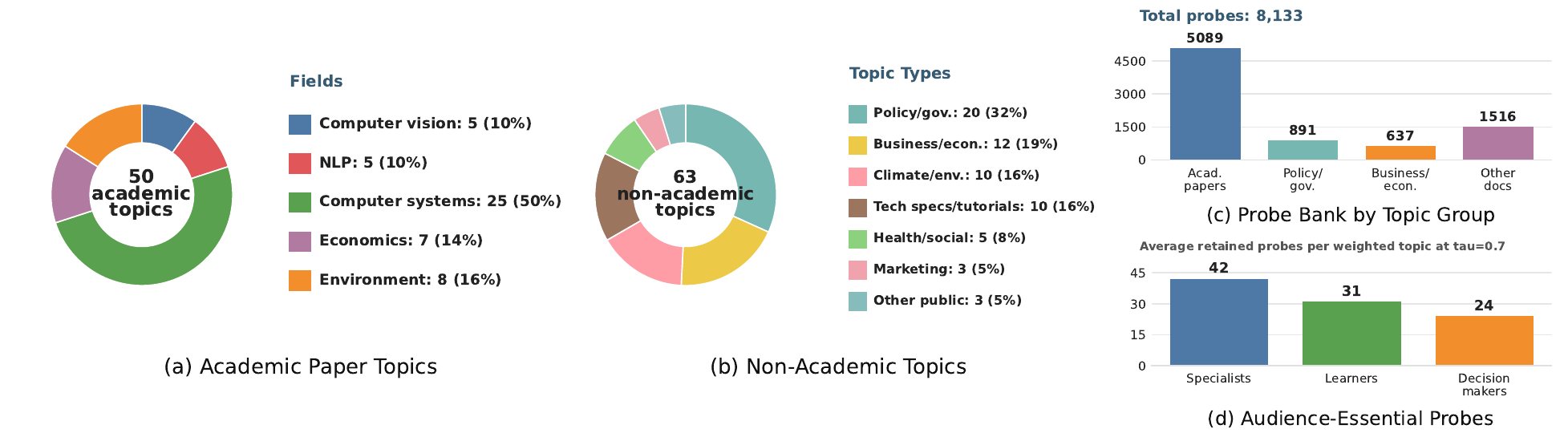}
    \caption{
        Source-topic composition of \oursys, split by academic fields, non-academic categories, and audience-conditioned probe scale.
    }
    \label{fig:source_dashboard}
\end{figure}

\hi{Source Diversity.}
The corpus spans academic papers, macroeconomic reports, corporate annual reports, technical standards, marketing materials, governance frameworks, and tutorials. Figure~\ref{fig:source_dashboard} summarizes this source-topic composition. This heterogeneity distinguishes \oursys from slide benchmarks that focus mainly on academic papers and more accurately reflects real presentation settings, where audience needs vary with both domain and scene.

\hi{Audience Profiles.}
We anchor the evaluation on three canonical audience profiles: \textit{Specialists}, \textit{Learners}, and \textit{Decision Makers}. Specialists intrinsically value granular technical details, methodological assumptions, and rigorous limitations. Learners prioritize high-level motivation, intuitive conceptual framing, and core empirical findings. Decision makers focus heavily on overarching conclusions, strategic risk assessments, and actionable implications. These profiles are intentionally designed to be broad yet distinct, enabling the unified benchmark framework to seamlessly generalize across highly disparate document categories.

\subsection{Probe Bank Statistics}
\label{subsec:probe_stats}

Table~\ref{tab:probe_artifacts} provides a statistical summary of the generated probe bank, categorized by source document group. Naturally, the absolute volume of probes correlates with the length and informational density of the underlying document. We observe distinct distributional trends: academic papers systematically yield a higher concentration of method, evidence, and implementation probes, whereas policy briefs or business reports predominantly generate context and implication-centric probes.

\begin{table}[!t]
\centering
\caption{Probe bank statistics by source group. Counts are before audience threshold filtering.}
\label{tab:probe_artifacts}
{\footnotesize
\setlength{\tabcolsep}{4pt}
\renewcommand{\arraystretch}{0.92}
\begin{tabular}{@{}lrrr@{}}
\toprule
\textbf{Group} & \textbf{Sources} & \textbf{Total probes} & \textbf{Median per source} \\
\midrule
Academic papers & 50 & 5,089 & 143 \\
Policy/governance reports & 20 & 891 & 20 \\
Business/economic reports & 12 & 637 & 21 \\
Technical specifications/tutorials & 10 & 602 & 22 \\
Climate/environment reports & 10 & 173 & 18 \\
Health/social-impact reports & 5 & 99 & 20 \\
Marketing/communication documents & 3 & 444 & 143 \\
Other public documents & 3 & 198 & 17 \\
\midrule
Total & 113 & 8,133 & 22 \\
\bottomrule
\end{tabular}
}
\end{table}

\hi{Depth Distribution.}
The stratification of probes into four distinct depth levels is what enables a single, unified probe bank to be effectively reused across diverse audiences. The utility weighting process can then emphasize Level 3 and Level 4 probes for specialist audiences, while assigning higher relative value to Level 1 and Level 2 probes for learners and decision makers when those probes better match the audience goal.

\hi{Domain Classification.}
Probe themes are mapped into six semantic domains: context, method, evidence, limitations, implementation, and implications. These domain labels are the foundation of the \textit{Domain-wise Coverage} metric defined in Algorithm~\ref{alg:audience_scoring} and analyzed in Section~\ref{subsec:exp_domain}. They identify whether a generation system systematically omits specific categories of useful information, even in cases where the aggregate Audience Coverage score appears satisfactory.

\hi{Evidence Modality.}
Furthermore, each probe explicitly records the modality of its supporting evidence within the source document—categorized as text, table, chart, figure, equation, or a multimodal mix. Tracking modality is critical for transparent evaluation, as contemporary LLM-based slide evaluators currently exhibit higher reliability when processing text and tabular data compared to complex, chart-heavy visual content.

\section{Experiments}
\label{sec:experiments}

We conduct a series of quantitative experiments to address three questions: (1) To what extent can current slide generation systems successfully communicate audience-essential information extracted from the source? (2) Does explicit audience conditioning measurably alter the distribution of covered information domains? (3) How profoundly do source-grounded correctness and the selected utility threshold influence the final interpretation of system performance?

\subsection{Experimental Setup}
\label{subsec:exp_setup}

\hi{Data Subset.}
The benchmark snapshot covers 113 topics. We report all completed deck-level evaluations for DeepPresenter and SlideTailor, including one audience-agnostic deck and three audience-conditioned decks for specialists, learners, and decision makers. We also report a focused NotebookLM ablation. The value of $n$ denotes valid source-topic evaluations in each row, so it can differ across systems when a generator does not produce a valid deck for every source (e.g., incompatible input file format).

\hi{Evaluated System and Backend Models.}
We benchmark DeepPresenter~\cite{PPTAgent,DeepPresenter} and SlideTailor~\cite{SlideTailor}, with NotebookLM included as an ablation. Slide and probe generation use Gemini 3.1 Pro Preview. Utility weighting, slide answerability, and correctness scoring use Gemini 3 Flash Preview. Correctness is scored at the claim level: the evaluator extracts checkable claims from visible deck text, verifies each claim against retrieved source evidence, and averages the support labels with importance weights.

\hi{Evaluation Metrics.}
We use the metrics defined in Algorithm~\ref{alg:audience_scoring}: \textit{Audience Coverage}, \textit{time-based Efficiency}, \textit{slide-based Efficiency}, claim-level \textit{Correctness}, and \textit{SafeEfficiency}. Our main results use the stringent threshold $\tau_A=0.7$, and we report a threshold sweep as sensitivity analysis. We also analyze \textit{Domain-wise Coverage} across six information domains: context, method, evidence, limitations, implementation, and implications.

\hi{Statistical Uncertainty}
\label{subsec:exp_uncertainty}
To account for document-level variability, we estimate statistical uncertainty by bootstrapping over the source-topic evaluation rows for each system, setting, and audience group. For every group, we resample the rows with replacement to create 10,000 bootstrap samples and report the 2.5th and 97.5th percentile endpoints. We use the percentile method because the distributions of our core metrics, especially Coverage and Correctness, are often skewed and non-normal, so the interval is not forced to be symmetric around the mean. Table~\ref{tab:main_results} reports these intervals together with the main point estimates.

\subsection{Main Results}
\label{subsec:exp_main}

Table~\ref{tab:main_results} presents the evaluation results evaluated at the $\tau_A=0.7$ threshold, together with bootstrap uncertainty for the main scored metrics. Each row reflects the performance scored explicitly against the utility weights corresponding to the target audience. Notably, the \textit{Agnostic} rows evaluate a single, identically generated baseline deck, but score its content independently against the distinct utility profiles of the three different audiences.

\hi{Finding 1: Essential information remains only partially covered.}
DeepPresenter and SlideTailor recover a substantial part of essential information, with best Audience Coverage scores of 0.714 and 0.594 over all completed cases. Specialist rows are lower than learner rows, showing that technical audiences require deeper source details. The NotebookLM ablation reaches 0.853 in the agnostic decision-maker setting and remains competitive for learner rows, although its image-based outputs are not directly comparable on editable-deck efficiency. The intervals for Audience Coverage show that learner-oriented information is usually easier to recover than specialist information, but document difficulty remains a major source of variation. 

\hi{Finding 2: Source-grounded Correctness fundamentally alters system ranking.}
Claim-level Correctness remains high for many generated decks, but it still changes how Efficiency should be interpreted. For example, SlideTailor's conditioned learner row has the strongest time-based Efficiency and a high Correctness score, which makes its SafeEfficiency the best among the main PPTX-generating systems. NotebookLM obtains useful Audience Coverage and high Correctness in this focused ablation, while Efficiency is not reported because its image-based PDF decks do not provide comparable editable slide statistics. The Correctness intervals are narrower after moving to claim-level verification, suggesting that most generated claims are source-supported while some decks still contain unverifiable or weakly supported claims. Audience conditioning should therefore be interpreted as improving information selection, not automatically improving all dimensions of presentation quality.

\begin{table}[!t]
\centering
\caption{Main results at $\tau_A=0.7$. AudCov., claim-level Correct., and SafeEff are reported as mean [2.5\%, 97.5\% percentile bootstrap interval] when enough rows are available. Higher is better for all metrics. ``--'' denotes an unreported metric.}
\label{tab:main_results}
\scriptsize
\setlength{\tabcolsep}{2.6pt}
\renewcommand{\arraystretch}{0.96}
\resizebox{\linewidth}{!}{
\begin{tabular}{@{}lllcccccc@{}}
\toprule
\textbf{System} & \textbf{Setting} & \textbf{Scored audience} & \textbf{$n$} & \textbf{AudCov.} & \textbf{Correct.} & \textbf{Eff-time} & \textbf{Eff-slide} & \textbf{SafeEff} \\
\midrule
DeepPresenter & Agnostic & Specialist & 113 & 0.413 [0.363, 0.464] & 0.842 [0.818, 0.864] & 2.505 & 1.822 & 2.248 [1.698, 2.831] \\
DeepPresenter & Agnostic & Learner & 113 & 0.658 [0.609, 0.705] & 0.842 [0.818, 0.864] & 2.560 & 1.848 & 2.281 [1.800, 2.784] \\
DeepPresenter & Agnostic & Decision maker & 113 & 0.622 [0.570, 0.672] & 0.842 [0.818, 0.864] & 1.876 & 1.343 & 1.663 [1.322, 2.026] \\
DeepPresenter & Conditioned & Specialist & 113 & 0.496 [0.445, 0.547] & 0.835 [0.813, 0.854] & 2.985 & 2.297 & 2.603 [2.017, 3.232] \\
DeepPresenter & Conditioned & Learner & 113 & 0.714 [0.672, 0.755] & 0.846 [0.826, 0.864] & 2.803 & 2.046 & 2.448 [1.964, 2.954] \\
DeepPresenter & Conditioned & Decision maker & 113 & 0.654 [0.607, 0.699] & 0.819 [0.795, 0.840] & 2.481 & 1.822 & 2.130 [1.702, 2.576] \\
\midrule
SlideTailor & Agnostic & Specialist & 100 & 0.257 [0.211, 0.305] & 0.823 [0.801, 0.843] & 2.225 & 1.062 & 1.885 [1.334, 2.524] \\
SlideTailor & Agnostic & Learner & 100 & 0.493 [0.435, 0.551] & 0.823 [0.801, 0.843] & 2.834 & 1.346 & 2.374 [1.758, 3.073] \\
SlideTailor & Agnostic & Decision maker & 100 & 0.469 [0.408, 0.530] & 0.823 [0.801, 0.843] & 2.203 & 1.051 & 1.846 [1.370, 2.391] \\
SlideTailor & Conditioned & Specialist & 100 & 0.331 [0.285, 0.378] & 0.823 [0.805, 0.840] & 3.798 & 1.785 & 3.231 [2.386, 4.160] \\
SlideTailor & Conditioned & Learner & 100 & 0.594 [0.548, 0.641] & 0.833 [0.815, 0.851] & 4.229 & 1.983 & 3.626 [2.847, 4.446] \\
SlideTailor & Conditioned & Decision maker & 100 & 0.540 [0.488, 0.591] & 0.818 [0.797, 0.837] & 3.171 & 1.495 & 2.665 [2.084, 3.292] \\
\midrule
NotebookLM & Agnostic & Specialist & 20 & 0.658 [0.561, 0.751] & 0.814 [0.786, 0.841] & -- & -- & -- \\
NotebookLM & Agnostic & Learner & 20 & 0.824 [0.742, 0.895] & 0.814 [0.786, 0.841] & -- & -- & -- \\
NotebookLM & Agnostic & Decision maker & 20 & 0.853 [0.755, 0.930] & 0.814 [0.786, 0.841] & -- & -- & -- \\
NotebookLM & Conditioned & Specialist & 20 & 0.730 [0.628, 0.822] & 0.809 [0.775, 0.840] & -- & -- & -- \\
NotebookLM & Conditioned & Learner & 20 & 0.744 [0.641, 0.835] & 0.812 [0.767, 0.850] & -- & -- & -- \\
NotebookLM & Conditioned & Decision maker & 20 & 0.786 [0.710, 0.853] & 0.805 [0.781, 0.827] & -- & -- & -- \\
\bottomrule
\end{tabular}
}
\end{table}

\subsection{Domain-wise Coverage}
\label{subsec:exp_domain}

Audience Coverage gives a macro signal, but hides which information types are recovered. We therefore stratify retained probes into six domains before scoring. In Table~\ref{tab:domain_results}, the percentage before the slash reports how much of the audience-essential utility belongs to that domain, and the number after the slash reports weighted coverage within that domain. Missing entries indicate that no retained probes from that domain pass the utility threshold for the audience and setting.

\hi{Finding 3: Audience utility weights shift the domain focus.}
The utility distribution reveals clear contrasts in audience priorities. Specialists allocate more essential utility to evidence, method, and implementation details, while decision makers shift utility away from implementation-heavy details toward context, implications, limitations, and evidence strength. Learners place more utility on context than specialists, but still keep a broad mix of method and evidence probes. This structural shift appears for both evaluated systems and highlights why a single Audience Coverage score is insufficient.

\hi{Finding 4: Domain-wise Coverage pinpoints targeting failures.}
Analyzing the shift from agnostic to conditioned generation reveals nuanced trade-offs. For DeepPresenter, conditioning improves Domain-wise Coverage across all six domains, with the largest learner gains in context and implications. SlideTailor shows the same positive direction, especially for learners and decision makers, but from a lower starting point. Domain-wise Coverage therefore shows which part of audience adaptation succeeds, instead of only reporting a macro average.

\begin{table}[!t]
\centering
\caption{Domain utility share and Domain-wise Coverage at $\tau_A=0.7$, where each cell is share (\%) / coverage.}
\label{tab:domain_results}
{\footnotesize
\setlength{\tabcolsep}{3pt}
\renewcommand{\arraystretch}{0.96}
\resizebox{\linewidth}{!}{
\begin{tabular}{@{}lllcccccc@{}}
\toprule
\textbf{System} & \textbf{Setting} & \textbf{Audience} & \textbf{Ctx.} & \textbf{Method} & \textbf{Evidence} & \textbf{Limit.} & \textbf{Impl.} & \textbf{Impact} \\
\midrule
DeepPresenter & Agnostic & Specialist & 10/0.54 & 5/0.52 & 25/0.43 & 12/0.48 & 24/0.49 & 24/0.40 \\
DeepPresenter & Agnostic & Learner & 26/0.75 & 6/0.66 & 19/0.56 & 10/0.64 & 16/0.72 & 23/0.55 \\
DeepPresenter & Agnostic & Decision maker & 26/0.77 & 3/0.64 & 18/0.53 & 14/0.60 & 11/0.69 & 29/0.48 \\
DeepPresenter & Conditioned & Specialist & 10/0.67 & 5/0.59 & 25/0.56 & 12/0.62 & 24/0.62 & 24/0.53 \\
DeepPresenter & Conditioned & Learner & 26/0.84 & 6/0.77 & 19/0.70 & 10/0.73 & 16/0.75 & 23/0.72 \\
DeepPresenter & Conditioned & Decision maker & 26/0.83 & 3/0.69 & 18/0.60 & 14/0.65 & 11/0.67 & 29/0.59 \\
SlideTailor & Agnostic & Specialist & 10/0.30 & 4/0.24 & 25/0.23 & 12/0.25 & 24/0.23 & 25/0.23 \\
SlideTailor & Agnostic & Learner & 26/0.53 & 5/0.49 & 19/0.39 & 10/0.43 & 16/0.41 & 24/0.32 \\
SlideTailor & Agnostic & Decision maker & 25/0.59 & 3/0.54 & 18/0.39 & 14/0.40 & 11/0.45 & 29/0.30 \\
SlideTailor & Conditioned & Specialist & 10/0.52 & 4/0.46 & 25/0.37 & 12/0.43 & 24/0.39 & 25/0.37 \\
SlideTailor & Conditioned & Learner & 26/0.75 & 5/0.65 & 19/0.53 & 10/0.64 & 16/0.62 & 24/0.55 \\
SlideTailor & Conditioned & Decision maker & 25/0.79 & 3/0.69 & 18/0.53 & 14/0.55 & 11/0.63 & 29/0.50 \\
\bottomrule
\end{tabular}
}
}
\end{table}

\subsection{Threshold Sensitivity}
\label{subsec:exp_threshold}

The audience-utility threshold $\tau_A$ decides which probes are treated as essential for the target audience. A lower threshold such as $\tau_A=0.3$ admits many background or secondary probes, while $\tau_A=0.7$ focuses on probes that the utility model regards as strongly required. Thus, $\tau_A=0.3$ asks whether a deck broadly touches useful source information, while $\tau_A=0.7$ asks whether it recovers the information that matters most for the target audience.

Table~\ref{tab:tau_sensitivity} summarizes the aggregate behavior across four threshold choices. As the threshold rises, fewer probes are retained and Audience Coverage usually increases because the denominator focuses on high-utility facts. Efficiency tends to decrease, since secondary useful details contribute less to the utility total. We use $\tau_A=0.7$ for the main claims because it isolates critical audience-required information.

\begin{table}[!t]
\centering
\caption{Sensitivity to the audience-utility threshold. Rank correlation compares Audience Coverage at each threshold with $\tau_A=0.7$.}
\label{tab:tau_sensitivity}
{\footnotesize
\setlength{\tabcolsep}{4pt}
\renewcommand{\arraystretch}{0.96}
\begin{tabular}{@{}rrrrrr@{}}
\toprule
\textbf{$\tau_A$} & \textbf{Rows} & \textbf{Retained} & \textbf{AudCov.} & \textbf{Eff-time} & \textbf{$\rho$ AudCov} \\
\midrule
0.3 & 1278 & 70.238 & 0.472 & 4.029 & 0.930 \\
0.5 & 1278 & 58.987 & 0.481 & 3.810 & 0.951 \\
0.7 & 1278 & 33.108 & 0.524 & 2.789 & 1.000 \\
0.9 & 1278 & 30.997 & 0.528 & 2.685 & 0.993 \\
\bottomrule
\end{tabular}
}
\end{table}

\subsection{Ablations and Diagnostics}
\label{subsec:exp_ablation}

We perform several ablation experiments to evaluate benchmark stability, identify key tradeoffs, and validate our primary implementation choices.

\hi{Deltas from matched audience-conditioning.}
We compare conditioned and agnostic decks across identical source cases to produce the paired deltas in Table~\ref{tab:additional_academic_deltas}. Audience conditioning gives positive shifts in Audience Coverage for all rows, especially for SlideTailor learners and DeepPresenter specialists. Correctness changes are small under claim-level verification, while SafeEfficiency improves for most rows because the utility gain is usually preserved after source verification. This suggests that audience adaptation mainly changes information selection, with grounding remaining a separate but compatible requirement.

\begin{table}[!t]
\centering
\caption{Paired conditioned-minus-agnostic deltas on academic topics.}
\label{tab:additional_academic_deltas}
{\footnotesize
\setlength{\tabcolsep}{3pt}
\renewcommand{\arraystretch}{0.96}
\begin{adjustbox}{max width=\linewidth}
\begin{tabular}{@{}llrrrr@{}}
\toprule
\textbf{System} & \textbf{Audience} & \textbf{Pairs} & \textbf{Delta AudCov.} & \textbf{Delta Correct.} & \textbf{Delta SafeEff} \\
\midrule
DeepPresenter & Specialist & 113 & 0.083 [0.045, 0.120] & -0.007 [-0.025, 0.011] & 0.354 [0.034, 0.691] \\
DeepPresenter & Learner & 113 & 0.056 [0.021, 0.091] & 0.004 [-0.014, 0.022] & 0.167 [-0.090, 0.446] \\
DeepPresenter & Decision maker & 113 & 0.032 [-0.004, 0.068] & -0.023 [-0.044, -0.002] & 0.467 [0.276, 0.673] \\
SlideTailor & Specialist & 100 & 0.074 [0.038, 0.111] & -0.000 [-0.020, 0.023] & 1.346 [0.727, 2.055] \\
SlideTailor & Learner & 100 & 0.101 [0.055, 0.150] & 0.010 [-0.011, 0.034] & 1.252 [0.719, 1.822] \\
SlideTailor & Decision maker & 100 & 0.071 [0.021, 0.122] & -0.005 [-0.028, 0.019] & 0.819 [0.447, 1.217] \\
\bottomrule
\end{tabular}
\end{adjustbox}
}
\end{table}

\hi{Ranking by different headline metrics.}
This diagnostic shows why the benchmark should avoid reducing slide quality to a single scalar. Table~\ref{tab:additional_ranking_tops} lists the top row selected by each headline metric among the PPTX-generating systems. Audience Coverage and Correctness agree on DeepPresenter's learner-conditioned row in this run, while Efficiency and SafeEfficiency prefer SlideTailor's learner-conditioned row. A broad but sparse deck and a compact high-utility deck should therefore lead to different conclusions.

\begin{table}[!t]
\centering
\caption{Ranking sensitivity under different headline metrics.}
\label{tab:additional_ranking_tops}
{\footnotesize
\setlength{\tabcolsep}{3pt}
\renewcommand{\arraystretch}{0.96}
\begin{adjustbox}{max width=\linewidth}
\begin{tabular}{@{}llrrrr@{}}
\toprule
\textbf{Ranking metric} & \textbf{Top row} & \textbf{Score} & \textbf{AudCov.} & \textbf{Correct.} & \textbf{SafeEff} \\
\midrule
Audience Coverage & DeepPresenter Conditioned / Learner & 0.714 & 0.714 & 0.846 & 2.448 \\
Efficiency-time & SlideTailor Conditioned / Learner & 4.229 & 0.594 & 0.833 & 3.626 \\
Correctness & DeepPresenter Conditioned / Learner & 0.846 & 0.714 & 0.846 & 2.448 \\
SafeEfficiency & SlideTailor Conditioned / Learner & 3.626 & 0.594 & 0.833 & 3.626 \\
\bottomrule
\end{tabular}
\end{adjustbox}
}
\end{table}

\hi{Agreement in metric order.}
The Spearman correlations in Table~\ref{tab:additional_metric_correlations} show how far the rankings of different metrics align among the PPTX-generating systems. Efficiency-time and SafeEfficiency are almost identical because Correctness varies less under claim-level verification. In contrast, Audience Coverage has weak rank agreement with Efficiency and SafeEfficiency. This supports reporting Coverage, Efficiency, and Correctness separately rather than hiding them inside one combined score.

\begin{table}[!t]
\centering
\caption{Spearman correlations between the aggregate metric rankings.}
\label{tab:additional_metric_correlations}
{\footnotesize
\setlength{\tabcolsep}{3pt}
\renewcommand{\arraystretch}{0.96}
\begin{adjustbox}{max width=\linewidth}
\begin{tabular}{@{}llr@{}}
\toprule
\textbf{Metric A} & \textbf{Metric B} & \textbf{Spearman rho} \\
\midrule
Audience Coverage & Efficiency-time & 0.000 \\
Audience Coverage & Correctness & 0.383 \\
Audience Coverage & SafeEfficiency & 0.049 \\
Efficiency-time & SafeEfficiency & 0.993 \\
Correctness & Efficiency-time & -0.220 \\
Correctness & SafeEfficiency & -0.170 \\
\bottomrule
\end{tabular}
\end{adjustbox}
}
\end{table}

\hi{Cross-audience scoring.}
We score each audience-conditioned deck from DeepPresenter against all three audience profiles to produce Table~\ref{tab:additional_cross_audience}, which reveals whether a prompt truly changes the information covered or merely changes surface wording. The decision-maker prompt shows a clear target advantage, while the learner prompt gives a smaller positive advantage and the specialist prompt is negative. While audience prompts are effective, they do not consistently align with the intended profile without audience-conditioned evaluation.

\begin{table}[!t]
\centering
\caption{Cross-audience scoring matrix for DeepPresenter decks conditioned on the audience.}
\label{tab:additional_cross_audience}
{\footnotesize
\setlength{\tabcolsep}{3pt}
\renewcommand{\arraystretch}{0.96}
\begin{adjustbox}{max width=\linewidth}
\begin{tabular}{@{}lrrrr@{}}
\toprule
\textbf{Prompt audience} & \textbf{Specialist} & \textbf{Learner} & \textbf{Decision maker} & \textbf{Target adv.} \\
\midrule
Specialist & 0.526 & 0.648 & 0.639 & -0.117 \\
Learner & 0.543 & 0.685 & 0.669 & 0.079 \\
Decision maker & 0.480 & 0.629 & 0.637 & 0.082 \\
\bottomrule
\end{tabular}
\end{adjustbox}
}
\end{table}

\hi{Ablation of the scene field in the generation prompt.}
We further remove the explicit scene description while keeping the audience profile. Table~\ref{tab:scene_prompt_ablation} reports the audience-only DeepPresenter rows and their paired deltas relative to the normal conditioned prompt. Audience-only prompts increase measured Audience Coverage in this controlled set, but they also reduce Correctness, especially for decision makers. This suggests that broad audience instructions can encourage wider information selection, while the scene field helps constrain the deck and should be reported as part of the generation condition.

\begin{table}[!t]
\centering
\caption{Effect of removing the explicit scene field from DeepPresenter prompts. Deltas are audience-only minus normal conditioned prompts.}
\label{tab:scene_prompt_ablation}
{\footnotesize
\setlength{\tabcolsep}{3pt}
\renewcommand{\arraystretch}{0.96}
\begin{adjustbox}{max width=\linewidth}
\begin{tabular}{@{}lrrrrrr@{}}
\toprule
\textbf{Audience} & \textbf{$n$} & \textbf{AudCov.} & \textbf{Correct.} & \textbf{SafeEff} & \textbf{$\Delta$ AudCov.} & \textbf{$\Delta$ Correct.} \\
\midrule
Specialist & 20 & 0.907 [0.849, 0.957] & 0.726 [0.643, 0.804] & 5.694 [4.331, 7.159] & 0.381 [0.265, 0.505] & -0.107 [-0.212, 0.006] \\
Learner & 20 & 0.926 [0.860, 0.969] & 0.796 [0.715, 0.866] & 5.397 [4.020, 6.927] & 0.240 [0.151, 0.337] & -0.061 [-0.149, 0.020] \\
Decision maker & 20 & 0.913 [0.843, 0.967] & 0.716 [0.642, 0.783] & 3.935 [2.968, 4.883] & 0.276 [0.162, 0.402] & -0.108 [-0.181, -0.037] \\
\bottomrule
\end{tabular}
\end{adjustbox}
}
\end{table}

\hi{Ablation of the generator family using NotebookLM.}
NotebookLM serves as a contrast within the family of generators. As shown in Table~\ref{tab:main_results}, it obtains competitive Audience Coverage for learner and decision-maker groups, while Correctness separates well-supported decks from more speculative ones.

\hi{Ablation of the practitioner profile.}
We use the practitioner profile solely as an ablation to test extensibility. Table~\ref{tab:additional_practitioner} reports the results under this fourth profile. The same audience-agnostic probes can be reweighted, and the retained probe count remains reasonable. Audience Coverage is high in this controlled set, while Correctness has wider dispersion than in the main benchmark rows. This shows that a new profile can expose source-support differences in implementation-oriented decks, so we use it as a stress test rather than a primary benchmark profile.

\begin{table}[!t]
\centering
\caption{Inspection of the practitioner profile.}
\label{tab:additional_practitioner}
{\footnotesize
\setlength{\tabcolsep}{3pt}
\renewcommand{\arraystretch}{0.96}
\begin{adjustbox}{max width=\linewidth}
\begin{tabular}{@{}lrrl@{}}
\toprule
\textbf{Metric} & \textbf{n} & \textbf{Mean / CI} & \textbf{Dispersion} \\
\midrule
Retained probes & 20 & 45.750 [37.100, 53.950] & med 46.500, sd 19.891, range 9.000--81.000 \\
Audience Coverage & 20 & 0.871 [0.806, 0.927] & med 0.925, sd 0.141, range 0.444--1.000 \\
Efficiency based on time & 20 & 5.074 [4.202, 5.985] & med 5.124, sd 2.076, range 1.512--8.926 \\
Correctness & 20 & 0.715 [0.639, 0.791] & med 0.726, sd 0.176, range 0.466--0.986 \\
SafeEfficiency & 20 & 3.715 [2.893, 4.579] & med 3.383, sd 1.991, range 0.983--8.380 \\
\bottomrule
\end{tabular}
\end{adjustbox}
}
\end{table}

\hi{Sensitivity of the evaluator.}
Paired evaluator diagnostics show that scores can vary with implementation. Table~\ref{tab:additional_evaluator} reports batching and input-modality contrasts on matched rows. In this larger diagnostic set, batch size 12 slightly lowers measured Audience Coverage compared with batch size 1, and image-based evaluation yields substantially lower Audience Coverage than deck-text evaluation. Correctness changes are small in both contrasts. Benchmark reports should therefore fix and disclose both input format and batching policy.

\begin{table}[!t]
\centering
\caption{Diagnostics for the sensitivity of paired evaluators.}
\label{tab:additional_evaluator}
{\footnotesize
\setlength{\tabcolsep}{3pt}
\renewcommand{\arraystretch}{0.96}
\begin{adjustbox}{max width=\linewidth}
\begin{tabular}{@{}llrrrr@{}}
\toprule
\textbf{Diagnostic} & \textbf{Contrast} & \textbf{Pairs} & \textbf{Delta AudCov.} & \textbf{Delta Correct.} & \textbf{Delta SafeEff} \\
\midrule
Batching of the judge & Batch size 12 - Batch size 1 & 17 & -0.072 [-0.110, -0.040] & -0.002 [-0.005, 0.000] & -0.602 [-0.820, -0.386] \\
Modality of the input & Images of slides - Text of the deck & 17 & -0.322 [-0.426, -0.226] & 0.022 [-0.009, 0.064] & -2.534 [-3.714, -1.581] \\
\bottomrule
\end{tabular}
\end{adjustbox}
}
\end{table}

\section{Conclusions}
\label{sec:conclusion}
\label{sec:broader_impacts}

We introduce \oursys, a benchmark for evaluating slide generation with audience-conditioned utility. It builds source-grounded probes without using audience information, assigns audience-specific utility weights to the same probes, and evaluates generated decks with Audience Coverage, Domain-wise Coverage, Efficiency, and Correctness. Our experiments show that current systems recover a substantial but still incomplete part of audience-essential information: DeepPresenter reaches a best Audience Coverage of 0.714 at $\tau_A=0.7$, SlideTailor reaches 0.594, and the NotebookLM ablation reaches 0.853 while showing clear grounding differences. NotebookLM can produce visually strong decks and competitive coverage in some settings, but several cases show weak source-grounded Correctness, so visual quality and broad topic coverage should not be treated as evidence support. \oursys can support research communication, education, policy briefing, and business decision support, but it should not be used as the only measure of presentation quality. Its current limitations are that LLM-generated probes and weights may miss facts in long or highly visual documents, three broad audience profiles may not fit all domains, and textual scoring does not directly evaluate graphs, diagrams, or layout-based explanations. Future work can extend \oursys to finer-grained profiles, stronger visual grounding, graded explanation quality, and multilingual or cross-cultural presentation settings.

\newpage

\begin{ack}
\end{ack}


\bibliographystyle{IEEEtran}
\bibliography{reference}



\newpage

\appendix
\section{Appendix}
\label{sec:appendix}

\subsection{Source Document Details}
\label{subsec:appendix_sources}

We provide the source document composition in Section~\ref{sec:construction}. To maintain the brevity of this appendix, we have included the complete list of documents, along with their metadata and source URLs, in our released benchmark files.

\subsection{Audience Profile Definitions}
\label{subsec:appendix_audiences}

We established the three audience profiles for the benchmark. By using identical profile descriptions for both the audience-conditioned slide prompts and the utility weighting, we ensure that generation and scoring remain closely aligned. Meanwhile, the probe bank itself stays independent of any specific audience. Table~\ref{tab:audience_profiles} details the exact configurations used.

\begin{table}[!h]
\centering
\caption{Audience profile configurations used in the main benchmark.}
\label{tab:audience_profiles}
{\scriptsize
\setlength{\tabcolsep}{3pt}
\renewcommand{\arraystretch}{1.08}
\begin{tabularx}{\linewidth}{@{}p{0.13\linewidth}p{0.18\linewidth}p{0.21\linewidth}XX@{}}
\toprule
\textbf{Audience} & \textbf{Role, expertise, time} & \textbf{Goal} & \textbf{High-value information} & \textbf{Low-value information} \\
\midrule
Specialist & Domain specialist or expert reviewer, expert, 12 min & Assess precision, evidence quality, caveats, and source-grounded claims. & Exact claims, definitions, assumptions, and scope, evidence quality, quantitative support, and source traceability, methods, mechanisms, and causal drivers, caveats, limitations, uncertainty, and edge cases. & Generic framing after the topic is established, unsupported persuasion or branding, visuals that replace evidence or mechanism detail. \\
\midrule
Learner & Learner or early-career audience, intermediate, 15 min & Understand the topic, key concepts, evidence, and reasoning process. & Context, prerequisite concepts, and motivating examples, terminology and metric explanations, step-by-step logic, main takeaways with supporting evidence, limitations and follow-up questions. & Unexplained jargon, acronyms, or specialized metrics, dense tables or formulas without interpretation, exhaustive implementation details before the core idea is clear. \\
\midrule
Decision maker & Decision maker or non-specialist stakeholder, nontechnical to mixed, 8 min & Identify source-supported implications, tradeoffs, risks, and next questions. & Plain-language problem and proposal, source-supported impact, evidence strength, and confidence, practical implications, risks, dependencies, and constraints, decision-relevant alternatives, uncertainty and next-step questions. & Implementation minutiae, equations, or internal mechanics without decision relevance, raw data without interpretation, field-specific jargon without translation, unsupported recommendations, timelines, or costs. \\
\bottomrule
\end{tabularx}
}
\end{table}

\subsection{Prompt Templates}
\label{subsec:appendix_prompts}

We provide the core prompt templates used in the benchmark pipeline below. The full prompts, which include all configuration details, are available in our released code repository.

\hi{Audience-Agnostic Probe Generation.}
\begin{framed}
\begin{small}
\noindent\textbf{Task:} Generate source-grounded probes from the provided document. Do not target any specific audience.

\noindent\textbf{Rules:}
1. Every probe must be answerable using only the source document.
2. Provide the question, expected answer, and the evidence span in the source.
3. Assign a depth level: 1 (context/motivation), 2 (main method/results), 3 (technical details/ablations), or 4 (implementation details/edge cases).
4. Cover all depth levels when the source contains relevant content.
5. Do not ask questions that can only be answered from citations or the reference list.
\end{small}
\end{framed}

\hi{Utility Weighting.}
\begin{framed}
\begin{small}
\noindent\textbf{Task:} Given a source probe, an audience profile, and a presentation scene, assign a utility weight.

\noindent\textbf{Rubric:}
1.0: Essential. The audience cannot achieve their goal without this information.
0.6: Important. This information significantly helps the audience.
0.3: Useful background. Nice to know but not required.
0.0: Irrelevant. This information does not help this audience.
\end{small}
\end{framed}

\hi{Slide Answerability Evaluation.}
\begin{framed}
\begin{small}
\noindent\textbf{Task:} Using only the visible content of the provided slides, answer the probe question. If the answer cannot be found in the slides, respond with ``not found.'' When an answer is found, cite the specific slide and content element that contains it.

\noindent The answer will be compared against the expected answer and source evidence. Answers that contradict the source evidence do not receive credit.
\end{small}
\end{framed}

\hi{Claim-level Correctness.}
\begin{framed}
\begin{small}
\noindent\textbf{Claim extraction task:} Extract atomic, checkable claims from the visible slide text. Keep only claims that can be verified against the source document, assign an importance level, and ignore decorative text, section titles, and generic presentation language.

\noindent\textbf{Claim verification task:} Given a batch of extracted claims and retrieved source evidence, label each claim as supported, weakly supported, unsupported, not verifiable, or contradicted. The final Correctness score is an importance-weighted aggregation of these labels.
\end{small}
\end{framed}

\subsection{Qualitative Deck Showcase}
\label{subsec:appendix_deck_showcase}

We visualize seven source cases in Figures~\ref{fig:deck_showcase_case019}--\ref{fig:deck_showcase_case016}. The examples are selected to show audience-profile effects and metric tradeoffs rather than visual style alone, including focused NotebookLM comparisons. In each row, the first image is always the first slide of the generated deck, followed by slides 2--5.

\newcommand{\deckshot}[1]{\includegraphics[width=0.192\linewidth]{#1}}
\newcommand{\deckshowcaserow}[6]{%
\noindent{\scriptsize\textbf{#1}}\\[0.30em]%
\noindent\deckshot{#2}\hfill\deckshot{#3}\hfill\deckshot{#4}\hfill\deckshot{#5}\hfill\deckshot{#6}\par\vspace{0.25em}%
}

\begin{figure}[!p]
\centering
{\setlength{\tabcolsep}{0pt}%
\deckshowcaserow{DeepPresenter / Specialist: AudCov. 0.694, Correct. 0.772, SafeEff 4.398}{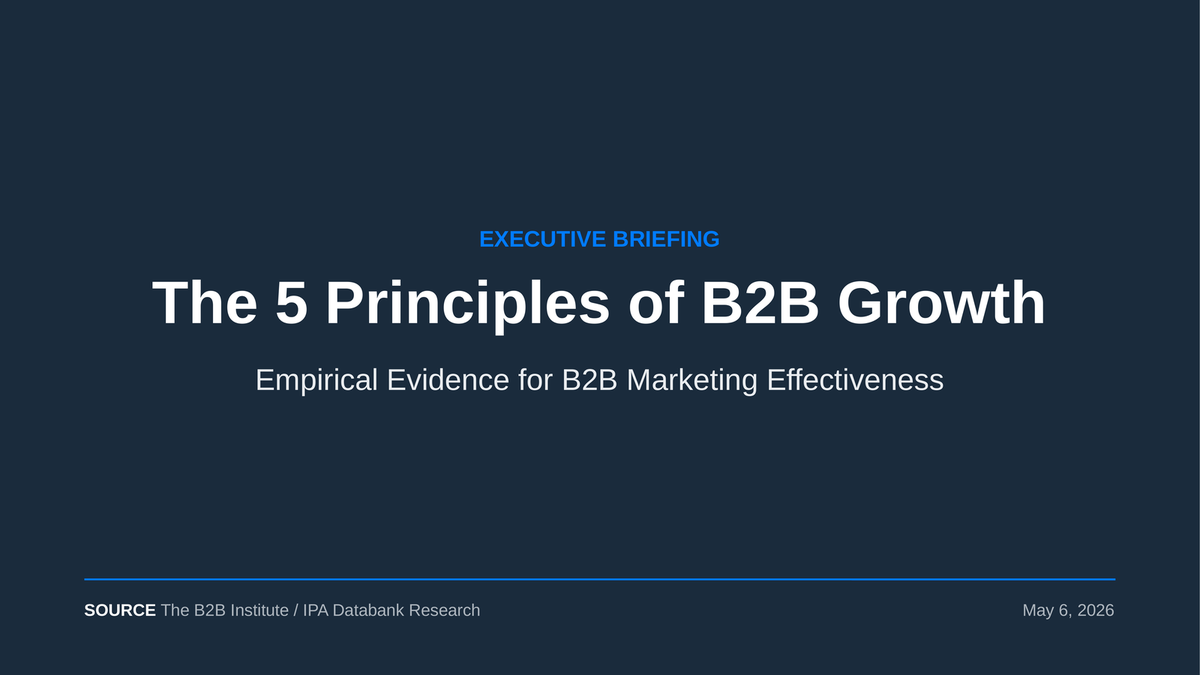}{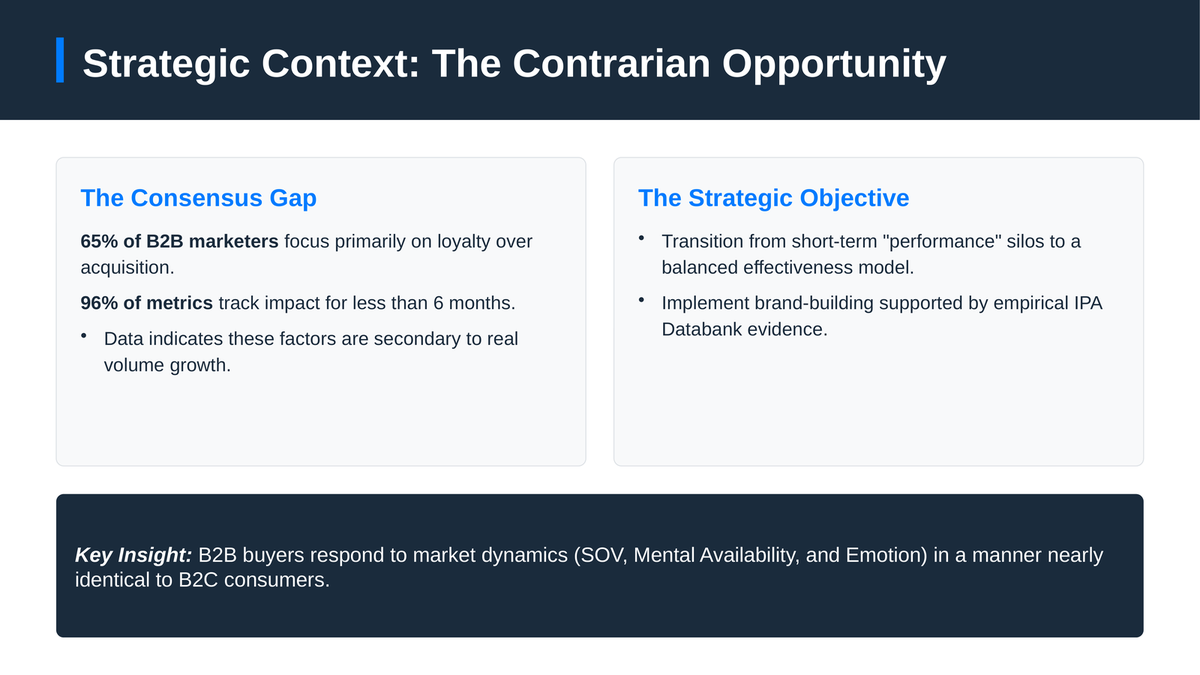}{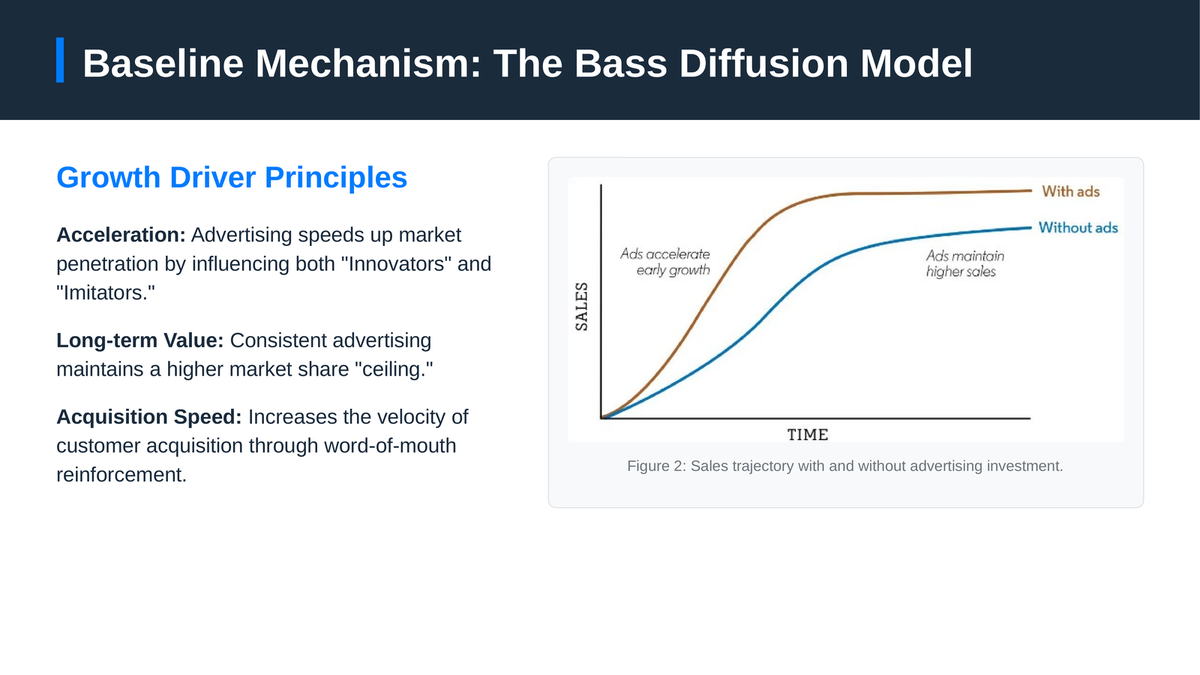}{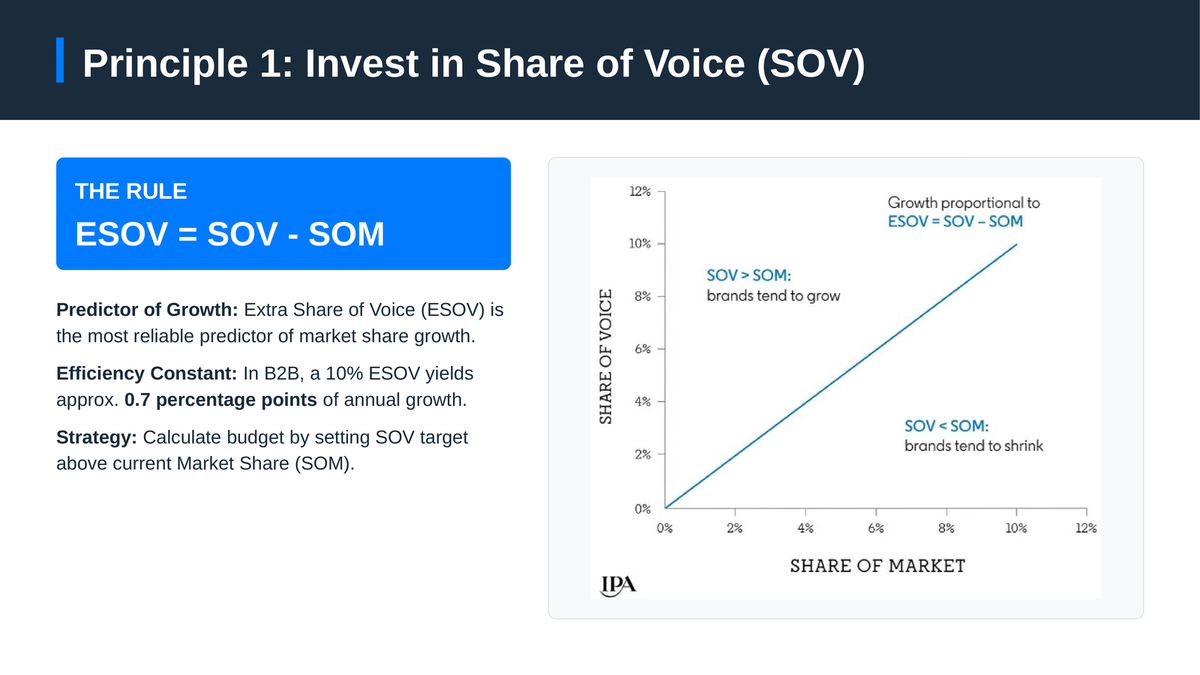}{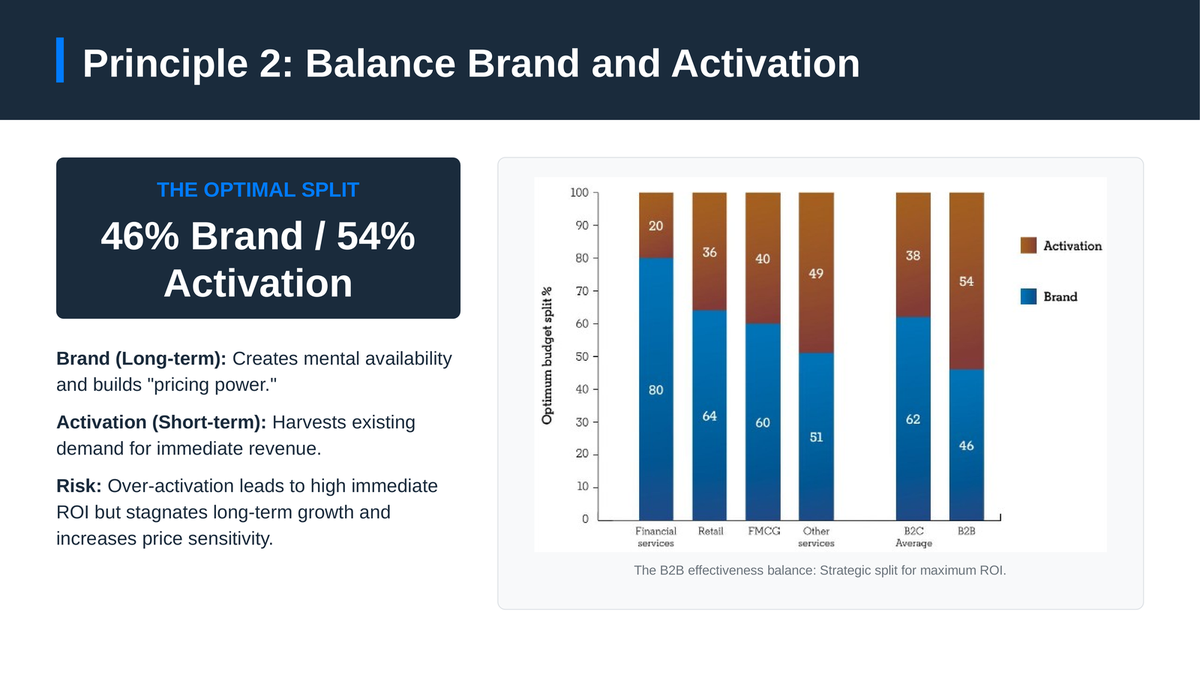}
\deckshowcaserow{DeepPresenter / Learner: AudCov. 0.859, Correct. 0.849, SafeEff 5.094}{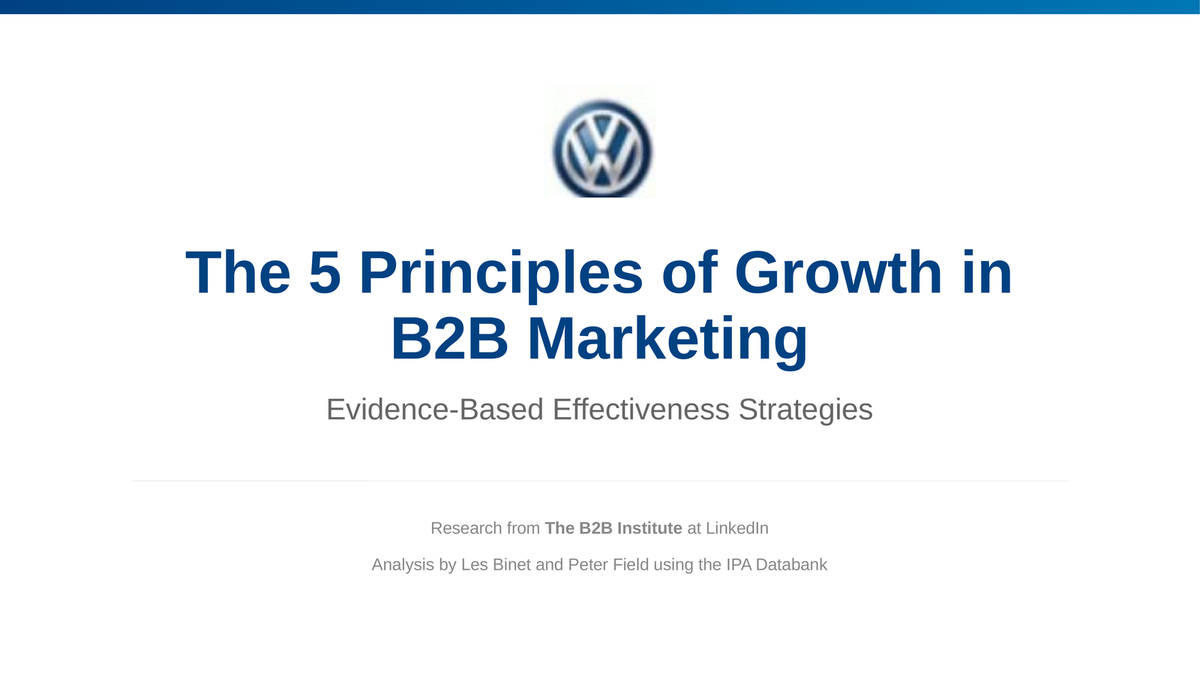}{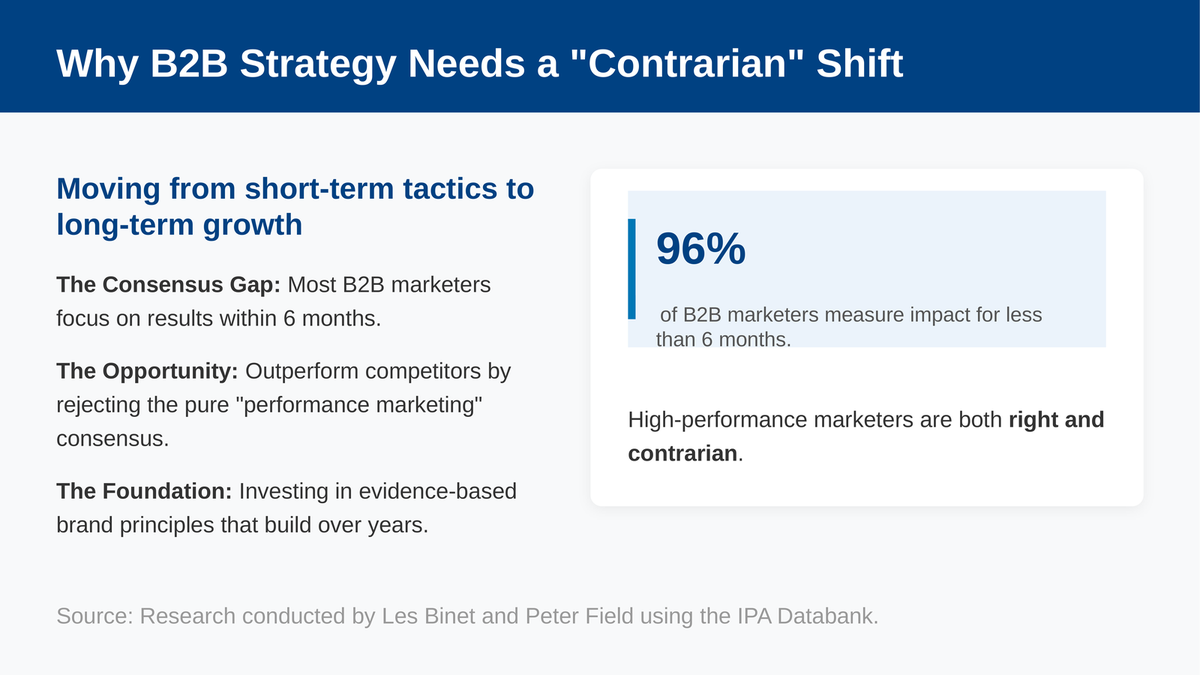}{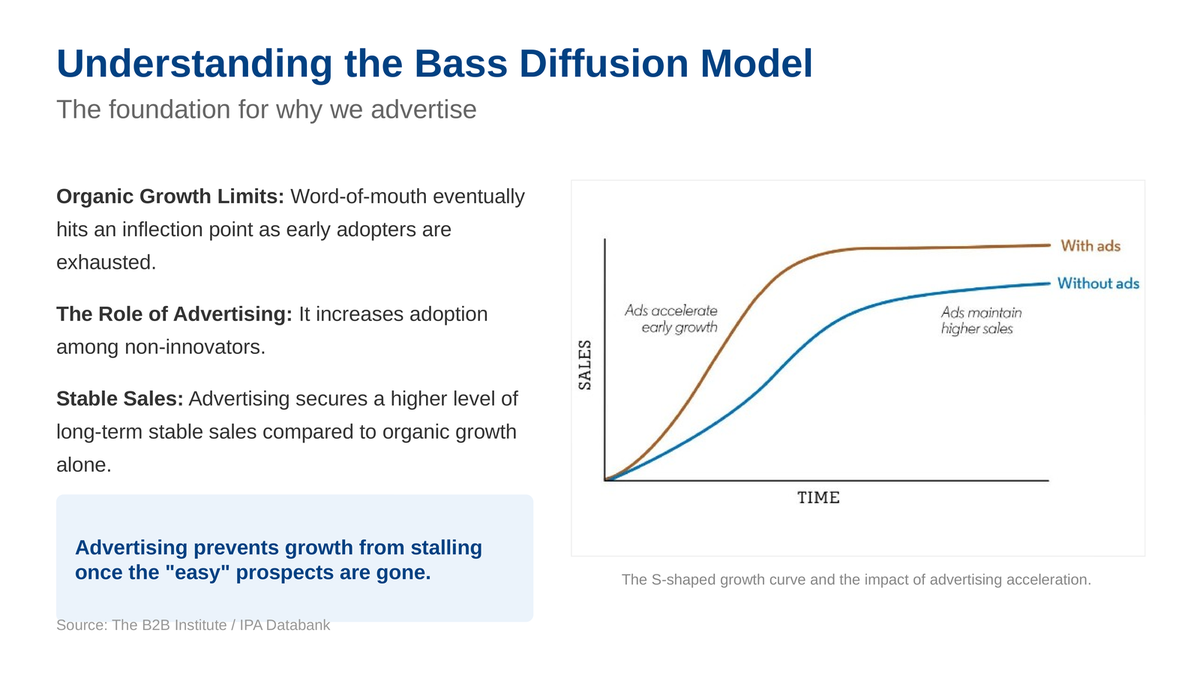}{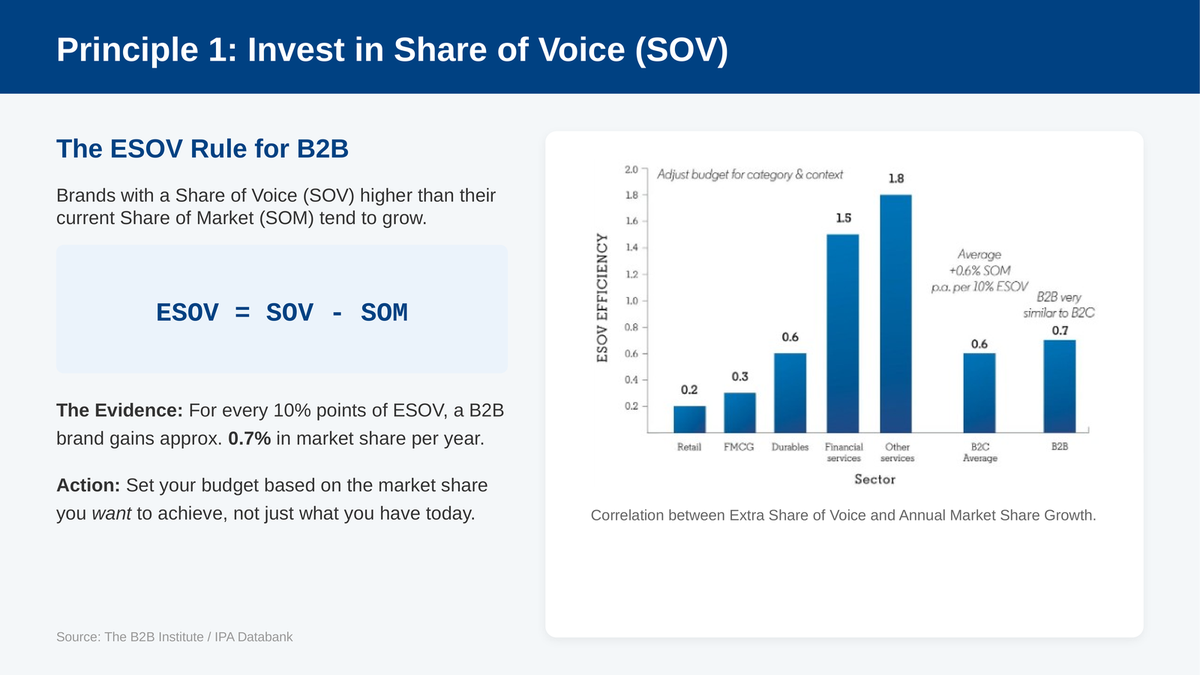}{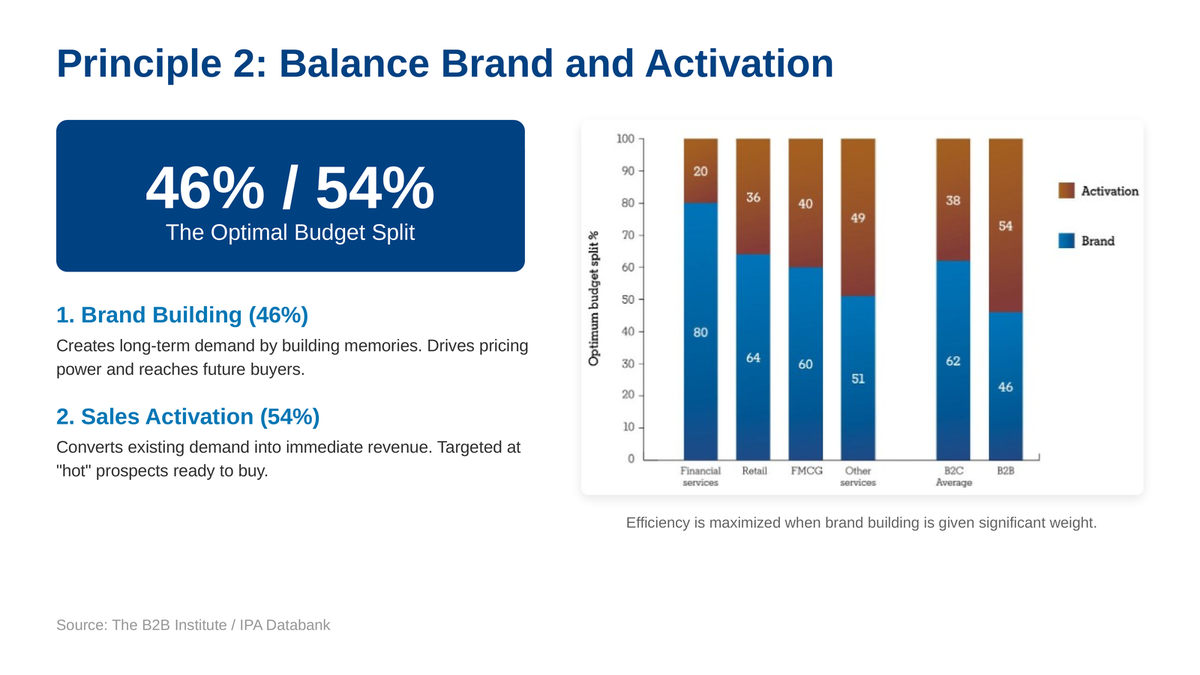}
\deckshowcaserow{DeepPresenter / Decision maker: AudCov. 0.774, Correct. 0.847, SafeEff 3.509}{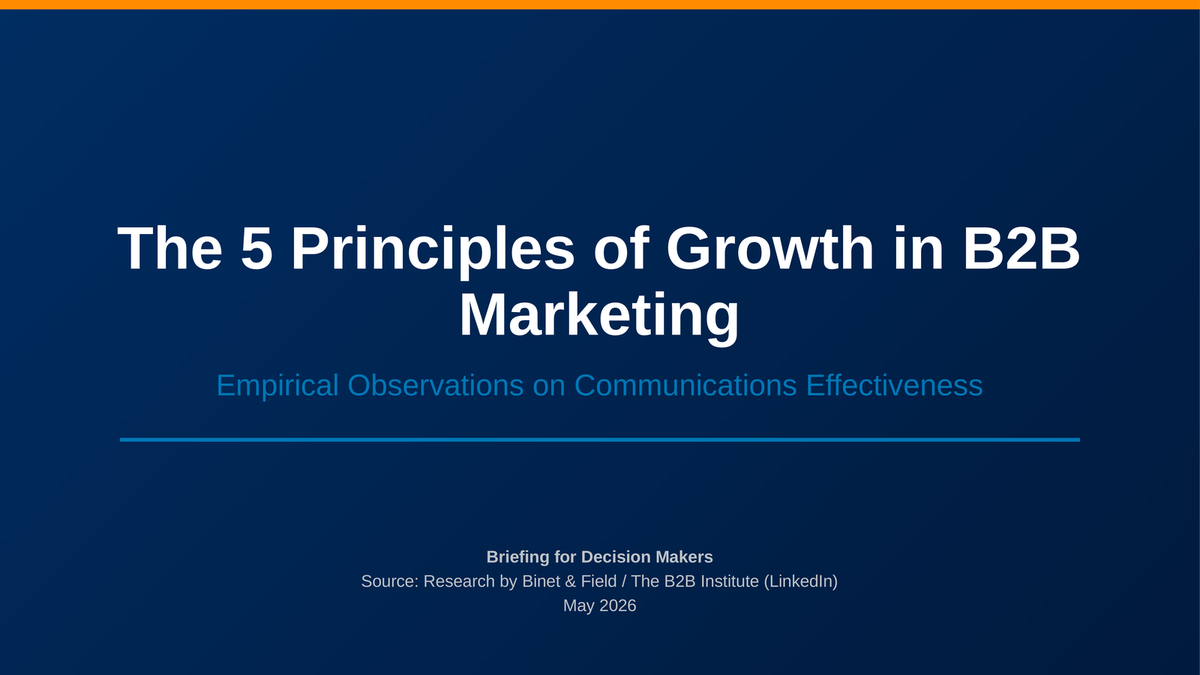}{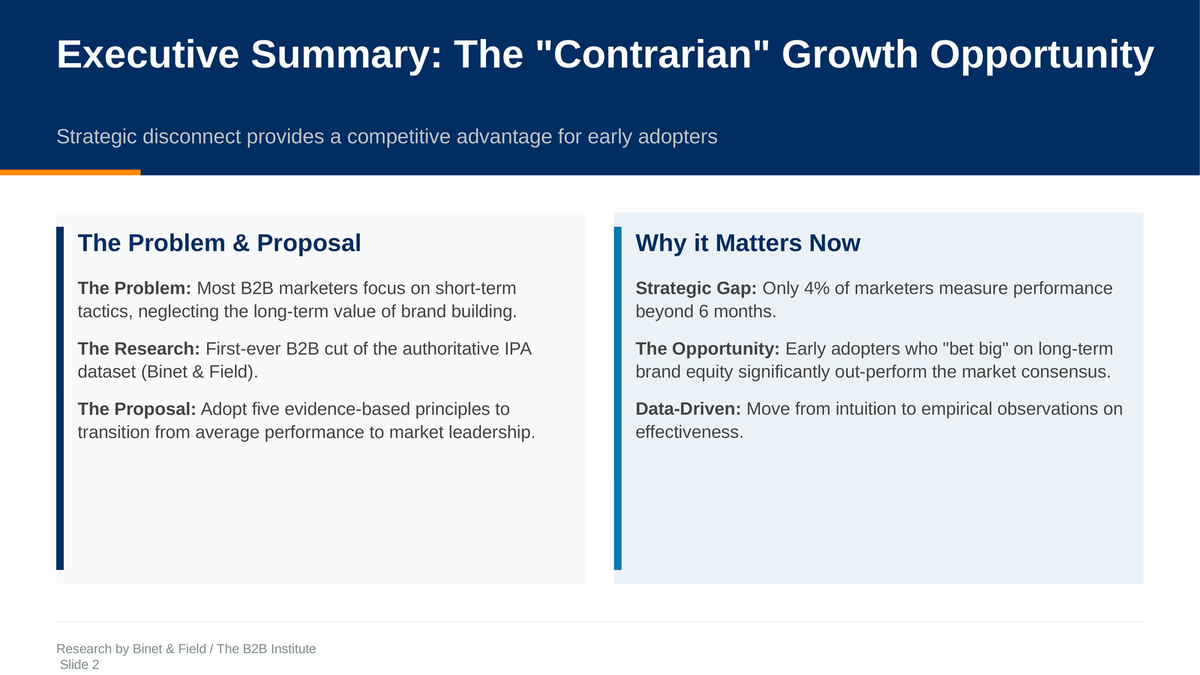}{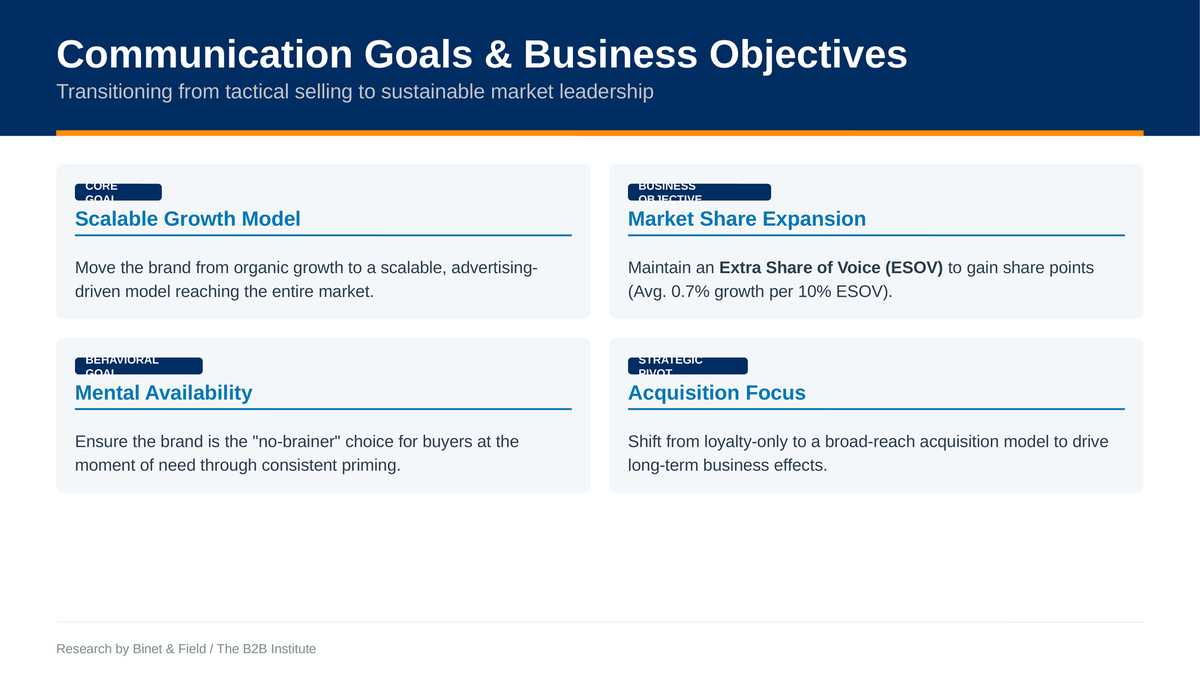}{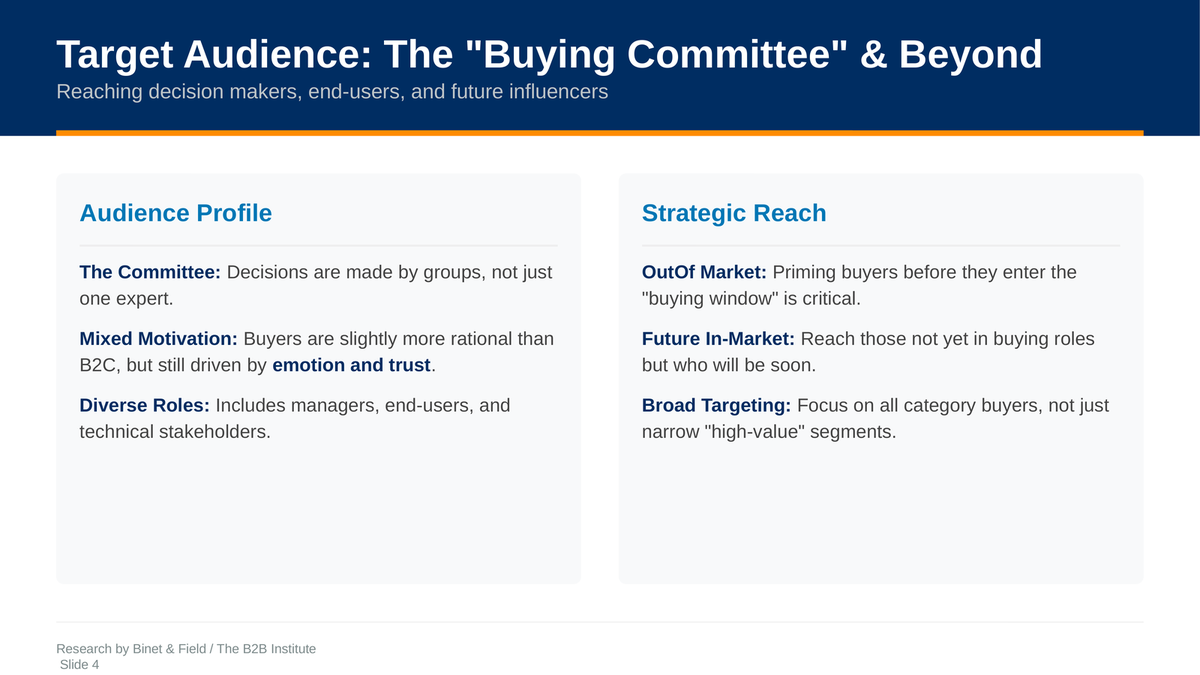}{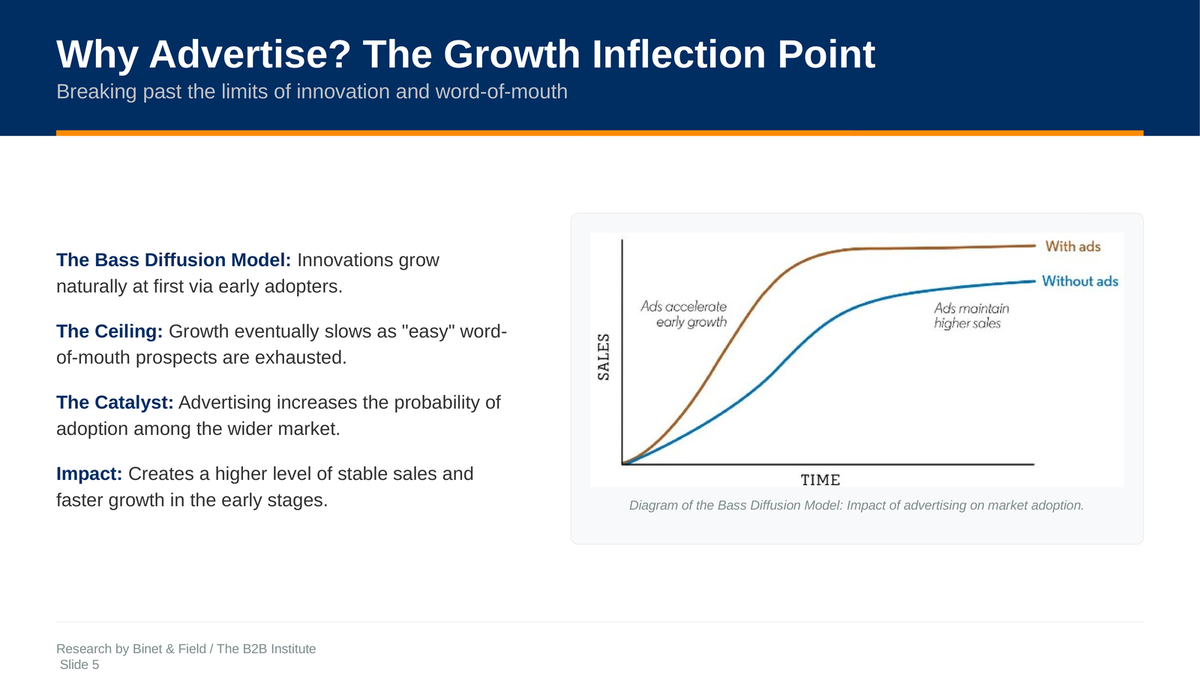}
\deckshowcaserow{SlideTailor / Specialist: AudCov. 0.484, Correct. 0.900, SafeEff 5.375}{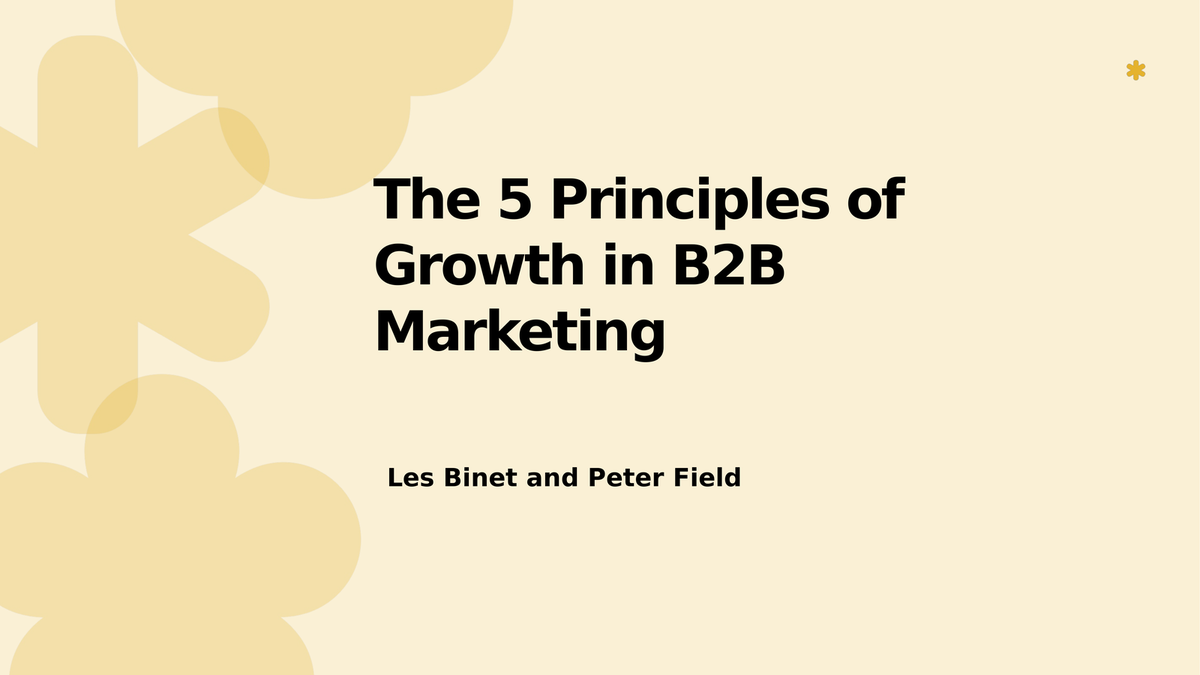}{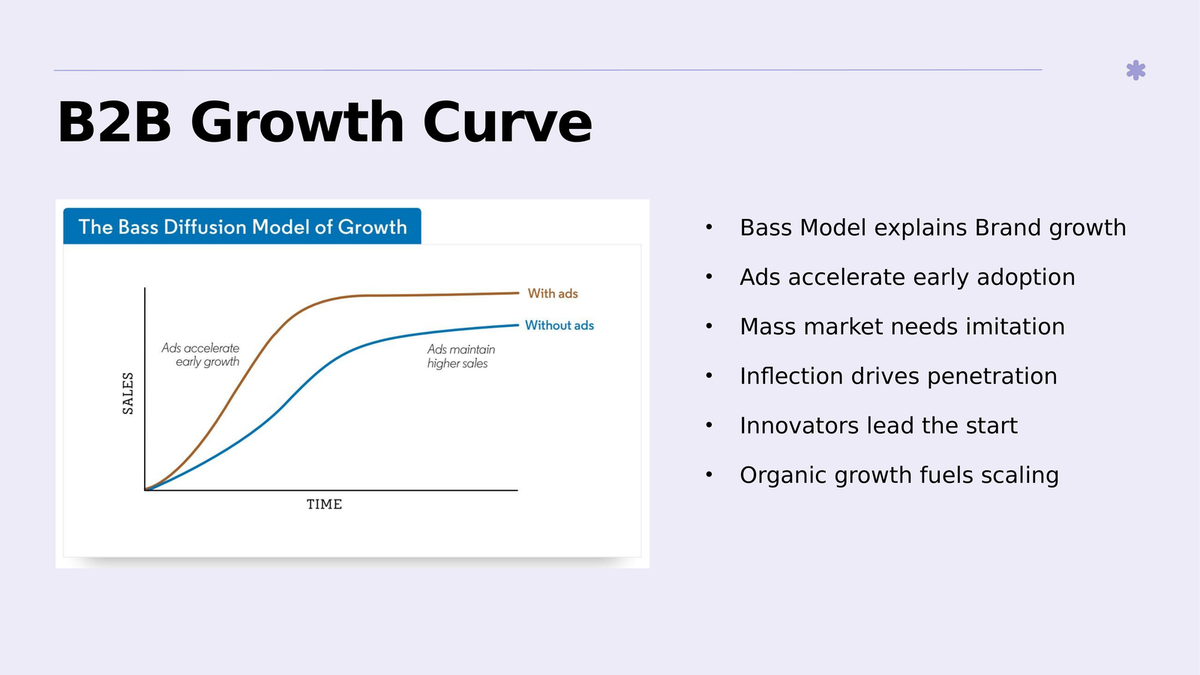}{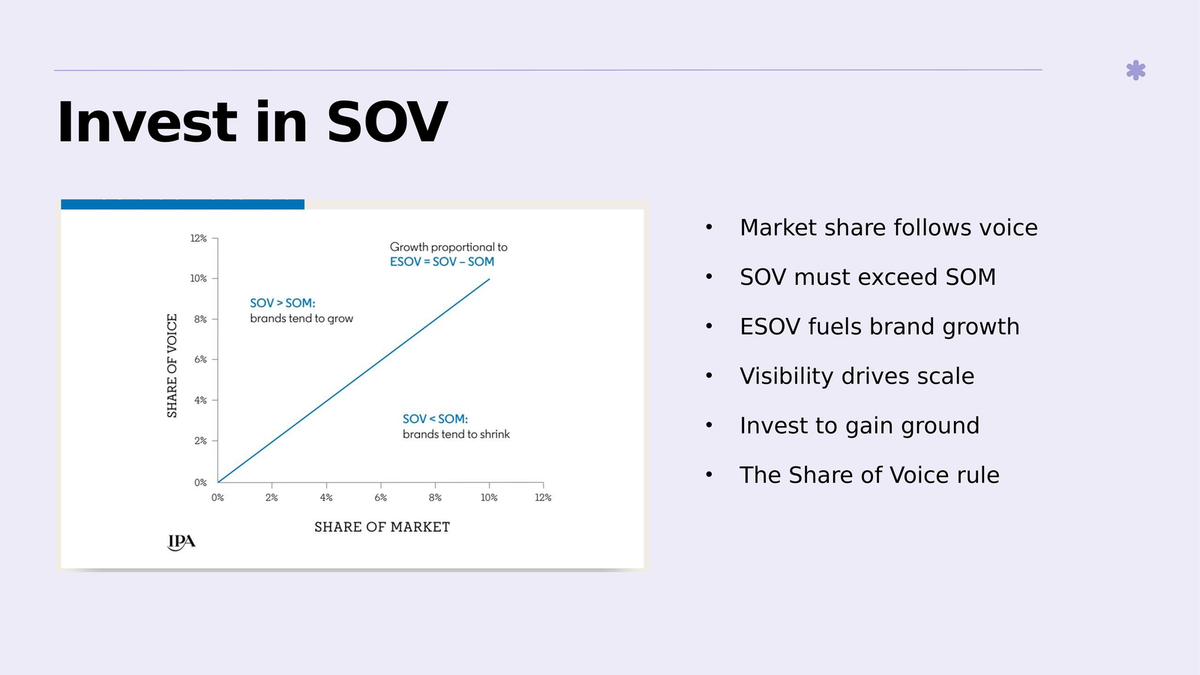}{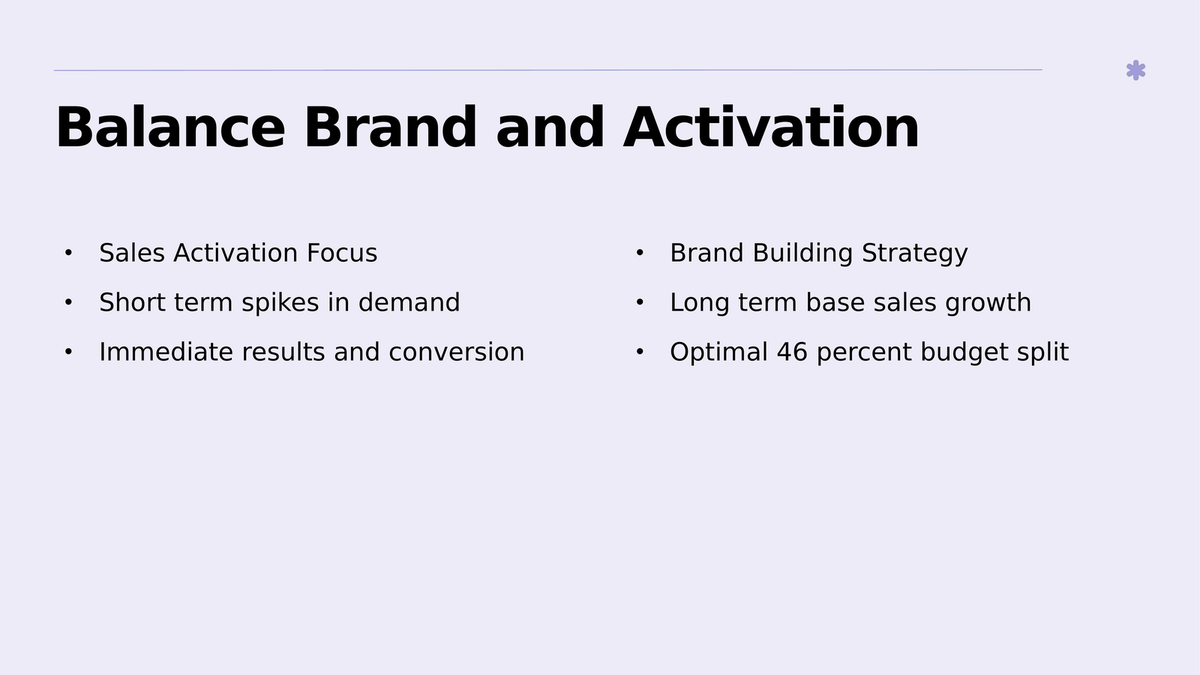}{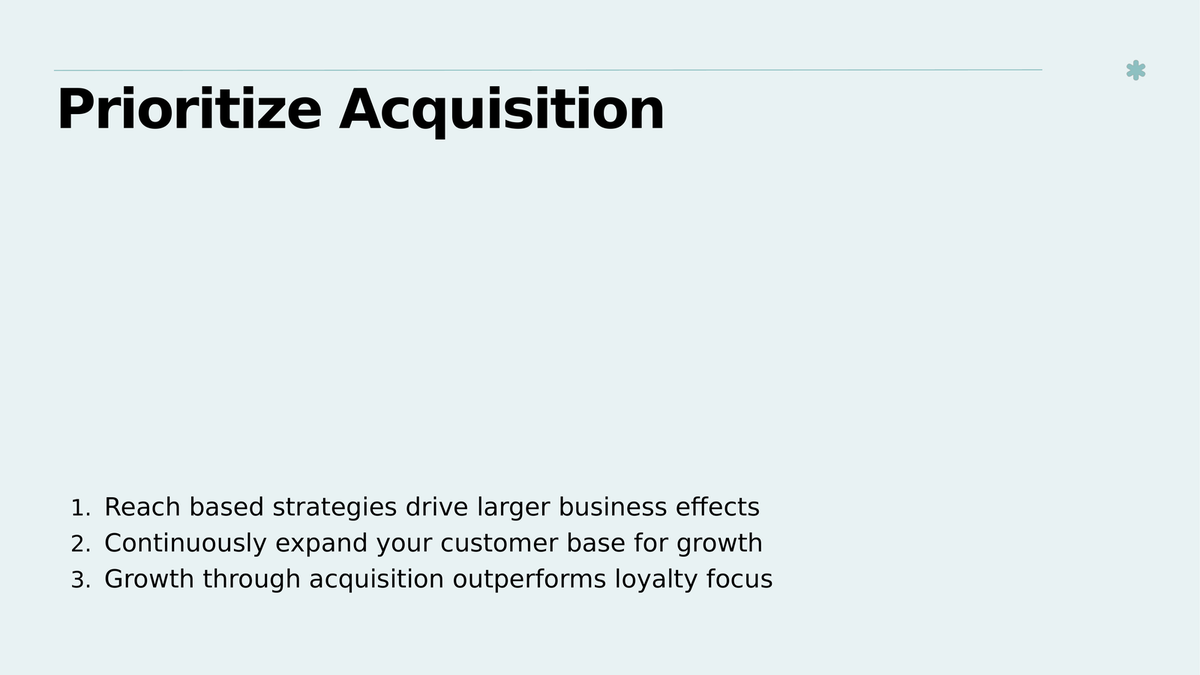}
\deckshowcaserow{SlideTailor / Learner: AudCov. 0.771, Correct. 0.899, SafeEff 9.790}{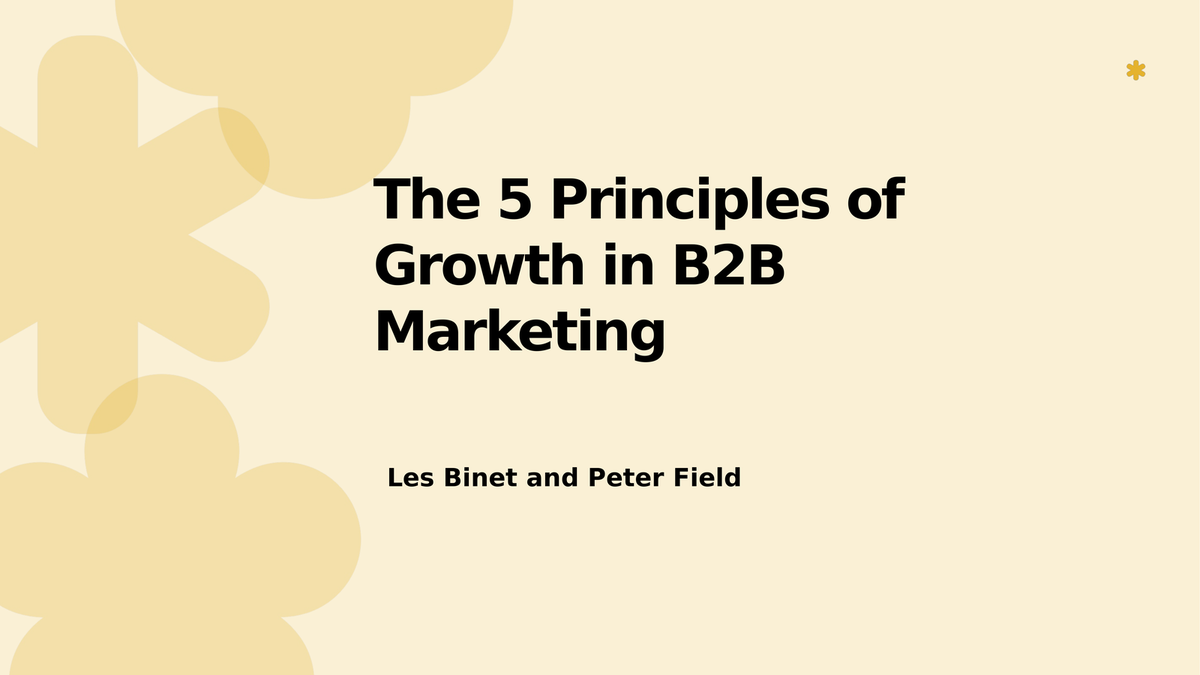}{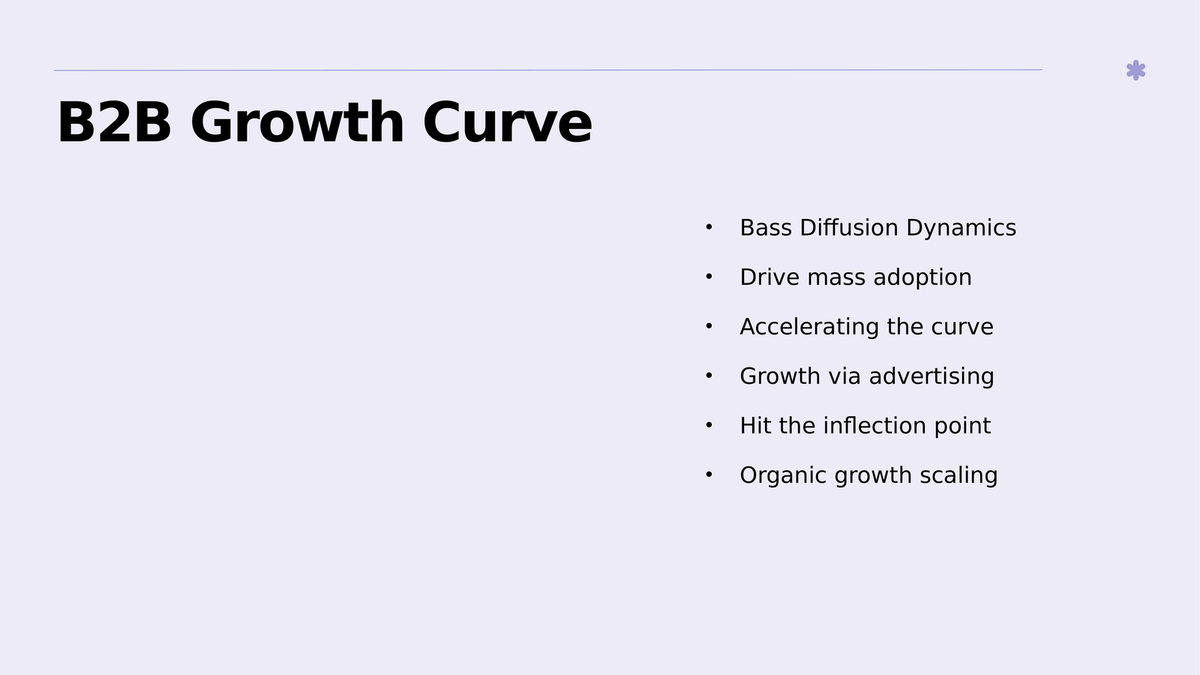}{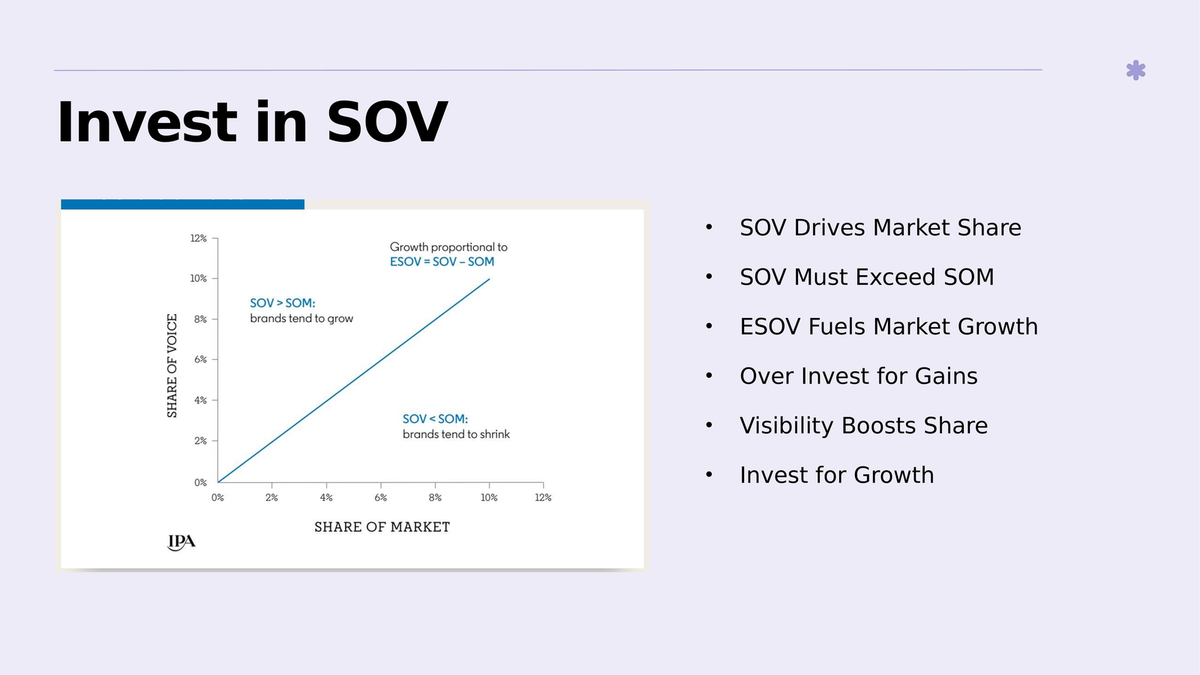}{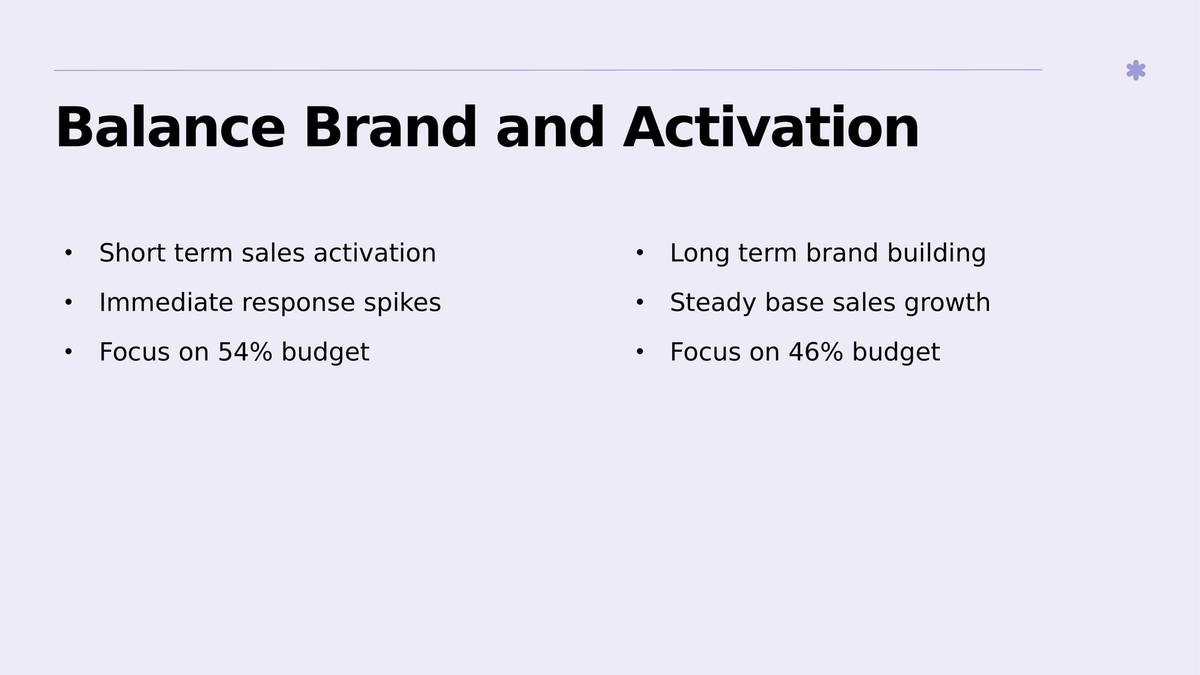}{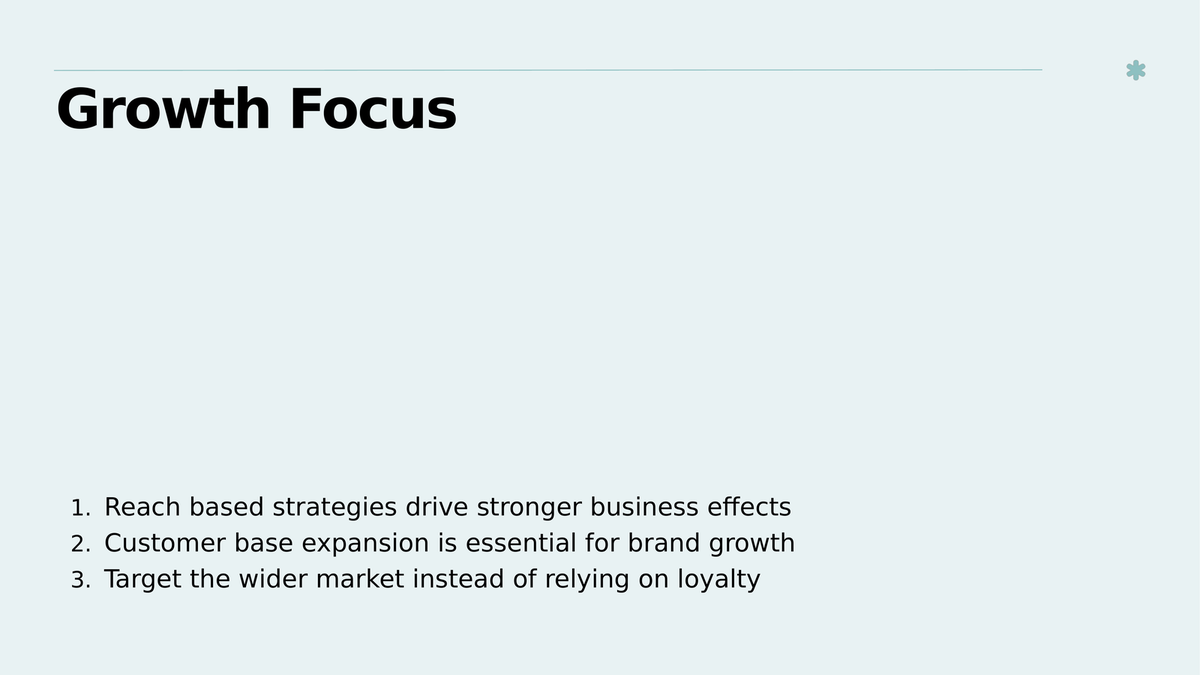}
\deckshowcaserow{SlideTailor / Decision maker: AudCov. 0.758, Correct. 0.857, SafeEff 7.840}{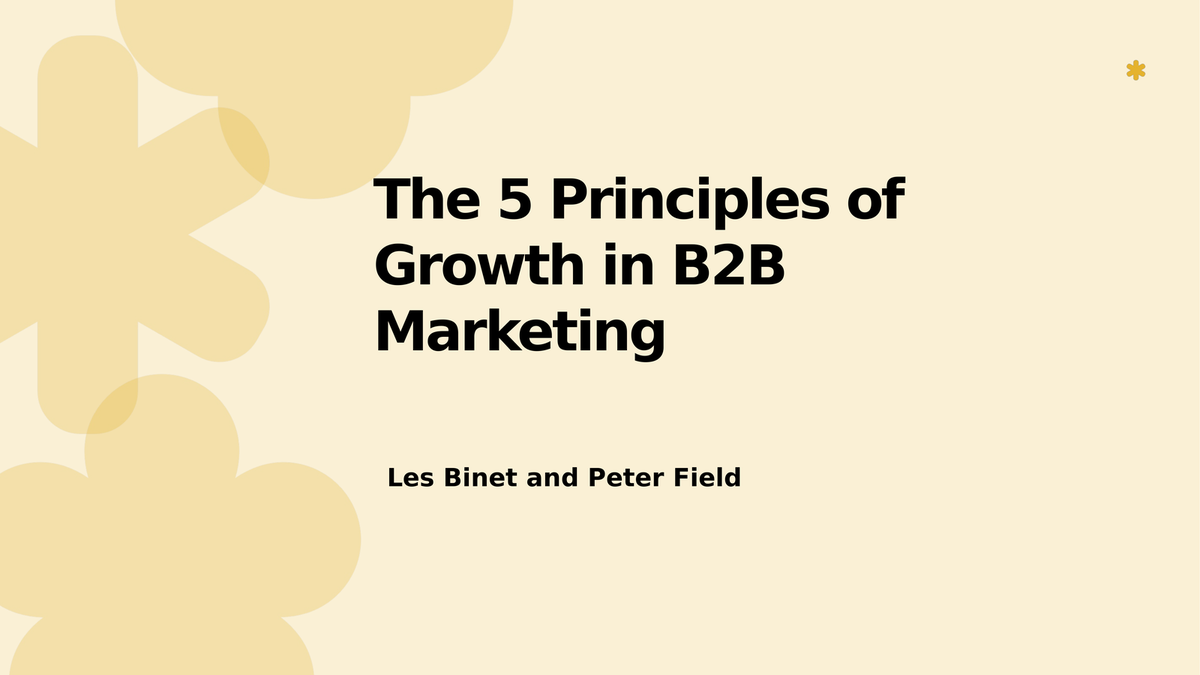}{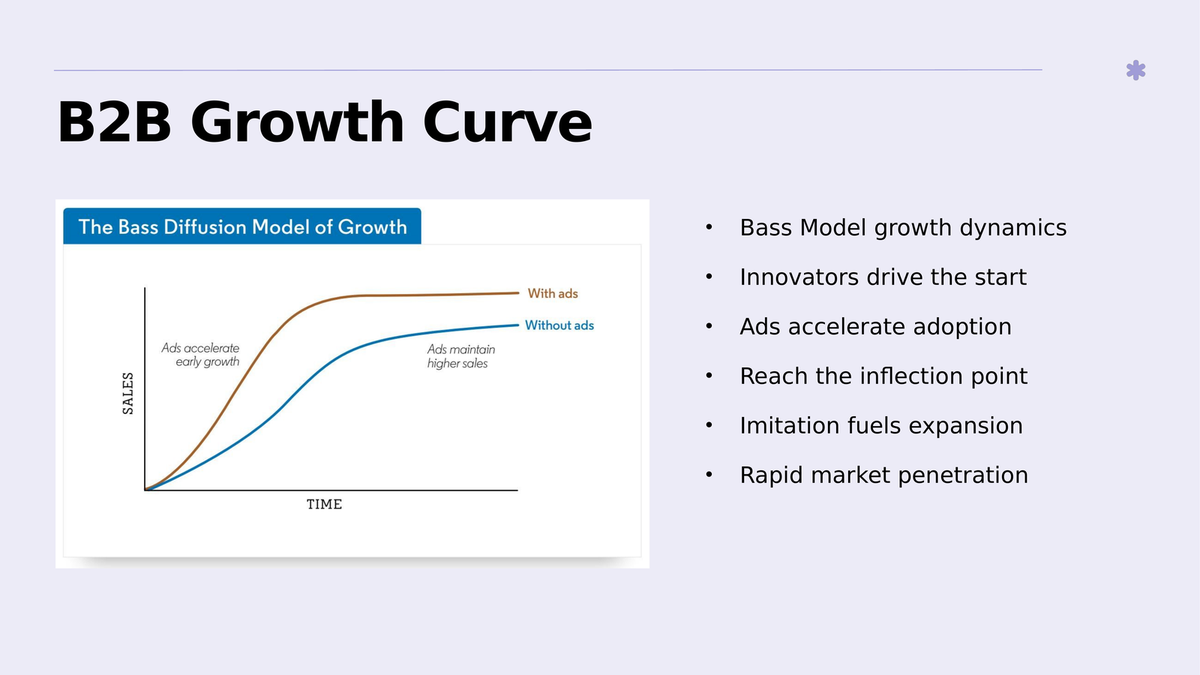}{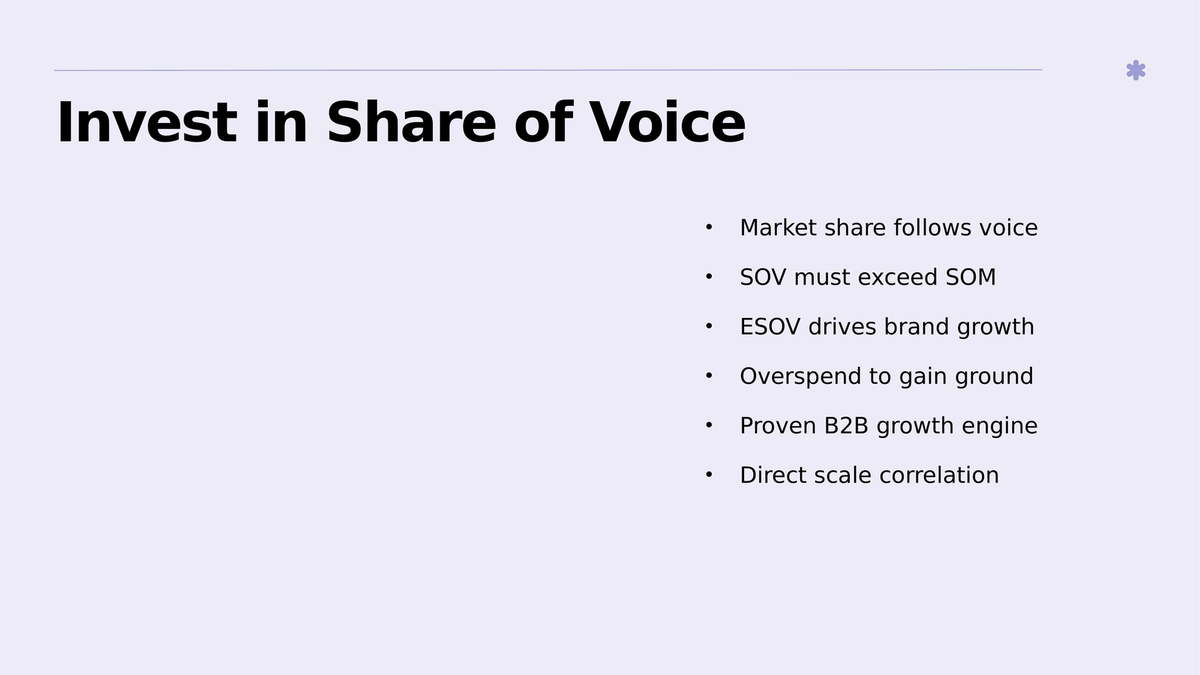}{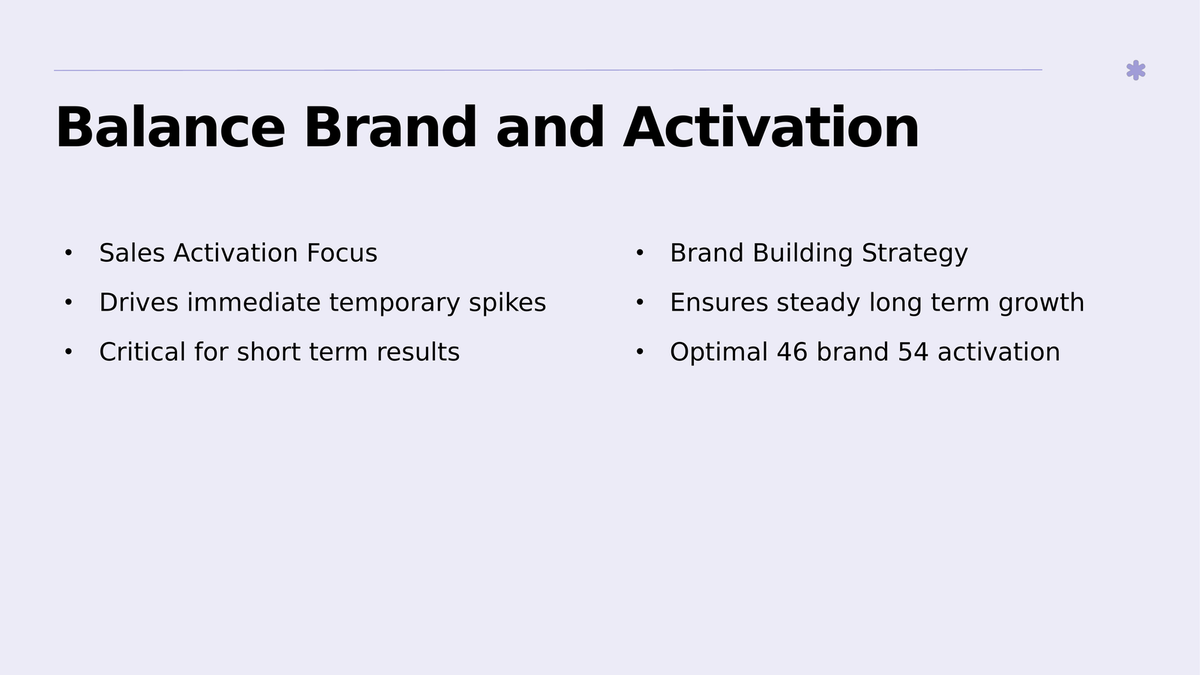}{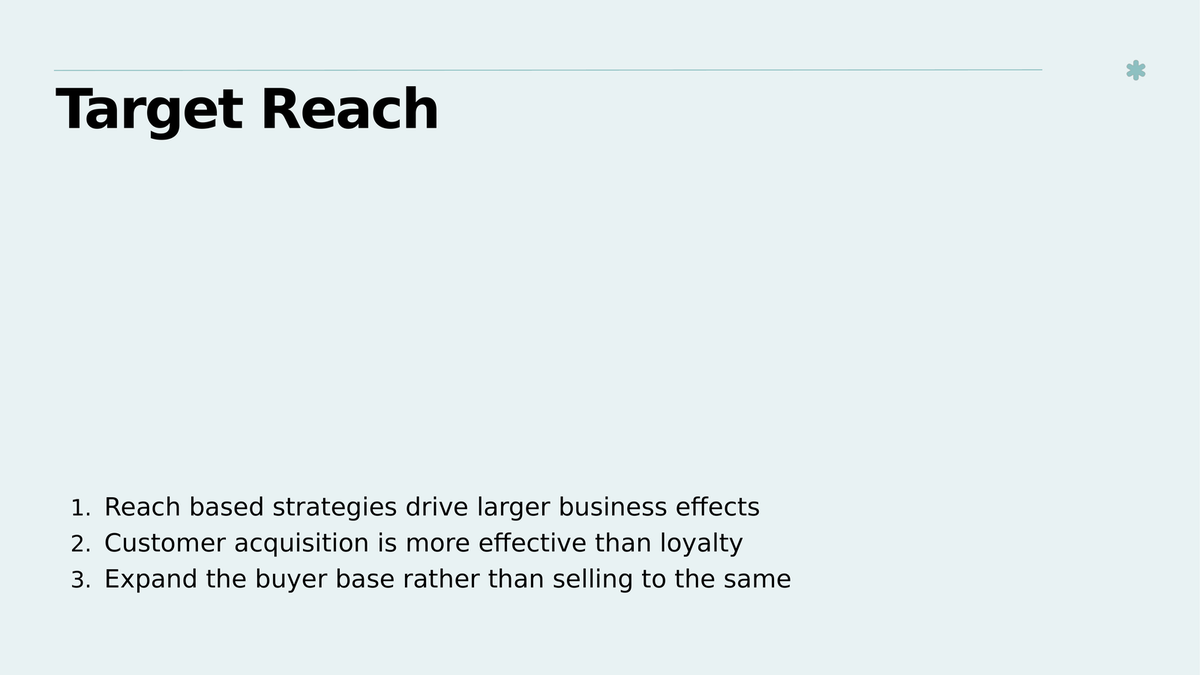}
}
\caption{Deck showcase for The 5 Principles of Growth in B2B Marketing. This full-profile example compares all three audiences for both DeepPresenter and SlideTailor. Metrics are computed for the matched audience.}
\label{fig:deck_showcase_case019}
\end{figure}

\begin{figure}[!p]
\centering
{\setlength{\tabcolsep}{0pt}%
\deckshowcaserow{DeepPresenter / Learner: AudCov. 0.673, Correct. 0.941, SafeEff 3.478}{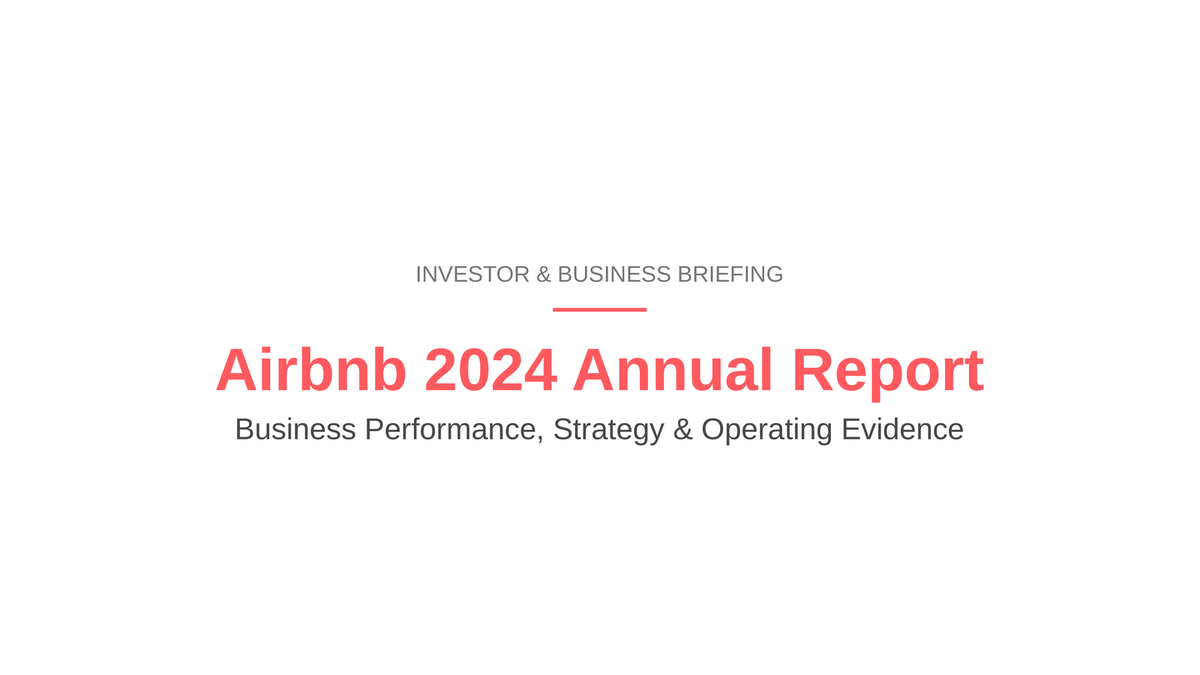}{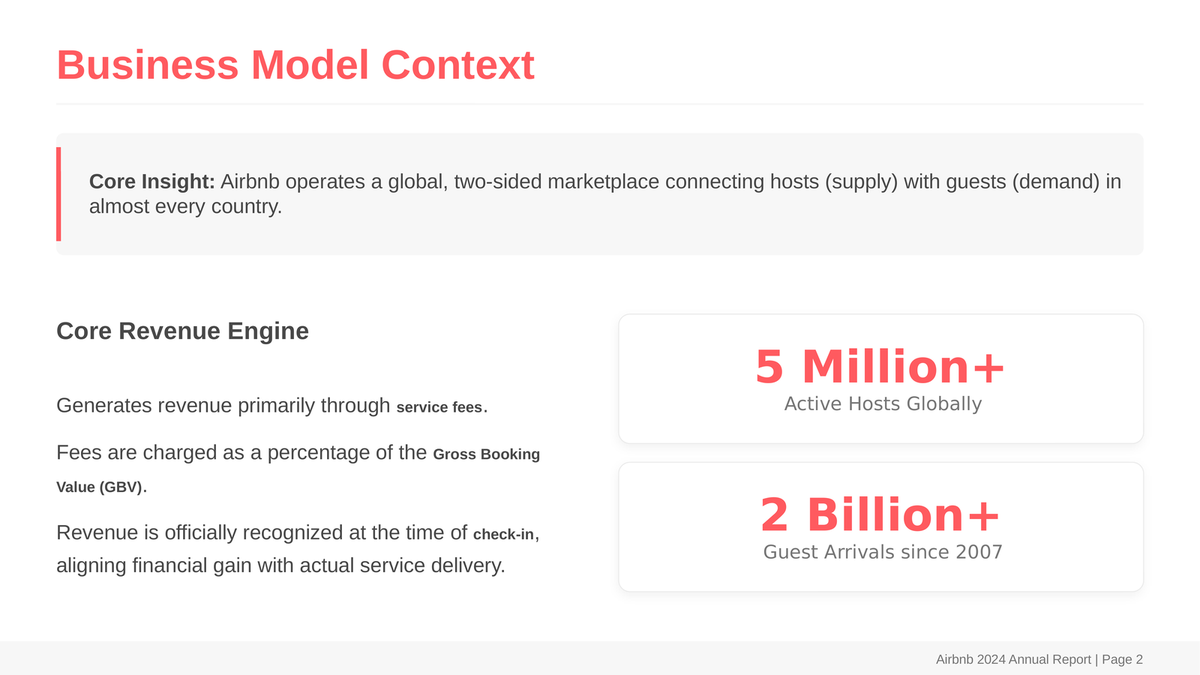}{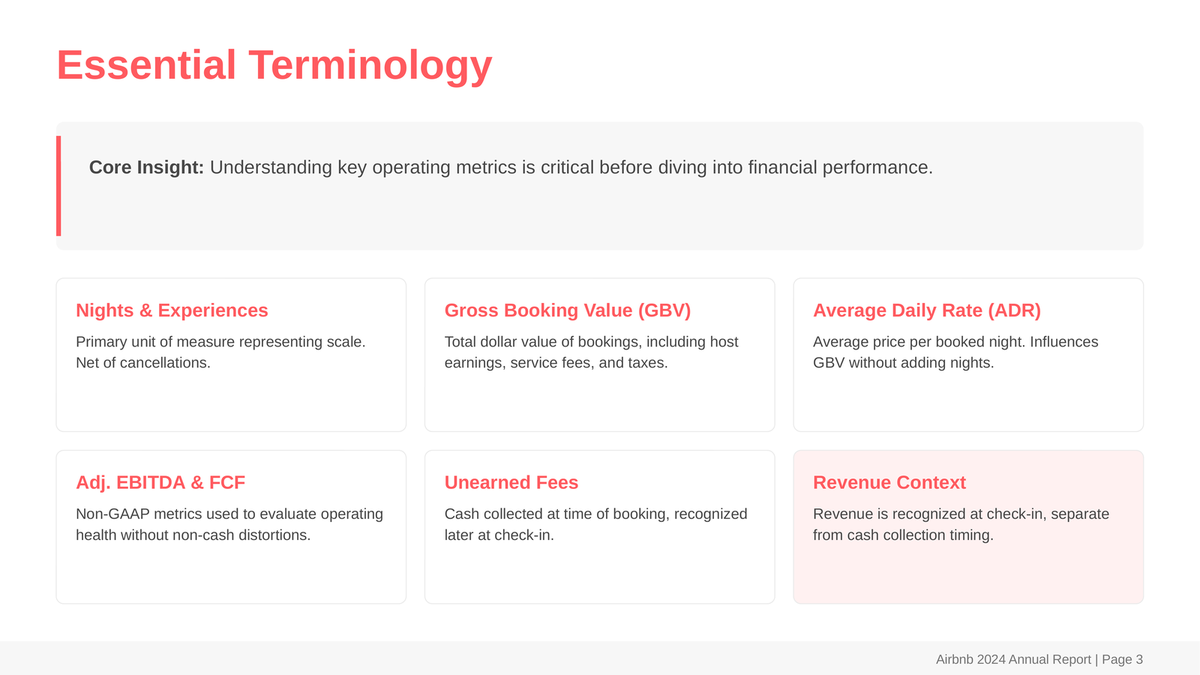}{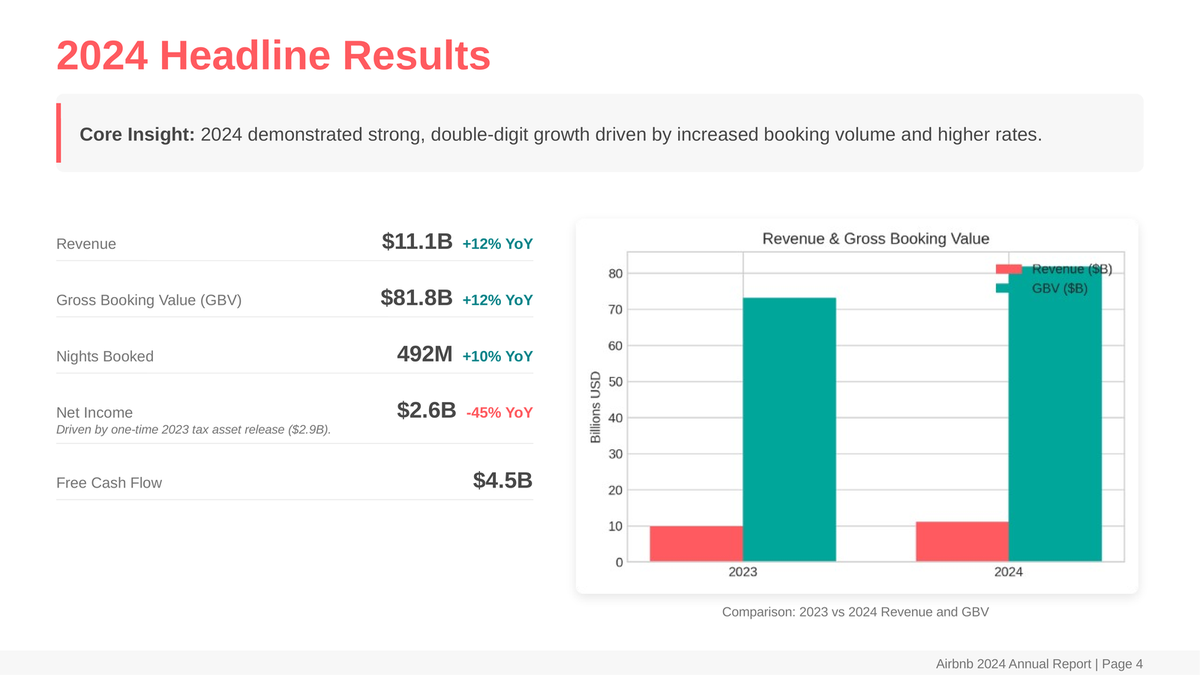}{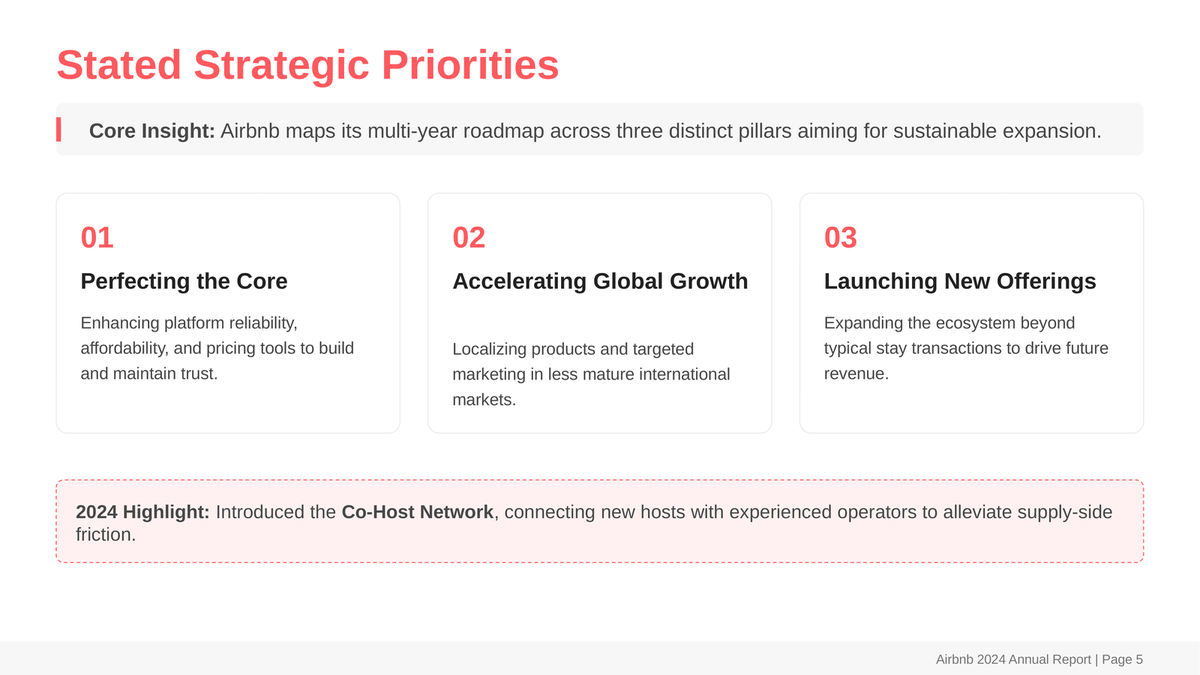}
\deckshowcaserow{DeepPresenter / Decision maker: AudCov. 0.692, Correct. 0.798, SafeEff 4.738}{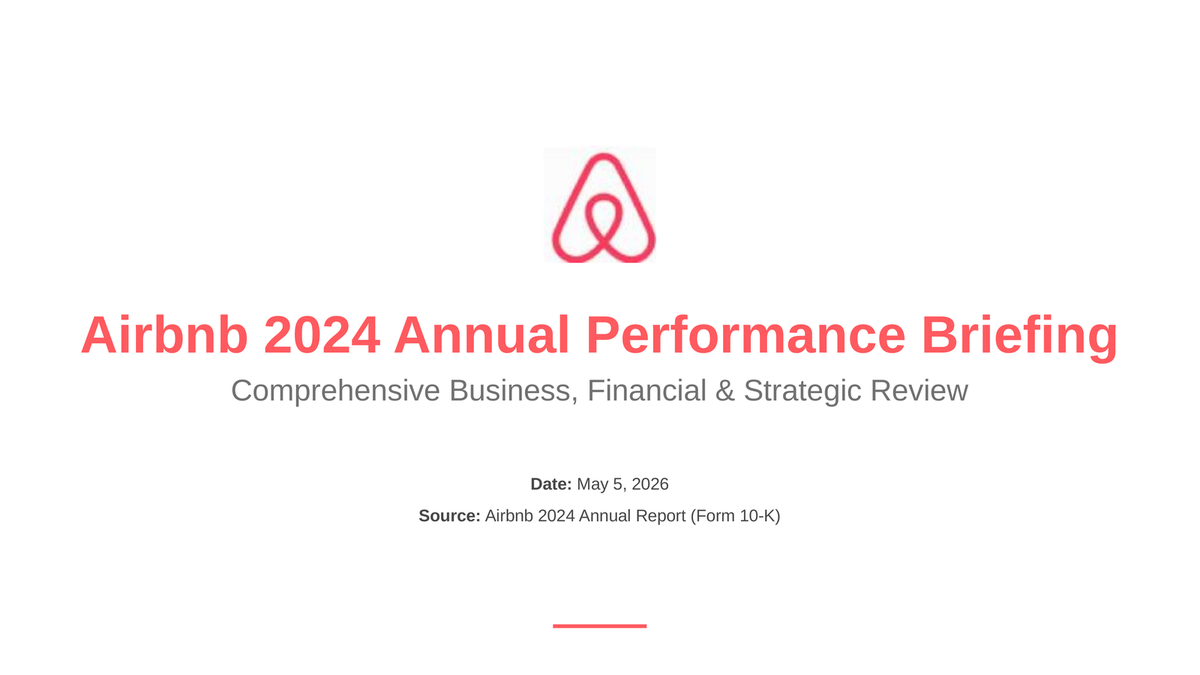}{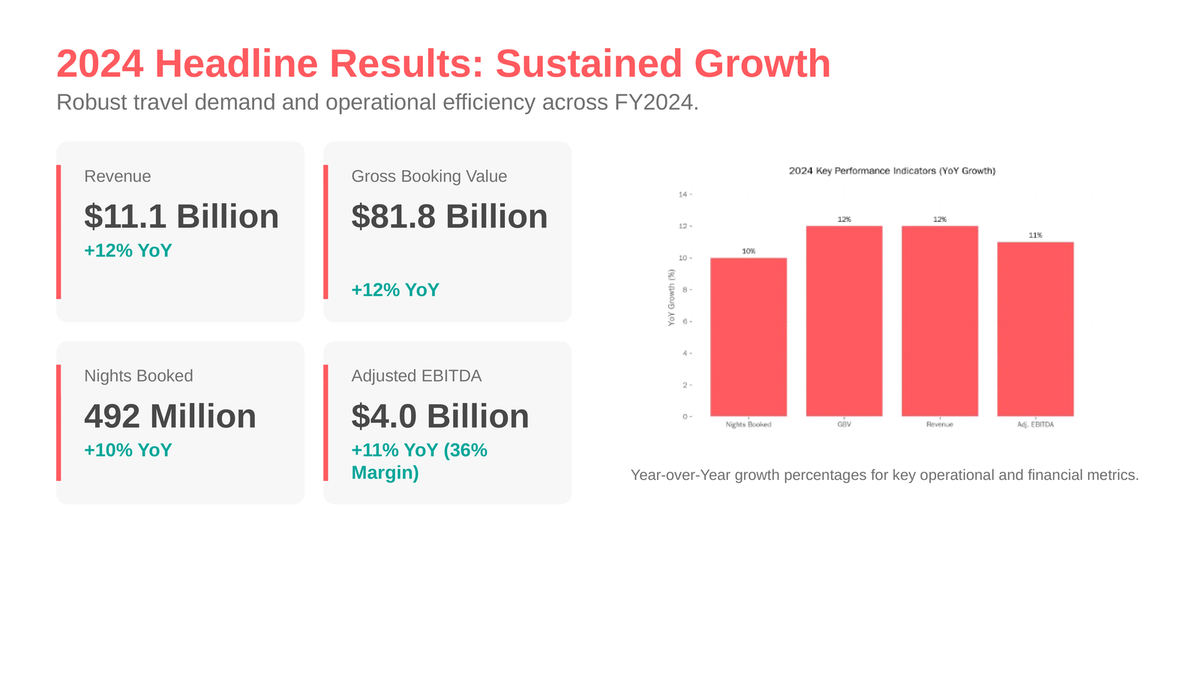}{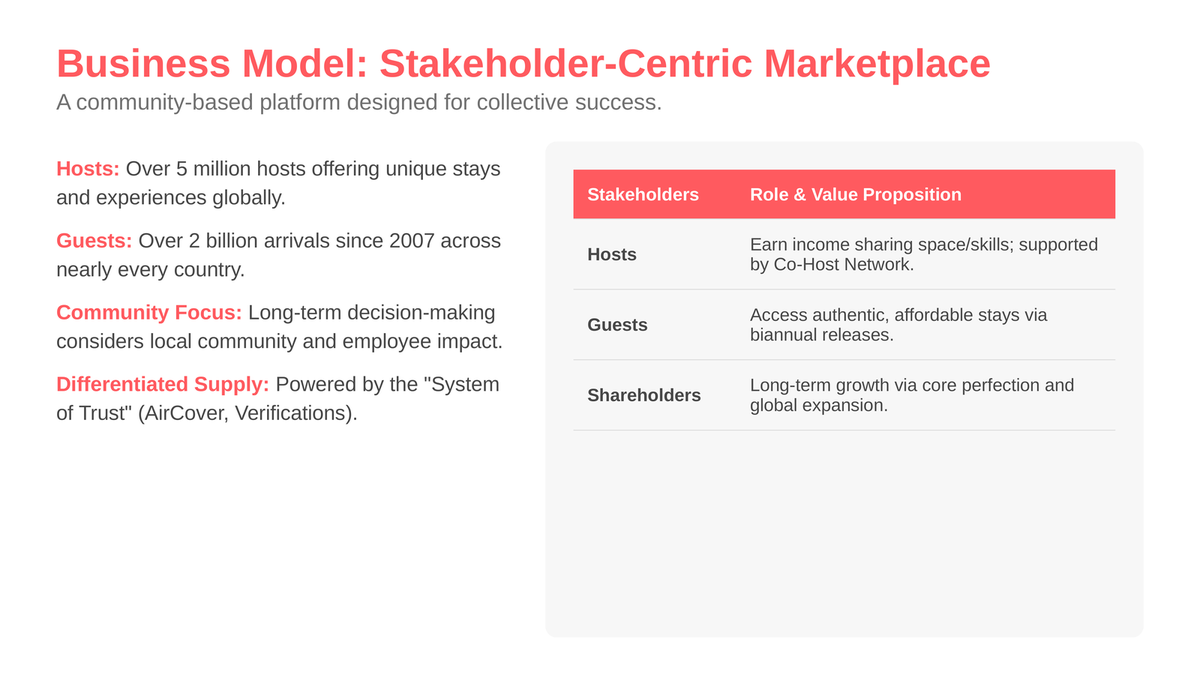}{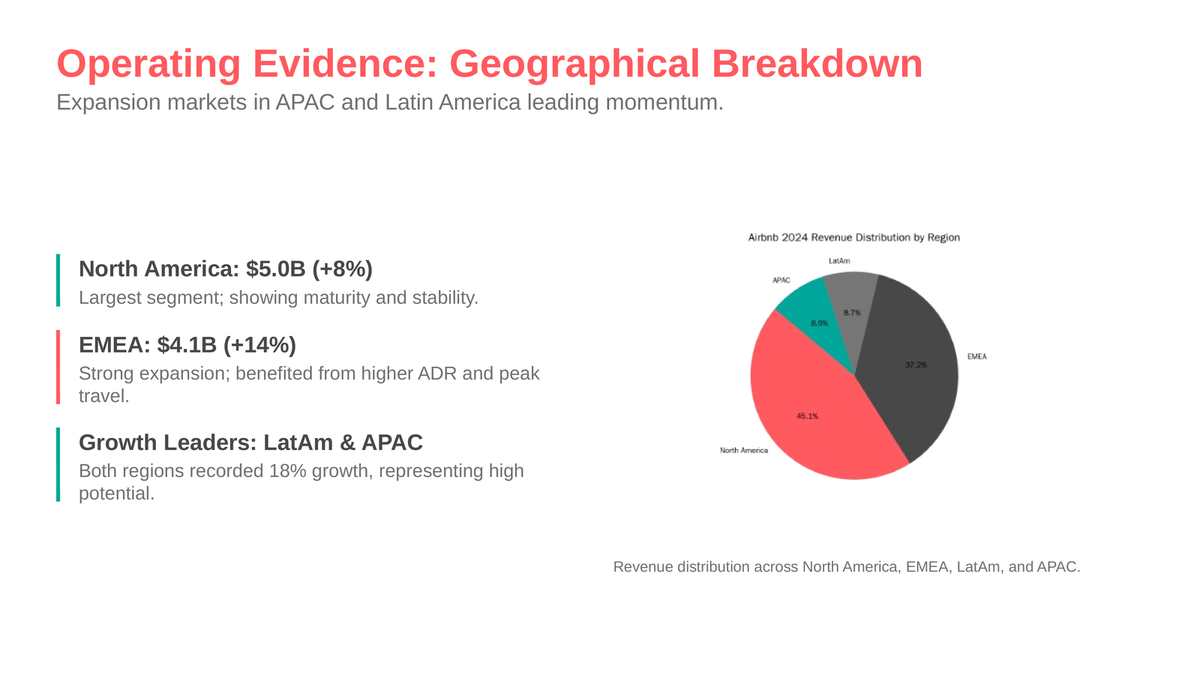}{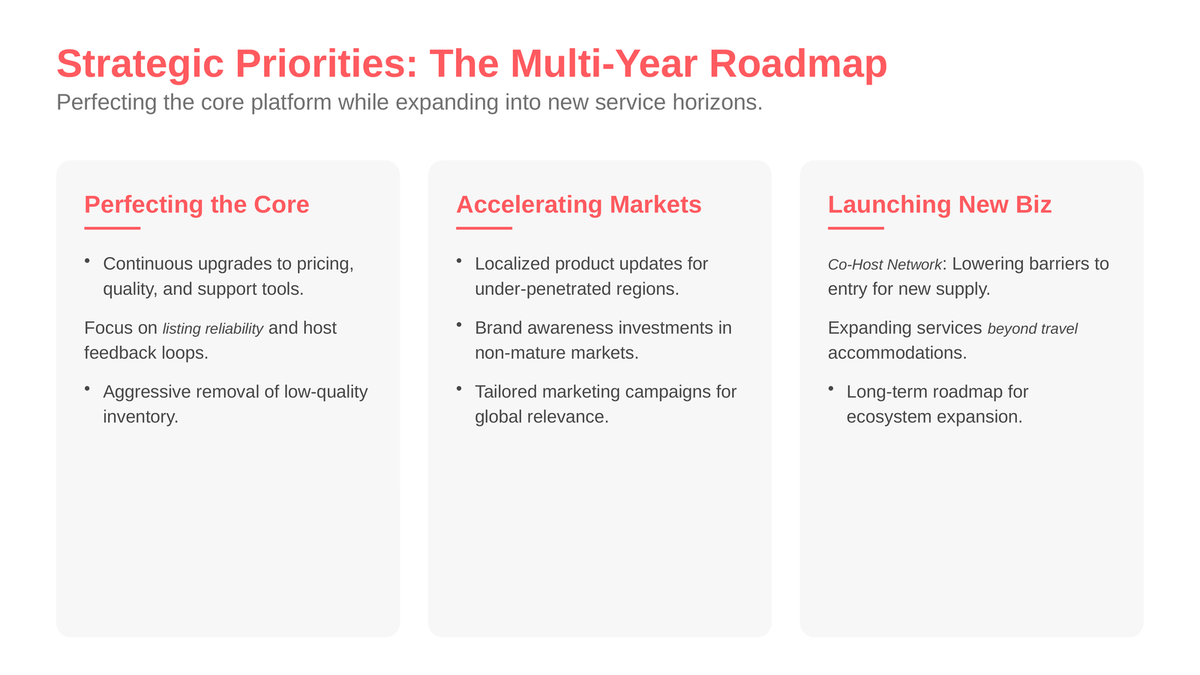}
\deckshowcaserow{SlideTailor / Learner: AudCov. 0.549, Correct. 0.611, SafeEff 2.901}{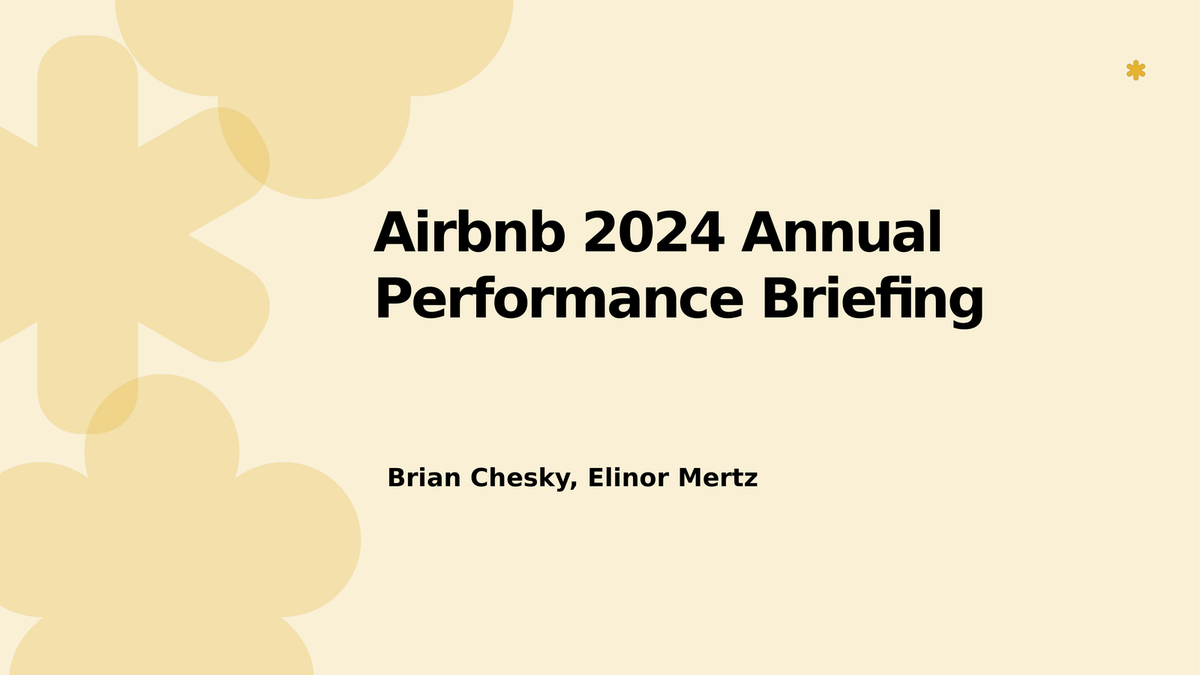}{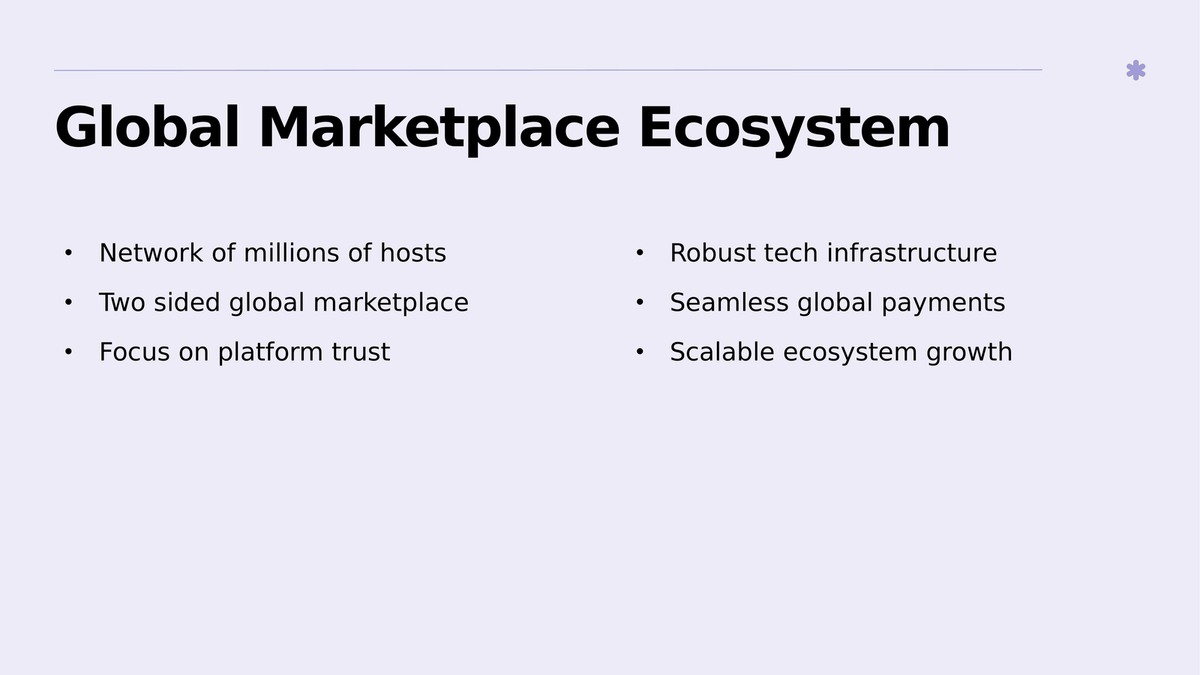}{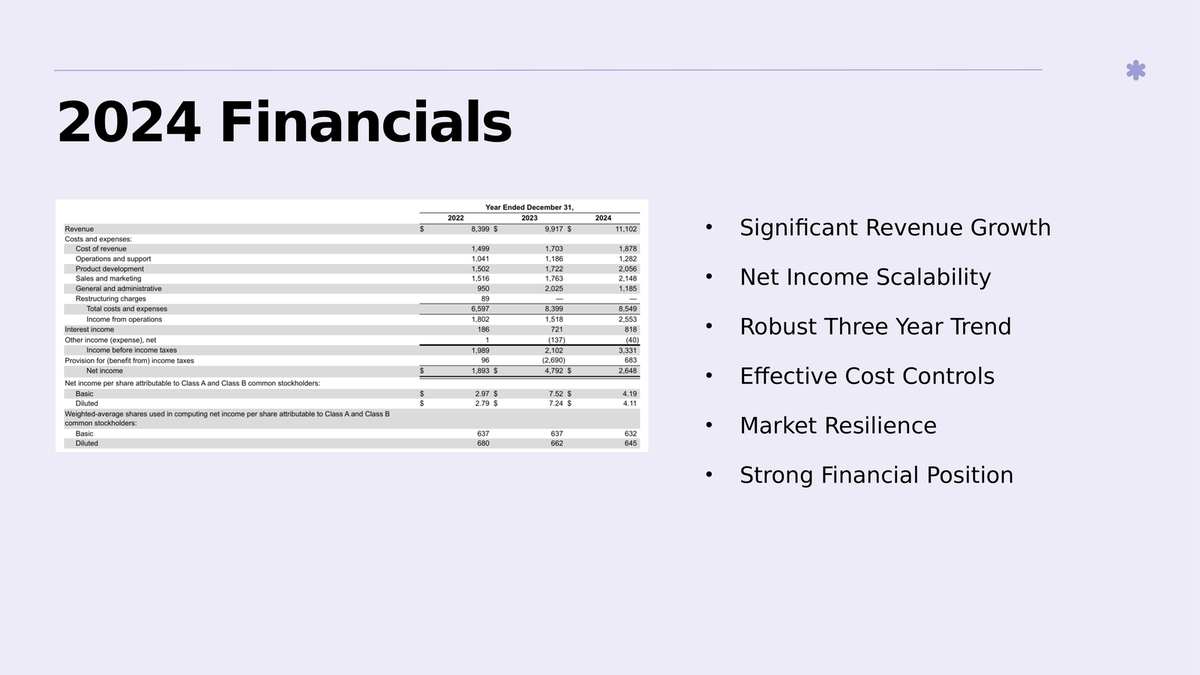}{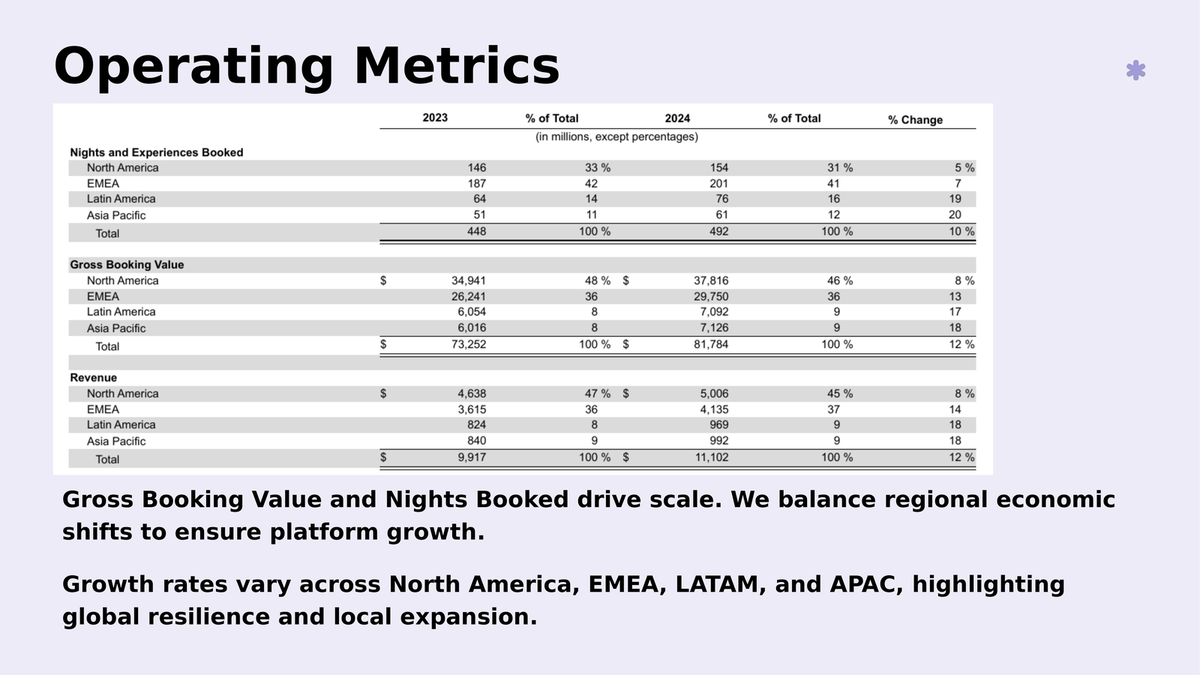}{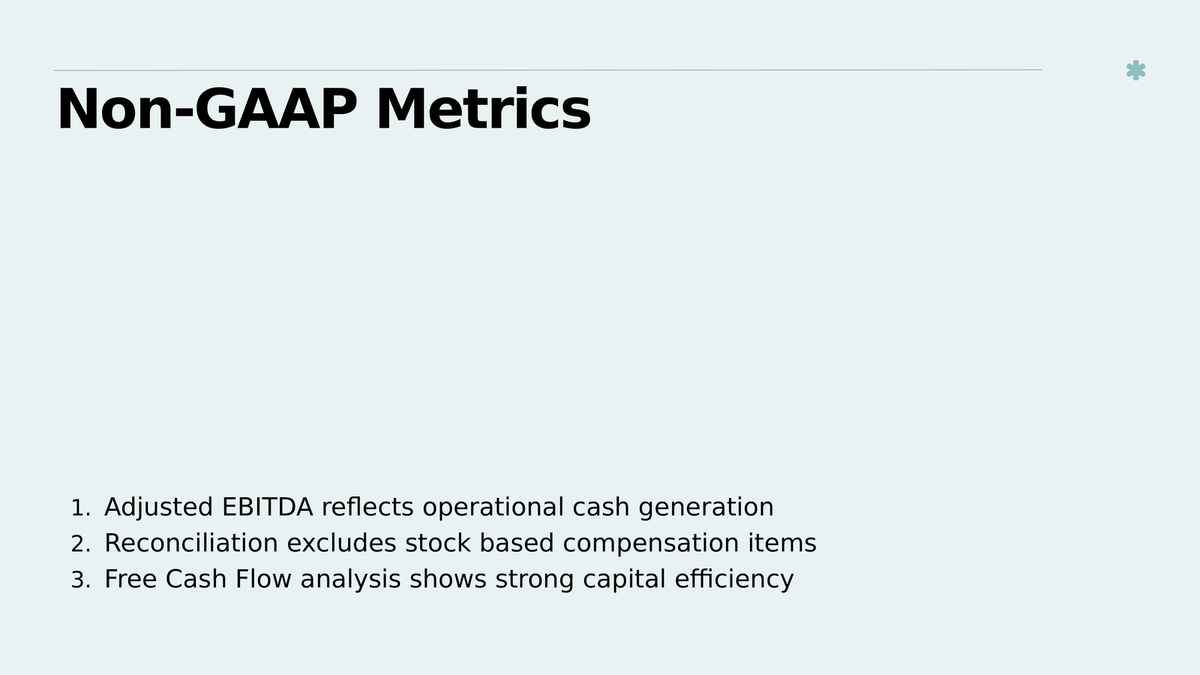}
\deckshowcaserow{NotebookLM / Specialist: AudCov. 0.685, Correct. 0.816, SafeEff 15.639}{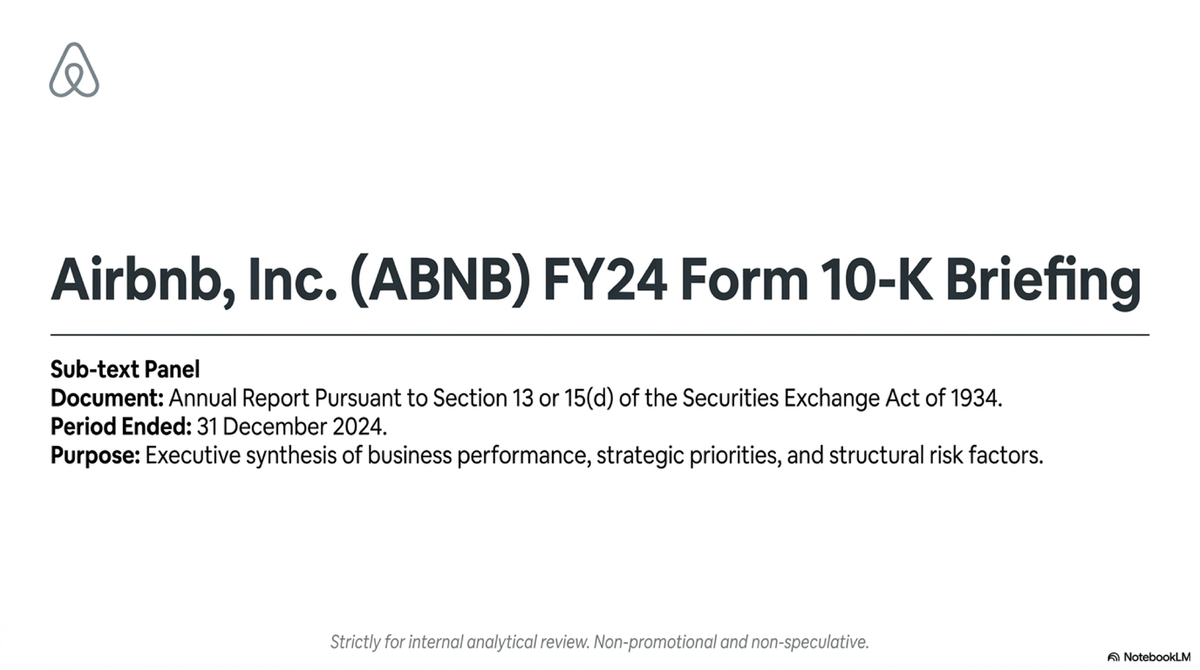}{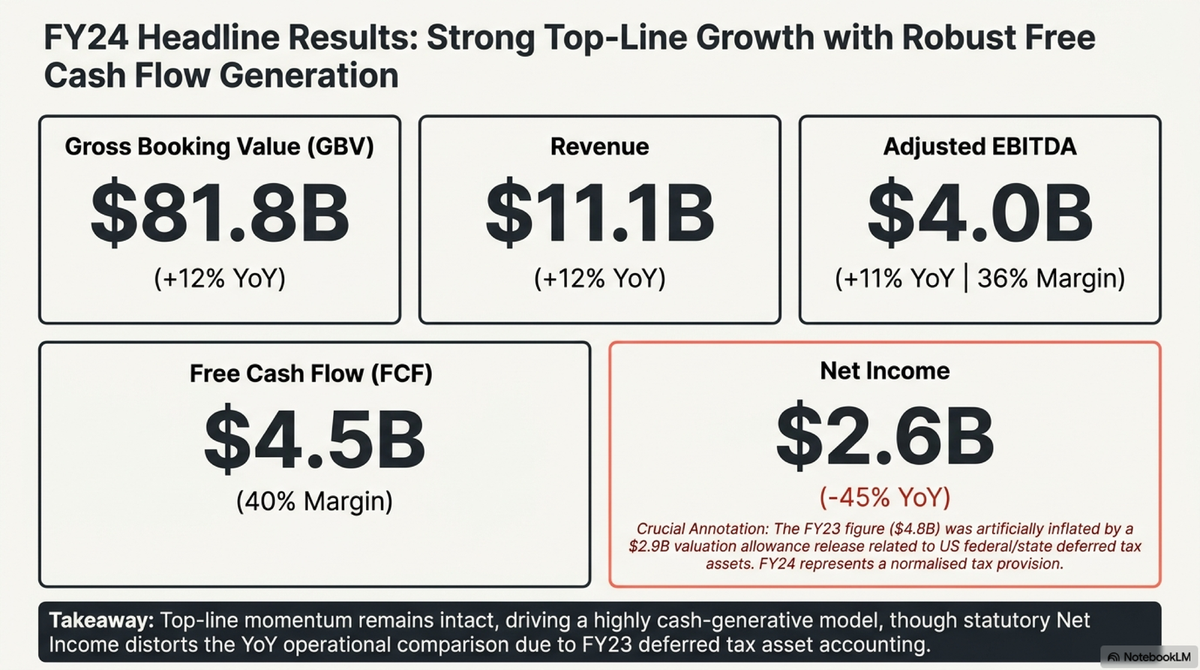}{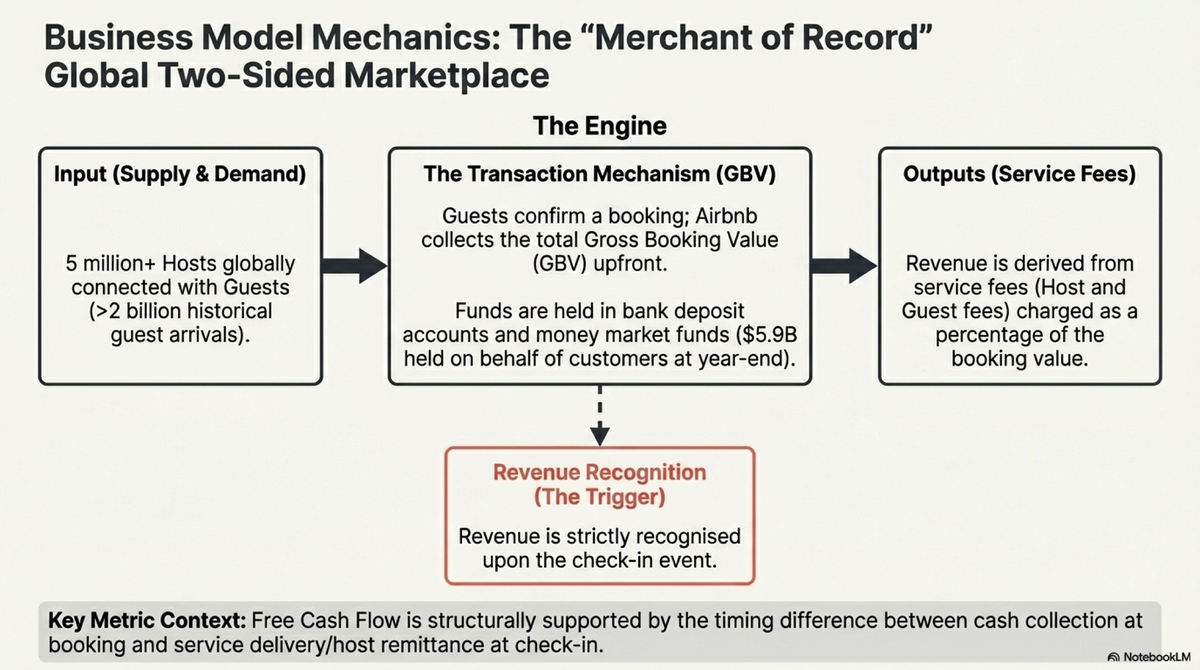}{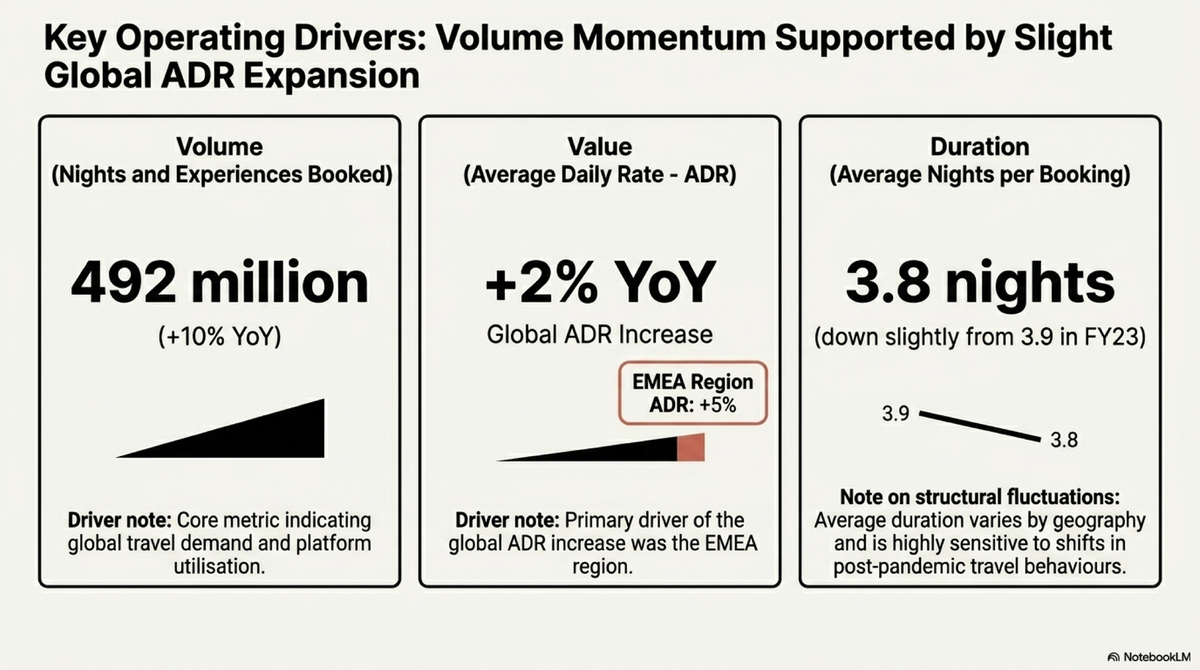}{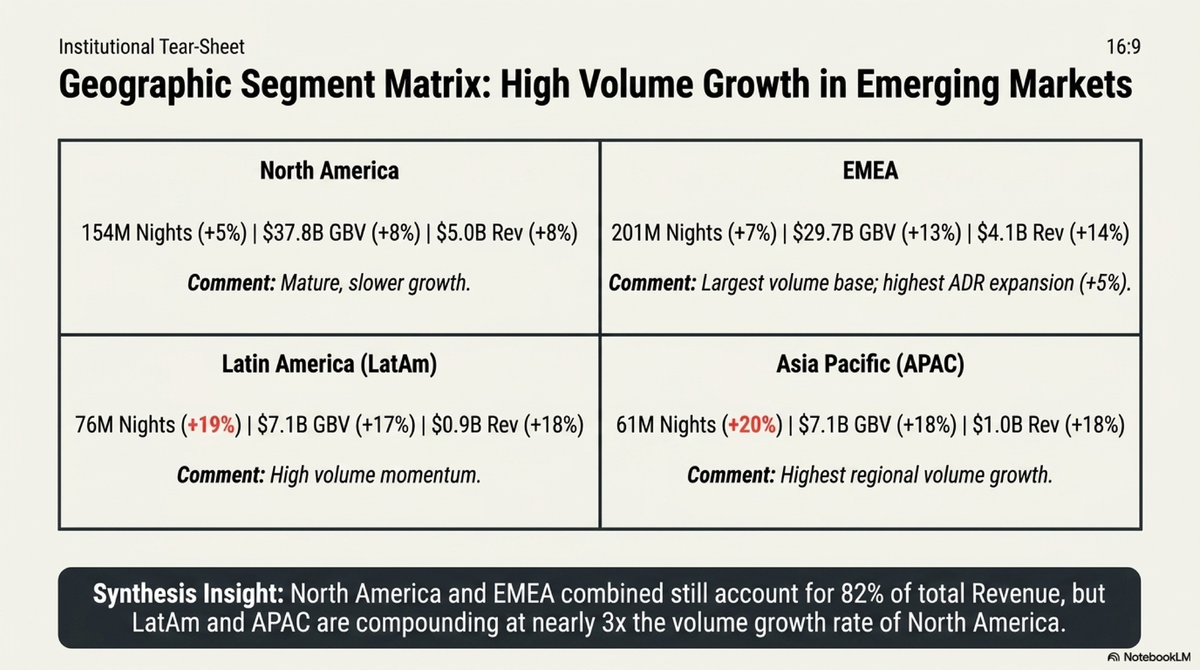}
\deckshowcaserow{NotebookLM / Learner: AudCov. 0.716, Correct. 0.921, SafeEff 8.143}{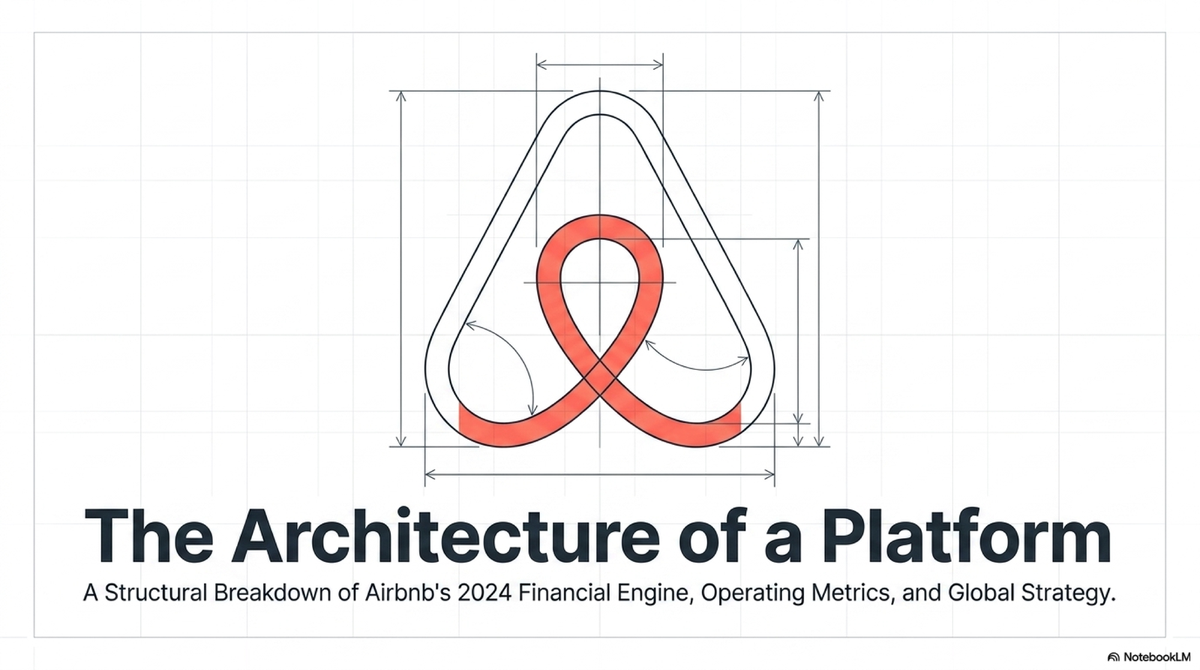}{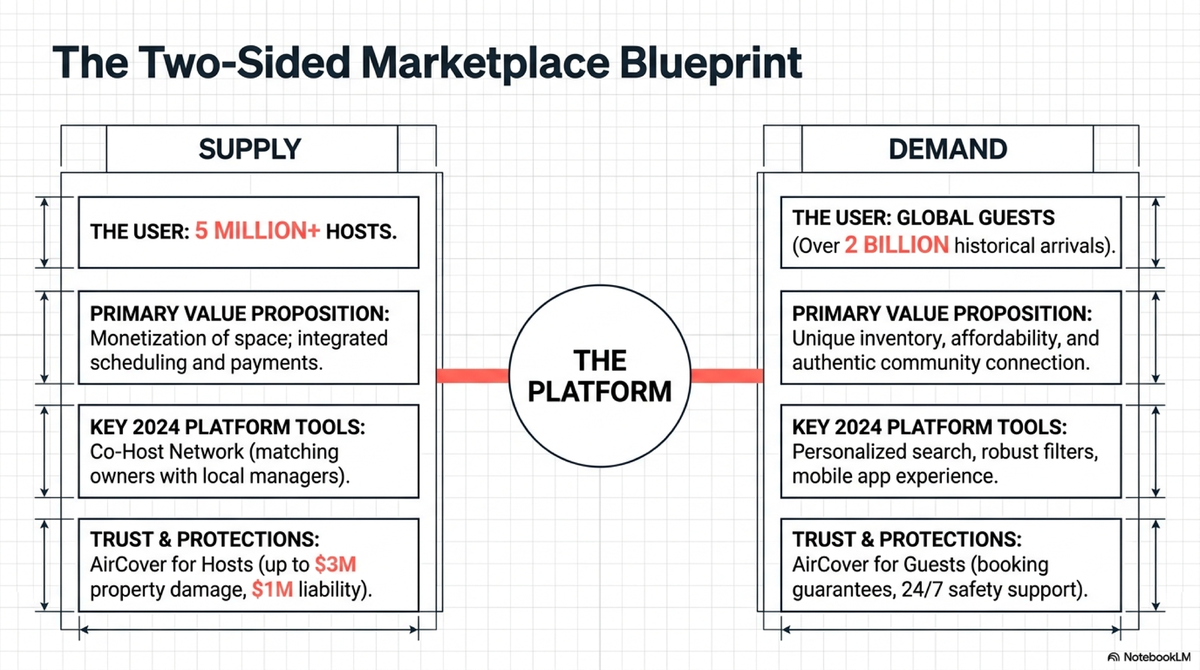}{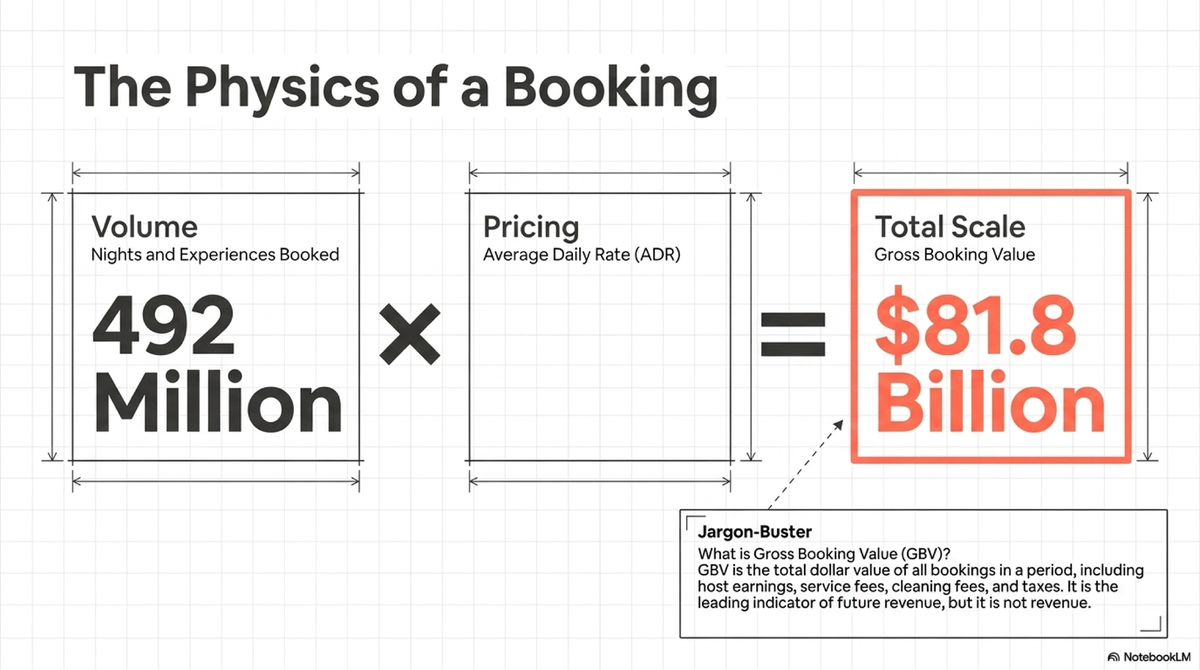}{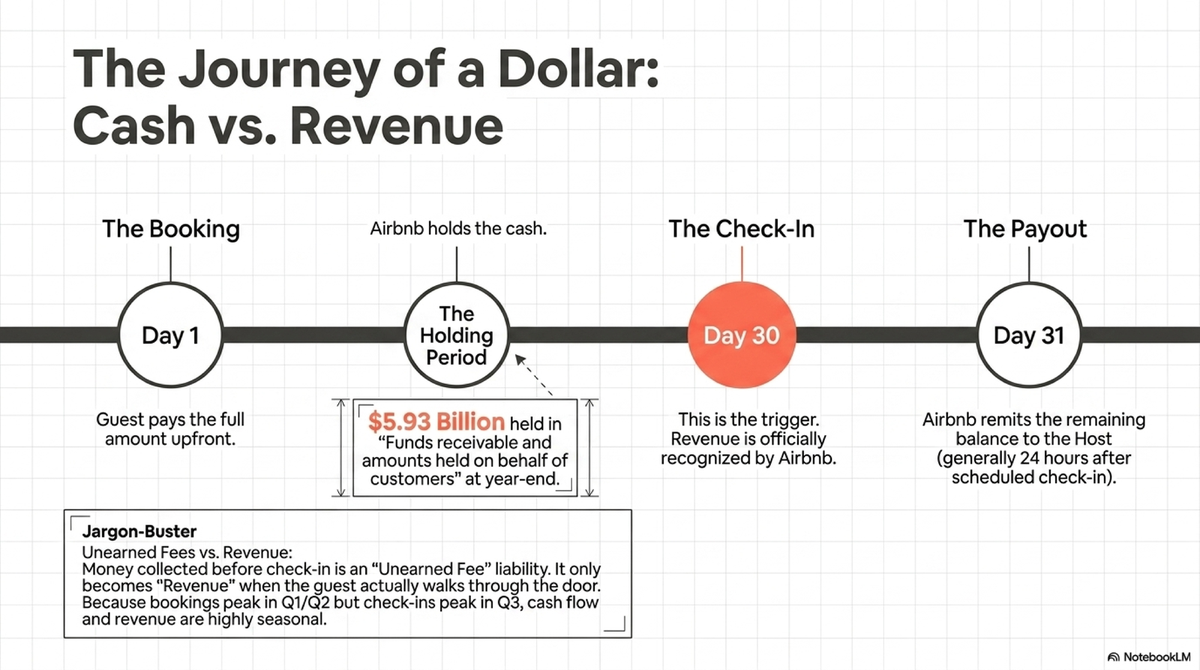}{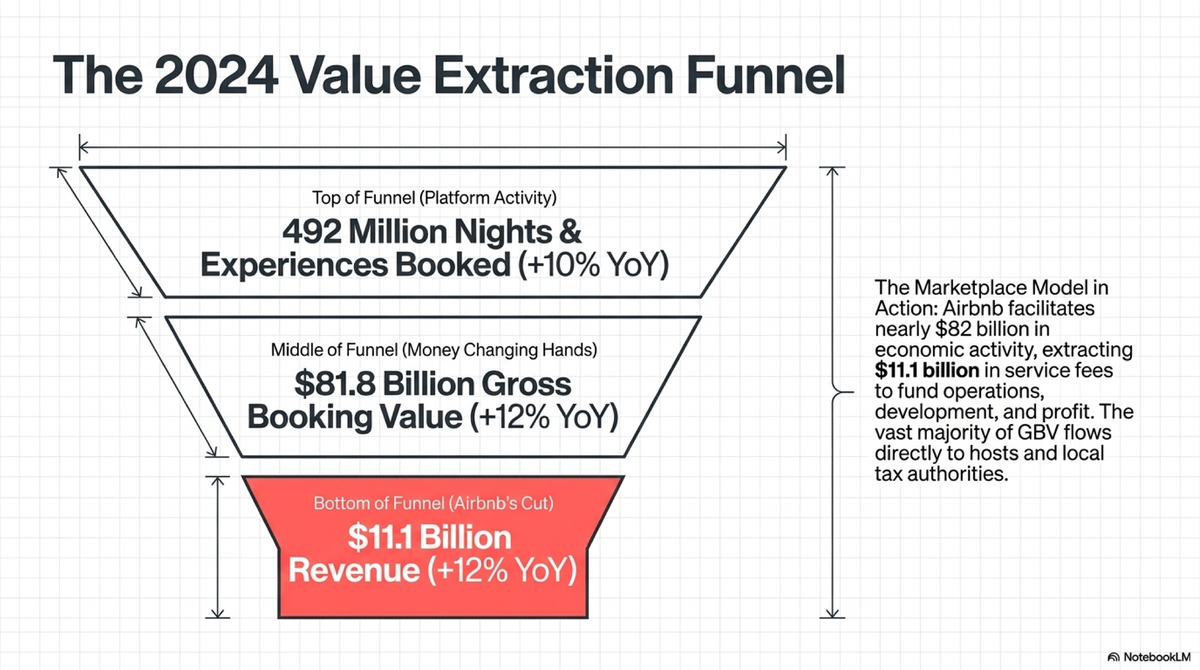}
\deckshowcaserow{NotebookLM / Decision maker: AudCov. 0.669, Correct. 0.875, SafeEff 16.147}{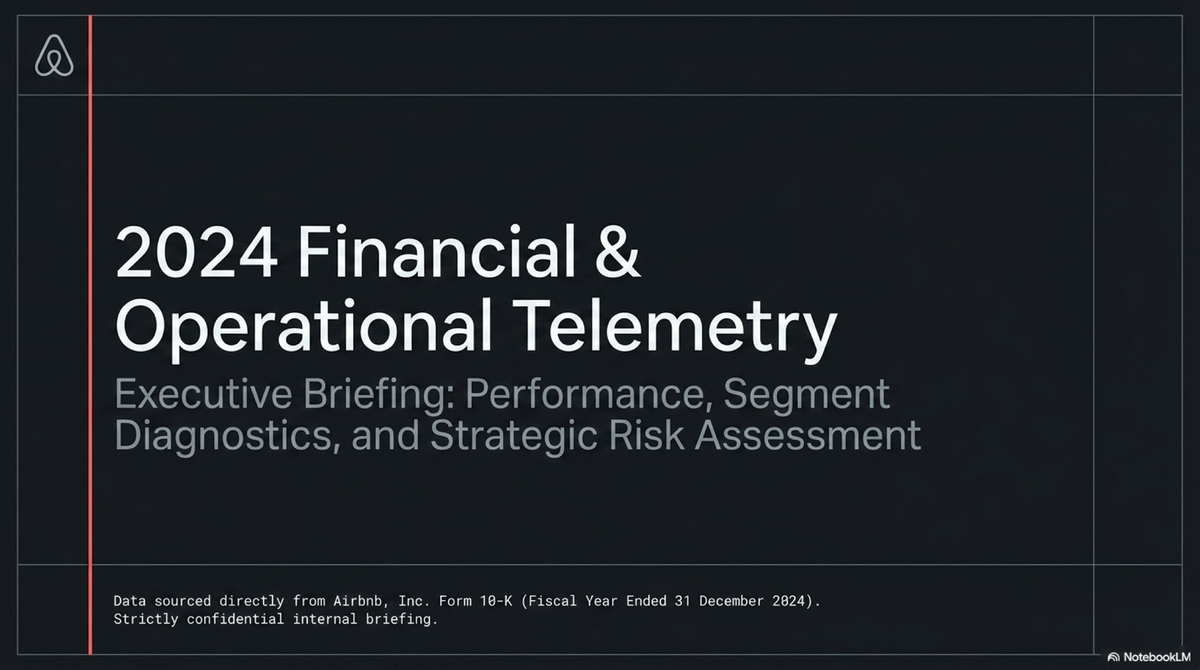}{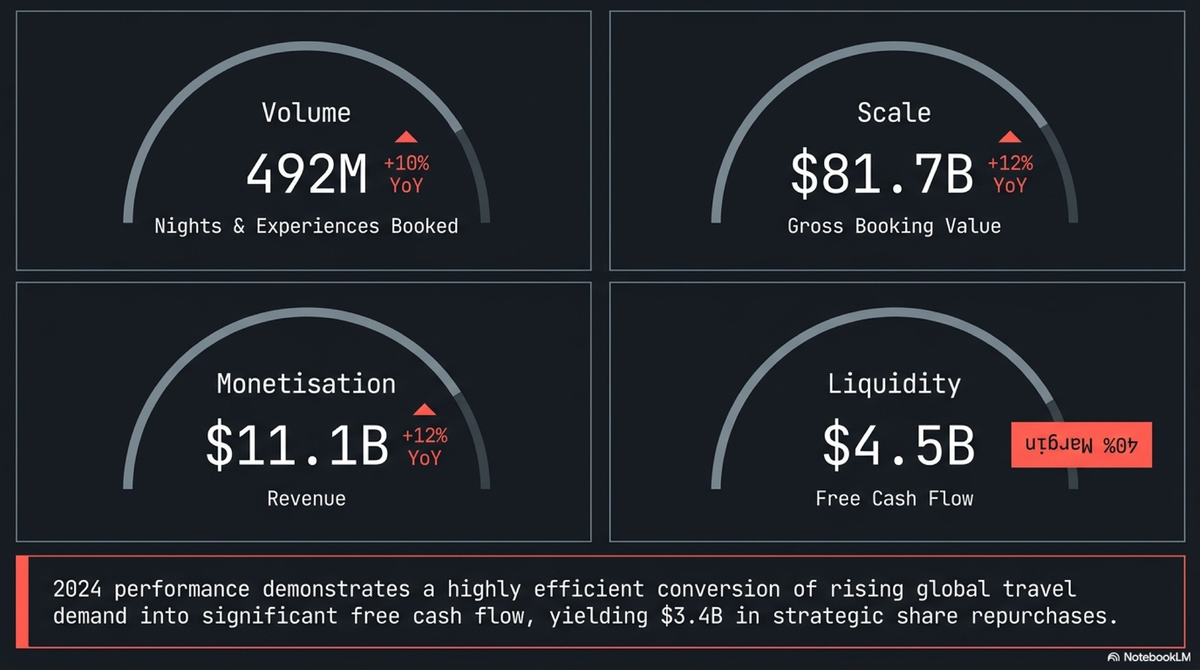}{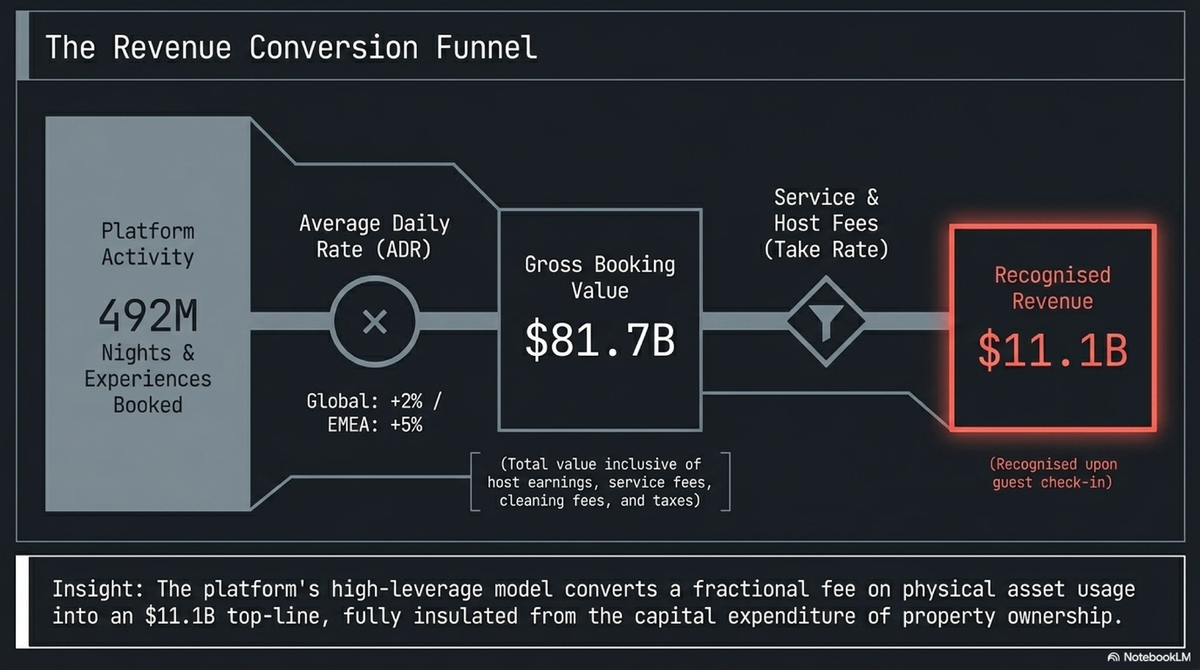}{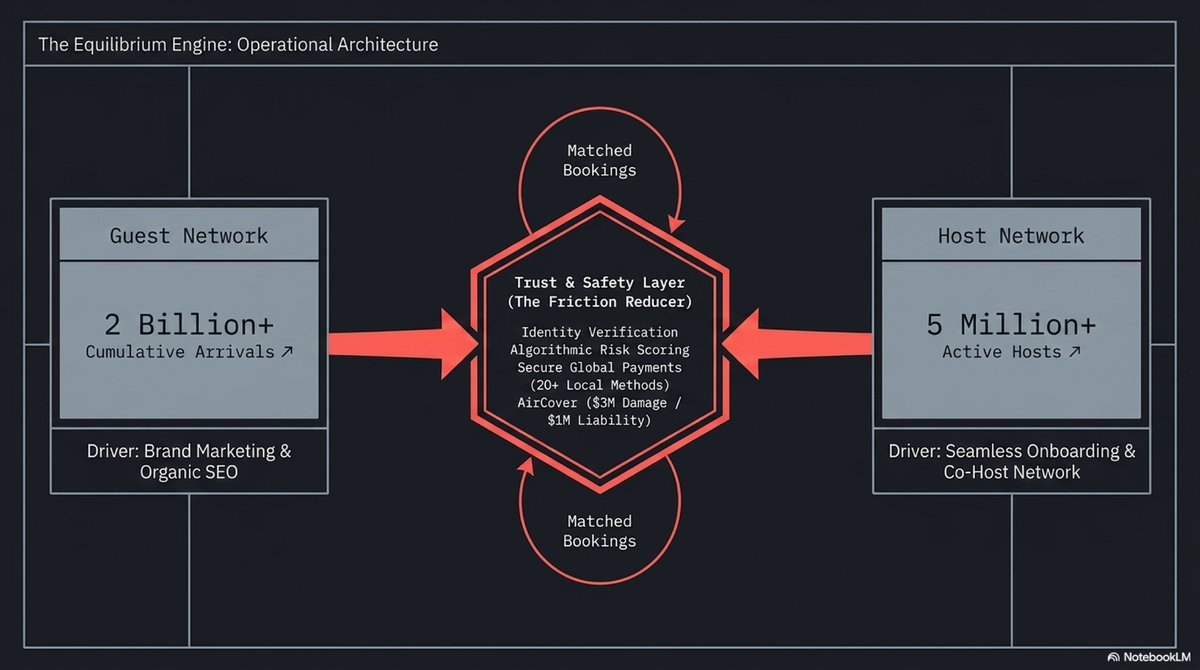}{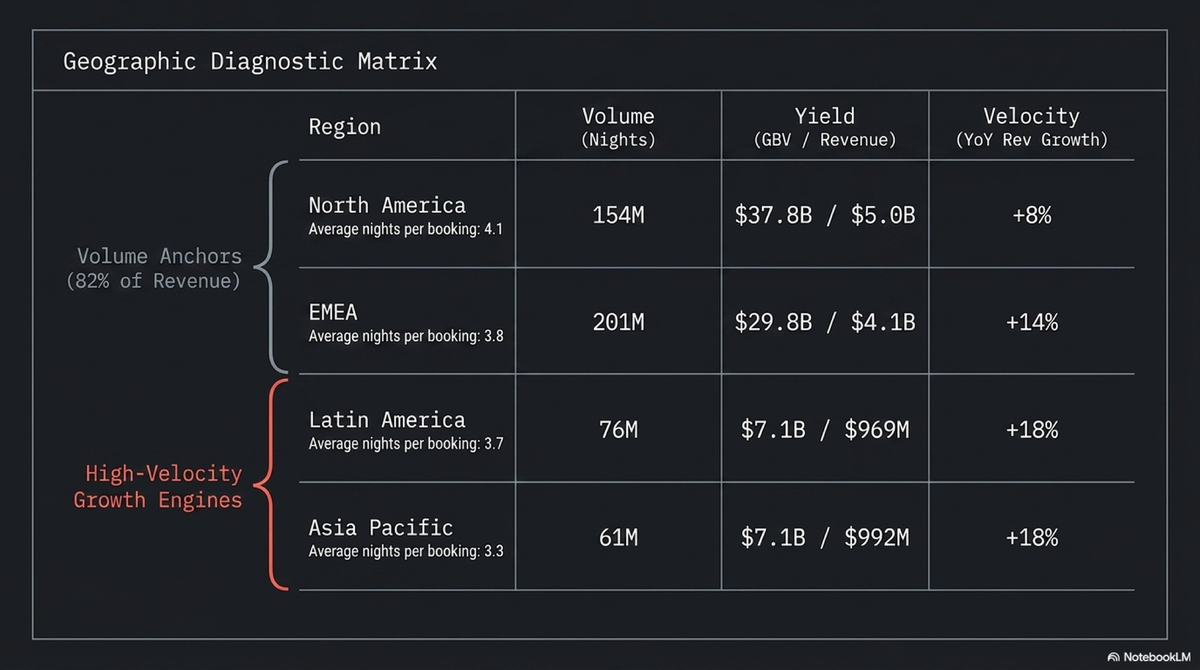}
}
\caption{Deck showcase for Airbnb 2024 Annual Report. This annual-report example shows audience-dependent information selection and the mixed grounding behavior of NotebookLM rows.}
\label{fig:deck_showcase_case010}
\end{figure}

\begin{figure}[!p]
\centering
{\setlength{\tabcolsep}{0pt}%
\deckshowcaserow{DeepPresenter / Agnostic, decision maker: AudCov. 0.184, Correct. 0.907, SafeEff 1.568}{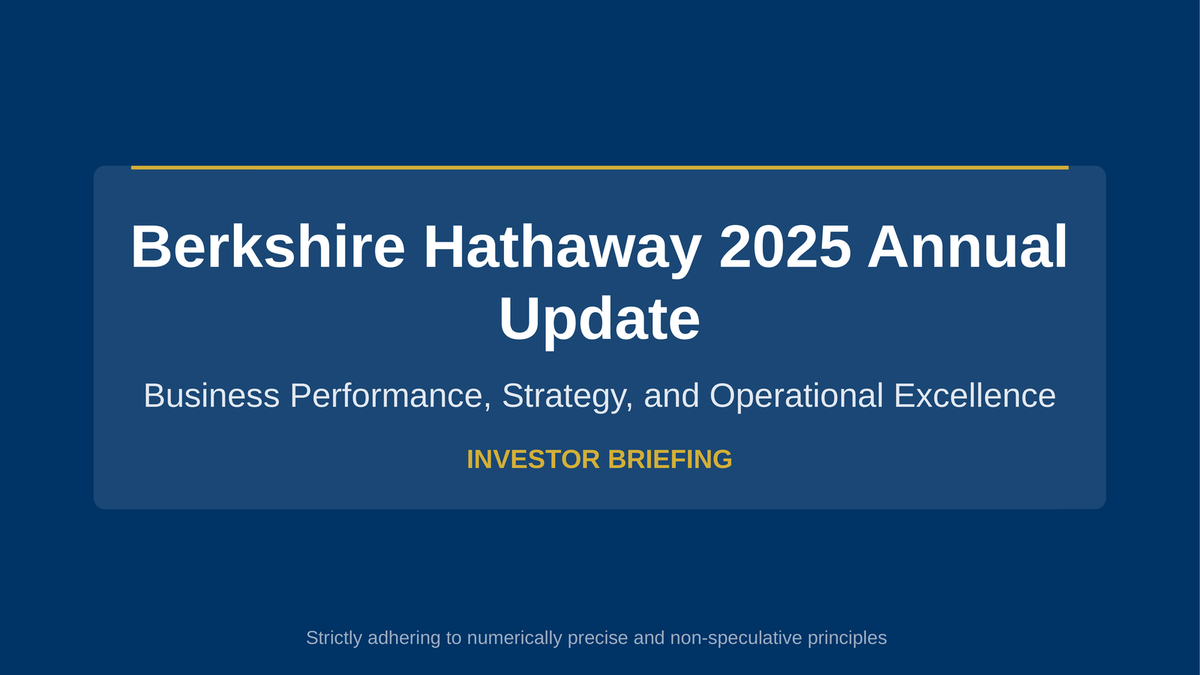}{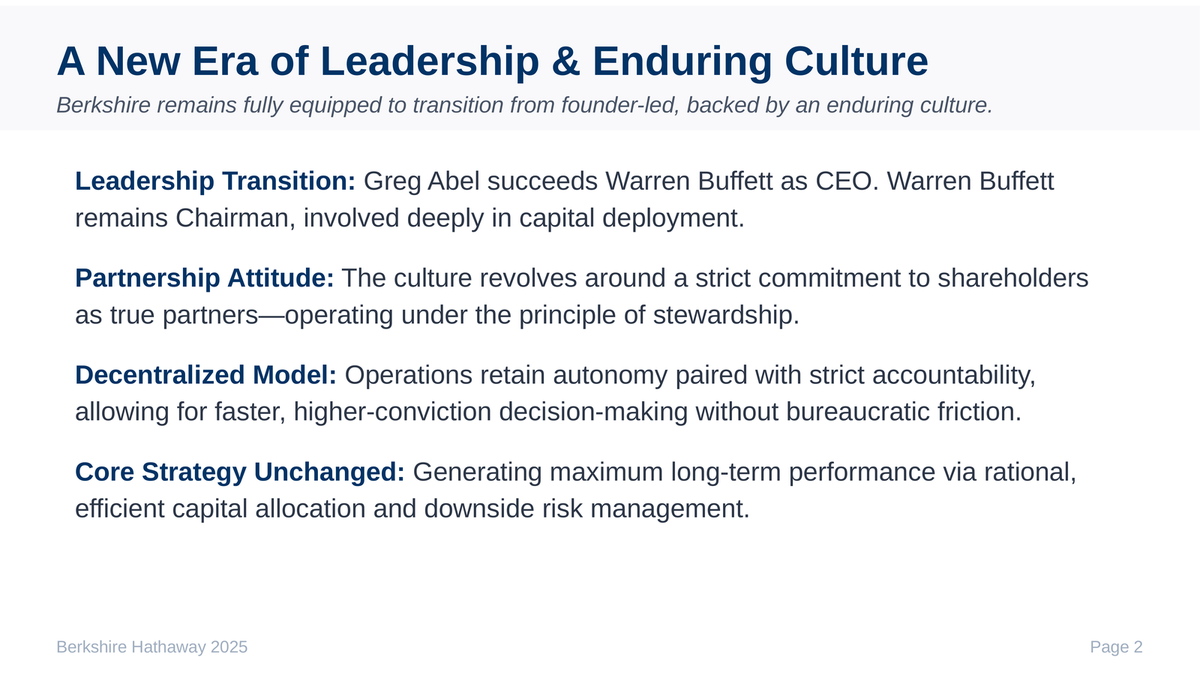}{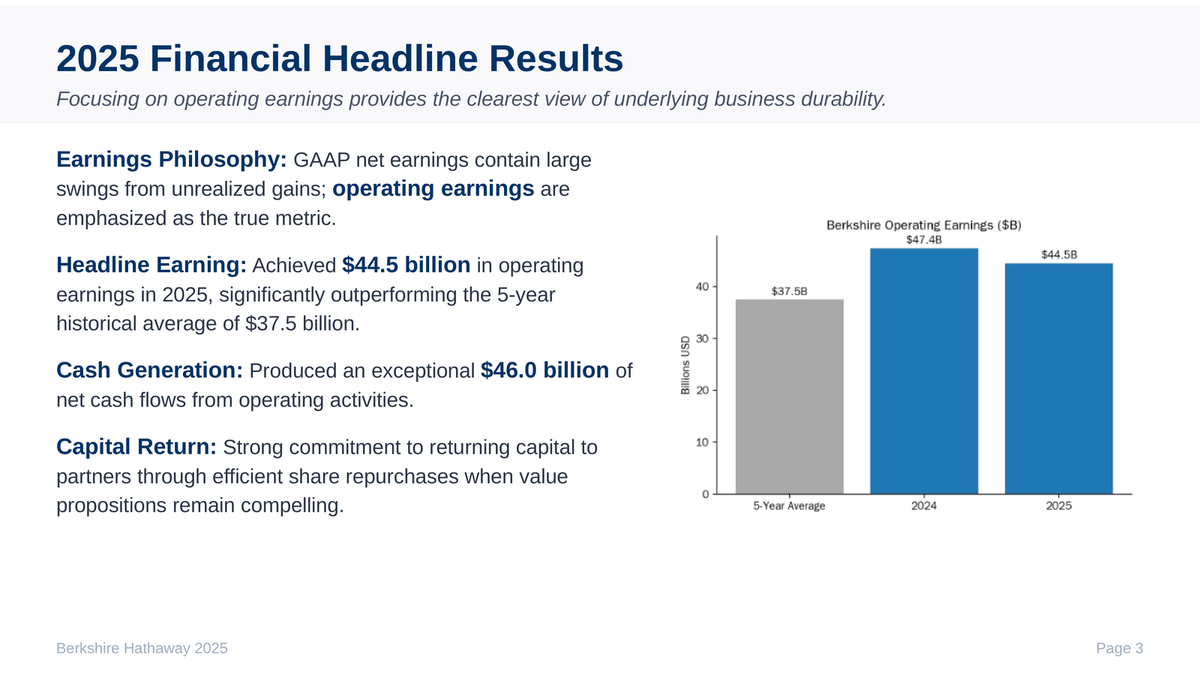}{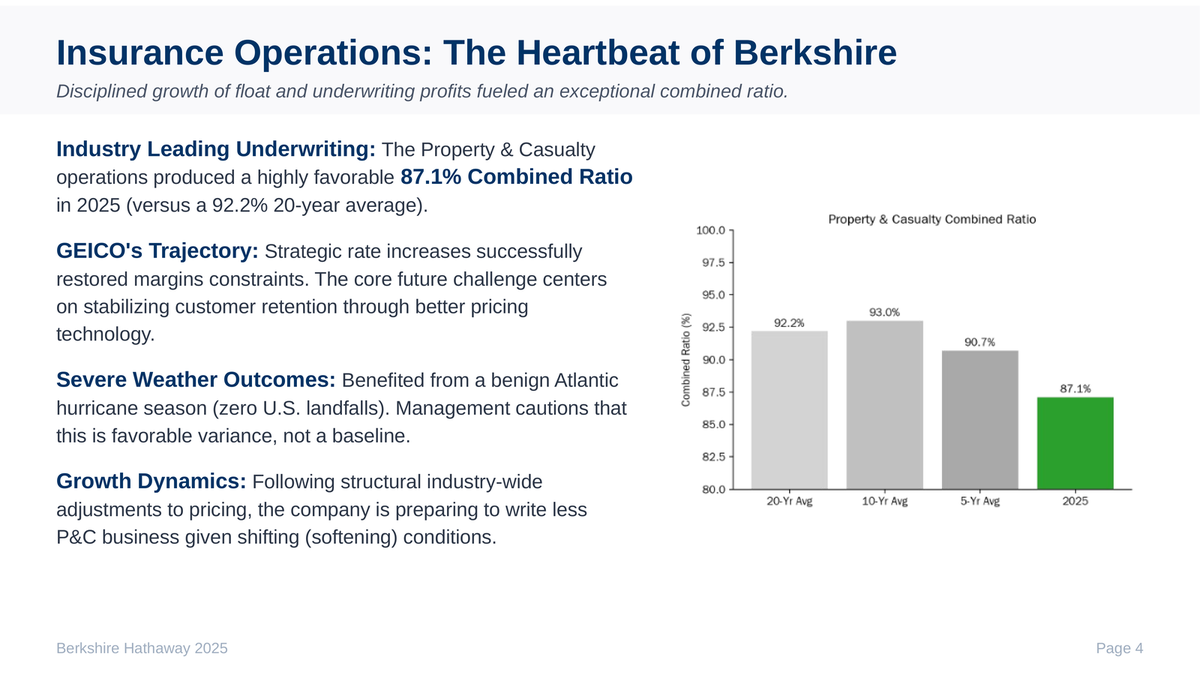}{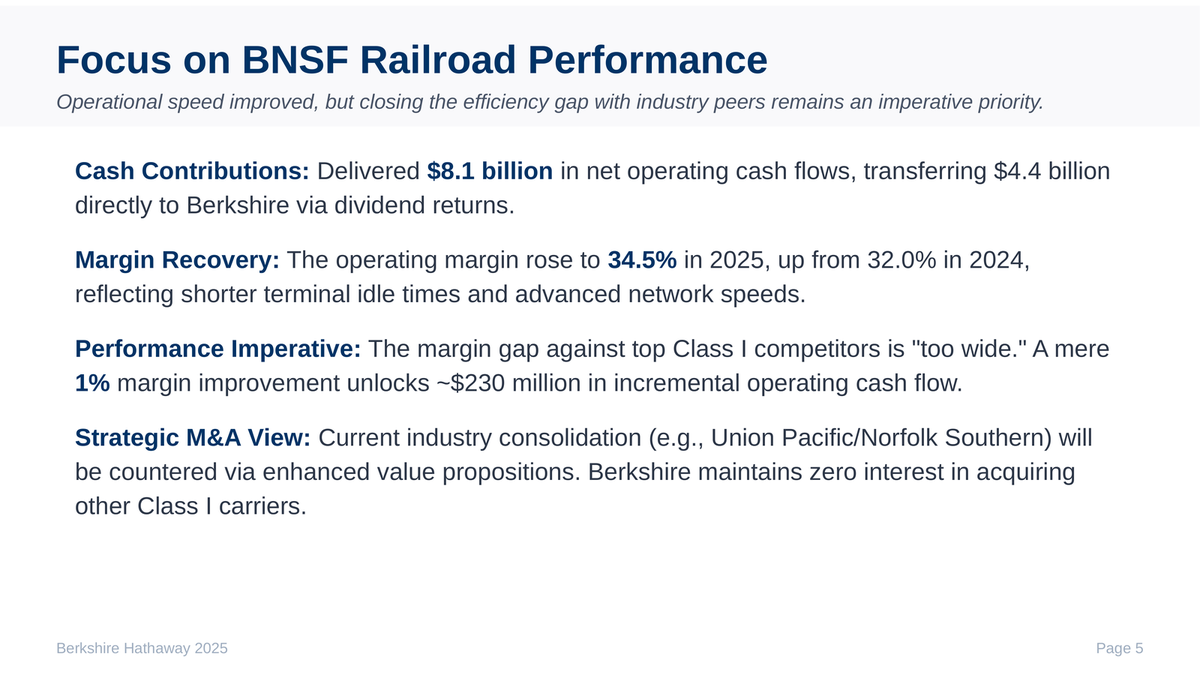}
\deckshowcaserow{SlideTailor / Agnostic, decision maker: AudCov. 0.011, Correct. 0.821, SafeEff 0.187}{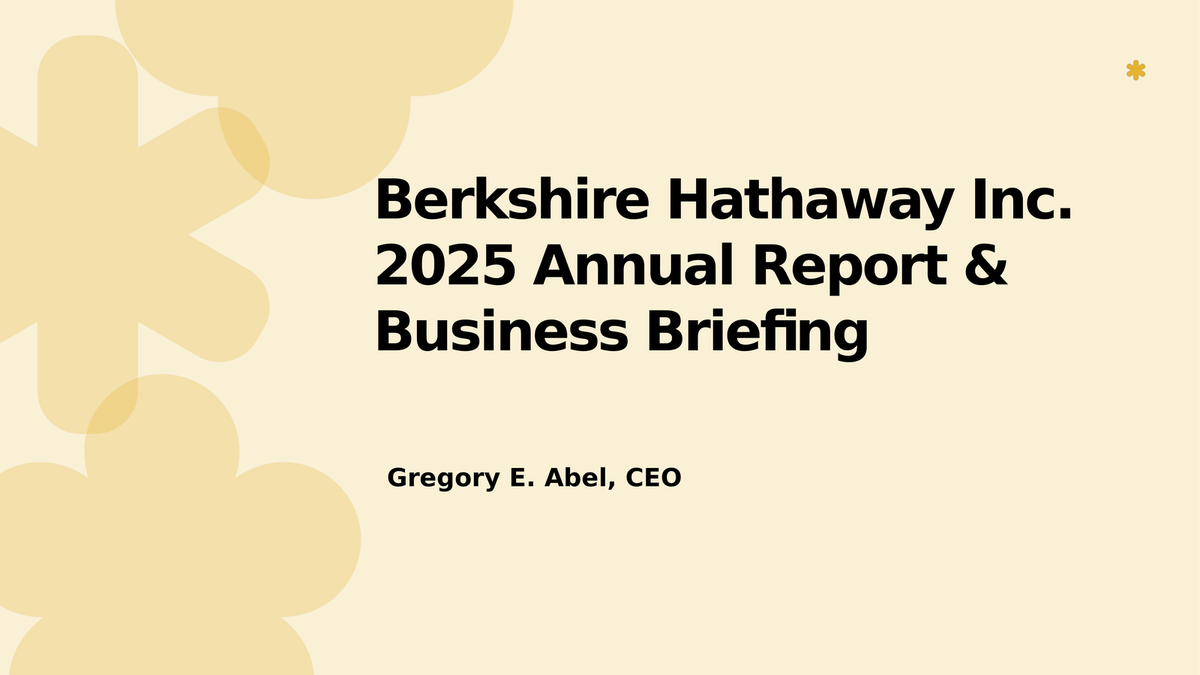}{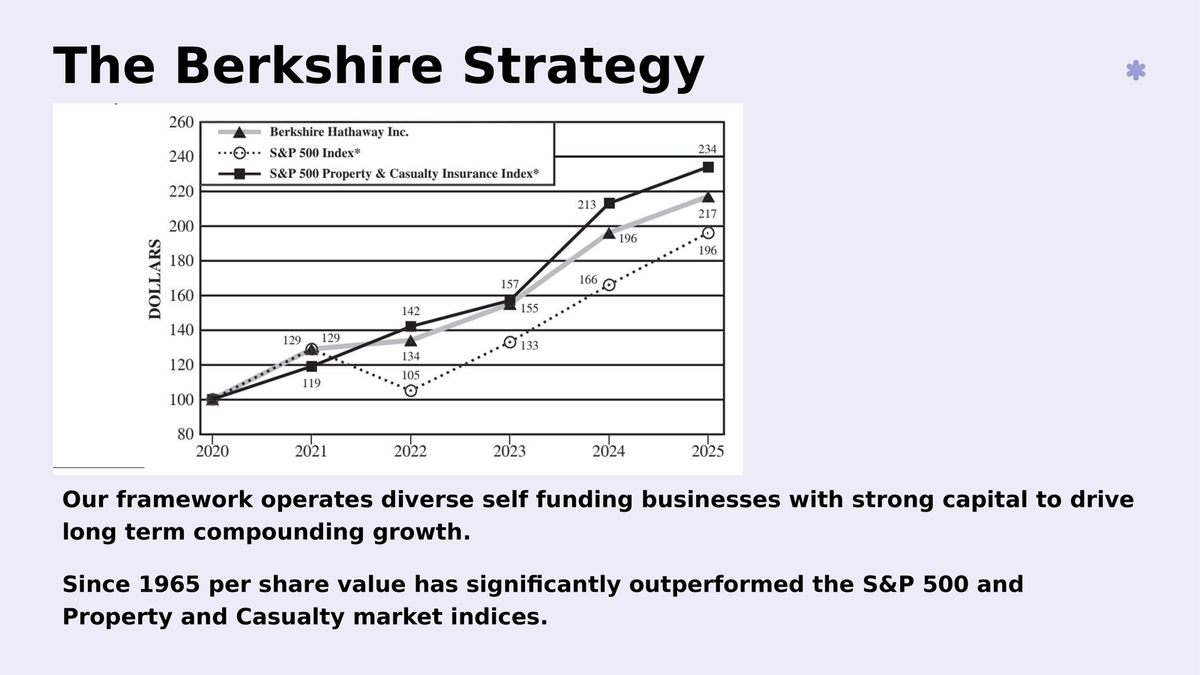}{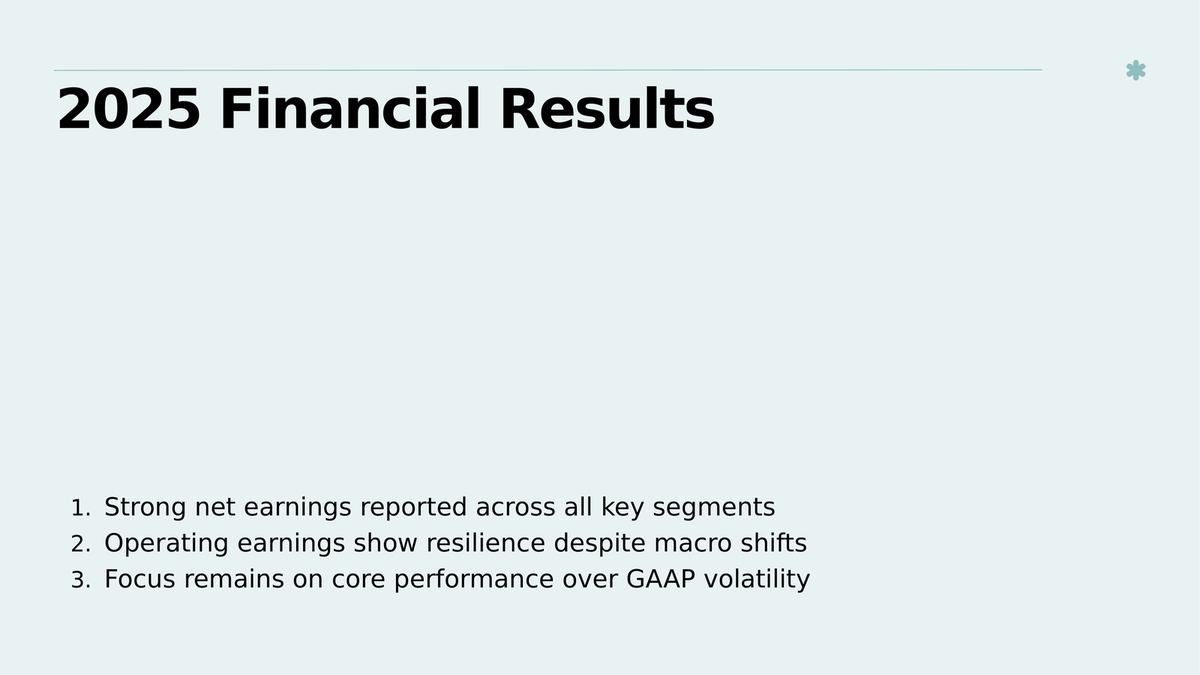}{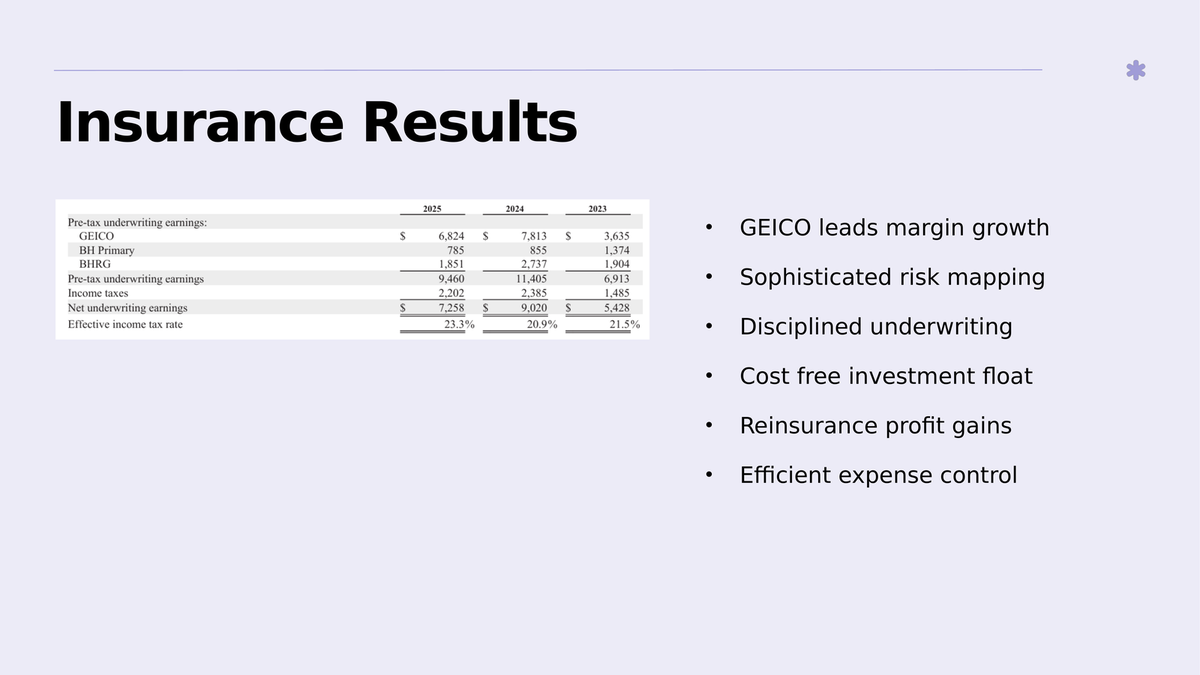}{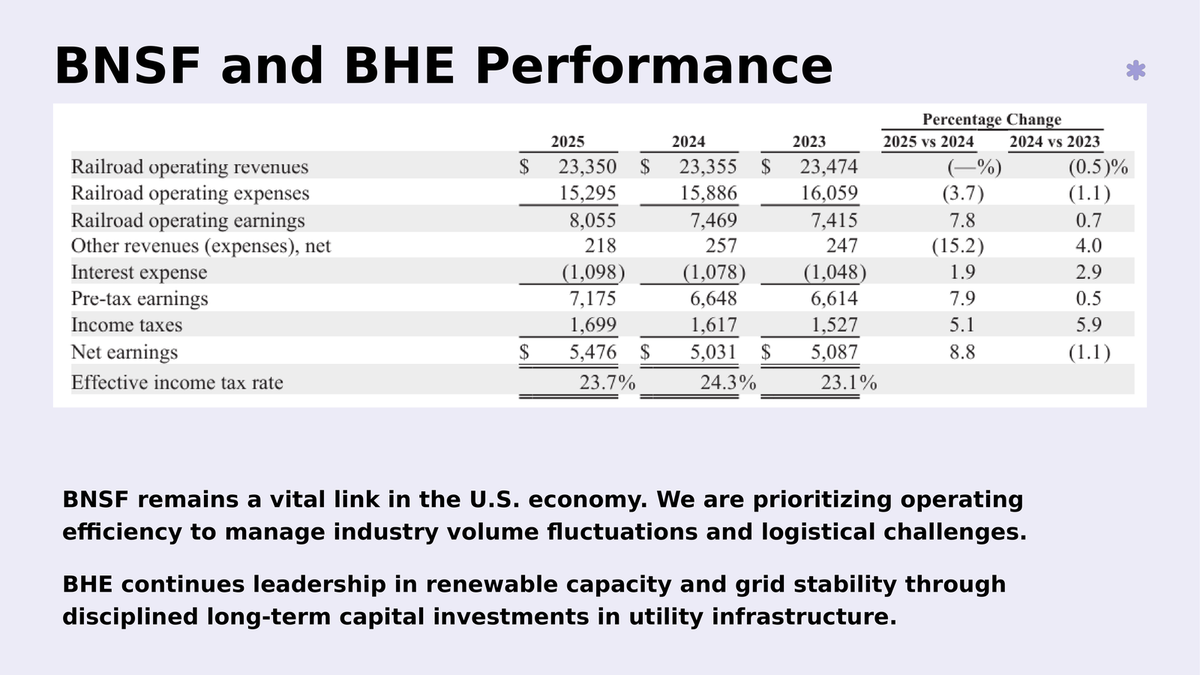}
\deckshowcaserow{NotebookLM / Agnostic, decision maker: AudCov. 0.848, Correct. 0.873, SafeEff 12.738}{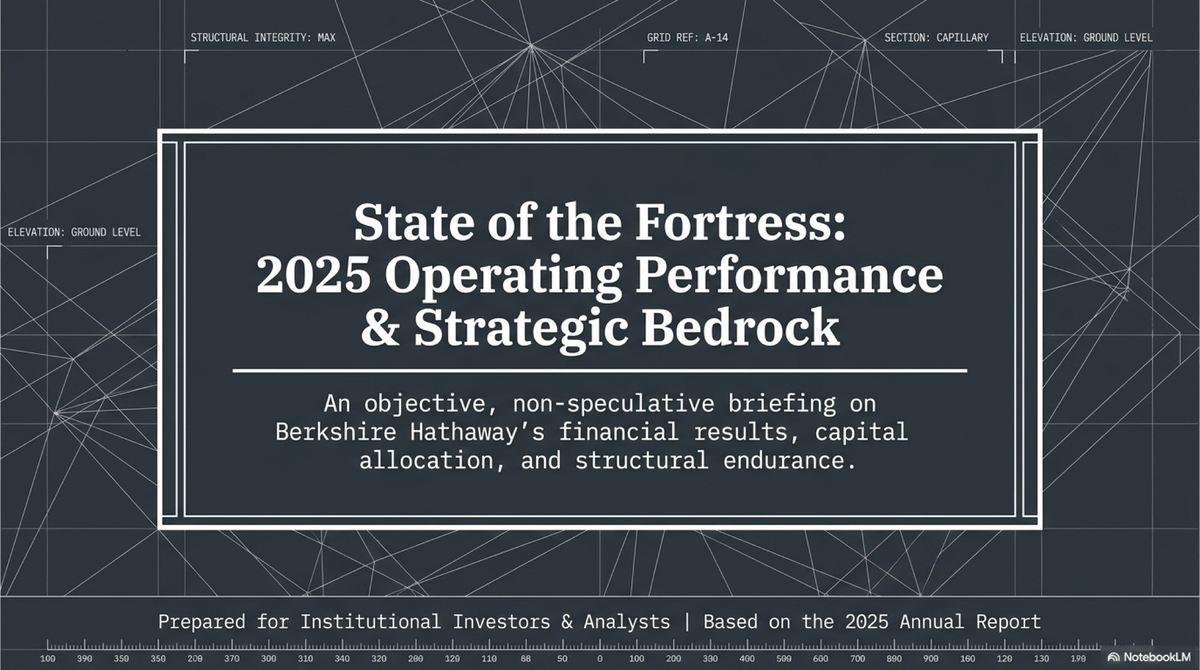}{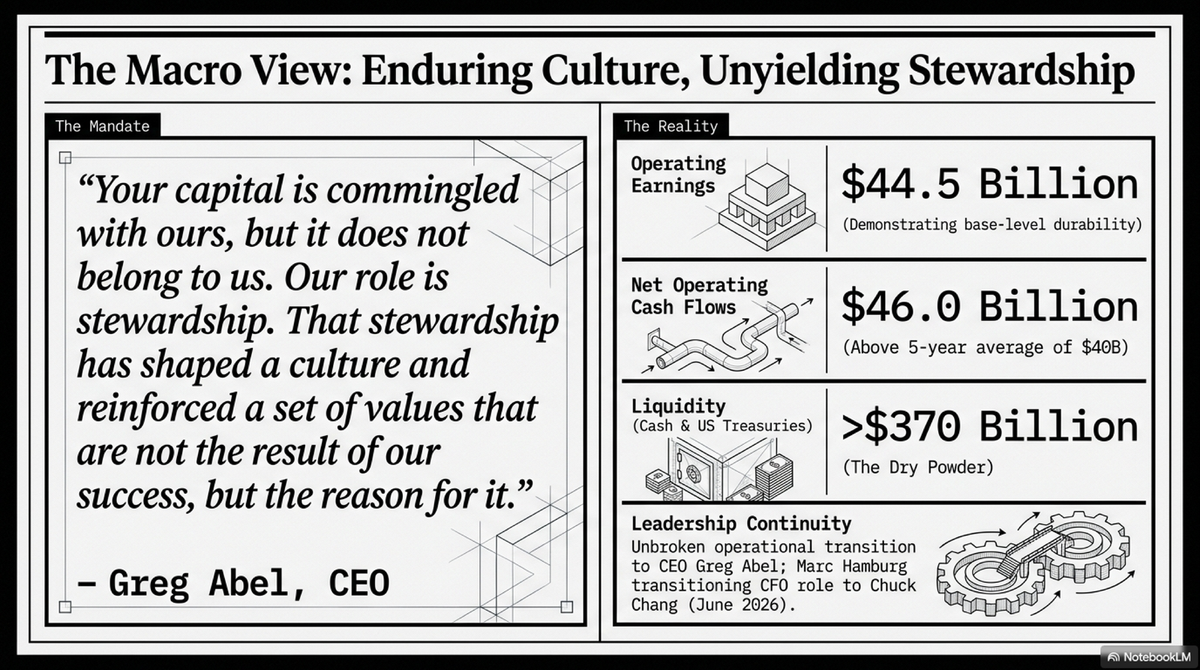}{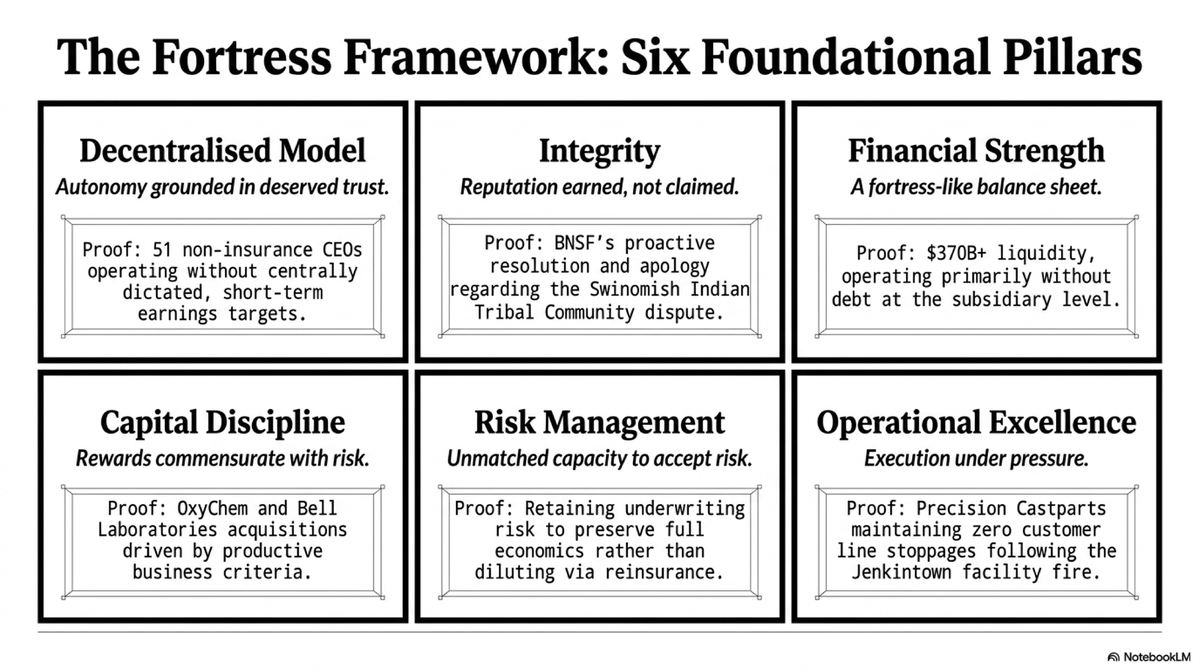}{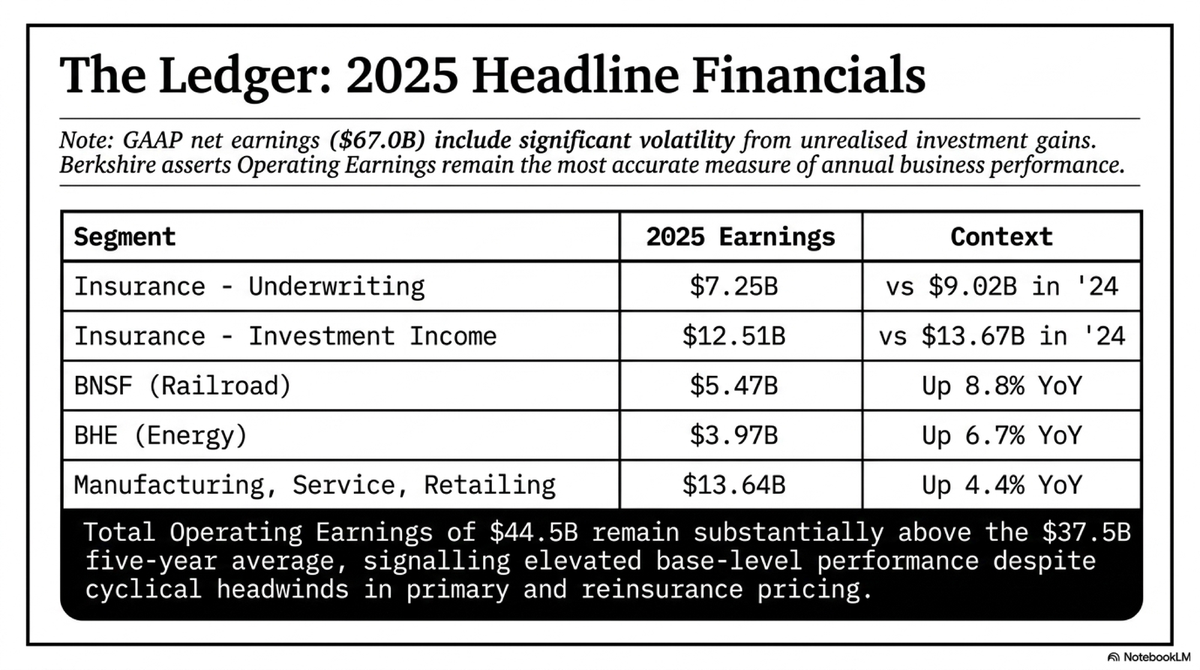}{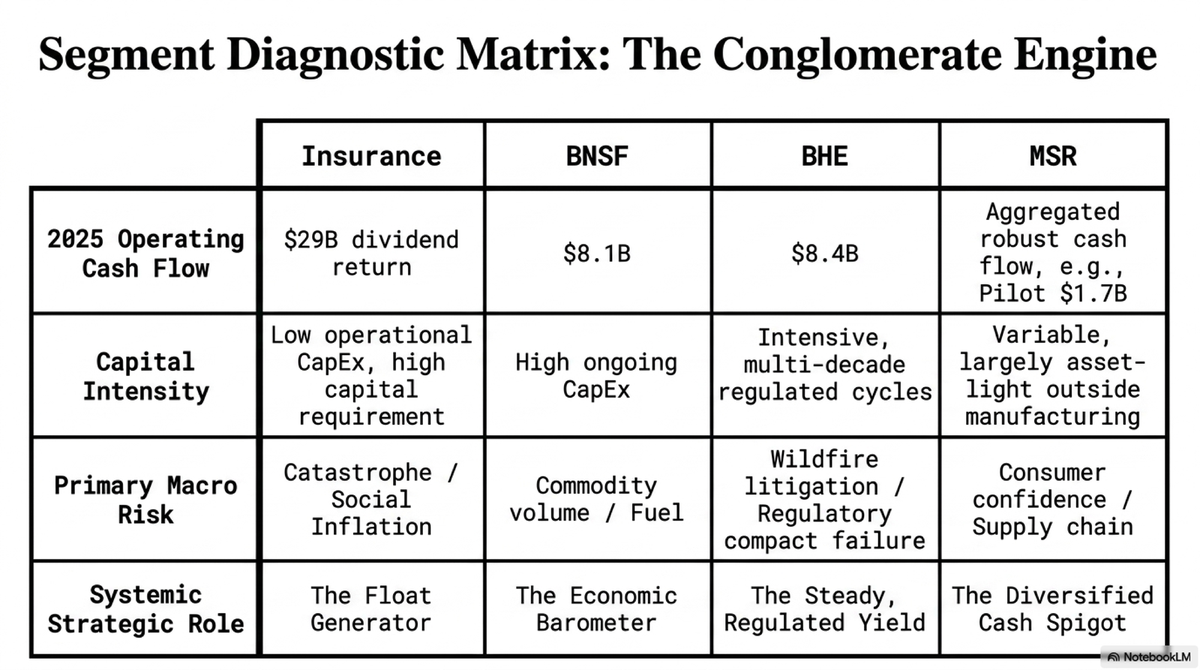}
}
\caption{Deck showcase for Berkshire Hathaway 2025 Annual Report. This agnostic decision-maker comparison shows that high Audience Coverage can still be separated from source-grounded Correctness.}
\label{fig:deck_showcase_case009}
\end{figure}

\begin{figure}[!p]
\centering
{\setlength{\tabcolsep}{0pt}%
\deckshowcaserow{DeepPresenter / Learner: AudCov. 0.864, Correct. 0.865, SafeEff 4.977}{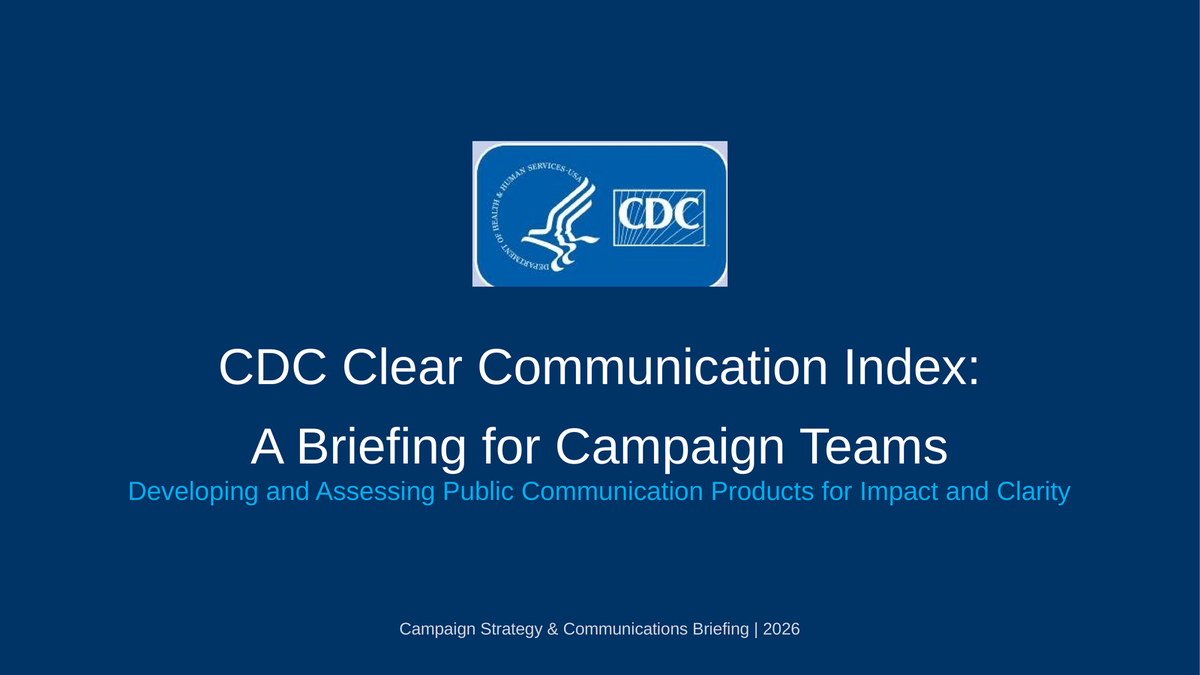}{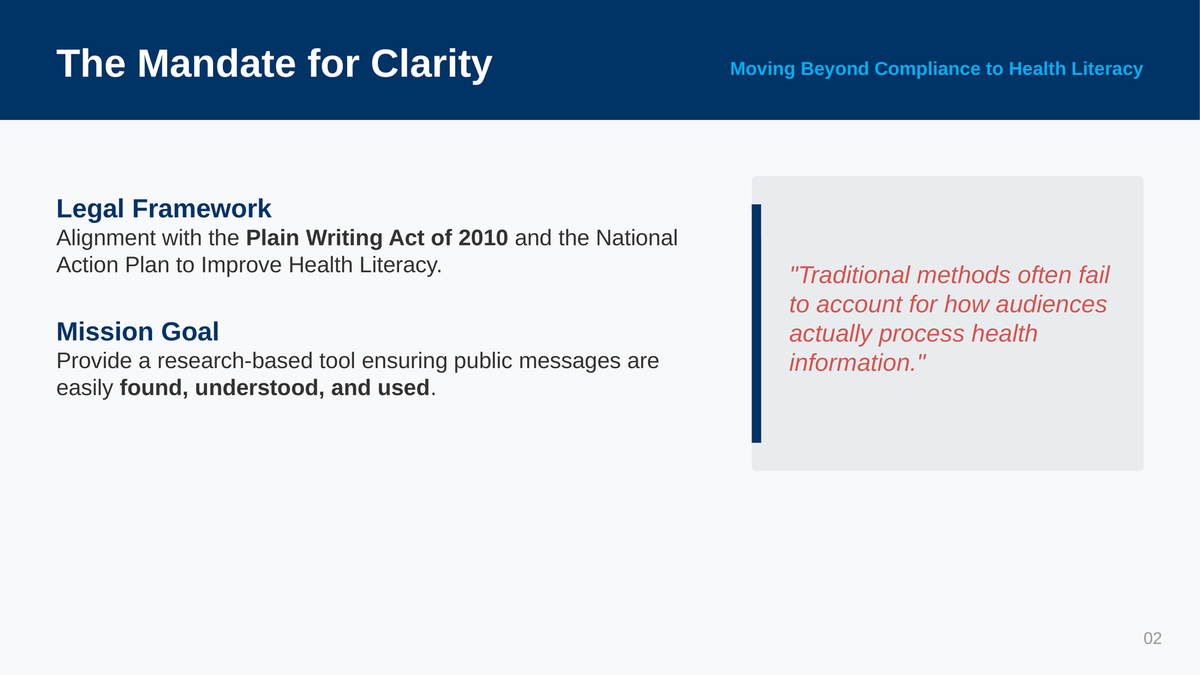}{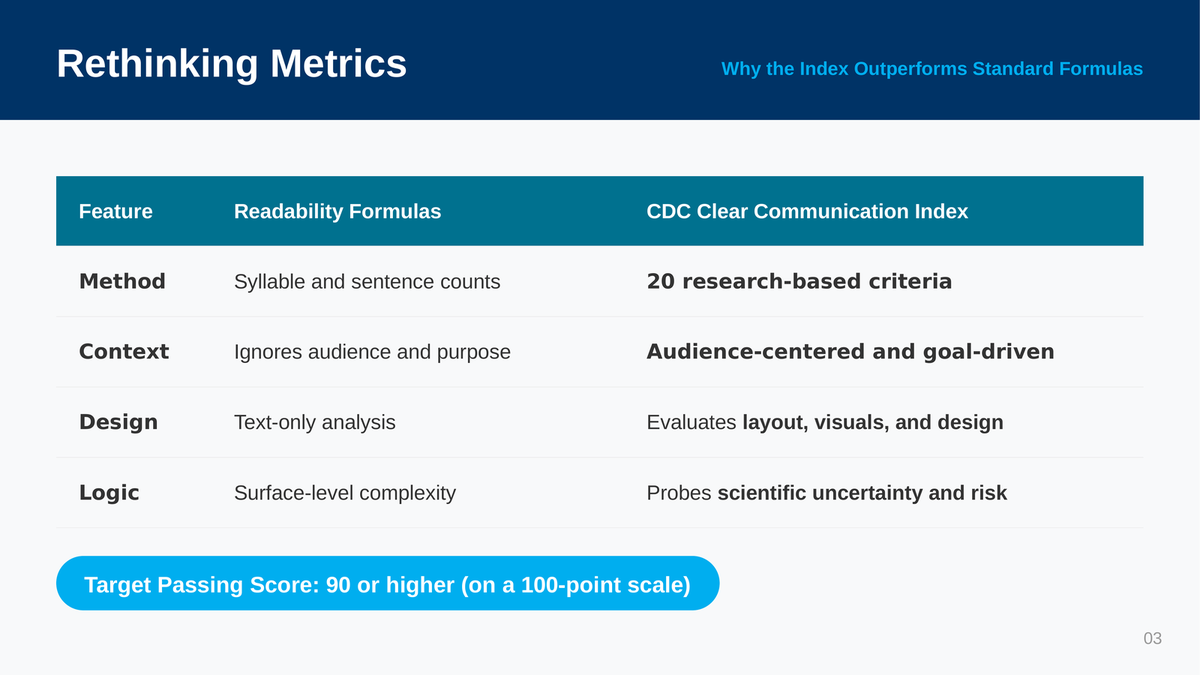}{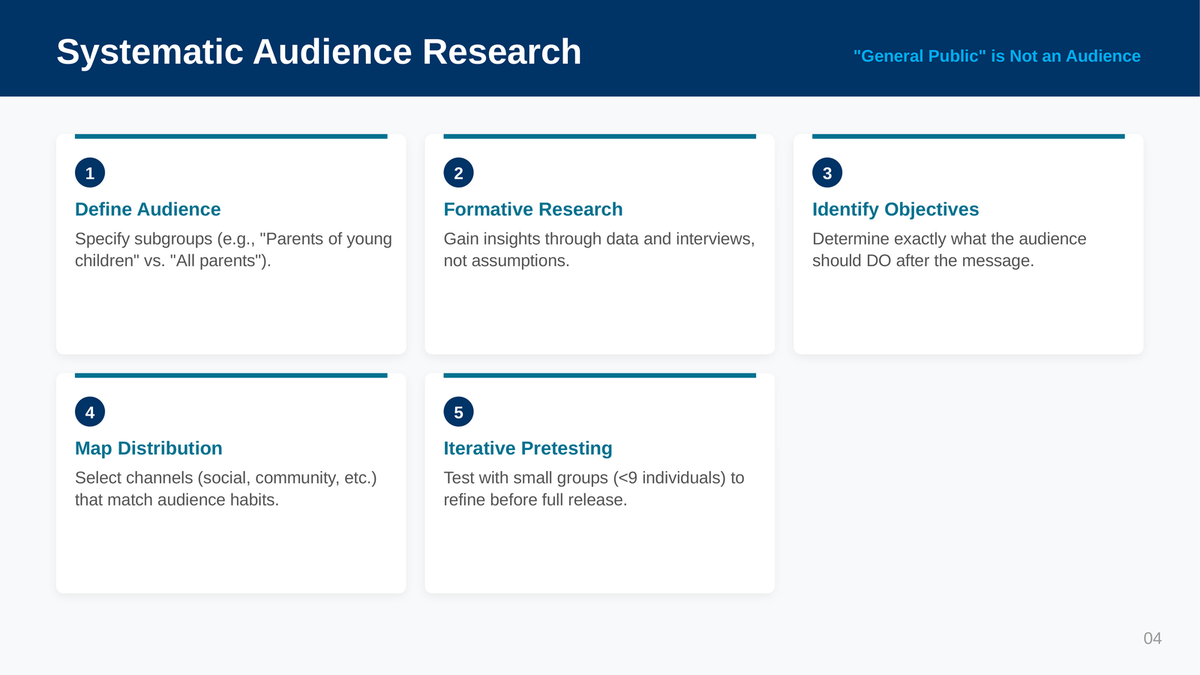}{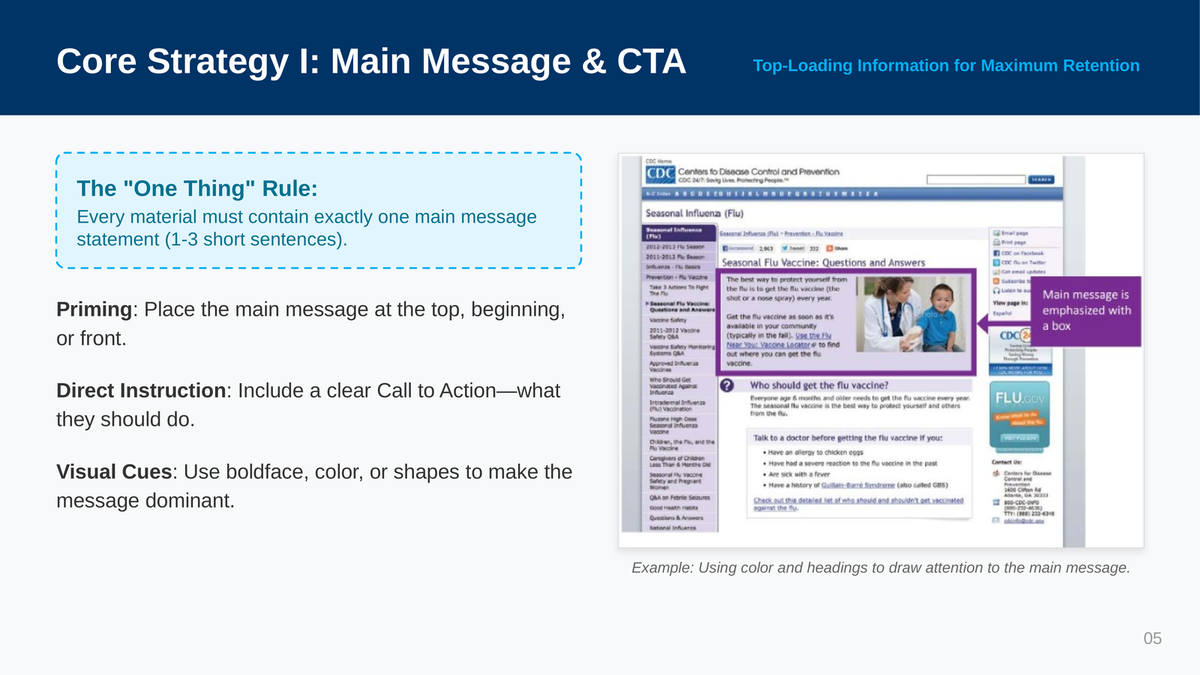}
\deckshowcaserow{DeepPresenter / Specialist: AudCov. 0.562, Correct. 0.923, SafeEff 2.855}{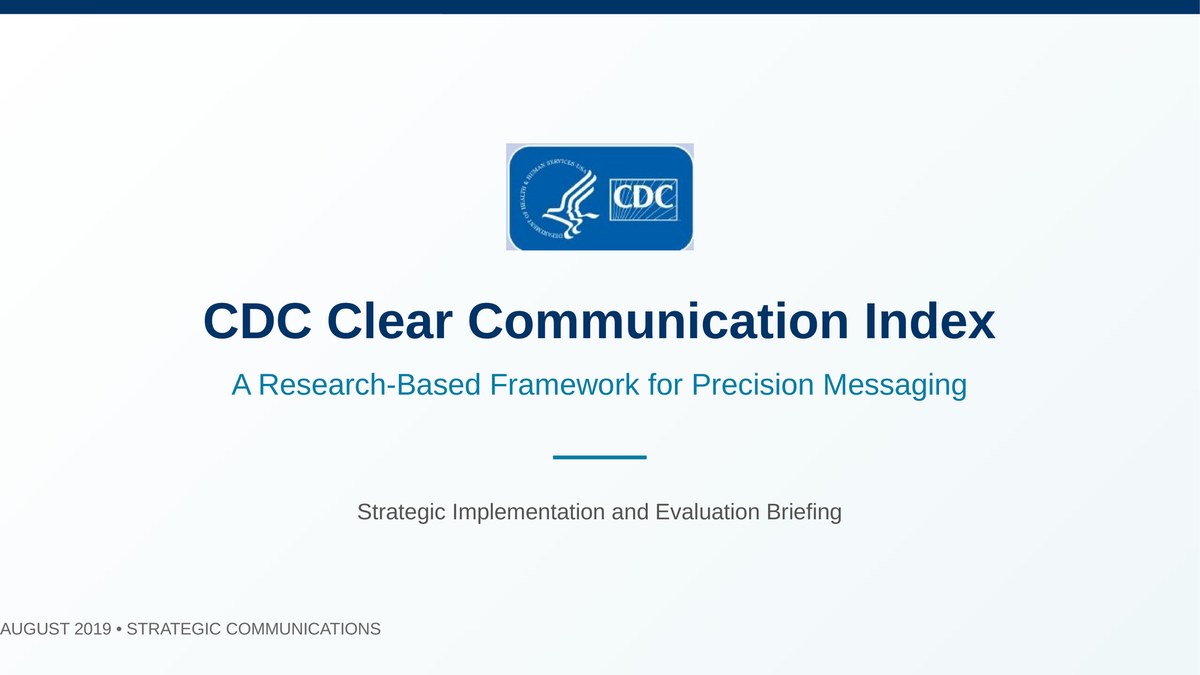}{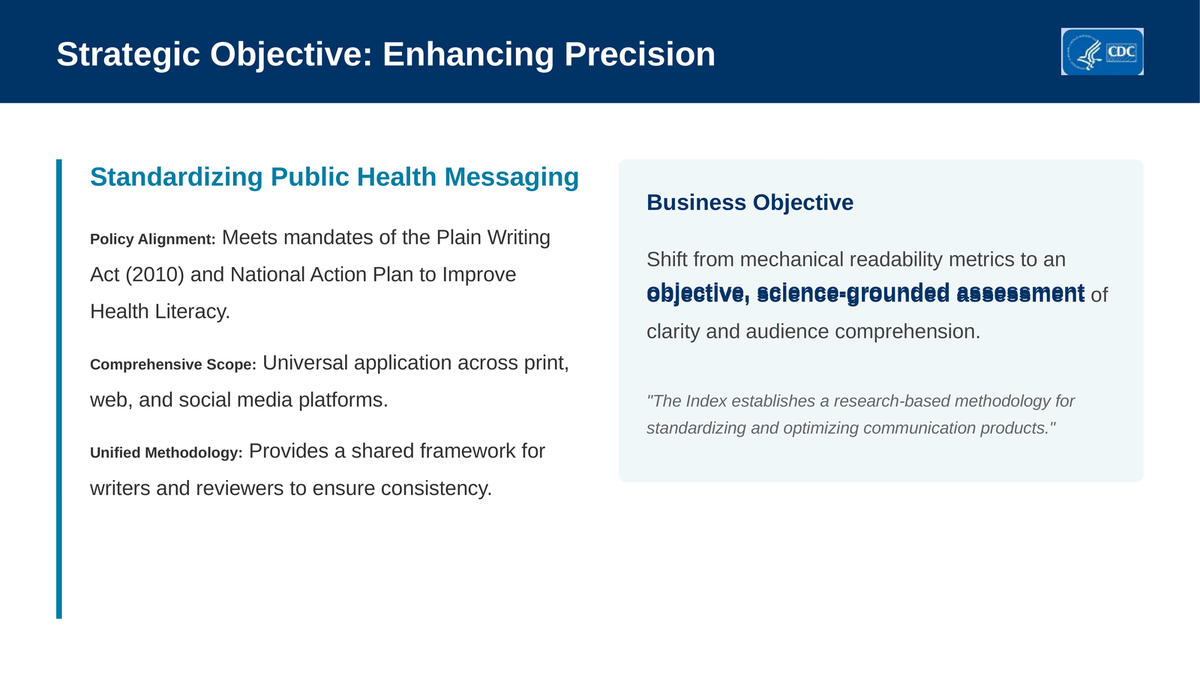}{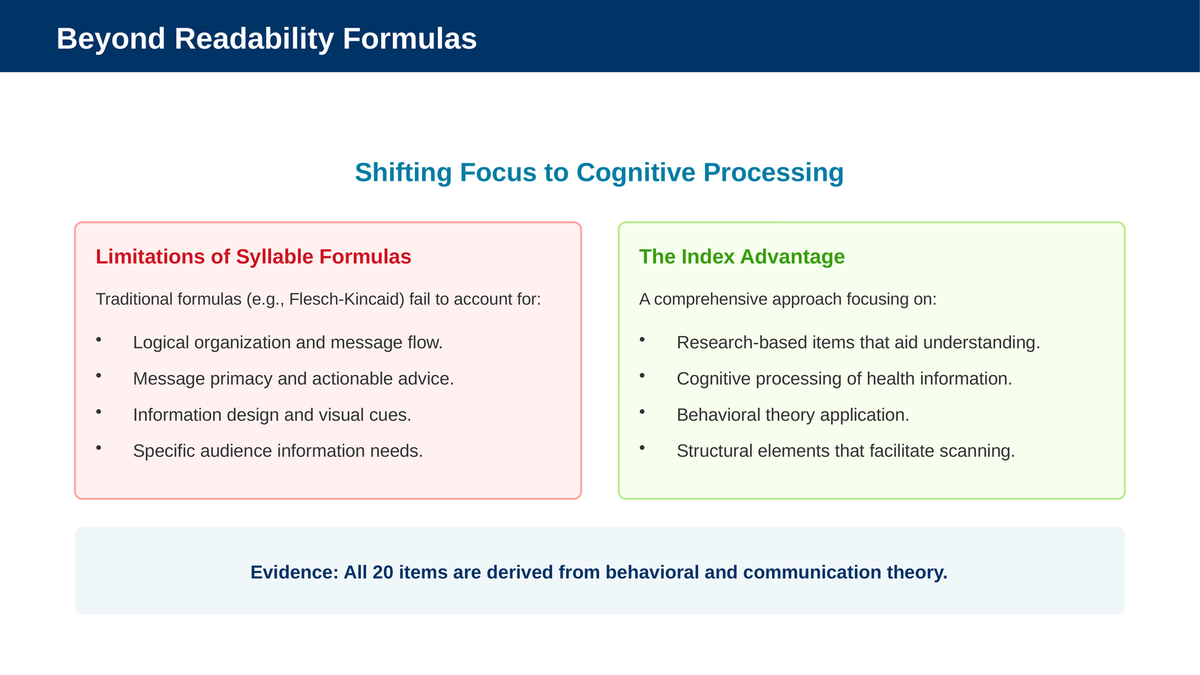}{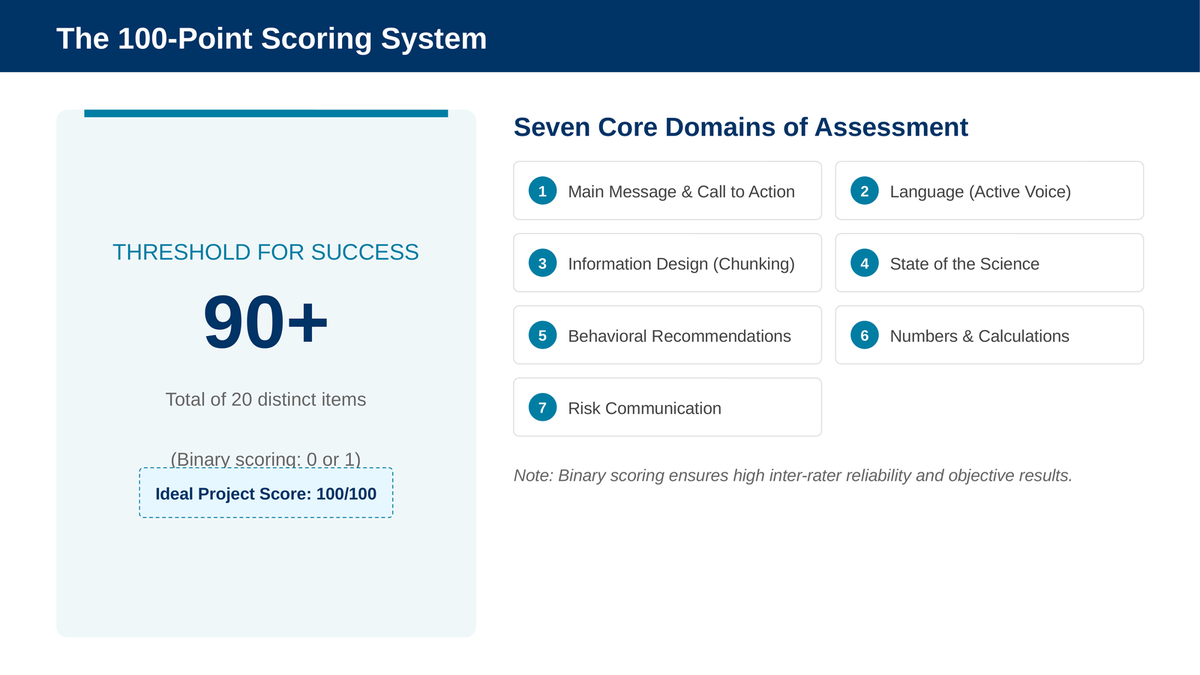}{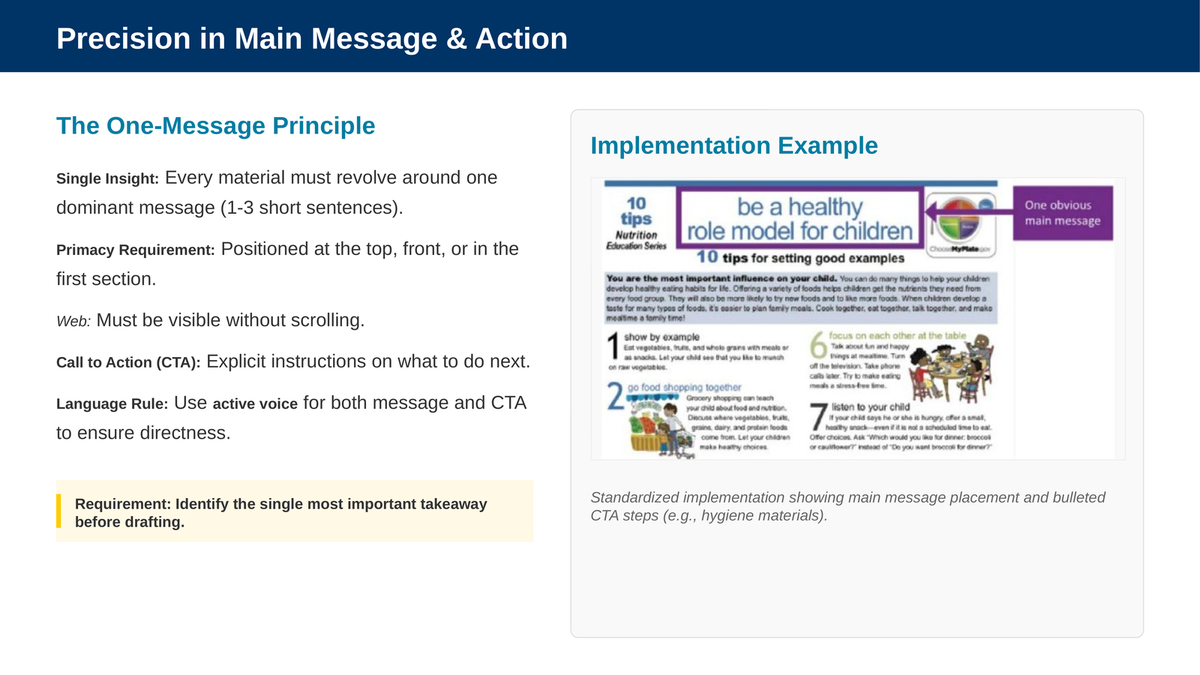}
\deckshowcaserow{SlideTailor / Specialist: AudCov. 0.292, Correct. 0.705, SafeEff 2.388}{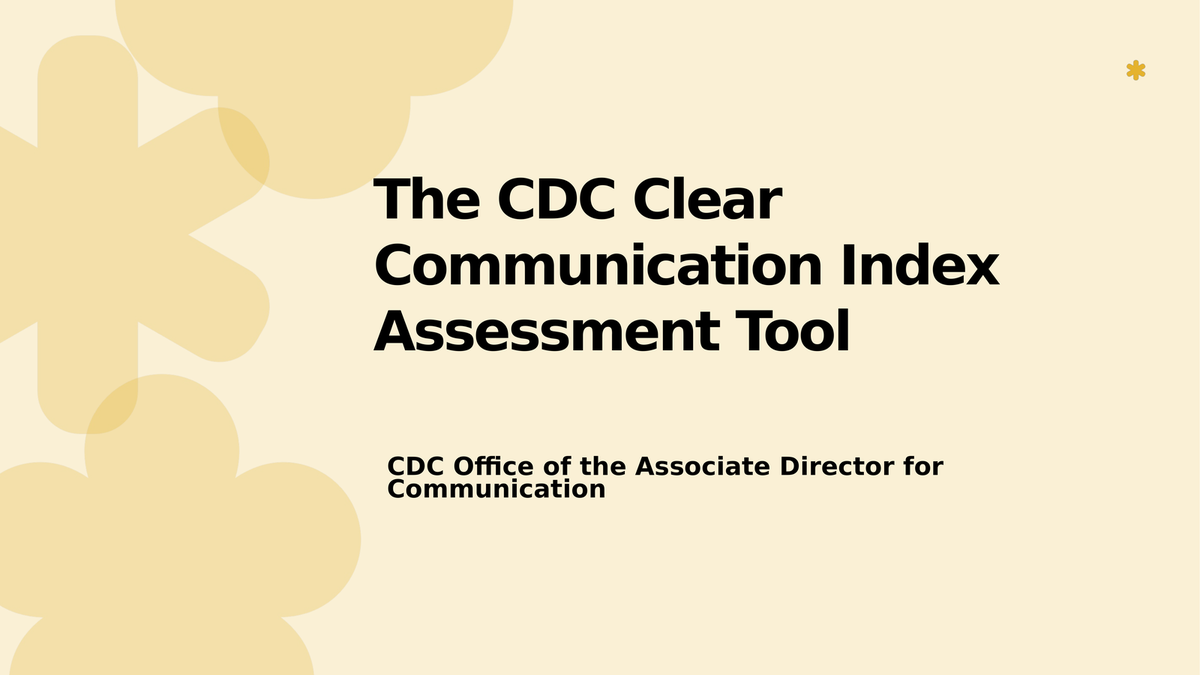}{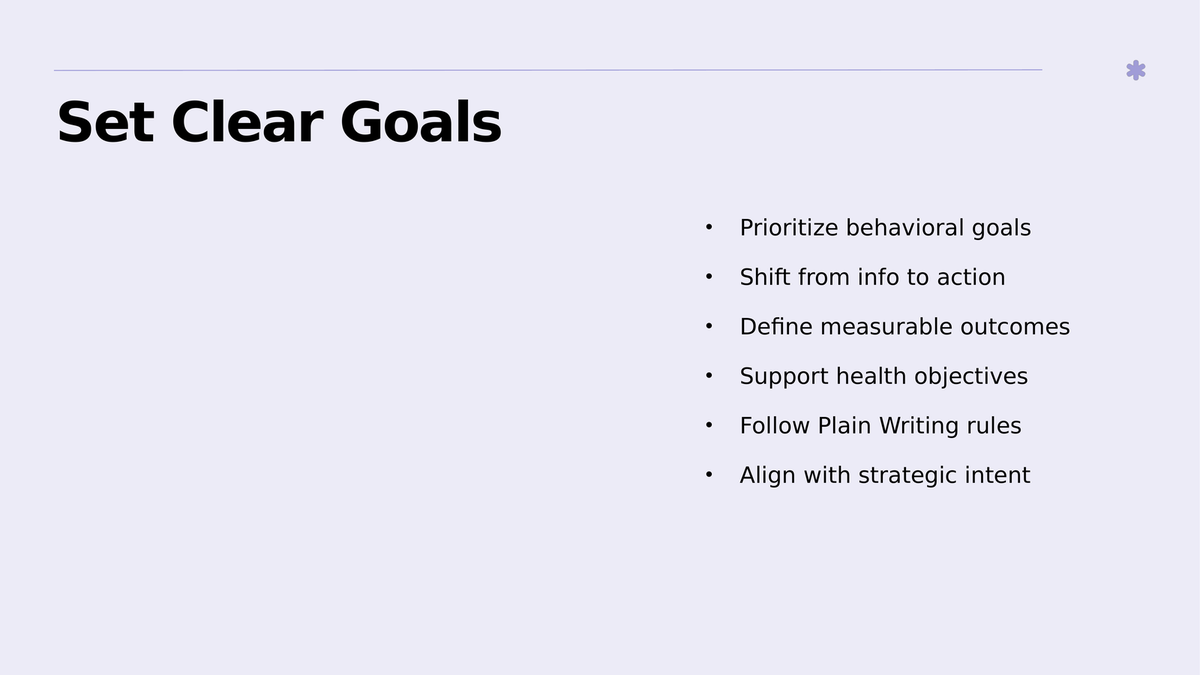}{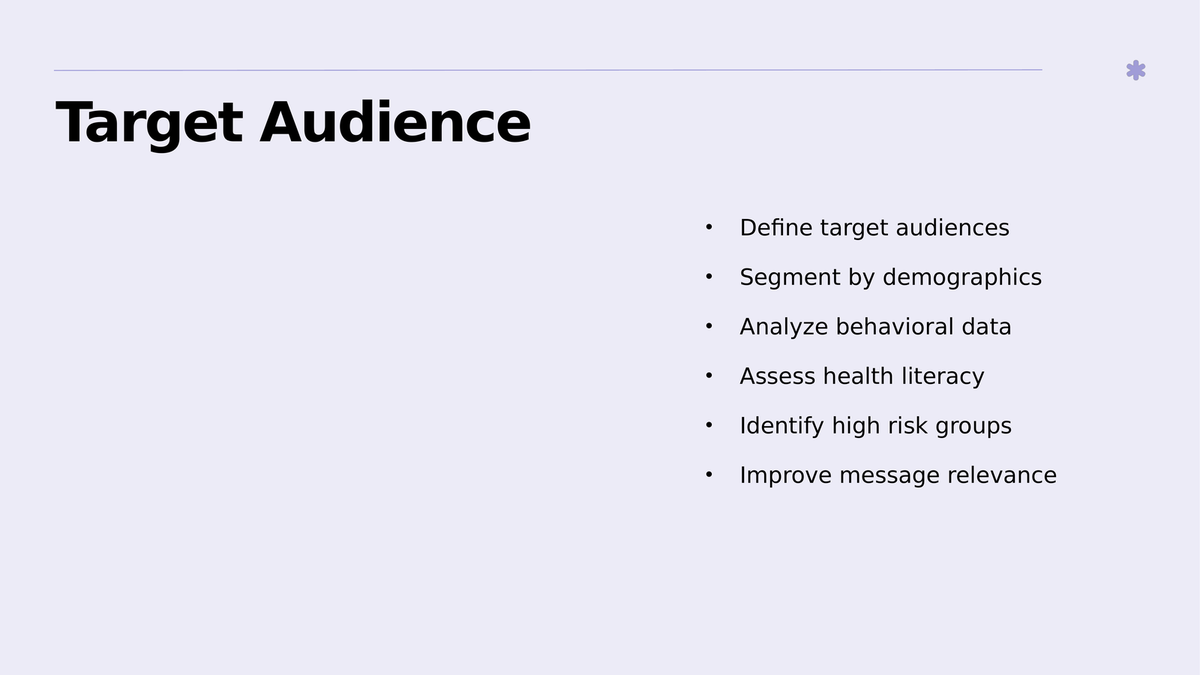}{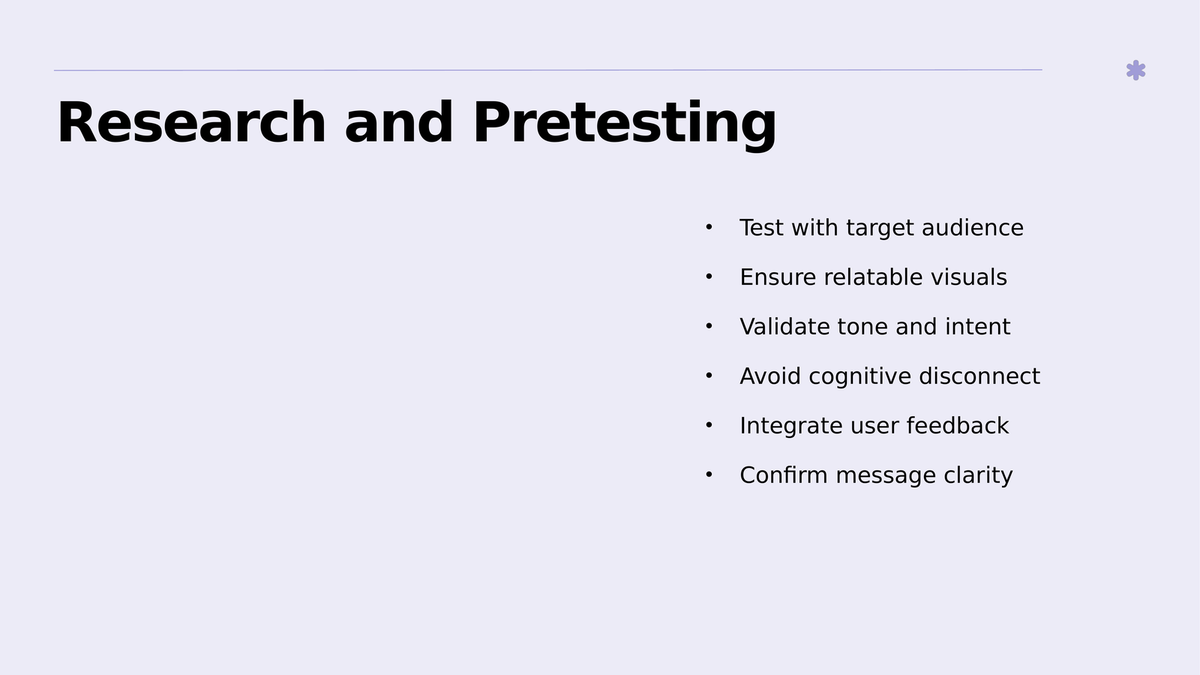}{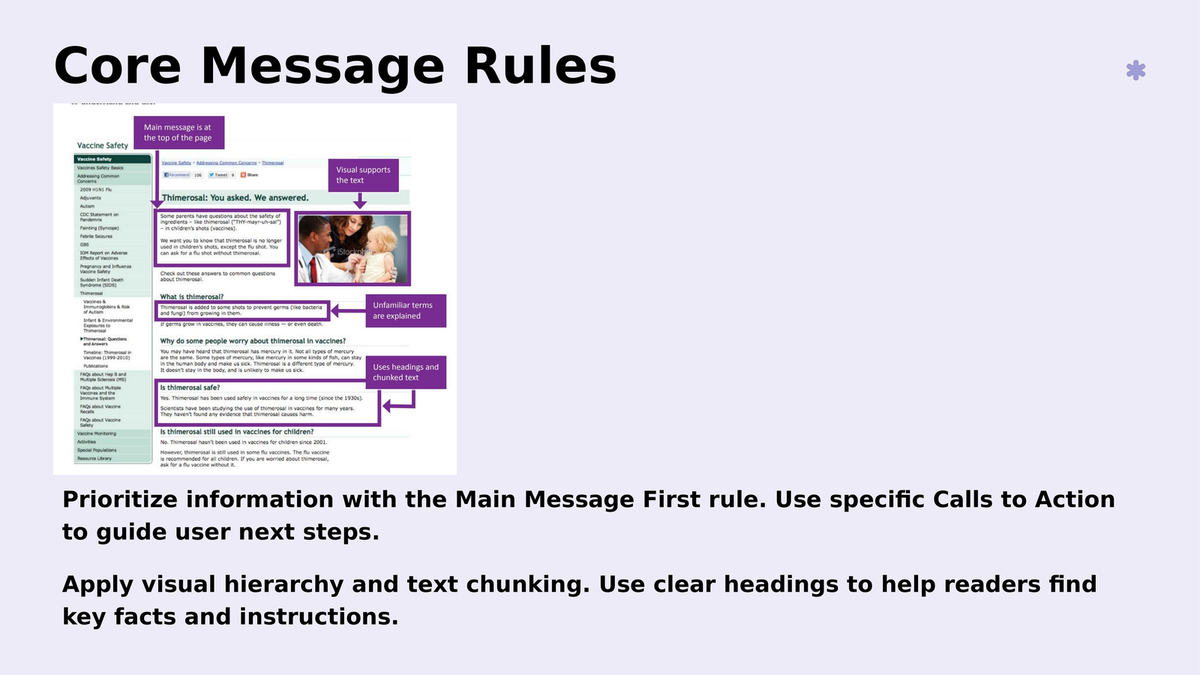}
}
\caption{Deck showcase for CDC Clear Communication Index User Guide. This communication-guidance case shows a depth and grounding tradeoff between learner and specialist presentations.}
\label{fig:deck_showcase_case018}
\end{figure}

\begin{figure}[!p]
\centering
{\setlength{\tabcolsep}{0pt}%
\deckshowcaserow{DeepPresenter / Specialist: AudCov. 0.125, Correct. 0.870, SafeEff 0.186}{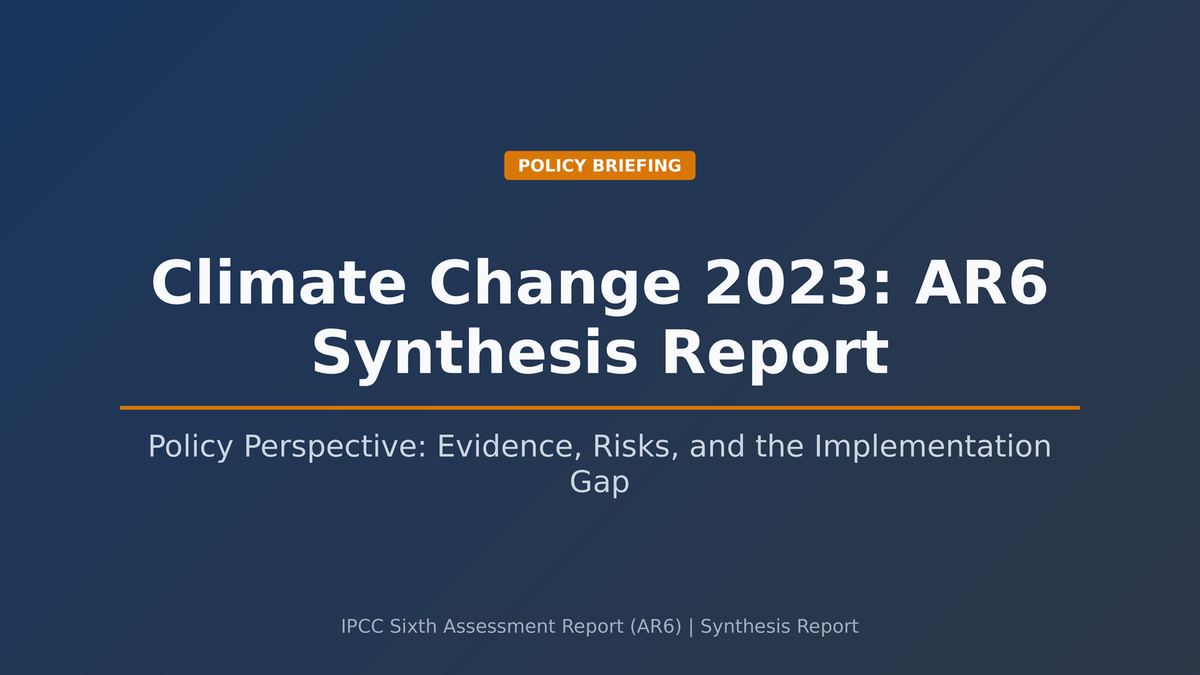}{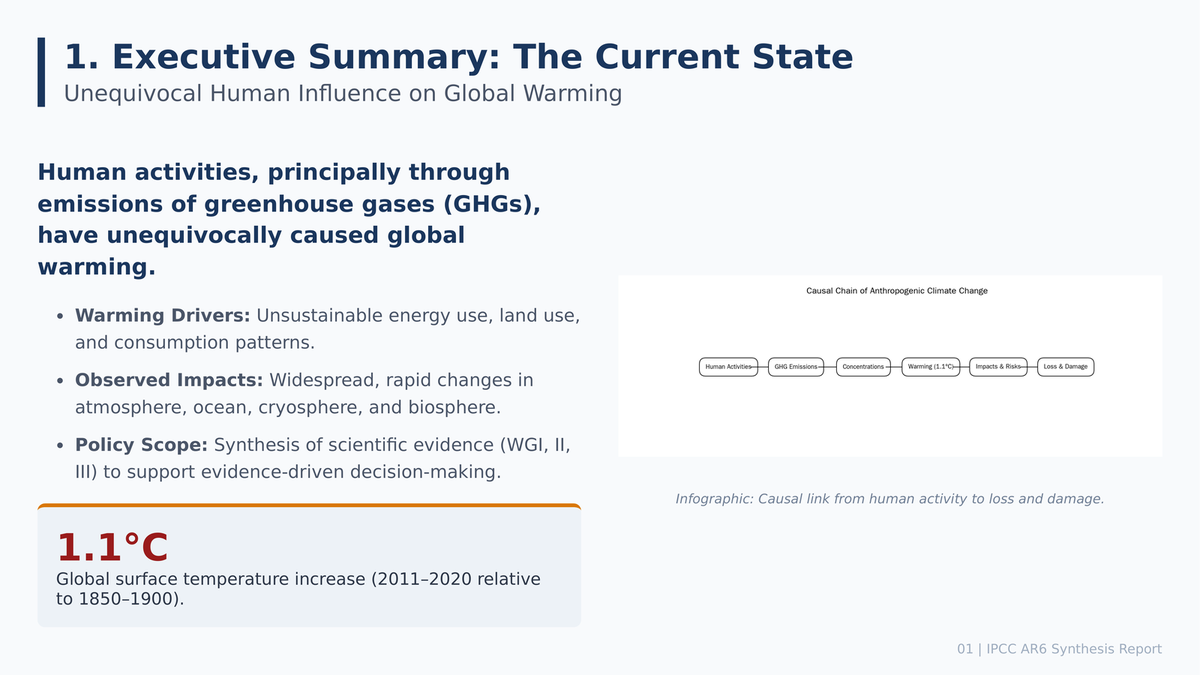}{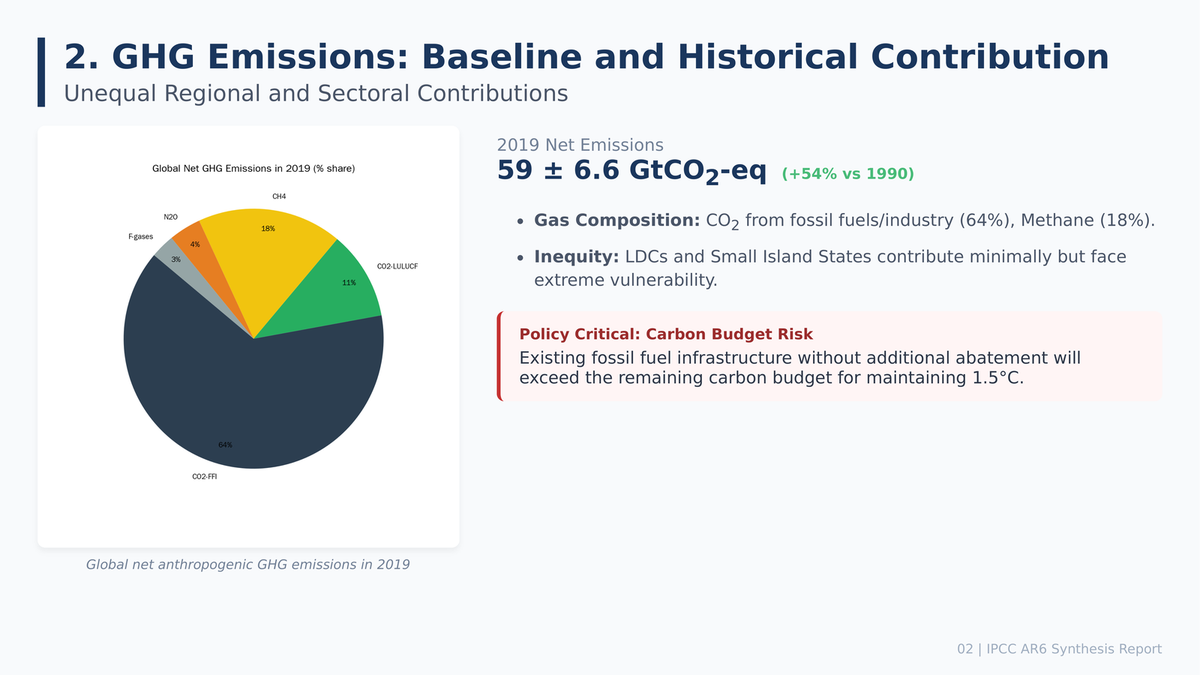}{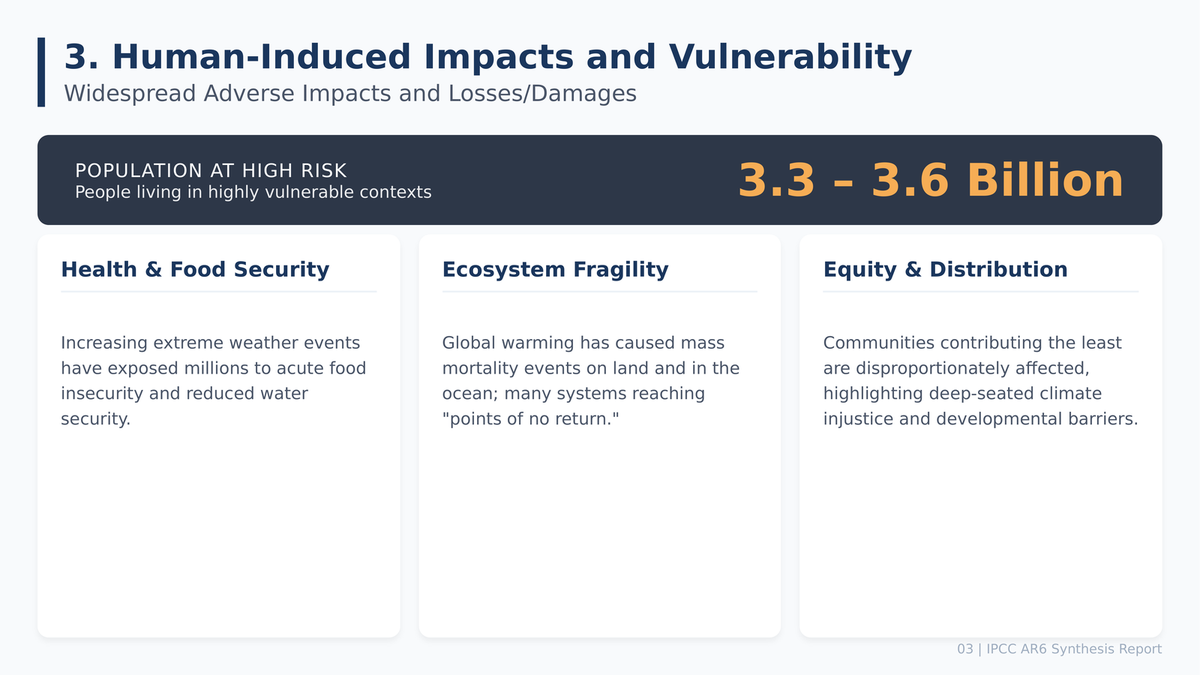}{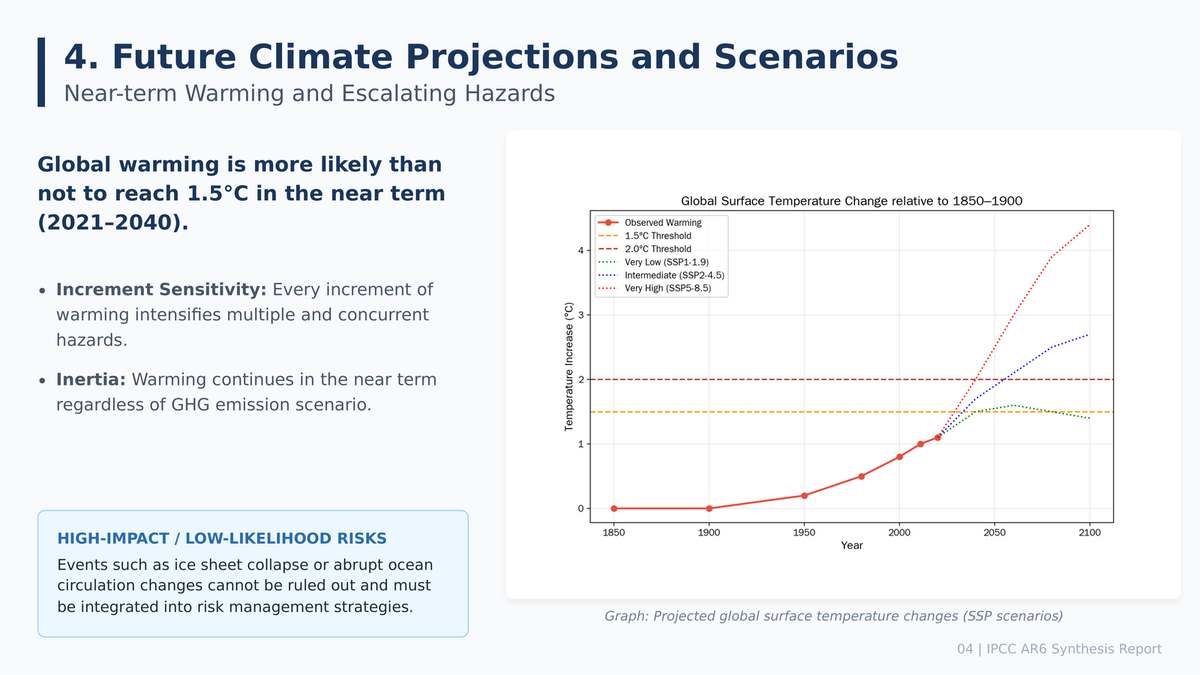}
\deckshowcaserow{DeepPresenter / Decision maker: AudCov. 0.077, Correct. 0.971, SafeEff 0.104}{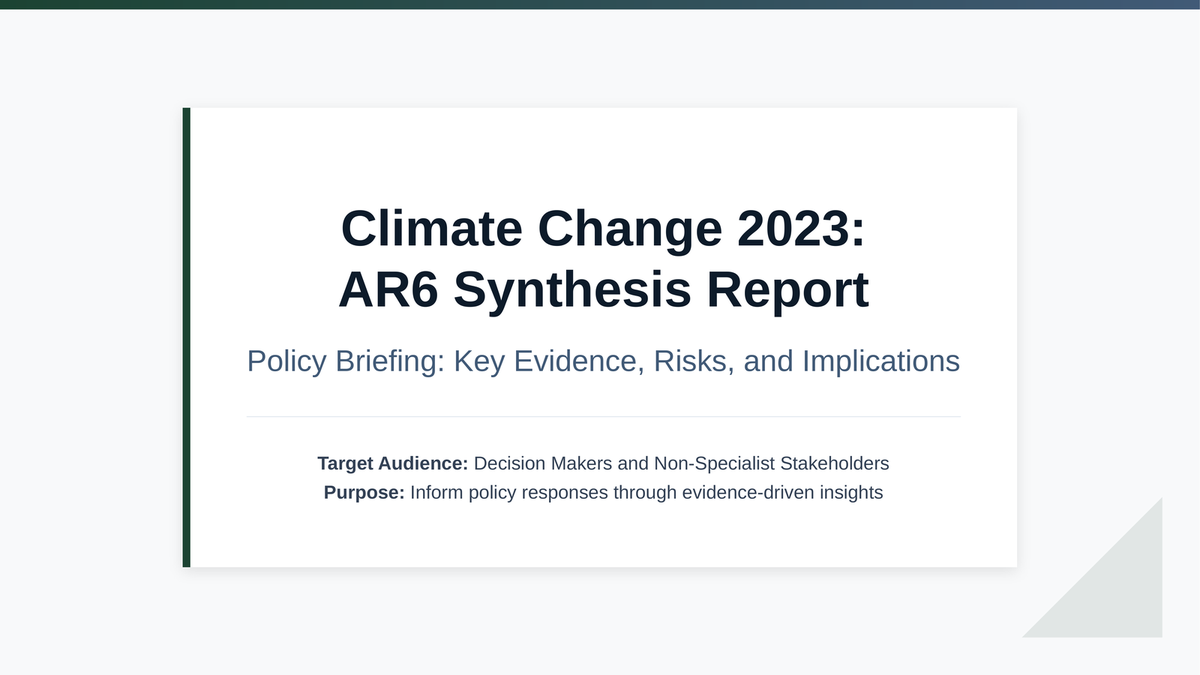}{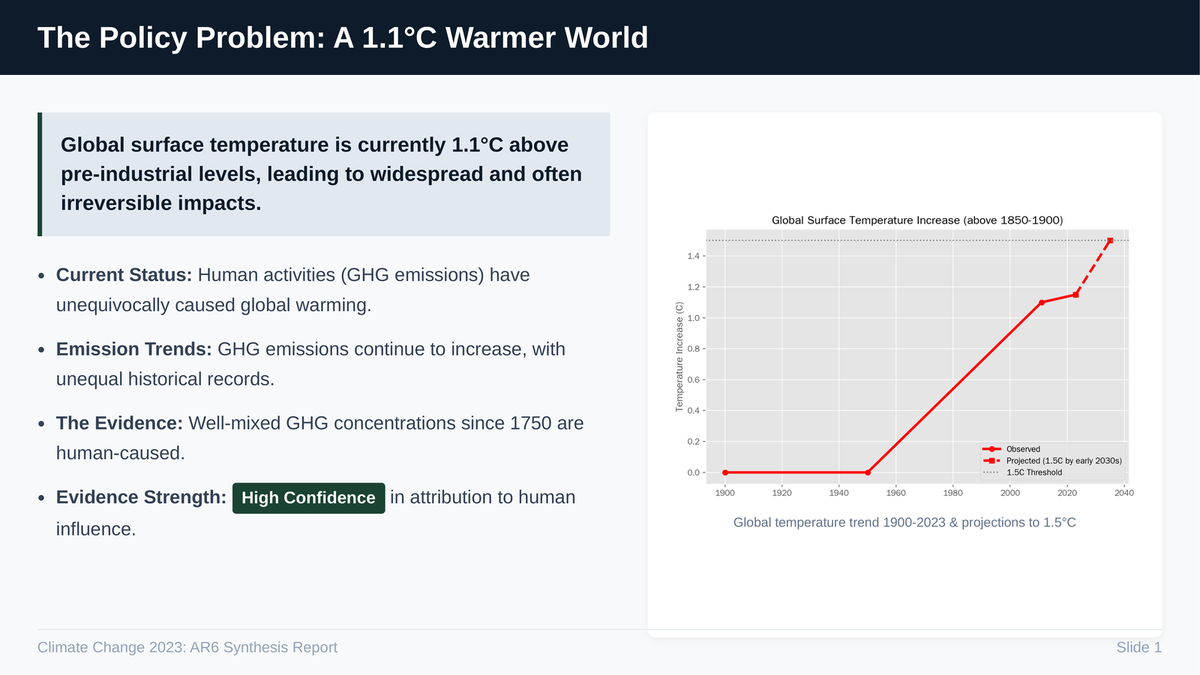}{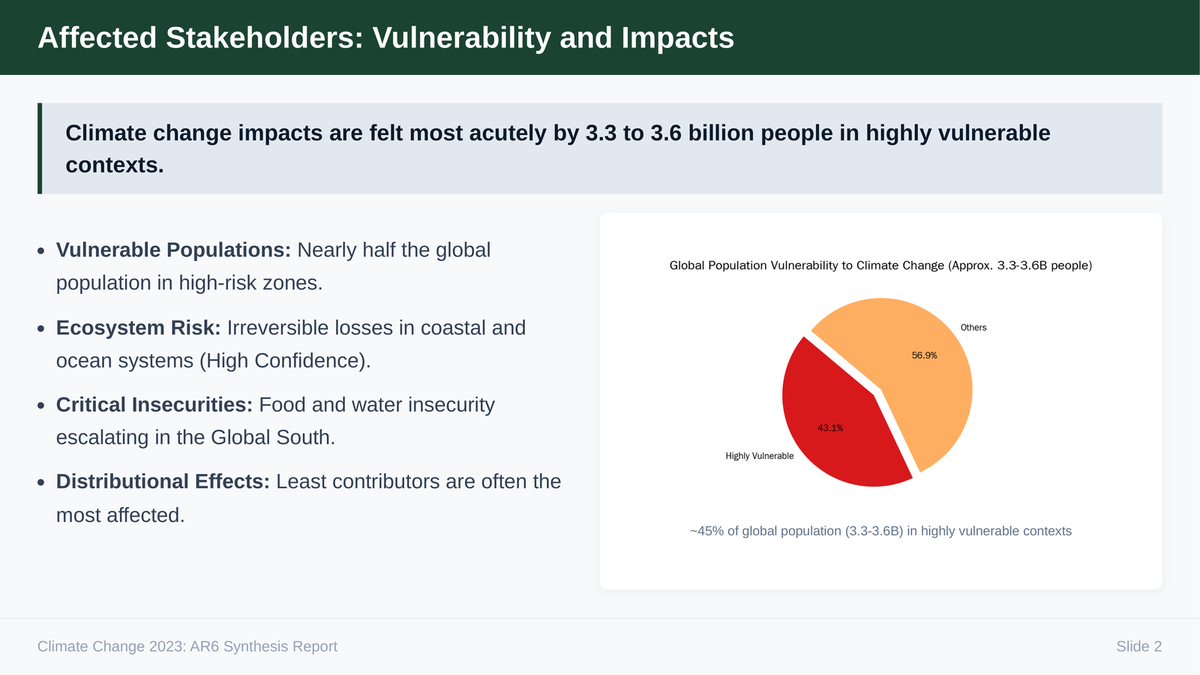}{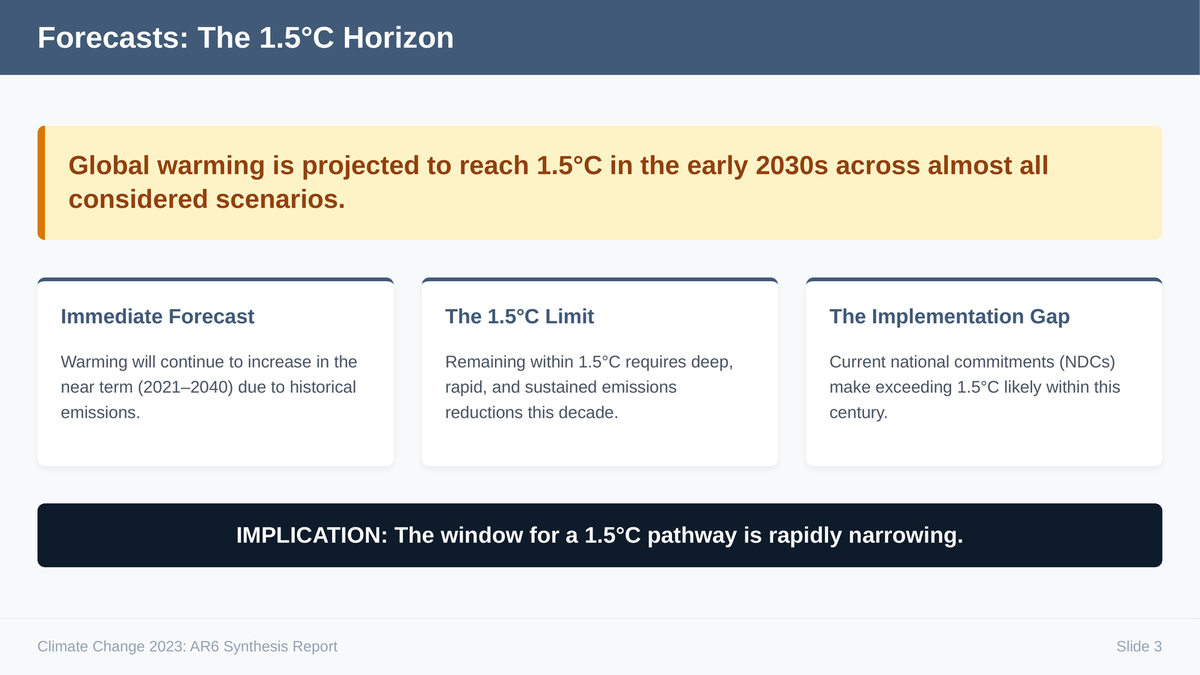}{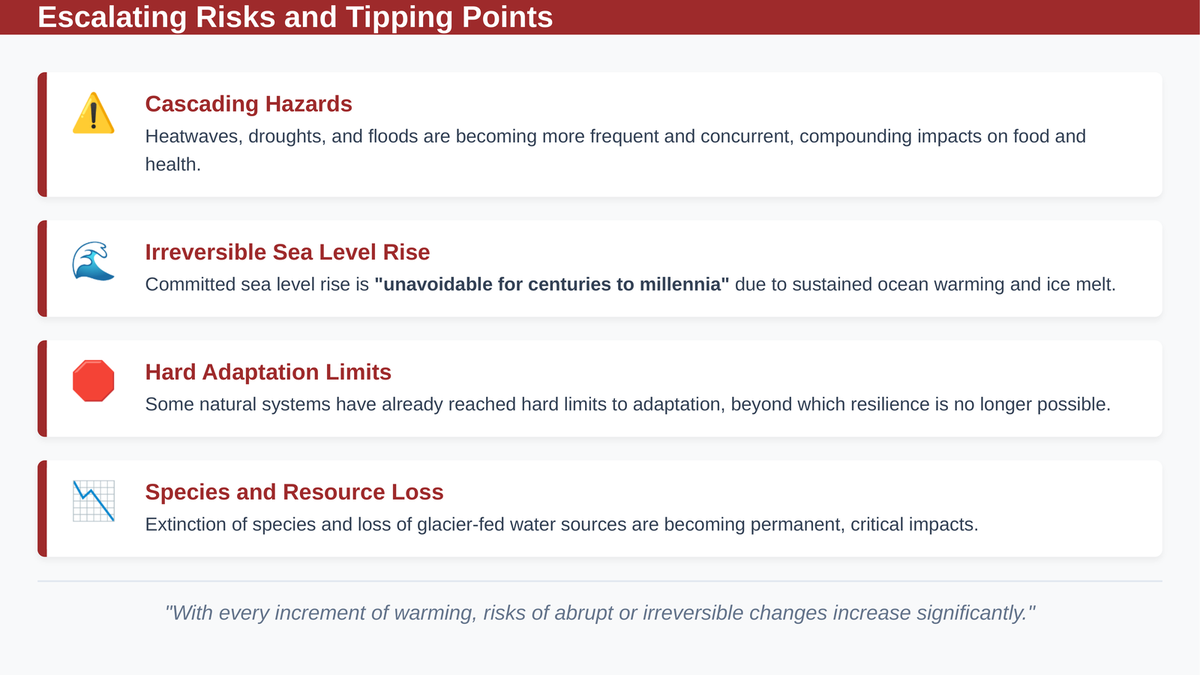}
\deckshowcaserow{SlideTailor / Decision maker: AudCov. 0.231, Correct. 0.908, SafeEff 0.525}{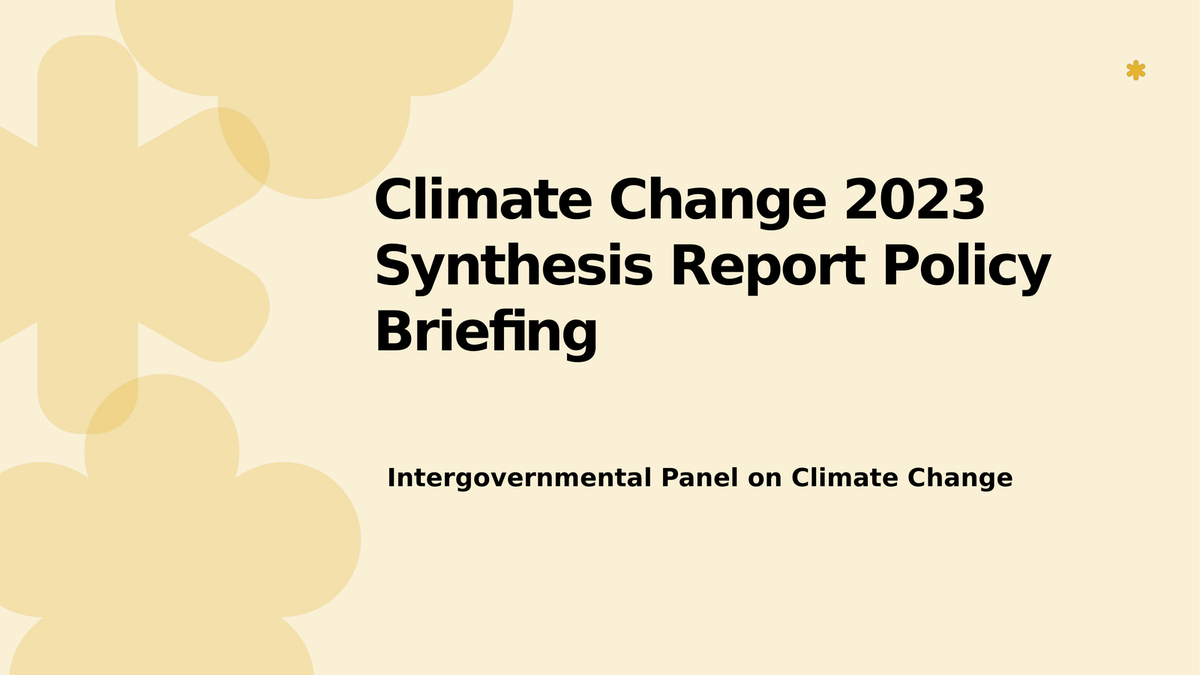}{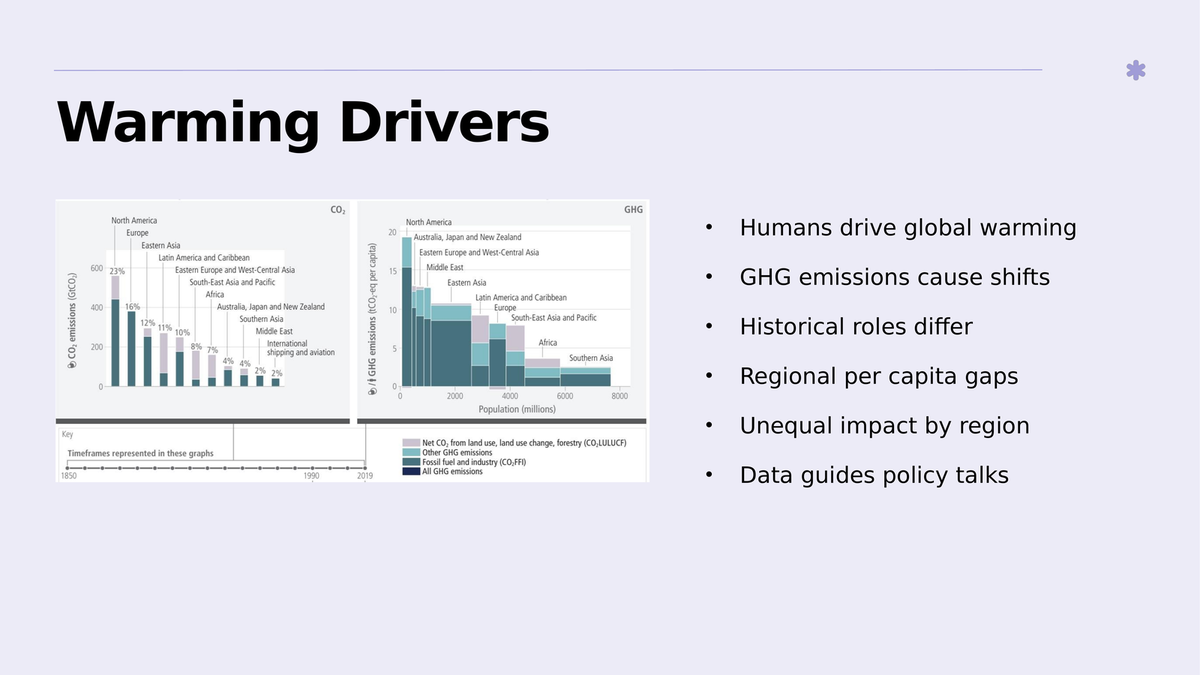}{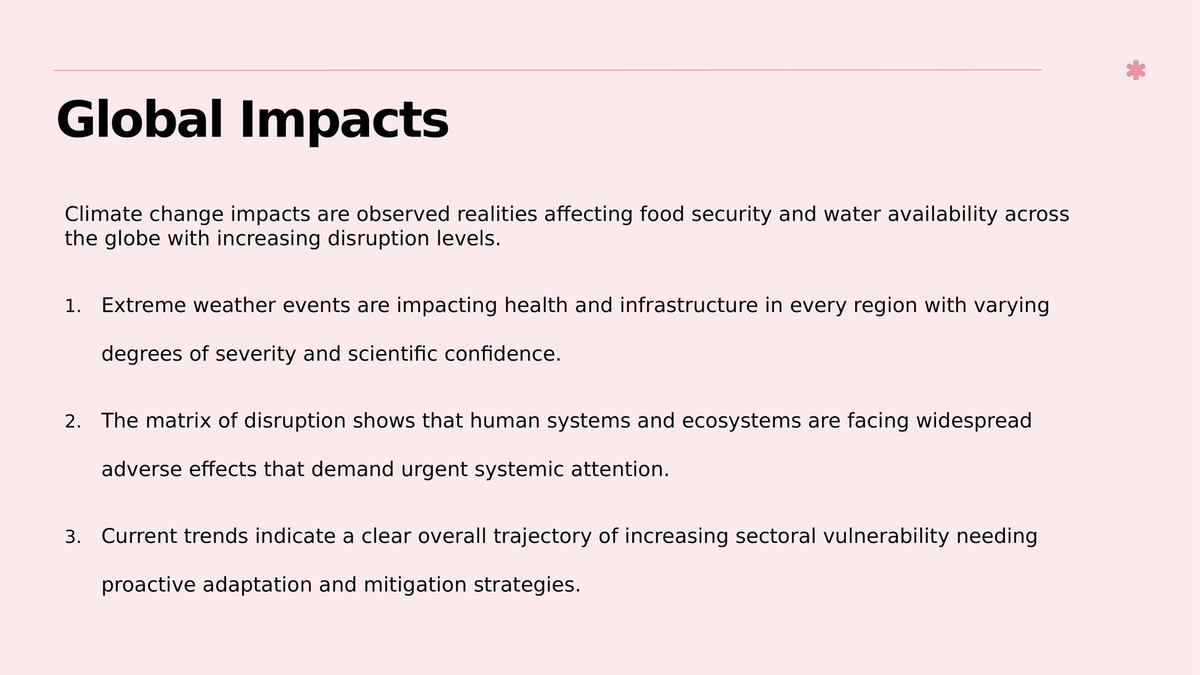}{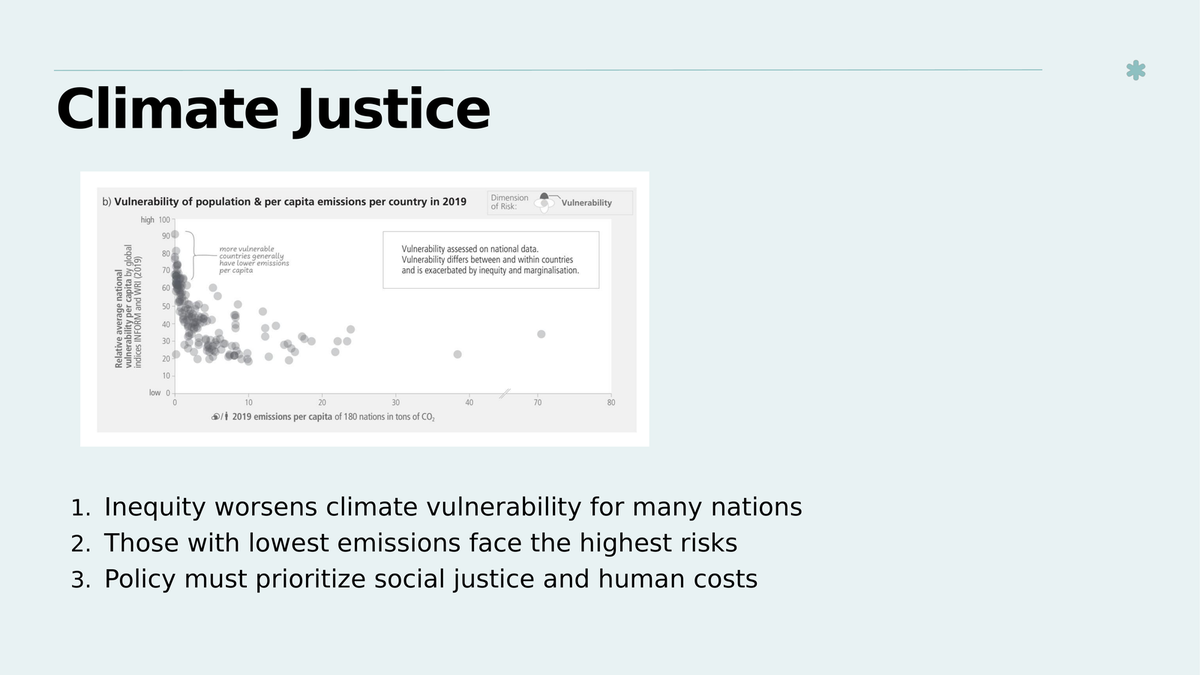}{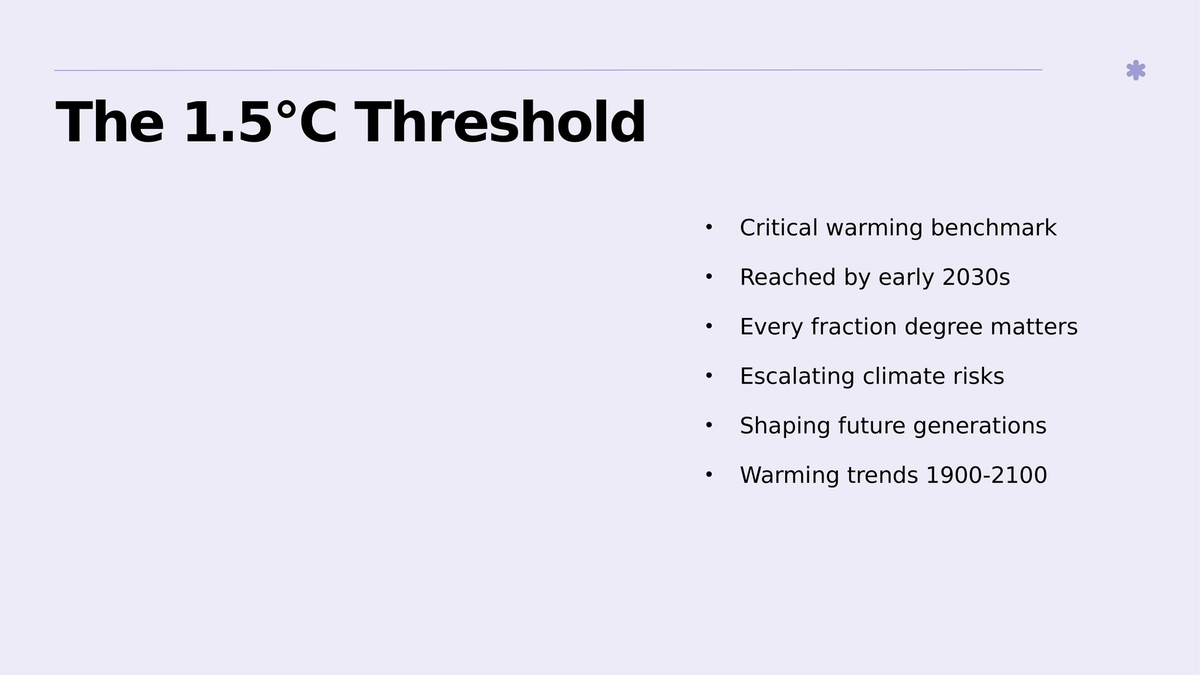}
\deckshowcaserow{NotebookLM / Specialist: AudCov. 0.750, Correct. 0.894, SafeEff 3.301}{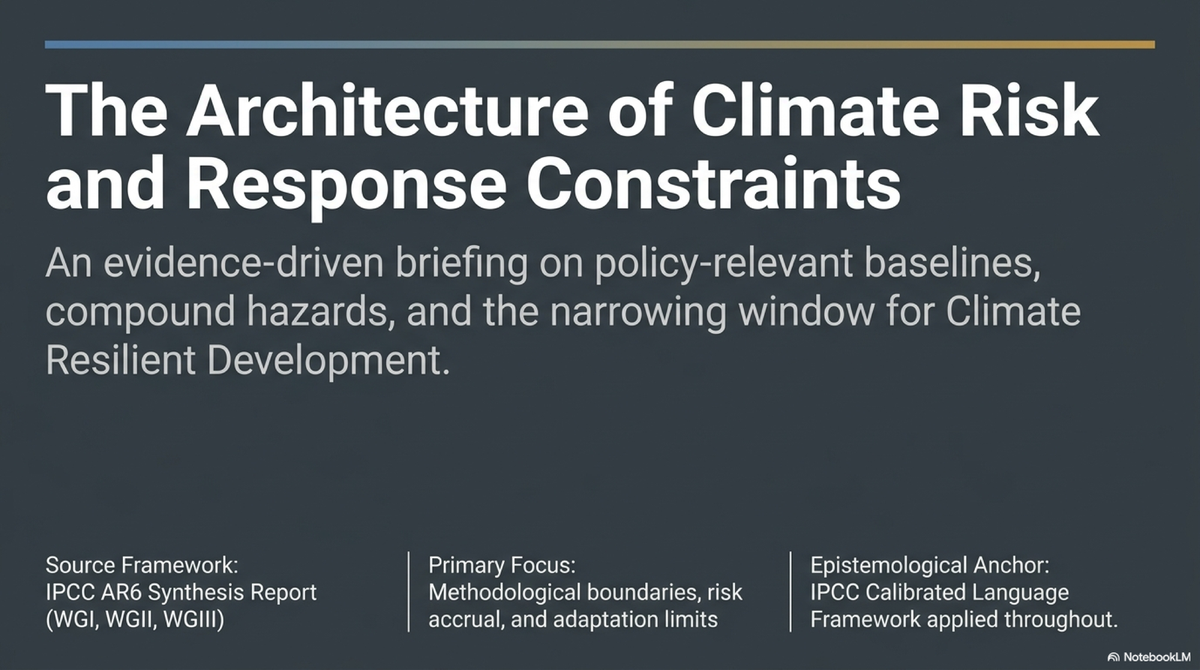}{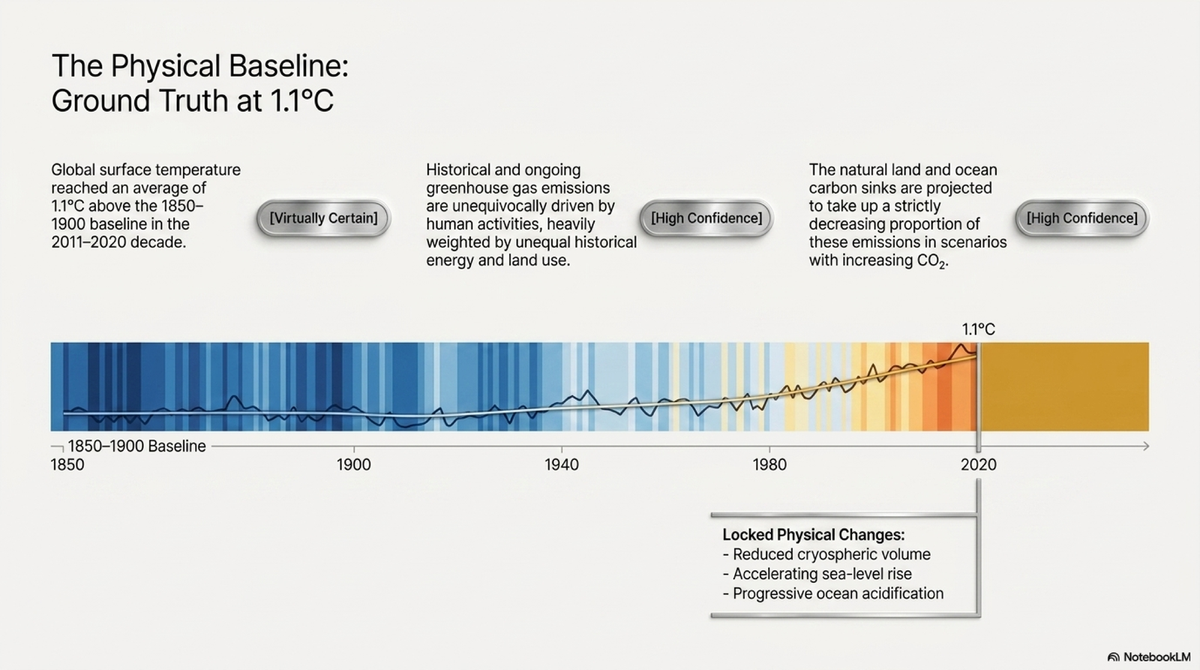}{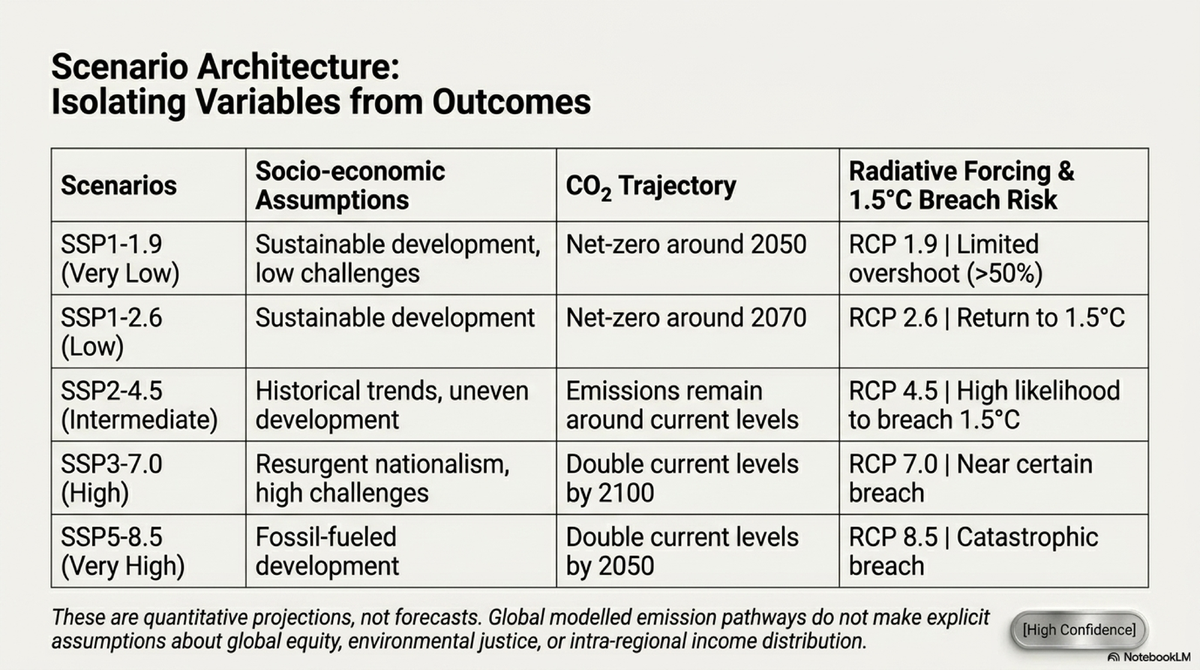}{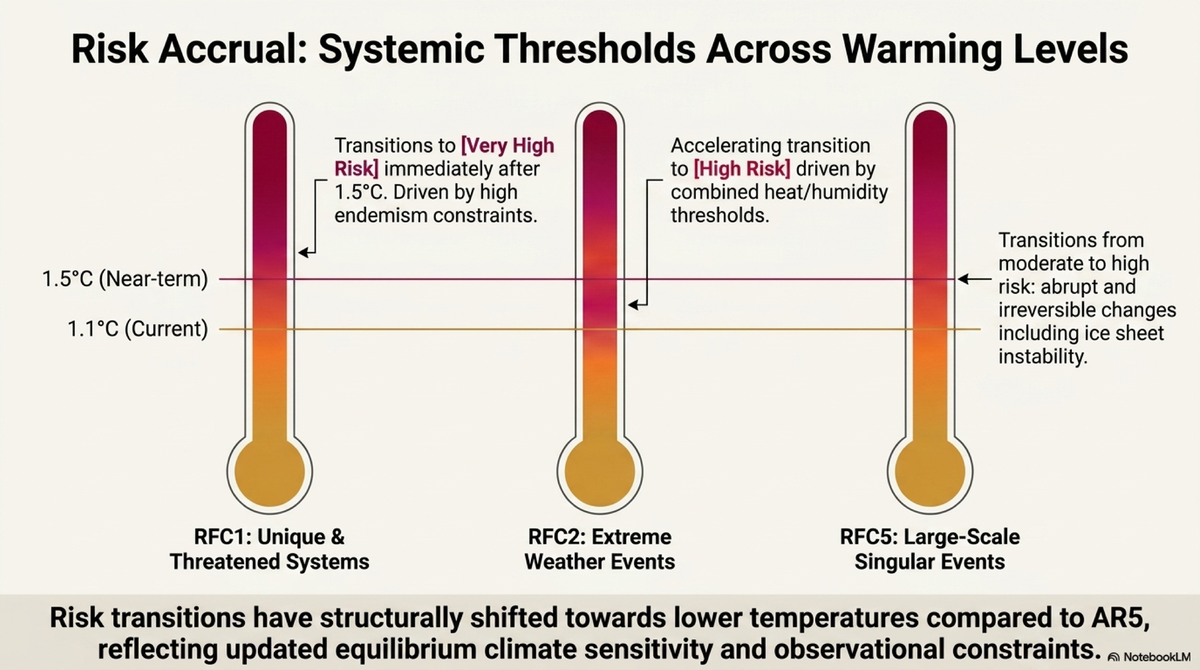}{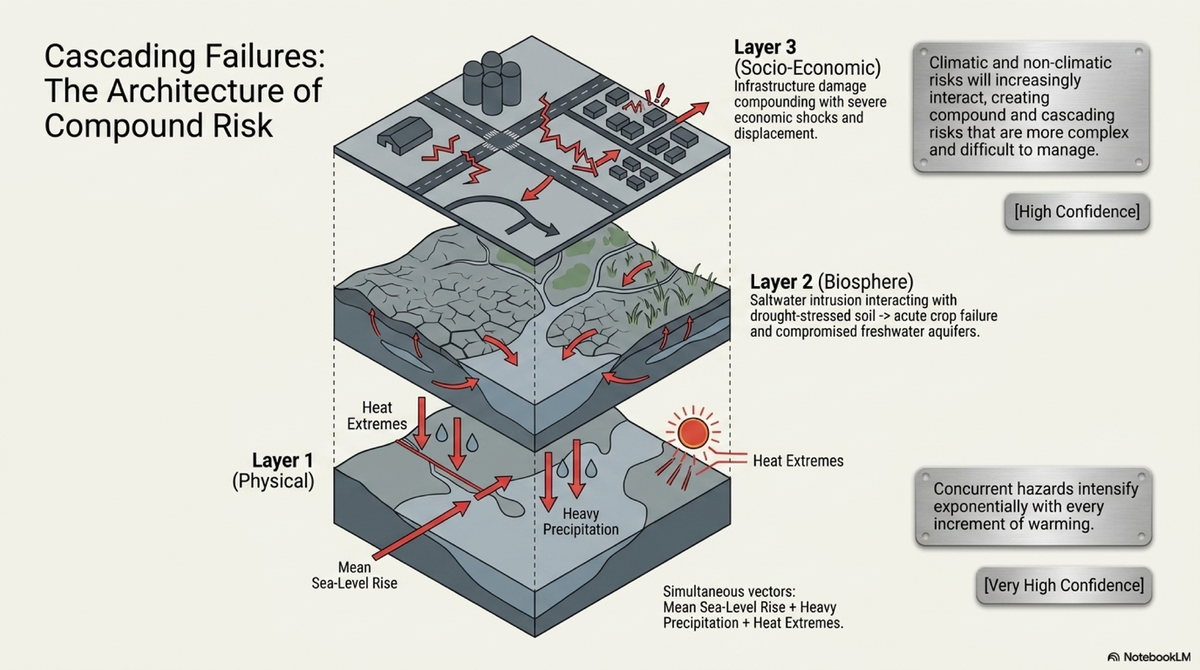}
\deckshowcaserow{NotebookLM / Decision maker: AudCov. 1.000, Correct. 0.877, SafeEff 3.257}{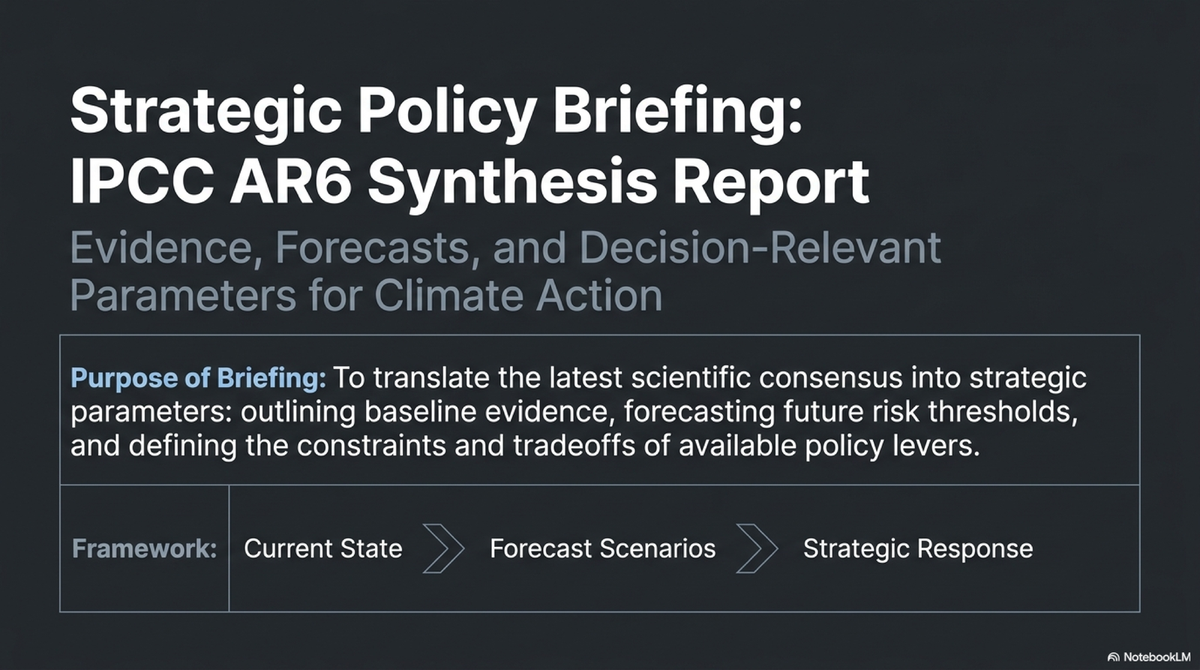}{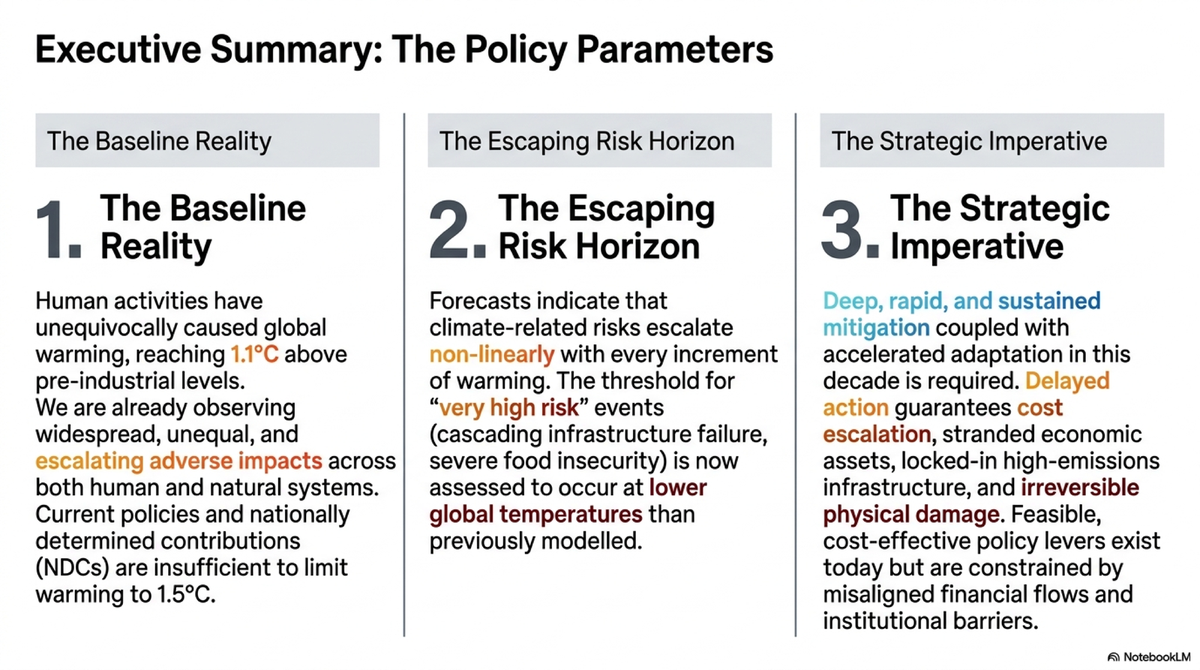}{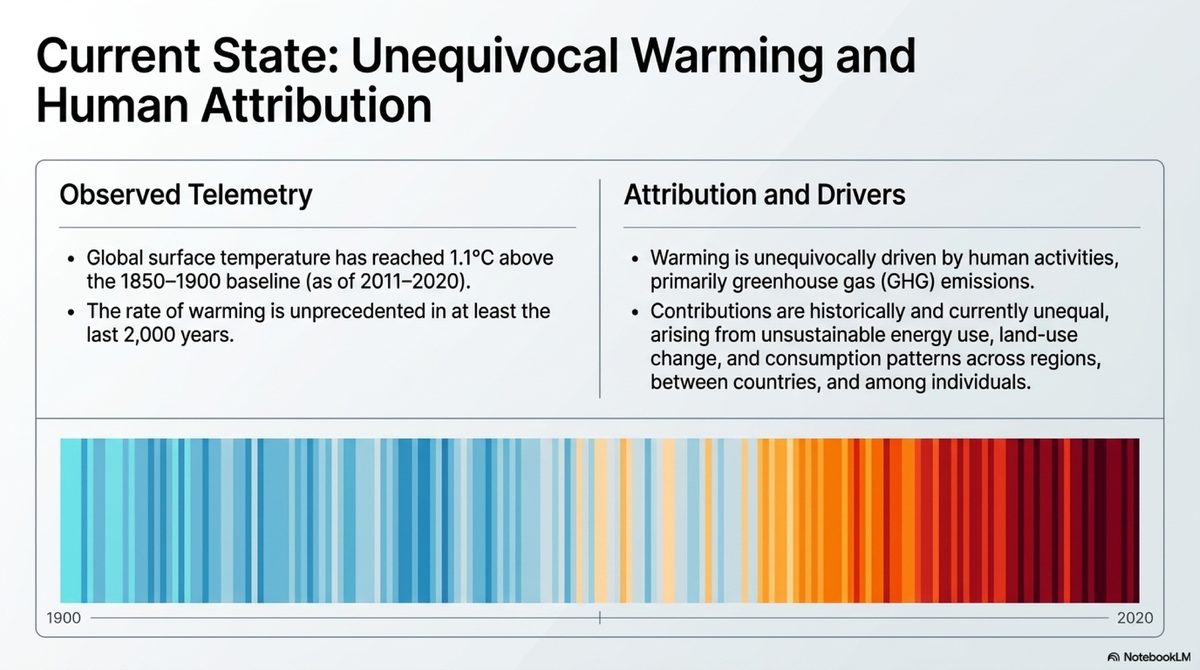}{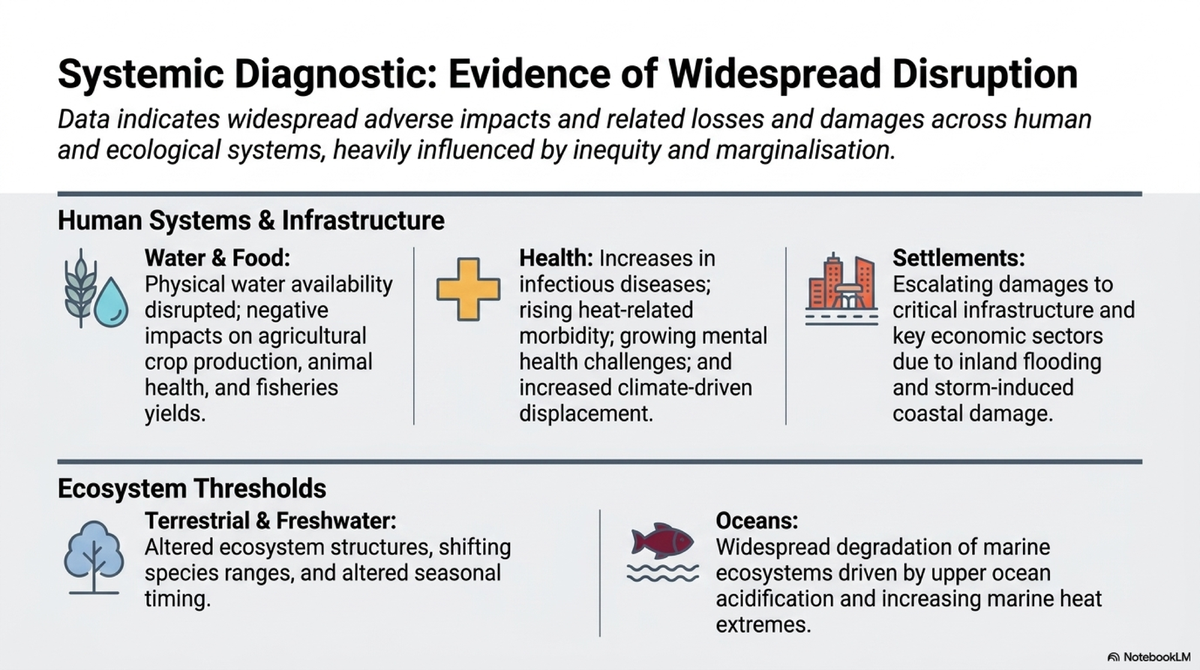}{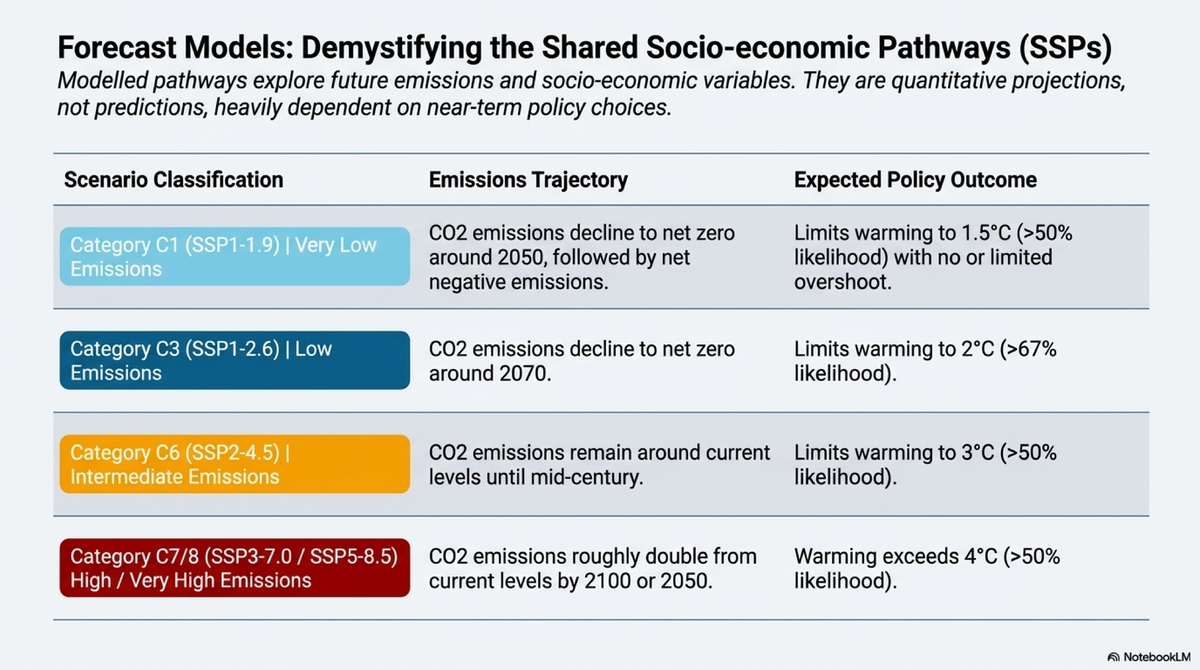}
}
\caption{Deck showcase for Climate Change 2023: AR6 Synthesis Report. This climate-policy case shows different recovered utility across audiences and generators, including a strongly grounded NotebookLM decision-maker row.}
\label{fig:deck_showcase_case085}
\end{figure}

\begin{figure}[!p]
\centering
{\setlength{\tabcolsep}{0pt}%
\deckshowcaserow{DeepPresenter / Learner: AudCov. 0.600, Correct. 0.814, SafeEff 0.315}{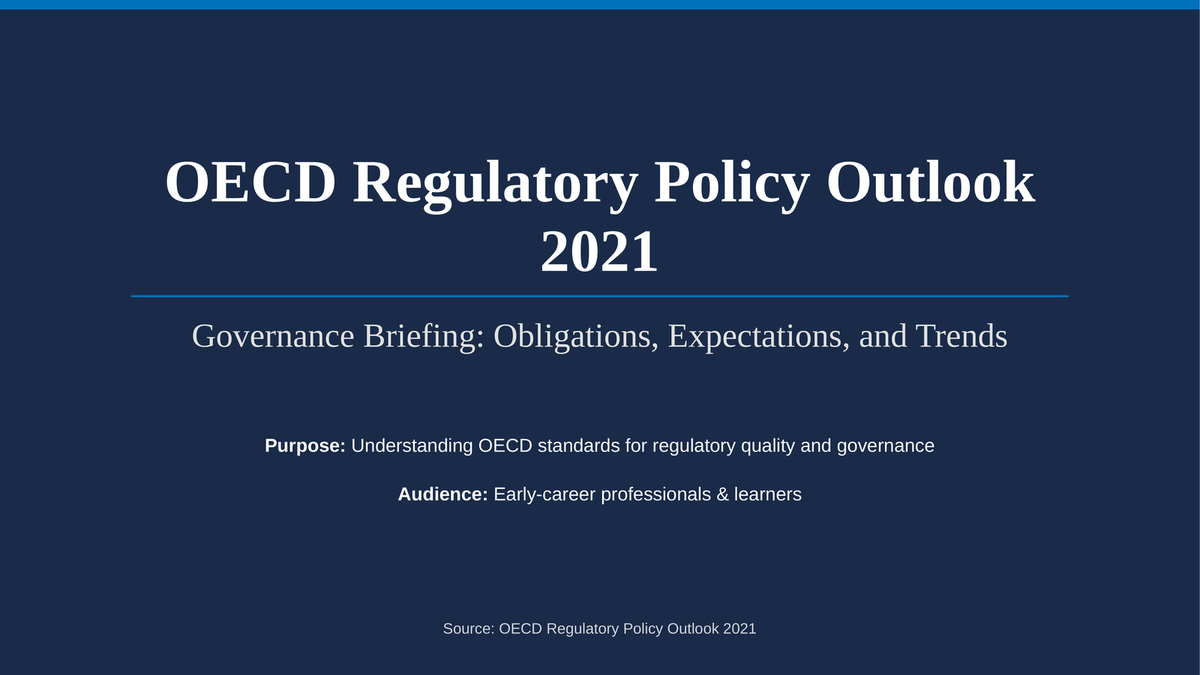}{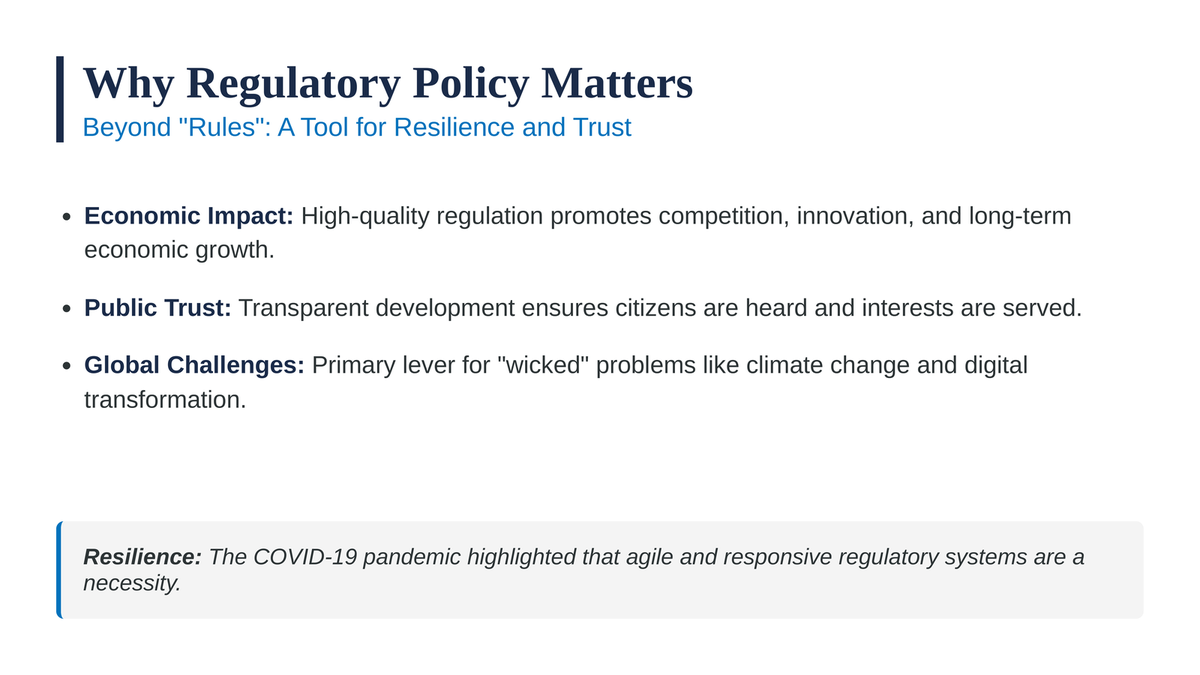}{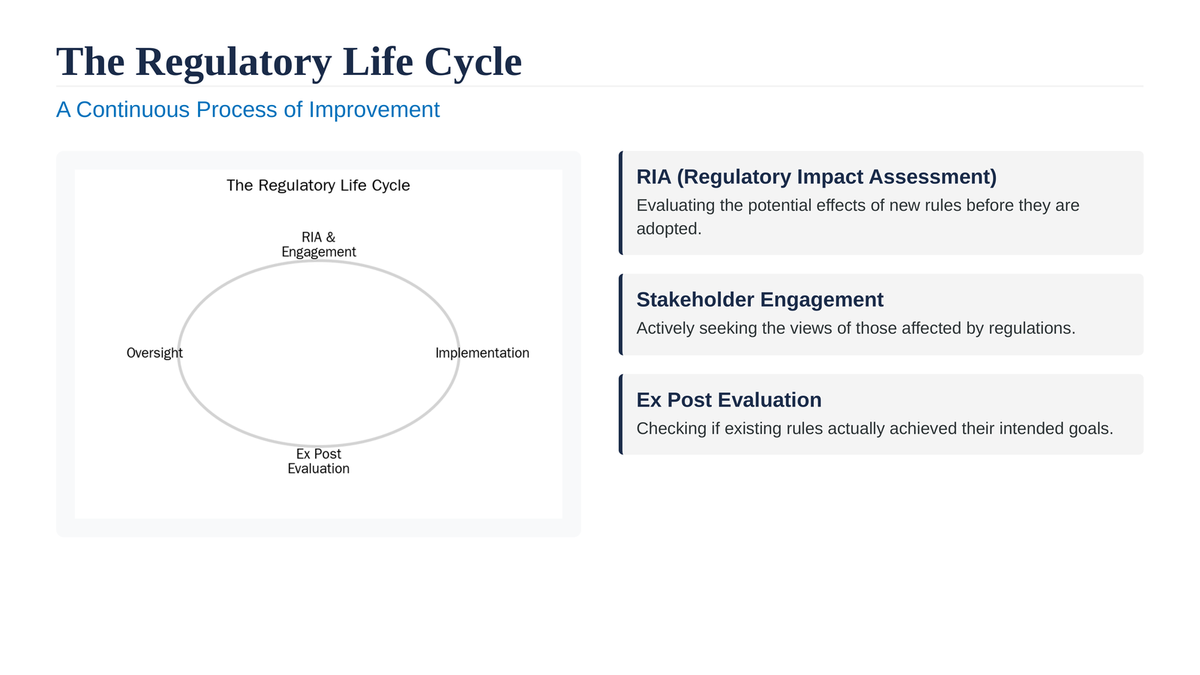}{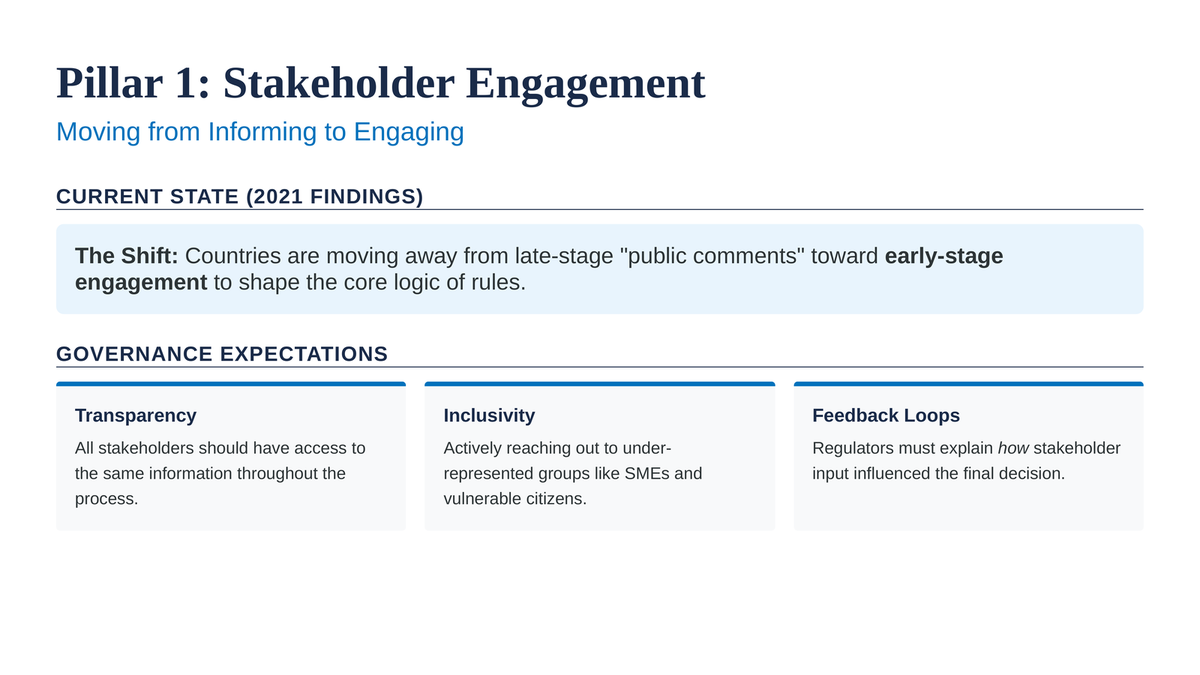}{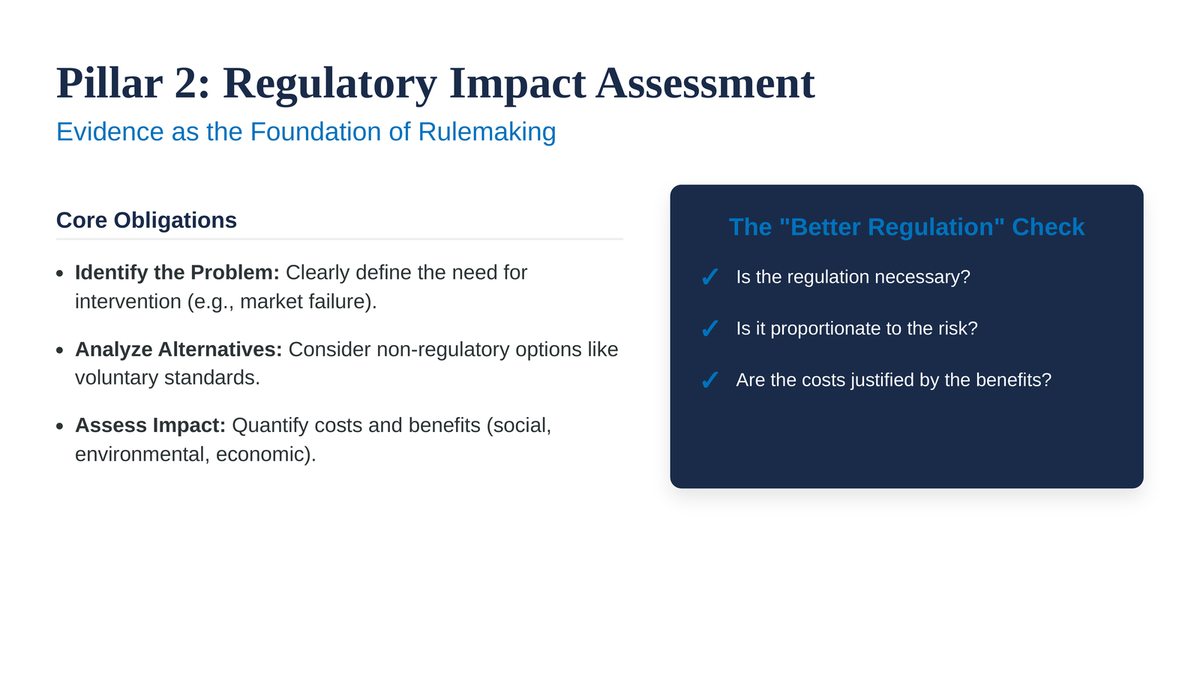}
\deckshowcaserow{DeepPresenter / Decision maker: AudCov. 0.833, Correct. 0.795, SafeEff 0.539}{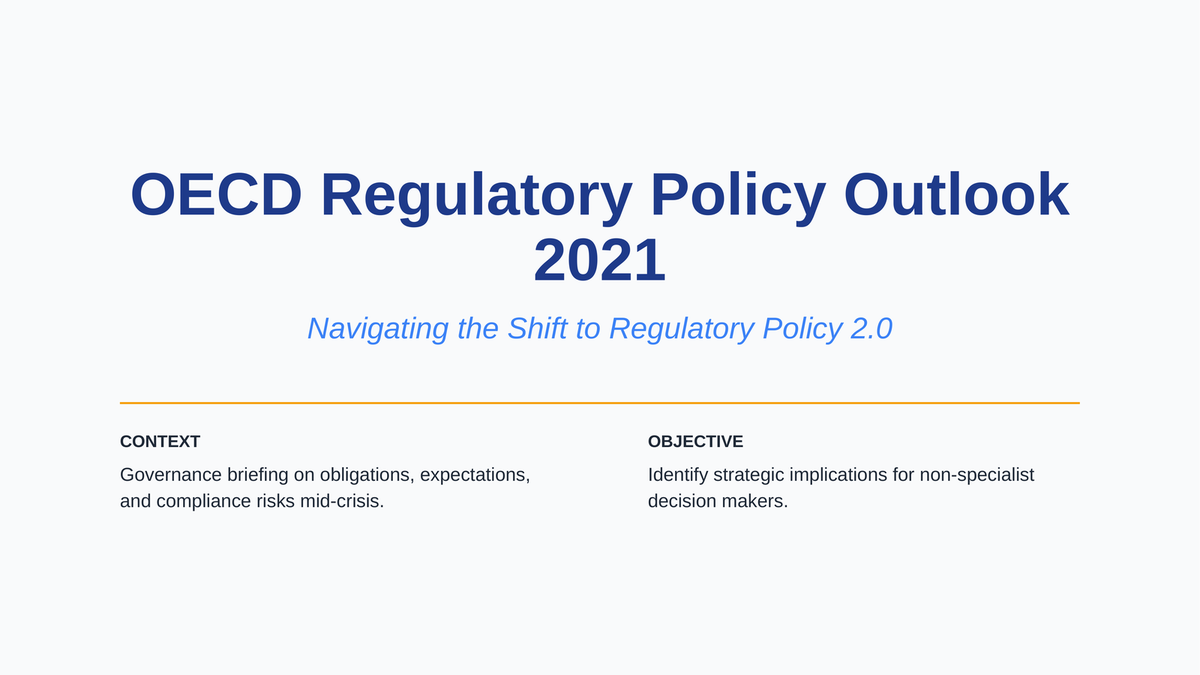}{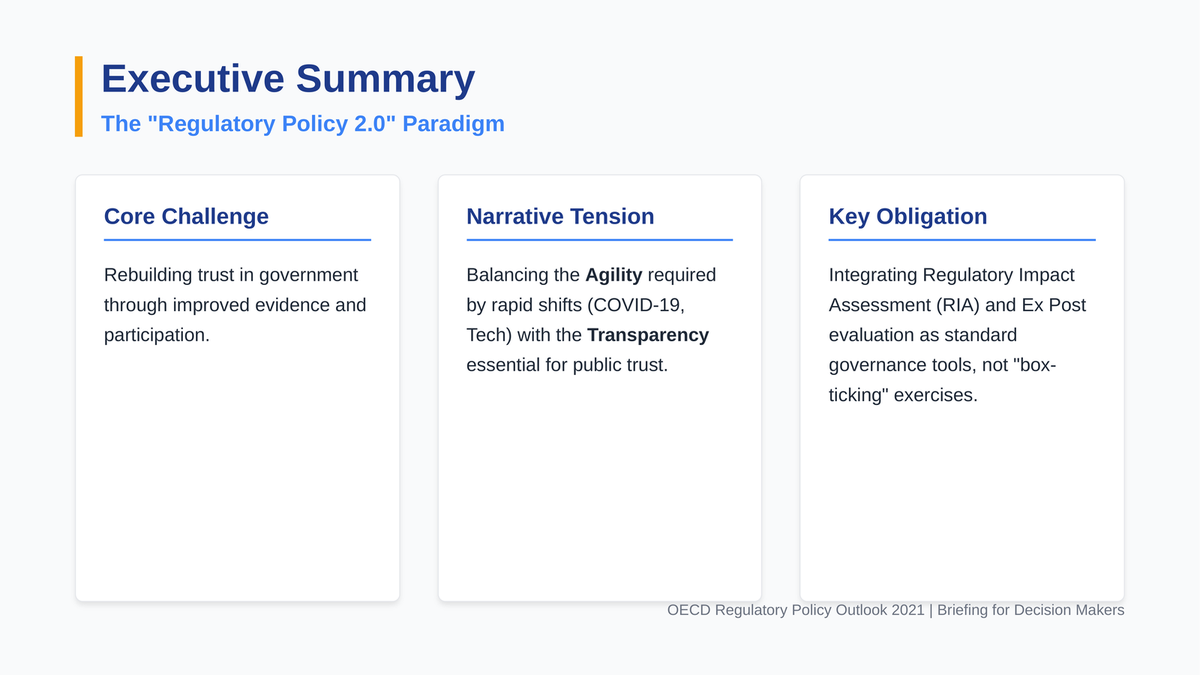}{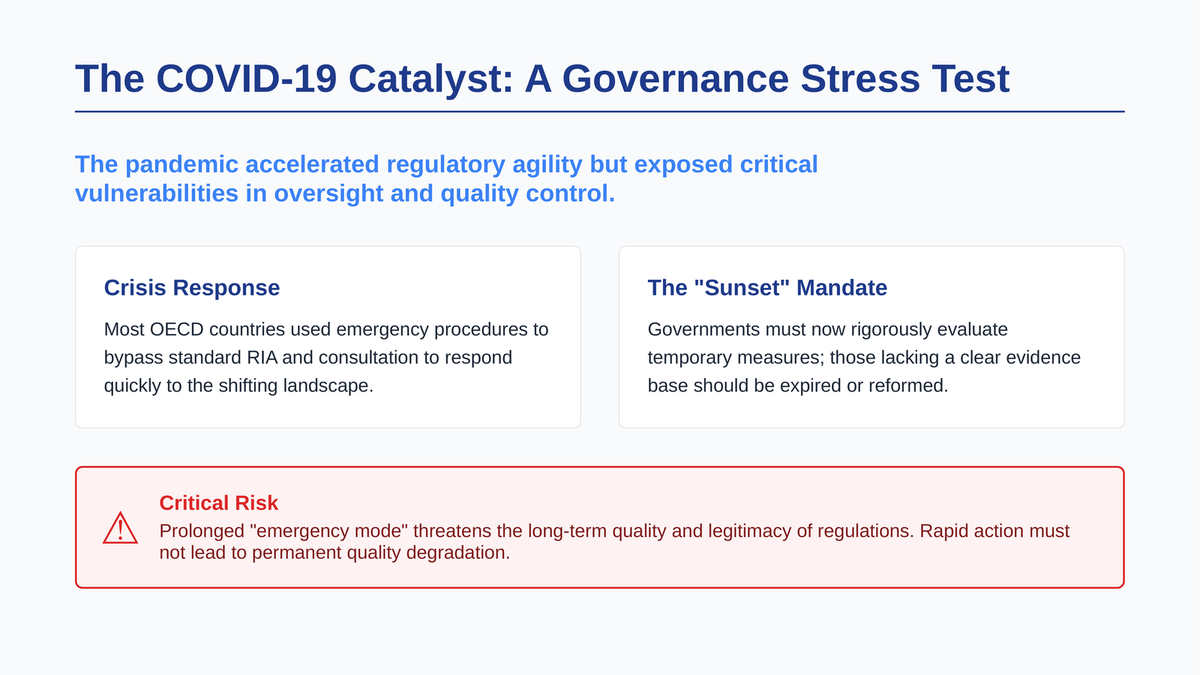}{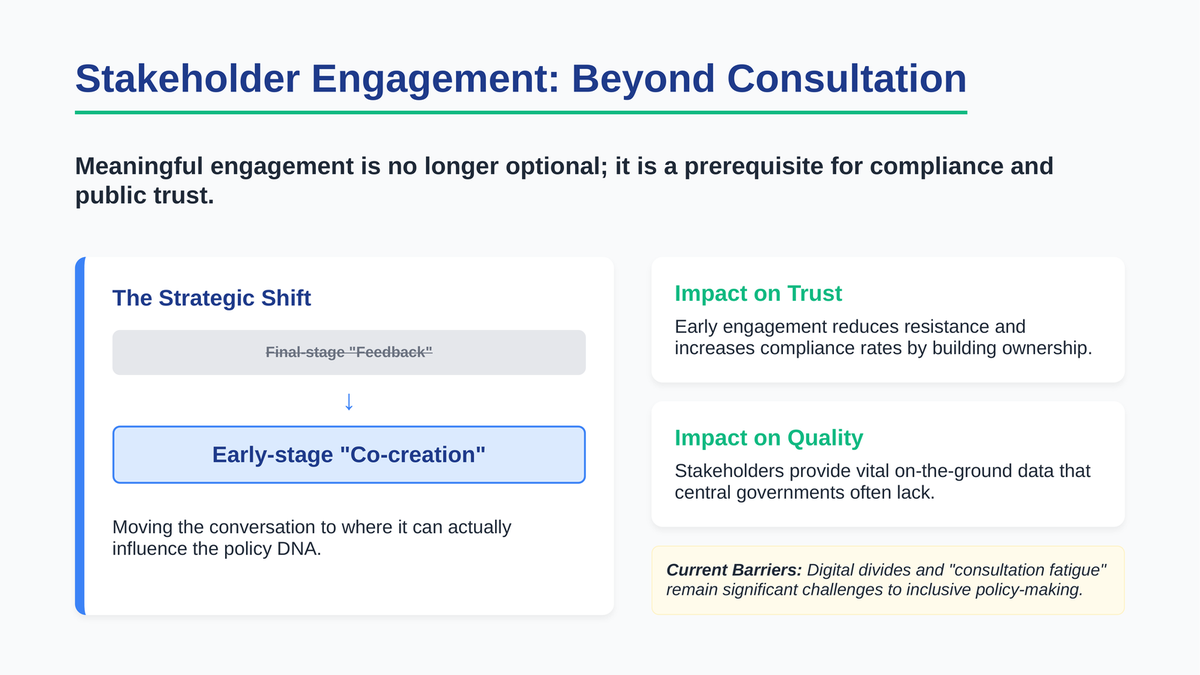}{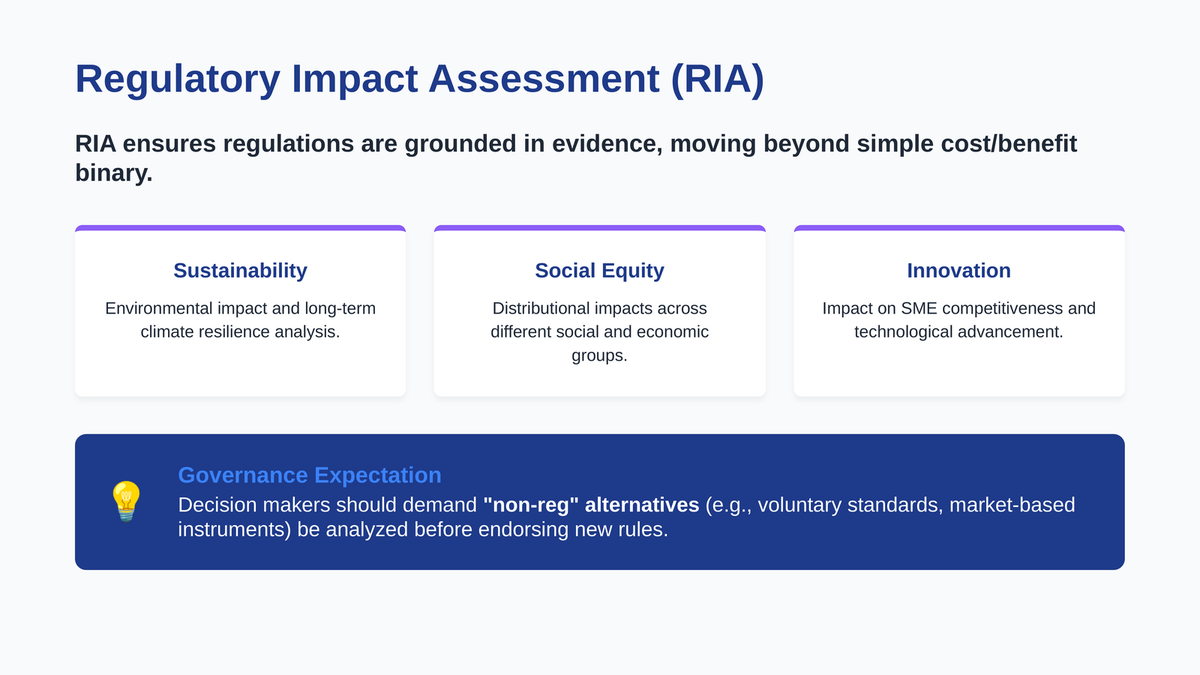}
\deckshowcaserow{SlideTailor / Learner: AudCov. 0.600, Correct. 0.927, SafeEff 0.605}{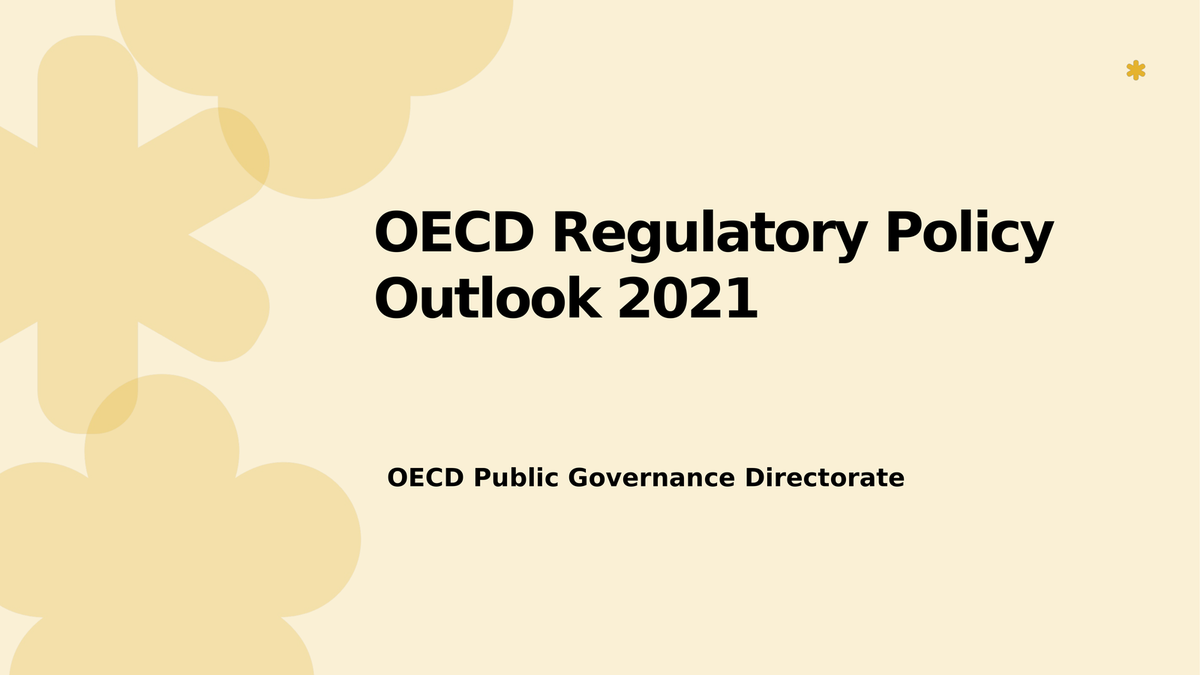}{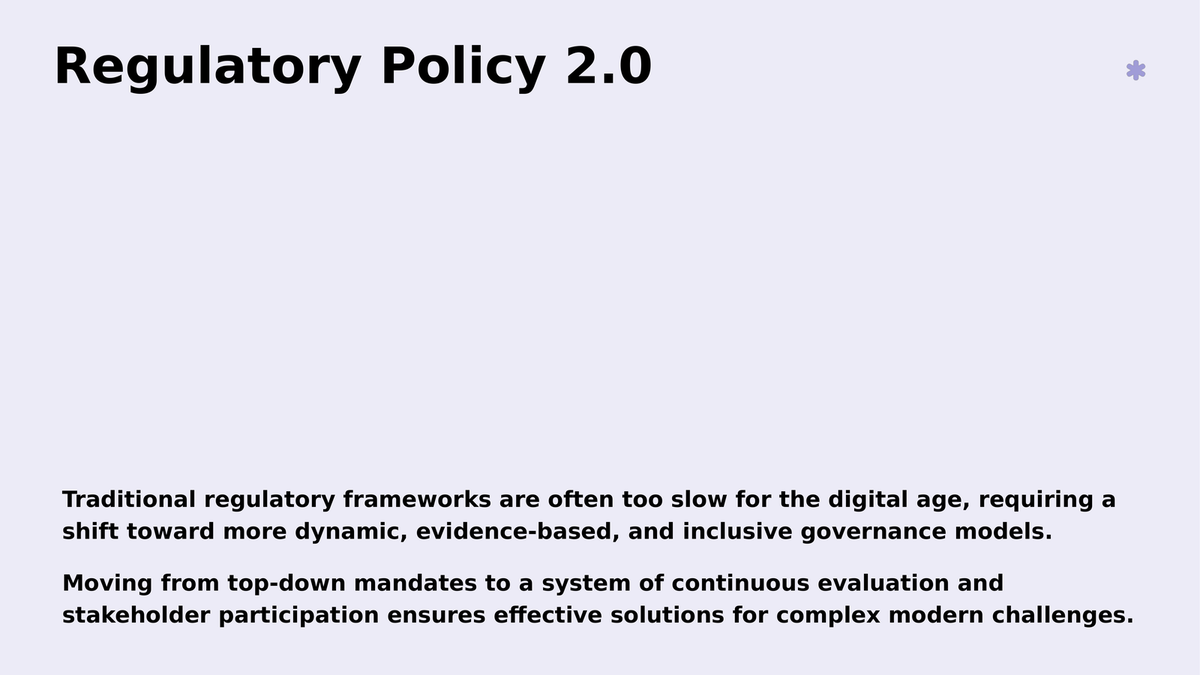}{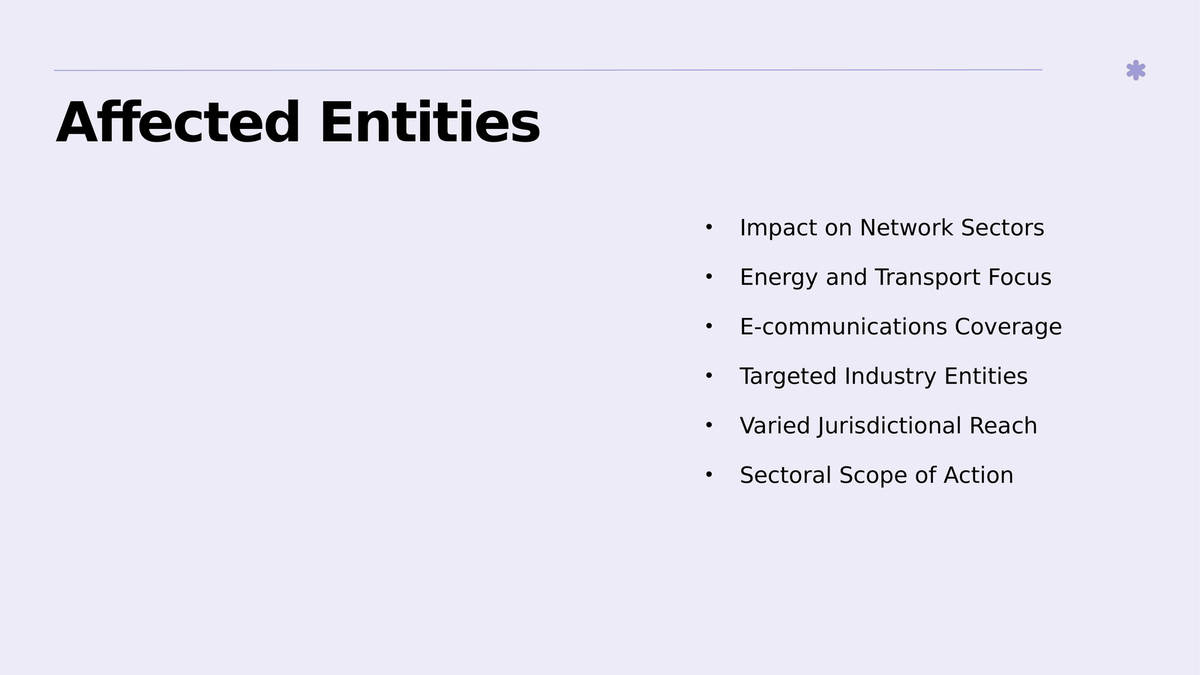}{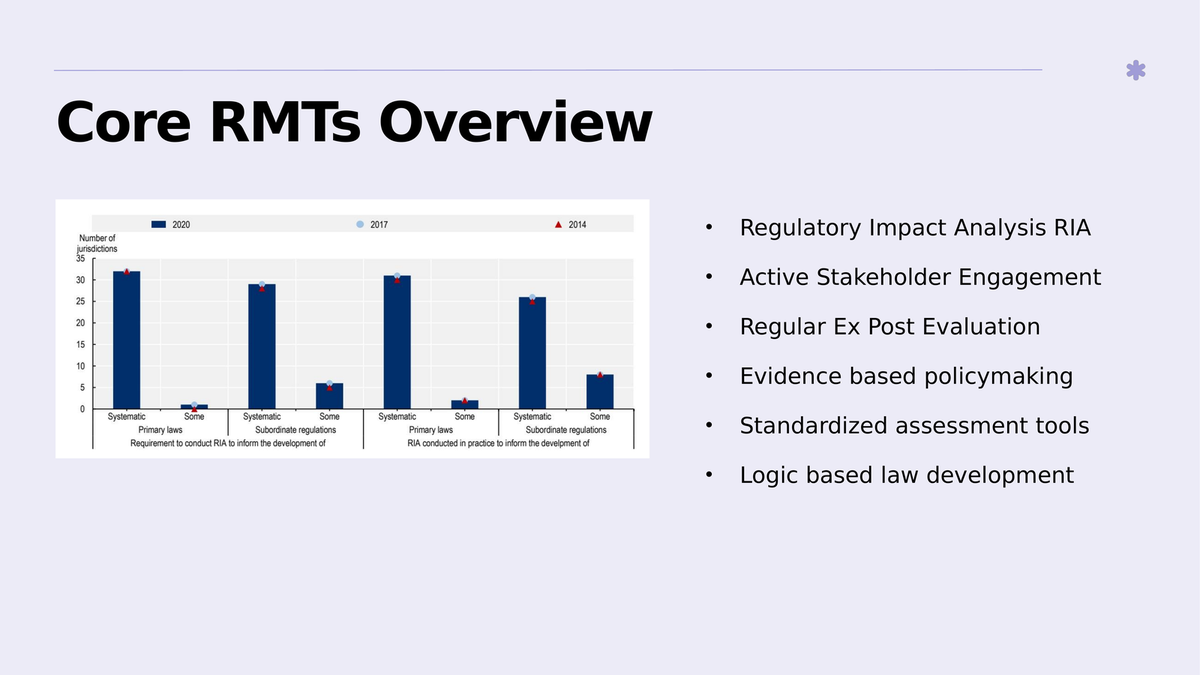}{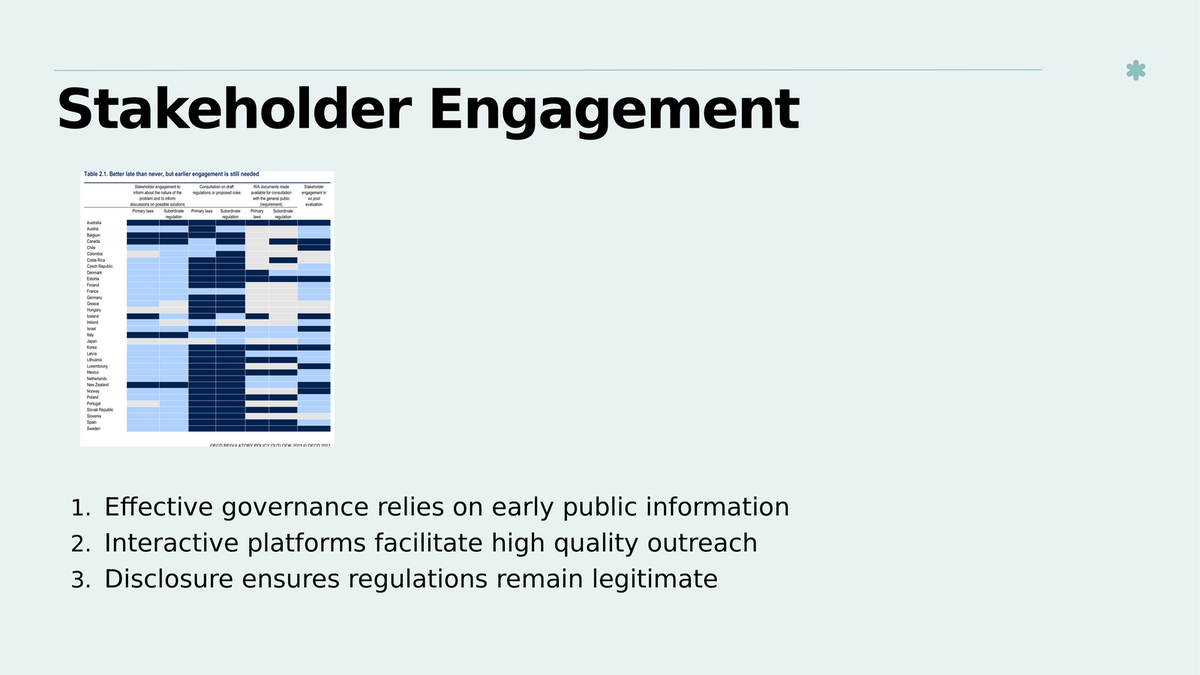}
}
\caption{Deck showcase for OECD Regulatory Policy Outlook 2021. This governance-report example compares learner and decision-maker decks.}
\label{fig:deck_showcase_case104}
\end{figure}

\begin{figure}[!p]
\centering
{\setlength{\tabcolsep}{0pt}%
\deckshowcaserow{DeepPresenter / Learner: AudCov. 0.631, Correct. 0.929, SafeEff 4.641}{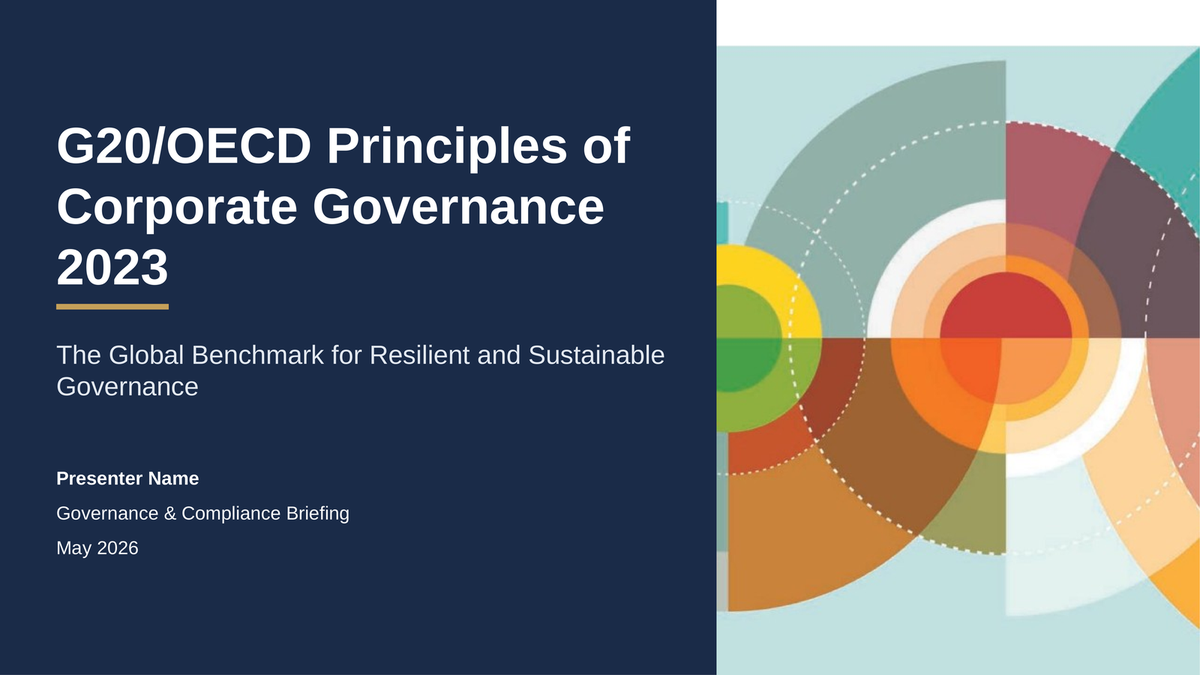}{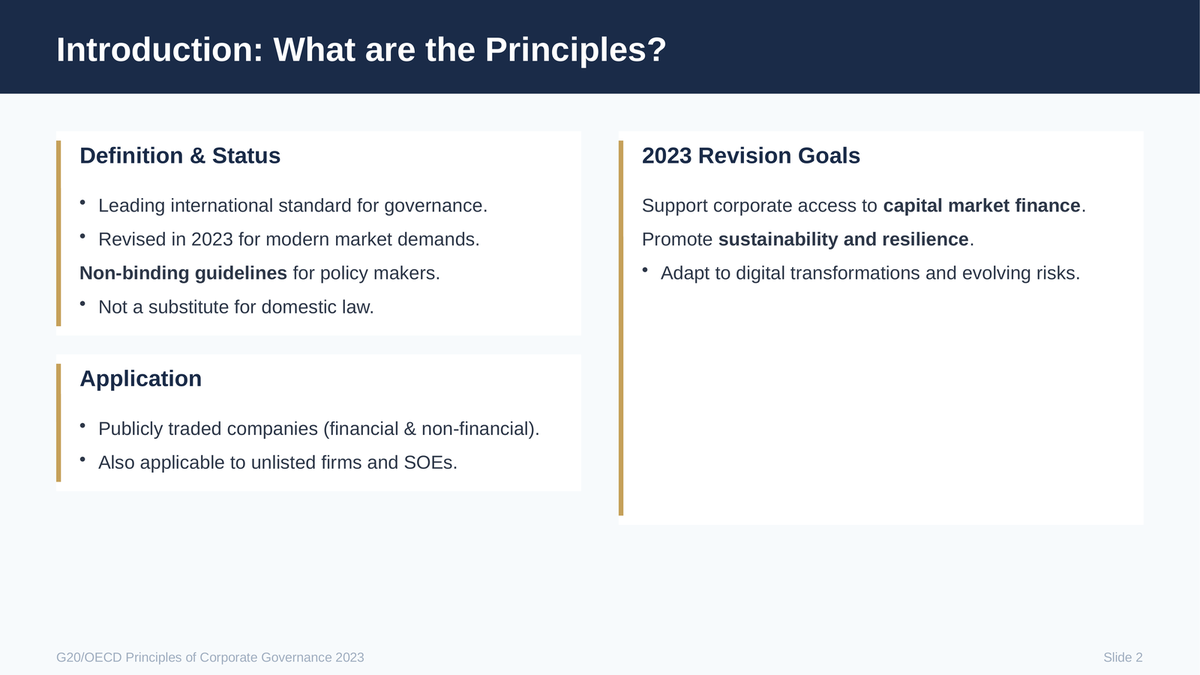}{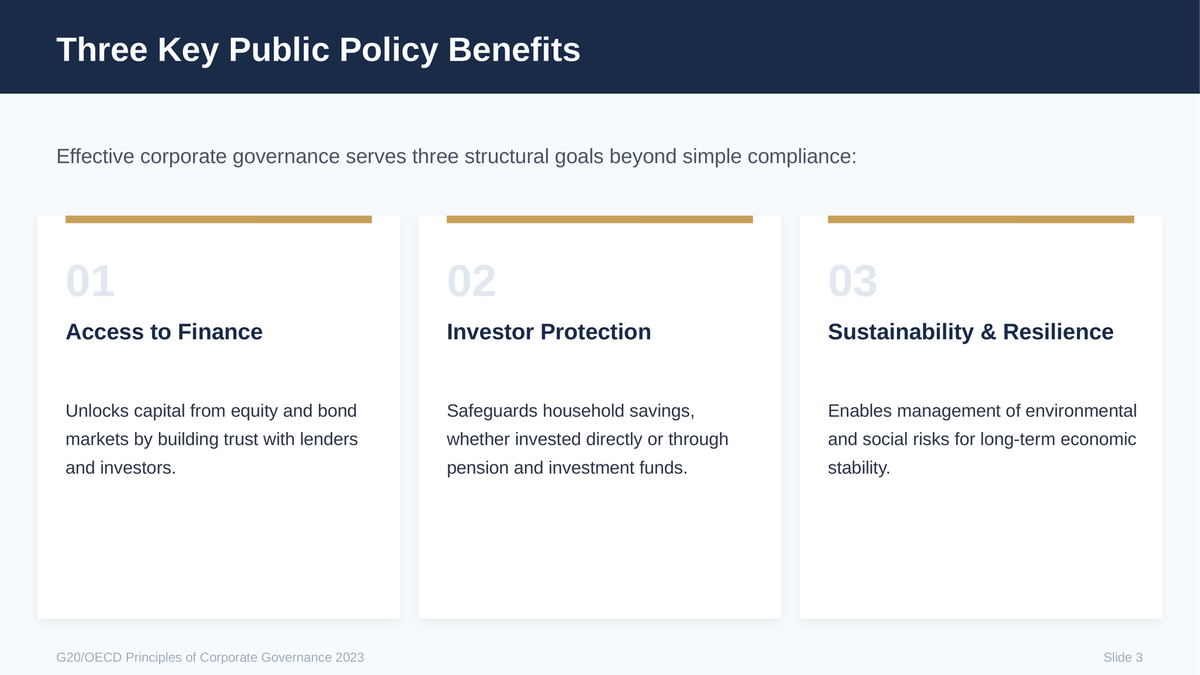}{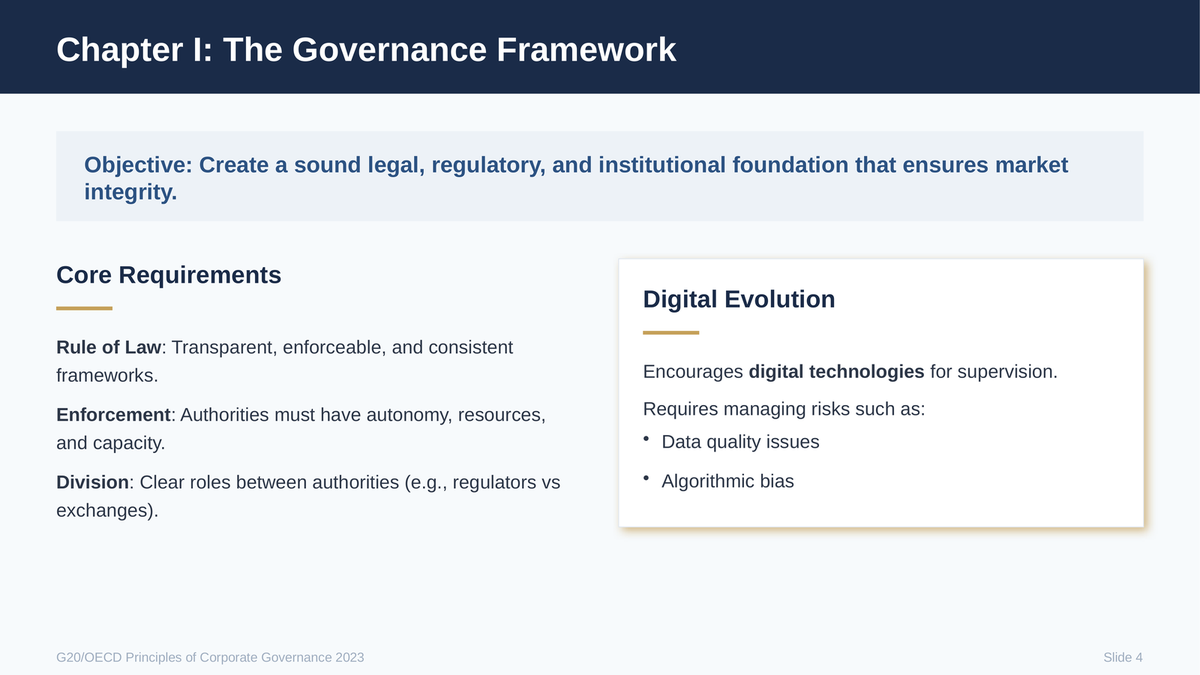}{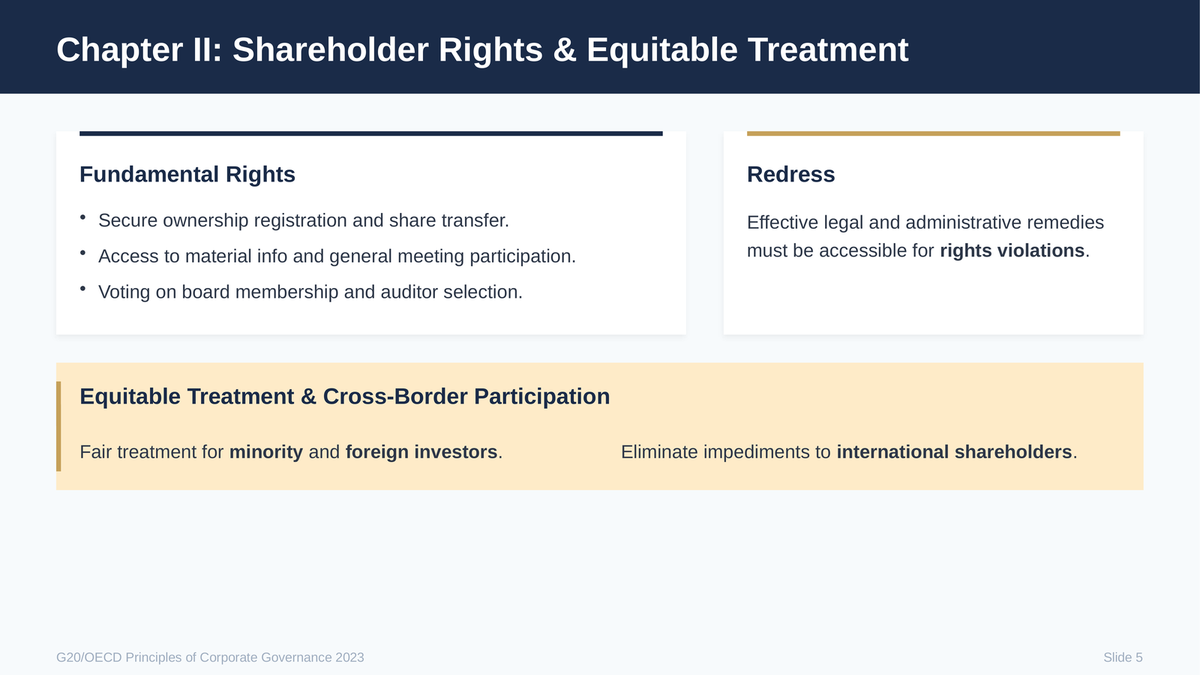}
\deckshowcaserow{DeepPresenter / Decision maker: AudCov. 0.514, Correct. 0.948, SafeEff 3.469}{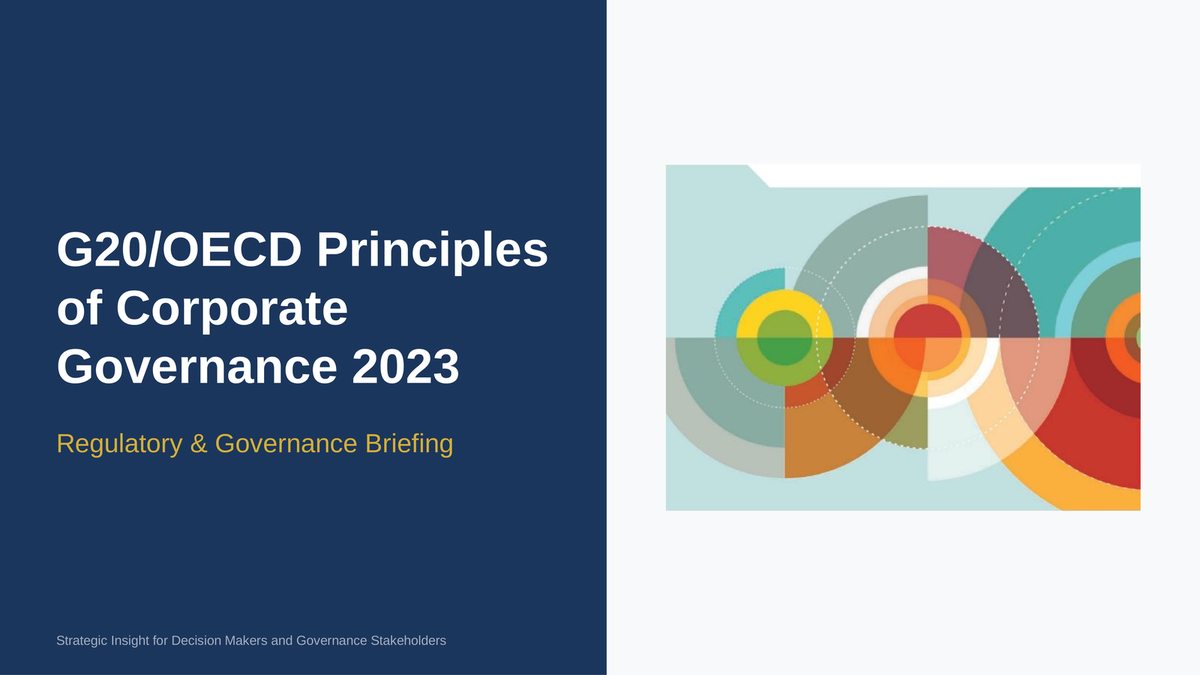}{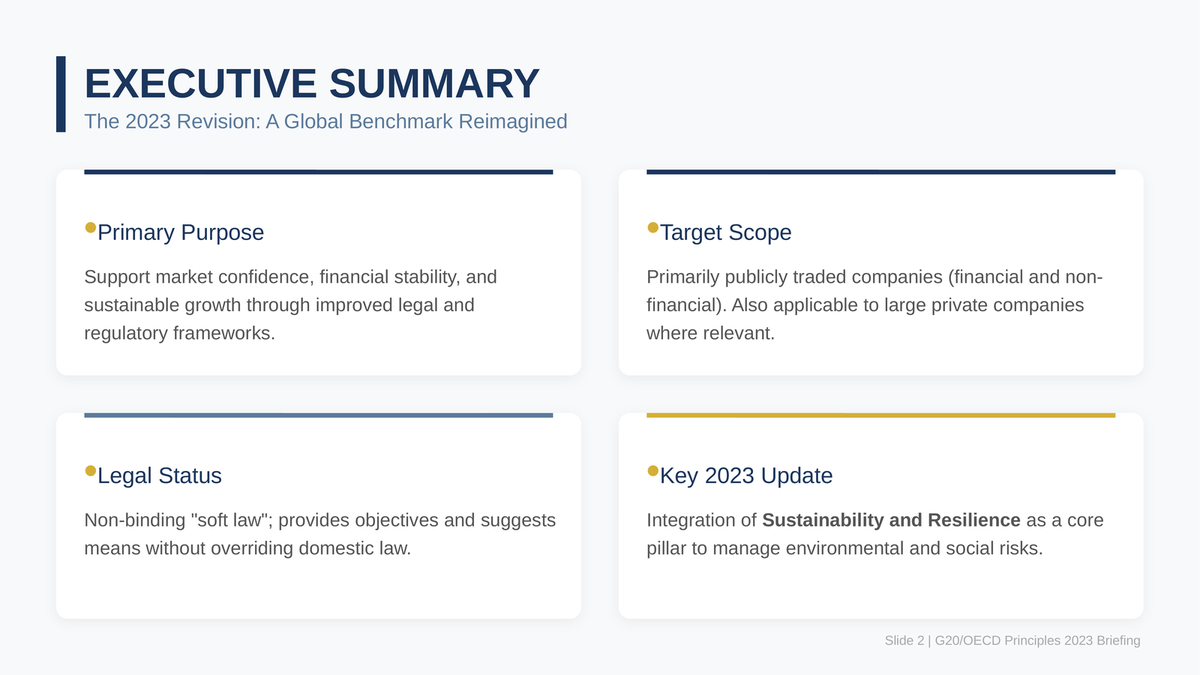}{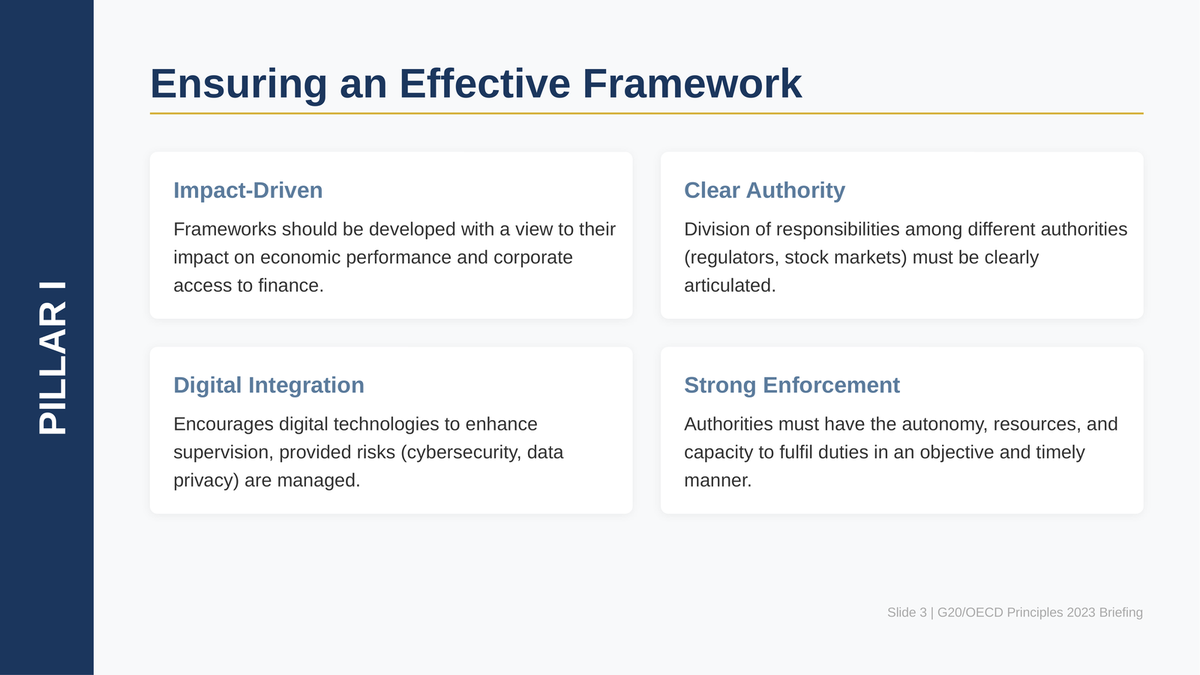}{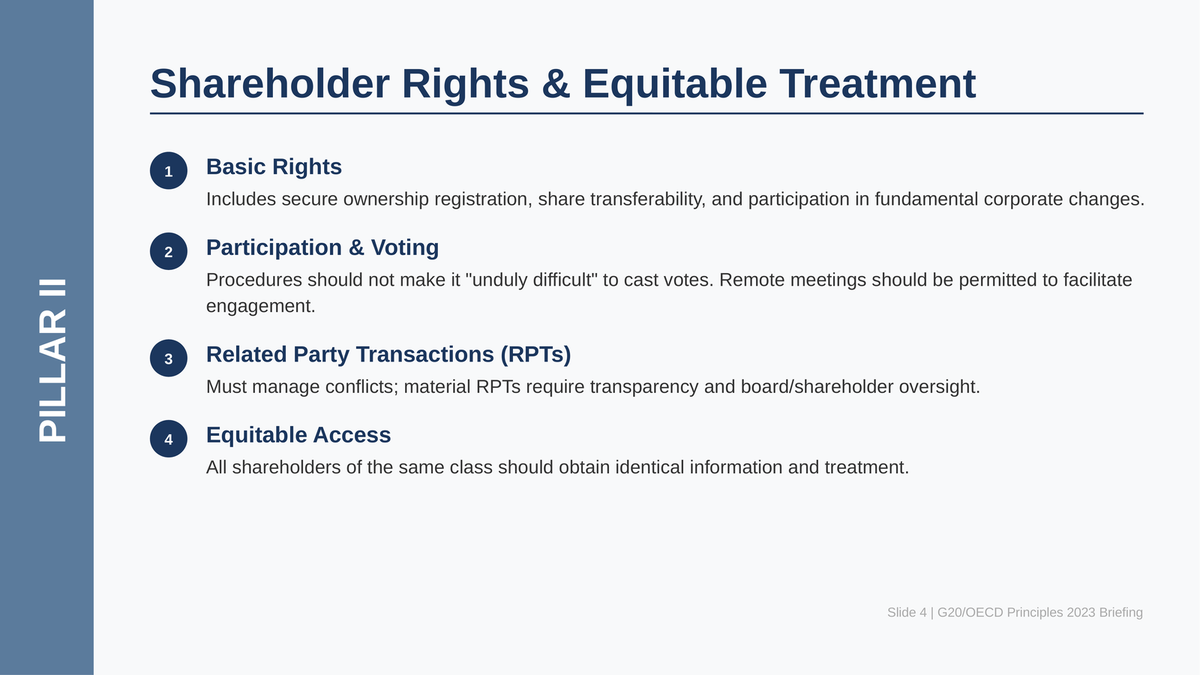}{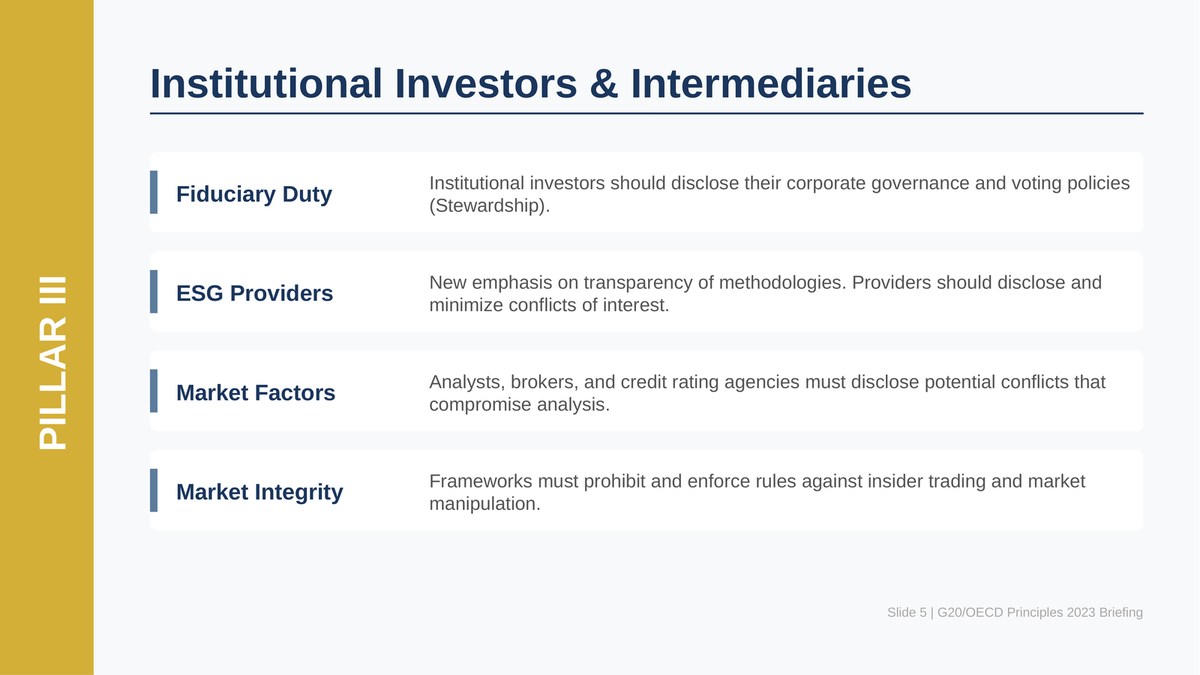}
\deckshowcaserow{SlideTailor / Decision maker: AudCov. 0.368, Correct. 0.926, SafeEff 4.391}{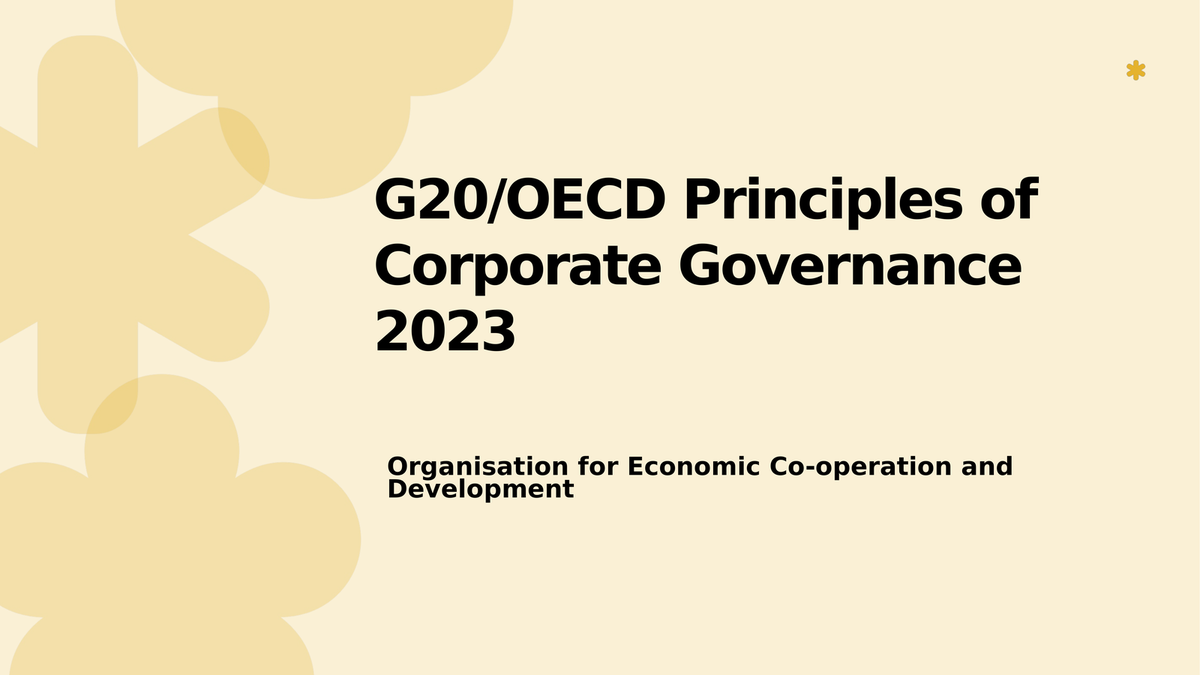}{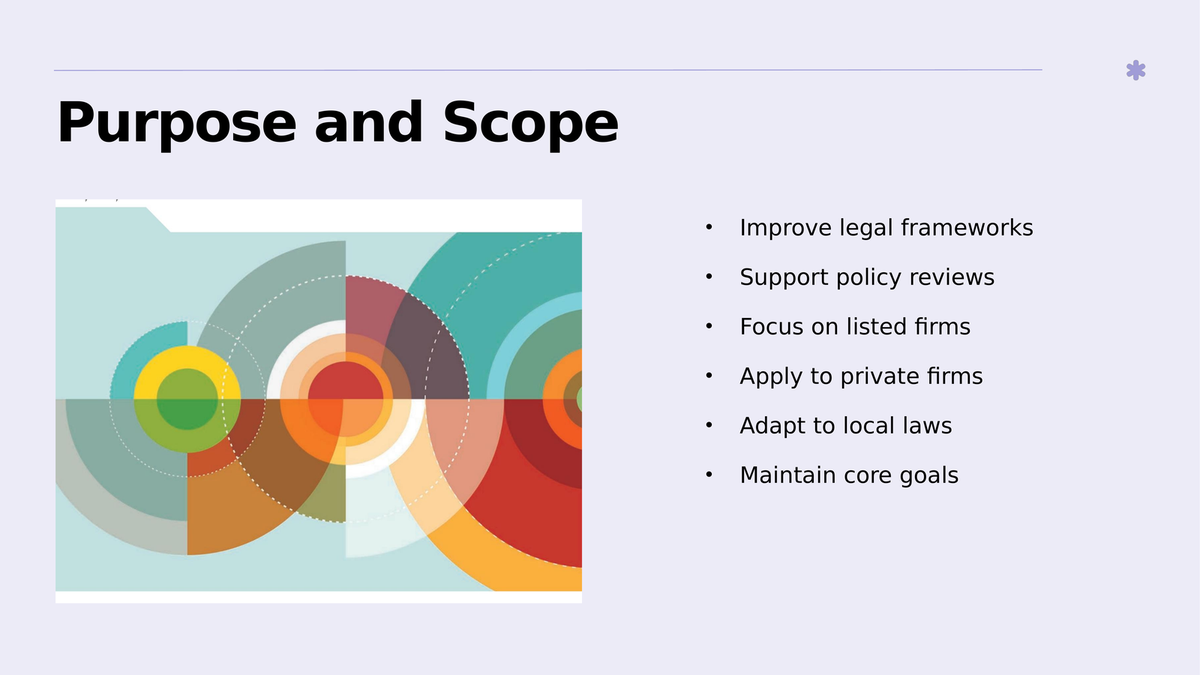}{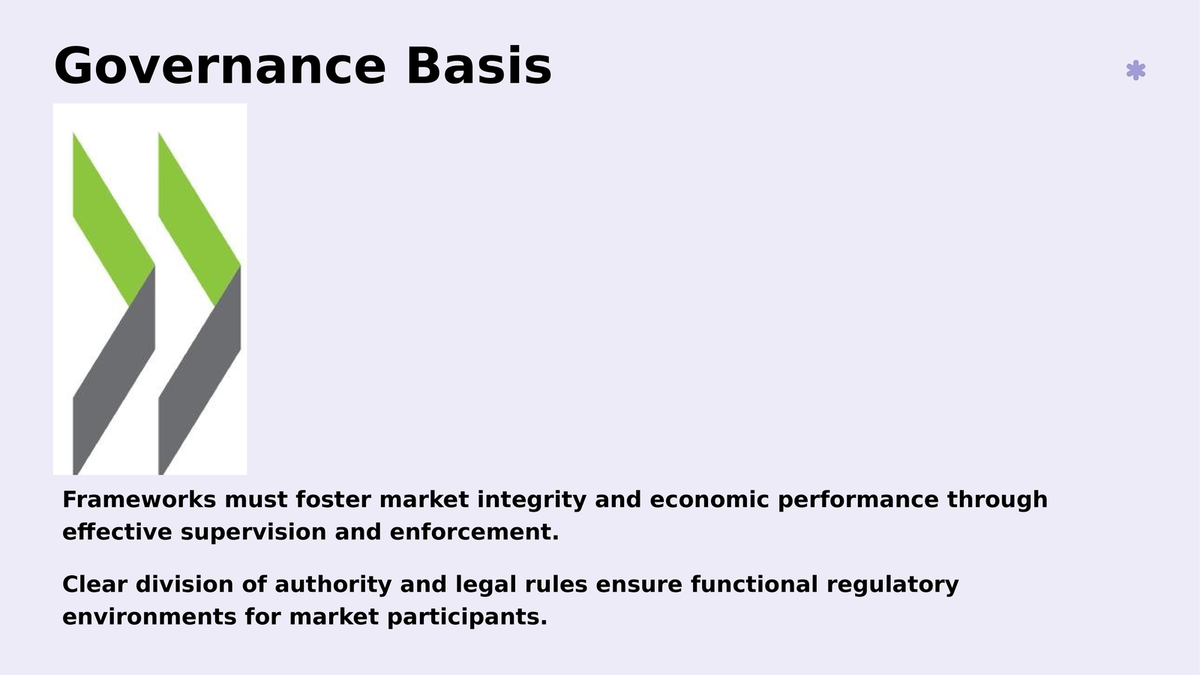}{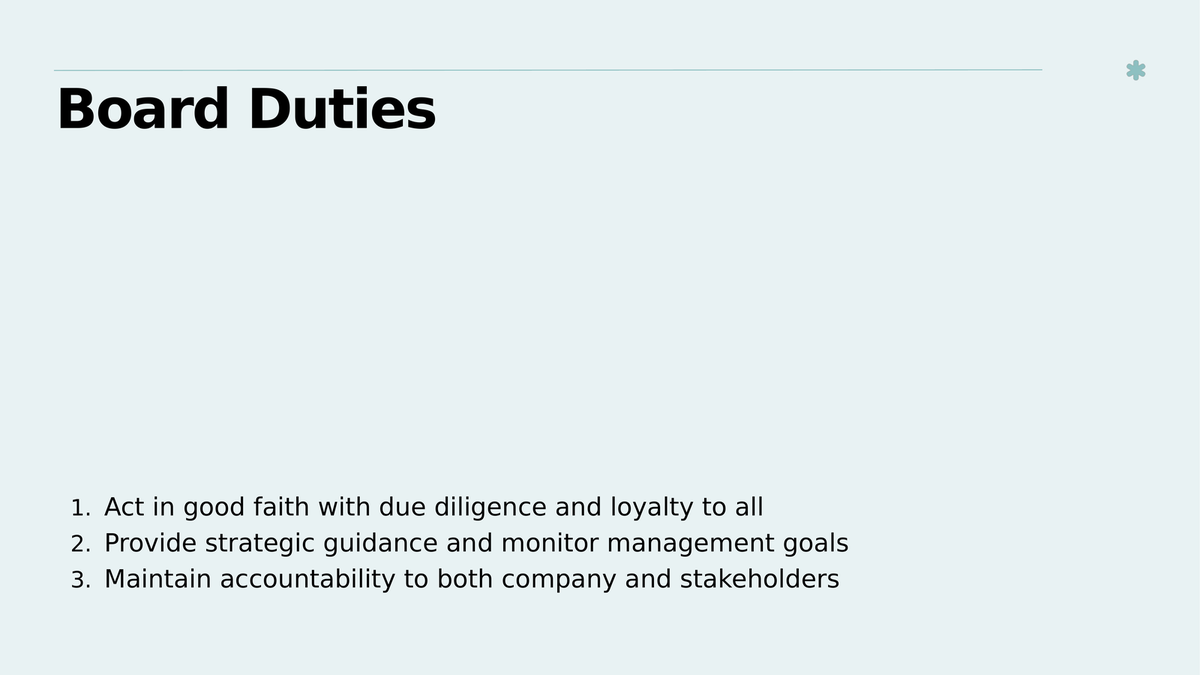}{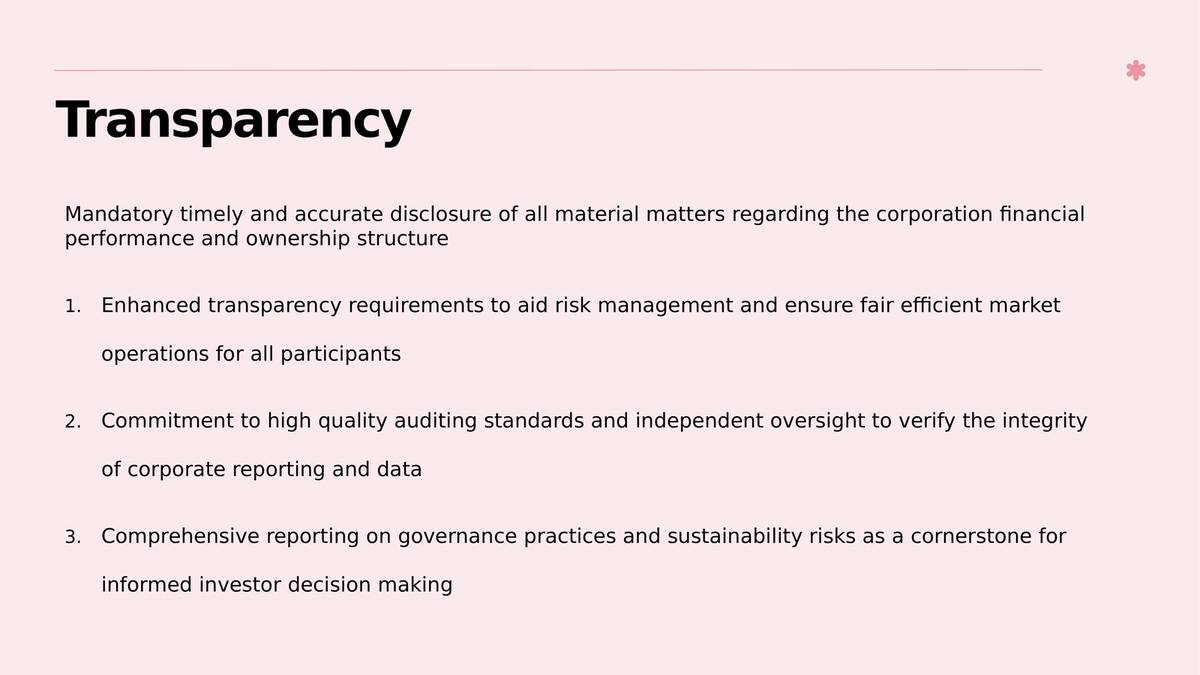}
}
\caption{Deck showcase for G20/OECD Principles of Corporate Governance 2023. This corporate-governance example shows a case where SlideTailor keeps higher correctness for the decision-maker row.}
\label{fig:deck_showcase_case016}
\end{figure}

\clearpage

\subsection{Compute Resources}
\label{subsec:appendix_compute}

We conducted our experiments on a local workstation equipped with two Intel Xeon Platinum 8457C CPUs, 192 logical CPU cores, 251 GiB of RAM, and 8 NVIDIA RTX 4090 GPUs. Tasks such as probe generation, utility weighting, slide answerability, and correctness scoring relied on external endpoints for LLMs. Local computing resources were primarily dedicated to parsing documents, rendering slides, orchestrating jobs, and executing SlideTailor. Table~\ref{tab:compute_resources} provides a summary of the completed jobs with valid timestamps for their start and finish. The summed time represents the wall time per job and is larger than the actual elapsed time because we executed many jobs in parallel.

\begin{table}[!t]
\centering
\caption{Compute time recorded for each stage of the pipeline. All times are in minutes except for the final column.}
\label{tab:compute_resources}
{\footnotesize
\setlength{\tabcolsep}{5pt}
\renewcommand{\arraystretch}{0.96}
\begin{tabular}{@{}lrrrr@{}}
\toprule
\textbf{Stage} & \textbf{Jobs} & \textbf{Median} & \textbf{Range} & \textbf{Summed hours} \\
\midrule
Generation of probes & 113 & 6.8 & 0.3--66.8 & 16.1 \\
Generation with DeepPresenter & 184 & 7.6 & 3.6--95.4 & 29.1 \\
Generation with SlideTailor & 168 & 5.2 & 2.4--25.2 & 18.2 \\
Evaluation of the deck & 129 & 36.5 & 2.9--2410.7 & 204.9 \\
\bottomrule
\end{tabular}
}
\end{table}

\subsection{Existing Assets, Licenses, and Release Policy}
\label{subsec:appendix_assets}

\oursys utilizes source documents in the public domain and existing systems for slide generation. We redistribute the PDFs of academic papers in our benchmark package, along with their original URLs and metadata. For non-academic documents with potential copyright limitations, we provide source URLs, metadata, document identifiers, and derived artifacts from the benchmark, such as probes and utility weights. Table~\ref{tab:asset_licenses} summarizes the assets we used for the experiments and the release.

\begin{table}[!t]
\centering
\caption{Assets used and their treatment for release.}
\label{tab:asset_licenses}
{\footnotesize
\setlength{\tabcolsep}{4pt}
\renewcommand{\arraystretch}{0.96}
\resizebox{\linewidth}{!}{
\begin{tabular}{@{}lll@{}}
\toprule
\textbf{Asset} & \textbf{Use in \oursys} & \textbf{License or release treatment} \\
\midrule
PPTAgent / DeepPresenter~\cite{PPTAgent,DeepPresenter} & Baseline for slide generation & The code for PPTAgent is under MIT, while DeepPresenter is cited as the successor system. \\
SlideTailor~\cite{SlideTailor} & Baseline for slide generation & MIT license within the released codebase. \\
PresentBench~\cite{PresentBench} & Reference for source and benchmark & Apache 2.0 license within the released codebase. \\
SlideCoder / SlideMaster~\cite{SlideCoder} & Reference for related implementation & Apache 2.0 license within the inspected codebase. \\
PPTBench-Eval~\cite{PPTBench} & Reference for related evaluation & MIT license within the inspected codebase. \\
Academic paper sources & Source documents and probes & Redistributed with original metadata and URLs. \\
Non-academic sources & Source documents and probes & Raw files not redistributed, while metadata, URLs, and probes are released. \\
LLM services & Generation, weighting, and scoring & Accessed via API terms from providers, where no model weights are redistributed. \\
\bottomrule
\end{tabular}
}
}
\end{table}




\end{document}